\documentclass[11pt,letterpaper]{article}
\usepackage[T1]{fontenc}
\usepackage[dvipsnames]{xcolor}
\usepackage[letterpaper,textwidth=7.05in,textheight=9.35in,top=0.80in,headheight=14pt,headsep=11pt,footskip=23pt]{geometry}
\usepackage{fancyhdr,titlesec,tgheros,tcolorbox}
\usepackage{newtxtext}
\usepackage{newtxmath}
\usepackage{graphicx,booktabs,xspace}
\usepackage[shortlabels,inline]{enumitem}
\usepackage[numbers,sort&compress]{natbib}
\usepackage[font=small,labelfont=bf,labelsep=period]{caption}
\usepackage[font=footnotesize]{subcaption}
\usepackage{listings}
\usepackage{microtype}
\usepackage{array,tabularx}
\usepackage{multirow}
\usepackage{pifont}
\usepackage{dblfloatfix} 
\AtBeginDocument{%
  \setlength{\abovedisplayskip}{5pt plus 1pt minus 2pt}%
  \setlength{\belowdisplayskip}{5pt plus 1pt minus 2pt}%
  \setlength{\abovedisplayshortskip}{3pt plus 1pt minus 2pt}%
  \setlength{\belowdisplayshortskip}{3pt plus 1pt minus 2pt}%
}
\definecolor{reportInk}{HTML}{2D263B}
\definecolor{reportAccent}{HTML}{5E4B81}
\definecolor{reportPanel}{HTML}{EFEDF3}
\definecolor{reportRule}{HTML}{D3CBDF}
\definecolor{reportSecondary}{HTML}{574B68}
\usepackage[pagebackref,breaklinks,colorlinks,allcolors=reportAccent]{hyperref}
\providecommand{\cmark}{\textcolor{green!60!black}{\ding{51}}}
\providecommand{\xmark}{\textcolor{red!75!black}{\ding{55}}}
\newcolumntype{Y}{>{\raggedright\arraybackslash}X}
\hypersetup{pdftitle={EmbodiedStudio: A Task-Driven Simulation Generation Platform for Embodied Learning and Evaluation},pdfauthor={Yang Zhao and Xubo Yang}}
\titleformat{\section}{\Large\sffamily\bfseries\color{reportInk}}{\thesection}{0.65em}{}
\titleformat{\subsection}{\large\sffamily\bfseries\color{reportInk}}{\thesubsection}{0.65em}{}
\titleformat{\subsubsection}{\normalsize\sffamily\bfseries\color{reportInk}}{\thesubsubsection}{0.65em}{}
\titlespacing*{\section}{0pt}{1.15em plus 2pt minus 2pt}{0.45em}
\titlespacing*{\subsection}{0pt}{0.9em plus 2pt minus 2pt}{0.35em}
\begin{document}
\thispagestyle{empty}
\enlargethispage{5pt}
\pdfbookmark[0]{Title and Abstract}{report-cover}
\begin{tcolorbox}[colback=reportPanel,colframe=reportPanel,boxrule=0pt,
  sharp corners,left=17pt,right=17pt,top=15pt,bottom=14pt,
  boxsep=0pt,before skip=0pt,after skip=0pt]
\setlength{\parskip}{0pt}
{\sffamily\bfseries\color{reportInk}\fontsize{36}{39}\selectfont EmbodiedStudio\par}
\vspace{6pt}
{\sffamily\color{reportInk}\fontsize{16}{20}\selectfont
A Task-Driven Simulation Generation Platform for\\
Embodied Learning and Evaluation\par}
\vspace{10pt}
{\sffamily\bfseries Yang Zhao\quad Xubo Yang\textsuperscript{*}\par}
\vspace{3pt}
{\small\sffamily\color{reportSecondary}Shanghai Jiao Tong University\quad
\textsuperscript{*}Correspondence: \href{mailto:yangxubo@sjtu.edu.cn}{yangxubo@sjtu.edu.cn}\par}
\vspace{9pt}
{\color{reportRule}\rule{\linewidth}{0.55pt}\par}
\vspace{7pt}
{\sffamily\bfseries\color{reportInk}Abstract\par}
\vspace{4pt}
{\small
Training and evaluating robots across diverse tasks requires a supply of simulation environments suited to those tasks. While recent scene and task generators offer a promising way to expand this supply, delivering a usable task environment still faces two practical gaps: generated scenes that look plausible in appearance often fail to support the intended task interactions or remain physically stable in simulation, and generated scoring programs frequently attempt task conditions without implementing them faithfully. We present \textbf{EmbodiedStudio}, a task-driven simulation generation platform that bridges these gaps by jointly authoring scene, task, robot-behavior, and rubric programs under closed-loop simulation feedback and independent review. Crucially, as the coding agent iteratively debugs scenes, behaviors, and stateful rubrics in simulation, this closed-loop process natively yields both forward \emph{construction experience} for learning to build environments and physical \emph{execution experience} for policy learning, all bound to frozen environment revisions for repeatable evaluation. Across scene generation, rubric evaluation, ablations, and learning studies, EmbodiedStudio substantially improves task readiness and rubric faithfulness while improving physical stability and retaining actionable supervision signals for layout authoring and policy adaptation.
\par}
\vspace{9pt}
{\small\sffamily\bfseries
\href{https://embodiedstudio.github.io/?view=paper}{Project page}
\quad\textcolor{reportSecondary}{\textbar}\quad
\href{https://embodiedstudio.github.io/?view=paper\#walkthrough}{Video walkthrough}
\quad\textcolor{reportSecondary}{\textbar}\quad
\href{https://embodiedstudio.github.io/?view=paper\#appendix-rollouts}{Robot recordings}\par}
\end{tcolorbox}

\vspace{8pt}
\noindent\begin{minipage}{\textwidth}
\centering
\includegraphics[width=\linewidth]{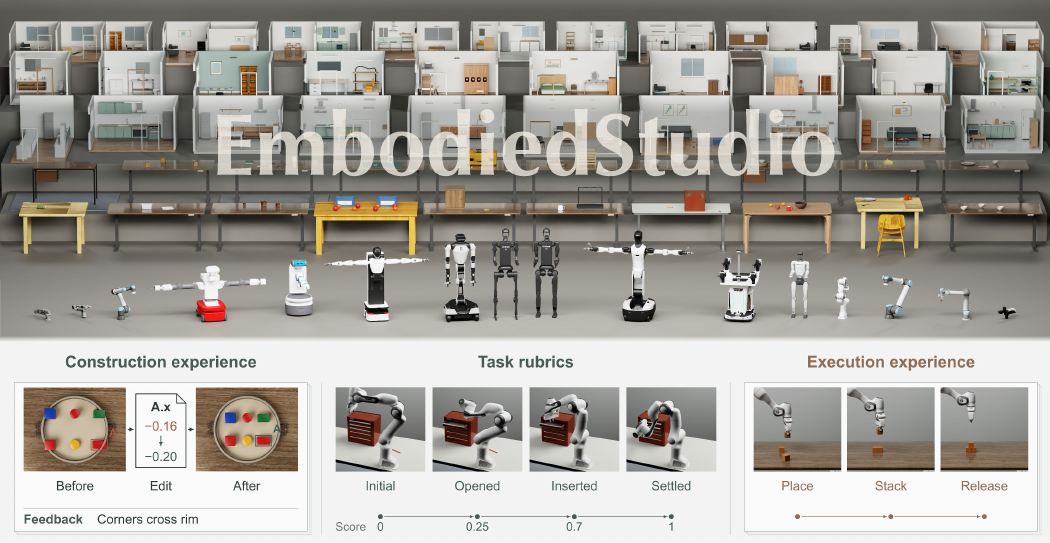}
\captionsetup{hypcap=false}
\captionof{figure}{\textbf{Task-conditioned simulation scenes and their associated experience.} Top: generated tabletop, single-room, and multi-room scenes, arranged in a shared display space with whole-scene scaling and architectural cutaways. The foreground shows 16 integrated robot models at a common display scale. Bottom: construction revisions and feedback (left), scene-bound task rubrics evaluated on recorded states (center), and robot execution experience (right).}
\label{fig:teaser}
\end{minipage}

\clearpage
\section{Introduction}
Robots must operate across varied objects, layouts, and interaction conditions. Simulation can supply environments for practicing and assessing these capabilities~\cite{procthor,robocasa}, yet each task scene requires suitable assets, physical relationships, and task definitions~\cite{simfoundry,embodiedgenv2}. Scaling this supply calls for generating complete, task-ready simulation environments rather than static 3D layouts alone.

Task readiness extends well beyond visual plausibility. Opening a drawer, placing a pen inside, and closing it requires an operable mechanism, unobstructed workspace, usable interior volume, and scoring logic that captures the ordered sequence. Although recent work advances scene and task generation from single-pass layout prediction to physics-aware optimization and multi-turn agentic refinement~\cite{physcene,sceneweaver,scenesmith,sage,scenerevis,gensim2}, two practical gaps remain between generating a scene and delivering a usable task environment. First, generated scenes that match object requests in appearance often lack task readiness or physical stability when stepped in simulation. Second, generated scoring programs frequently attempt the requested conditions yet fail to implement them faithfully, leaving a large gap between requirement coverage and correct expression.

Bridging these gaps requires shifting from standalone scene synthesis to joint authoring of scenes, robot behaviors, and task rubrics under closed-loop simulation feedback and independent review, verifying physical readiness and scoring fidelity together. Moreover, this closed-loop process unifies two forms of training experience that prior pipelines obtain separately or synthetically. Whereas prior work on learning iterative scene construction relies on reverse-engineered editing trajectories~\cite{scenerevis} and robot data generators typically collect trajectories in pre-built or separately defined tasks~\cite{mimicgen,embodiedswe}, an agent that iteratively debugs scenes, behaviors, and rubrics in simulation natively produces both: \emph{construction experience}, which records forward authoring actions and diagnostic feedback to supervise environment-generation models, and \emph{execution experience}, which records physical robot interactions in the resulting scene to train visuomotor policies. Associating both system-generated experiences with the frozen scene, task, and executable rubric turns task-driven scene generation into a unified process for authoring evaluable environments while capturing learnable construction and execution supervision.

To this end, we introduce \textbf{EmbodiedStudio}, a task-driven simulation generation platform centered on the executable task scene (Figure~\ref{fig:teaser}). Given a natural-language request, a coding agent develops coupled scene, task, robot-behavior, and rubric programs through simulator observations, physical measurements, execution feedback, and an independent internal reviewer. The system retains forward construction histories, executes and records physical robot trajectories when requested, and binds stateful rubrics to committed scene facts. Freezing environment revisions decouples scene construction from subsequent behavior updates and external policy evaluation under identical task conditions.

Our main contributions are:
\begin{itemize}[leftmargin=*,nosep,topsep=2pt]
  \item \textbf{Role-separated authoring platform:} We introduce EmbodiedStudio, a closed-loop platform that jointly authors task-ready scenes, robot behaviors, and rubrics from natural-language requests while separating program authoring from physics-grounded measurement and independent review---preventing author self-evaluation bias and unifying debugging with forward construction and execution experience capture.
  \item \textbf{Decoupled representation and stateful rubrics:} We formulate environments over a factorized representation and frozen revisions that decouple spatial ownership, physical relations, task bindings, and robot control from access-isolated stateful rubrics, supporting multi-scale synthesis, layout-invariant evaluation, and replayable supervision ($\mathcal H_x, \mathcal D_x$).
  \item \textbf{Empirical study across scenes, rubrics, and learning:} Comprehensive evaluations across multi-scale scenes, shared-scene rubric faithfulness, mechanism ablations, multi-platform robot rollouts, and downstream layout and policy learning demonstrate consistent gains over baselines and verify how simulation feedback and independent review drive physical stability.
\end{itemize}

\section{Related Work}
\begin{table*}[!t]
\centering
\caption{\textbf{High-level capability comparison across representative scene, task, and embodied simulation generators.}}
\label{tab:capability_comparison}
\setlength{\tabcolsep}{4.2pt}
\renewcommand{\arraystretch}{0.86}
\resizebox{\textwidth}{!}{%
\begin{tabular}{llccccccccc}
\toprule
\textbf{\shortstack{Primary\\Focus}} & \textbf{Method} & \textbf{Tabletop} & \textbf{\shortstack{Single\\Room}} & \textbf{\shortstack{Multi\\Room}} & \textbf{\shortstack{Sim-Refined\\Scene}} & \textbf{\shortstack{Task-Driven\\Scene}} & \textbf{\shortstack{Behavior Program\\Generation ($B$)}} & \textbf{\shortstack{Stateful\\Rubric ($R$)}} & \textbf{\shortstack{Construction\\Exp. ($\mathcal{H}_x$)}} & \textbf{\shortstack{Execution\\Exp. ($\mathcal{D}_x$)}} \\
\midrule
\multirow{6}{*}{Scene Synthesis}
 & Holodeck~\cite{holodeck} & \cmark & \cmark & \cmark & \xmark & \xmark & \xmark & \xmark & \xmark & \xmark \\
 & LayoutVLM~\cite{layoutvlm} & \cmark & \cmark & \xmark & \xmark & \xmark & \xmark & \xmark & \xmark & \xmark \\
 & SceneReVis~\cite{scenerevis} & \xmark & \cmark & \xmark & \xmark & \xmark & \xmark & \xmark & \cmark & \xmark \\
 & SceneSmith~\cite{scenesmith} & \cmark & \cmark & \cmark & \cmark & \cmark & \xmark & \xmark & \xmark & \xmark \\
 & EmbodiedGen V2~\cite{embodiedgenv2} & \cmark & \cmark & \cmark & \cmark & \cmark & \xmark & \xmark & \xmark & \xmark \\
 & SAGE~\cite{sage} & \cmark & \cmark & \cmark & \cmark & \cmark & \xmark & \xmark & \xmark & \cmark \\
\midrule
\multirow{6}{*}{Task \& Simulation}
 & GenSim~\cite{gensim} & \cmark & \xmark & \xmark & \xmark & \cmark & \xmark & \cmark & \xmark & \cmark \\
 & RoboGen~\cite{robogen} & \cmark & \xmark & \xmark & \xmark & \cmark & \cmark & \xmark & \xmark & \cmark \\
 & SimFoundry~\cite{simfoundry} & \cmark & \xmark & \xmark & \cmark & \cmark & \xmark & \xmark & \xmark & \cmark \\
 & RoboLab~\cite{robolab} & \cmark & \xmark & \xmark & \cmark & \xmark & \xmark & \cmark & \xmark & \xmark \\
 & EmbodiedSmith~\cite{embodiedsmith} & \cmark & \cmark & \xmark & \cmark & \cmark & \cmark & \cmark & \xmark & \cmark \\
\cmidrule{2-11}
 & \textbf{EmbodiedStudio (Ours)} & \textbf{\cmark} & \textbf{\cmark} & \textbf{\cmark} & \textbf{\cmark} & \textbf{\cmark} & \textbf{\cmark} & \textbf{\cmark} & \textbf{\cmark} & \textbf{\cmark} \\
\bottomrule
\end{tabular}%
}
\vspace{-10pt}
\end{table*}

\paragraph{Scene generation and quality.}
Procedural systems such as ProcTHOR and Infinigen Indoors scale indoor creation via rule-based synthesis~\cite{procthor,infinigenindoors}, while data-driven models learn layout distributions or semantic graph conditioning~\cite{atiss,diffuscene,instructscene}. For open-ended instructions, one-pass and optimization-based methods map language to spatial layouts via LLM planning, relational graphs, or differentiable objectives~\cite{layoutgpt,holodeck,idesign,layoutvlm}, and recent works enforce physical interactivity or structural constraints in PhyScene, PhyScene3D, HOG-Layout, ZoneMaestro, and VoxScene~\cite{physcene,physcene3d,hoglayout,zonemaestro,voxscene}.

Beyond single-pass prediction, agentic frameworks iteratively inspect and refine scenes: SceneWeaver, SceneSmith, SAGE, and EmbodiedGen V2 equip agents with visual and simulation tools~\cite{sceneweaver,scenesmith,sage,embodiedgenv2}, SceneReVis trains a reflective model on reverse-engineered editing trajectories via multi-turn RL~\cite{scenerevis}, and SceneEval evaluates text--scene semantic alignment~\cite{sceneeval}. While recent agentic synthesizers such as SceneSmith and SAGE~\cite{scenesmith,sage} refine layouts with visual and physics feedback, geometric plausibility or static stability still does not ensure task-ready environments (Table~\ref{tab:scene}). By combining factorized spatial--physical--task grounding across scales with role-separated physical measurement and independent read-only review, EmbodiedStudio bridges plausible layout synthesis and simulation-ready task scenes (Table~\ref{tab:capability_comparison}).

\paragraph{Task construction, feedback, and evaluation.}
Generative simulation frameworks synthesize task programs, demonstrations, or training supervision, as in GenSim, GenSim2, RoboGen, and FactorSim~\cite{gensim,gensim2,robogen,factorsim}. Whereas Eureka optimizes dense reward functions for RL exploration~\cite{eureka}, task evaluation requires scoring logic that faithfully verifies completion conditions. Standard benchmarks---RLBench, LIBERO, BEHAVIOR-1K, and RoboCasa---provide curated task suites~\cite{rlbench,libero,behavior1k,robocasa}. Toward automated task supply, SimFoundry constructs real-to-sim scenes for policy learning, RoboLab generates simulation tasks to benchmark generalist policies, and concurrent EmbodiedSmith jointly synthesizes assets, scenes, tasks, and demonstrations~\cite{simfoundry,robolab,embodiedsmith}. Because generated success code often attempts task conditions without implementing them faithfully---and prior task generators either target tabletop scales~\cite{gensim,robolab}, omit stateful rubrics and multi-room scenes~\cite{robogen}, or couple generation and evaluation within single rooms~\cite{embodiedsmith}---we formulate evaluation as access-isolated, stateful rubrics grounded in measured physical facts over decoupled task bindings and frozen revisions, preserving faithfulness across scales and layout variations.

\paragraph{Experience generation for policy and layout learning.}
Action generators such as Code as Policies, VoxPoser, and ReKep synthesize control programs or keypoint constraints~\cite{codeaspolicies,voxposer,rekep}, while Scaling Up and Distilling Down couples trajectory generation with policy distillation~\cite{scalingup}. MimicGen, DemoGen, and RoboTwin 2.0 expand demonstrations across spatial variations or domain randomization~\cite{mimicgen,demogen,robotwin2}, and concurrent EmbodiedSWE uses coding agents to collect long-horizon demonstrations~\cite{embodiedswe}. Whereas demonstration pipelines such as MimicGen, RoboTwin~2.0, and EmbodiedSWE~\cite{mimicgen,robotwin2,embodiedswe} operate in pre-built or separately defined tasks---and SceneReVis~\cite{scenerevis} learns from reverse-engineered static edits---EmbodiedStudio unifies interactive authoring and physical execution over task-agnostic primitives, natively yielding leak-free forward construction histories ($\mathcal{H}_x$) alongside rubric-verified execution trajectories ($\mathcal{D}_x$) across diverse robot embodiments on the same frozen revisions.

\begin{figure*}[!t]
  \centering
  \includegraphics[width=\textwidth]{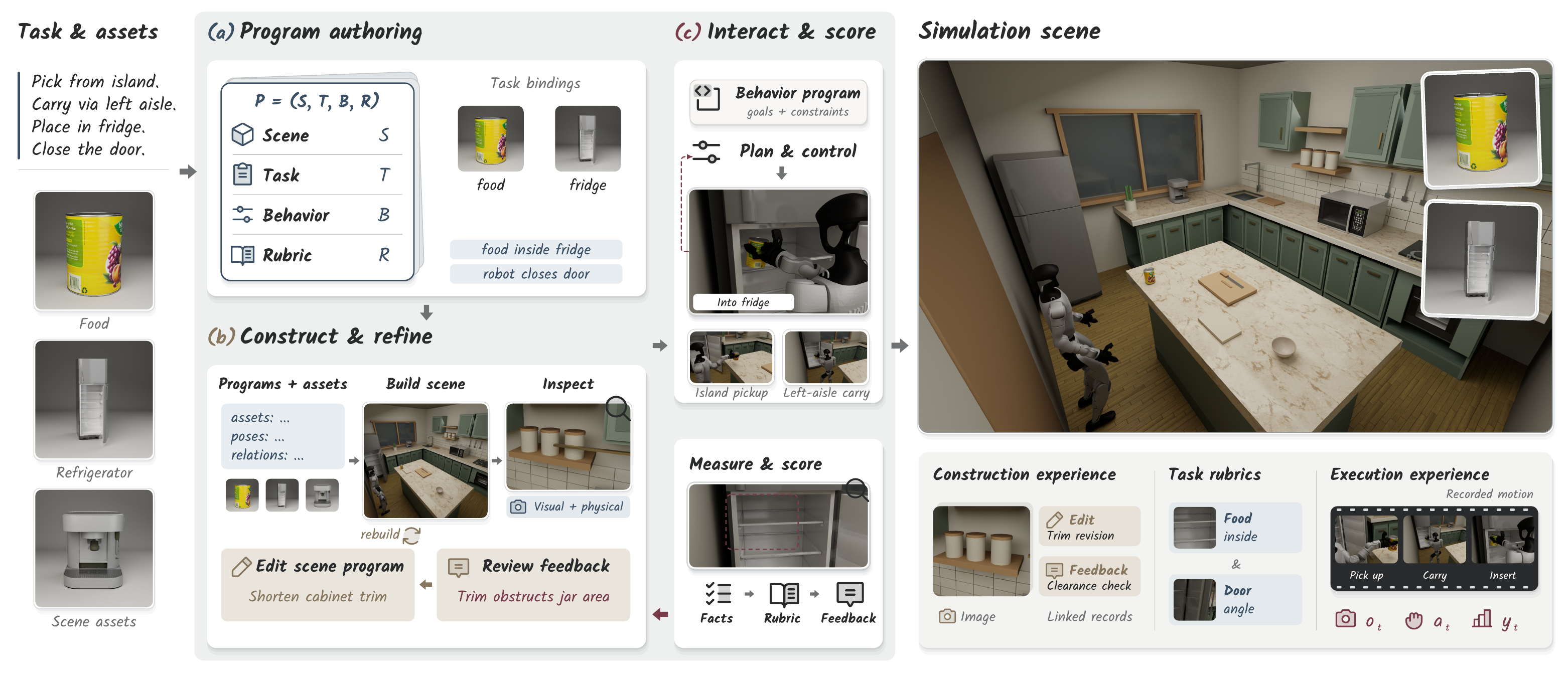}
  \caption{\textbf{Scene-centered construction, execution, and feedback.} A coding agent develops scene and task programs through simulator observations and internal review. Construction experience records this process; execution experience records robot interactions in the resulting scene. Task rubrics read measured facts and remain bound to the frozen scene and task during subsequent use.}
  \label{fig:method}
  \vspace{-4pt}
\end{figure*}

\section{Framework and System Design}
\label{sec:method}
Given a task request and asset library, EmbodiedStudio enforces role separation across authoring, measurement, and verification over minimal task-agnostic primitives while keeping task semantics in versioned programs centered on an executable scene (Figure~\ref{fig:method}). The platform operates through decoupled authoring (Figure~\ref{fig:method}a), closed-loop construction and review (Figure~\ref{fig:method}b), and grounded execution, scoring, and consumer acceptance (Figure~\ref{fig:method}c).

\subsection{Task conditions and decoupled representation}
Given a request $x$ and available asset resources $\mathcal A$, the agent represents a task instance as a coupled program tuple (Figure~\ref{fig:method}a):
\begin{equation}
P=(S,T,B,R),
\end{equation}
where $S$ defines scene entities, physical properties, and spatial relationships; $T$ binds task-relevant entities and parameters (e.g., binding \texttt{food} and \texttt{fridge} in Figure~\ref{fig:method}a); $B$ specifies robot behavior when requested; and $R$ defines task progress, completion, and termination conditions. Across tabletop, single-room, and multi-room scales, $S$ organizes entities into a unified hierarchical scene tree with stable entity identities, spatial bounds, and functional roles. Unlike rigid single-parent scene trees that conflate geometric ownership with physical support, spatial ownership, coordinate frames, physical support or containment relations, and task bindings are factorized separately---so an object inside an articulated drawer is supported by it without becoming its rigid kinematic child, and layout seeds can vary without rewriting scoring code.

Compiling $(S,T,R)$ produces an immutable revision $W_v=\operatorname{Build}(S,T,R;\mathcal A)$ as the canonical visual--physics authority, grounding geometry, rendered views, task bindings, rubric logic, and embodiment interfaces in native bindings (decoupling task semantics from robot morphologies and controllers $B$). Parameterized ranges in $S$ compile distinct layout seeds into separate revisions $W_v$ (e.g., train/test splits), while $B$ evolves independently against fixed $W_v$.

\subsection{Scene construction and construction experience}
During construction (Figure~\ref{fig:method}b), the native coding agent operates on the same unified build-and-execution interface used for production, sourcing assets, authoring spatial relations, and configuring robots without a separate trace pipeline. Asset sourcing prioritizes simulation-ready articulated and curated libraries before general 3D repositories, falling back to self-contained online retrieval or procedural asset scripts when gaps arise. At iteration $k$, conditioned on $x$, current programs $P_k$, and evidence $E_k$, the agent selects an edit or tool action $u_k=\mathcal G_\theta(x,P_k,E_k)$, observing simulator or review feedback $v_k$ (build diagnostics, renders, physical measurements, and rollouts) into $E_{k+1}$ and applying diffs $\Delta P_k$ to obtain $P_{k+1}$. Rather than author self-approval, an independent reviewer inspects renders, measurements, and bindings in $P_{k+1},E_{k+1}$ from an isolated read-only context against $x$ (without edit access or authority to bypass measurements, e.g., flagging cabinet trim obstructing the jar area in Figure~\ref{fig:method}b), prompting revisions before delivery.

We retain the observable forward construction history $\mathcal H_x=\bigl(x,\{(u_k,v_k,\Delta P_k)\}_{k=1}^{K}\bigr)$, where trace references link rendered images, code revisions, and diagnostic outcomes across all attempts, including failed or interrupted steps. To prevent future-information leakage when supervising environment-authoring models (Section~\ref{sec:learning}), each training input exposes only context visible prior to decision $u_k$, while subsequent consequences are recorded separately.

\subsection{Robot execution and interaction experience}
For robot execution (Figure~\ref{fig:method}c, top), the agent authors manipulation stages, spatial goals, motion constraints, and recovery strategies in $B$ using task-agnostic motion planning and control primitives. Candidate plans execute inside an isolated boundary via closed-loop steps receiving only declared observations $o_t$ (never privileged simulator handles or state overrides), stepping $W_v$ under frozen behavior $B_b$ to yield physical trajectories $\tau_{v,b}=\operatorname{Run}(W_v,B_b)$. Each episode records step-wise observations $o_t$, requested and executed control actions $a_t$, measured physical states, and rubric outputs $y_t$, linked to revisions $(v,b)$ and the initial state alongside cameras, action conventions, and simulation clocks. Failed and interrupted rollouts are preserved explicitly, while successful episodes form a rubric-verified demonstration subset within $\mathcal D_x$ for policy learning (Section~\ref{sec:policy-learning}).

\subsection{Task rubrics and environment use}
To evaluate task execution faithfully (Figure~\ref{fig:method}c, bottom), the agent authors the rubric program $R$ from task requirements and scene bindings prior to scoring rollouts. For multi-stage tasks—such as opening a drawer, inserting a pen, and closing it, or placing food inside a refrigerator and closing the door (Figure~\ref{fig:method})—the rubric tracks ordered events and persistent conditions through a finite-state transition:
\begin{equation}
(z_{t+1},y_t)=R(T,f_t,z_t),
\end{equation}
where $f_t$ denotes committed simulator measurements at step $t$, $z_t$ is the rubric's internal finite state, and $y_t$ reports step-wise progress, reward, completion, termination, and diagnostic details. $R$ executes in a read-only isolated process with no simulator control handles, separating policy observations $o_t$ from privileged measurements $f_t$. Because $W_v$ and $R$ are frozen, scores can be deterministically recomputed from logged states and audited independently (Section~\ref{sec:rubric-eval}).

Finally, an evaluable package $\mathcal E_x$ bundles $(W_v,T,R)$ with a standard rollout interface and is accepted only when independent downstream consumers reload, replay, and recompute records in fresh processes, linking $\mathcal E_x$ to $\mathcal H_x$ and $\mathcal D_x$.

\section{Experiments}

\subsection{Experimental setup}
\label{sec:setup}
Our evaluation answers four questions: \textbf{(Q1)} whether joint authoring improves both semantic alignment and task readiness over existing scene generators (Section~\ref{sec:scene-eval}); \textbf{(Q2)} whether scene-bound rubrics faithfully express original task requirements (Section~\ref{sec:rubric-eval}); \textbf{(Q3)} how simulation-feedback revision and independent review contribute to generation quality (Section~\ref{sec:ablations}); and \textbf{(Q4)} whether the resulting physical interactions and construction histories support downstream robot execution and learning (Sections~\ref{sec:interactions}, \ref{sec:learning}--\ref{sec:policy-learning}).

\paragraph{Requests.}
The common suite contains 50 natural-language task requests: 20 tabletop, 20 single-room, and 10 multi-room. The rubric study reuses the 20 tabletop environments. The complete request list is provided in the supplementary material.

\paragraph{Baselines.}
For scene generation, we compare against language-conditioned layout methods (Holodeck~\cite{holodeck}, LayoutVLM~\cite{layoutvlm}) and agentic simulation-scene generators (SAGE~\cite{sage}, EmbodiedGen V2~\cite{embodiedgenv2}, SceneSmith~\cite{scenesmith}). Each baseline retains its own asset pipeline, stopping rules, and iteration limits. For rubric generation, GenSim~\cite{gensim}, RoboGen~\cite{robogen}, and RoboLab~\cite{robolab} author scoring programs for shared frozen scenes (Section~\ref{sec:rubric-eval}).

\paragraph{Metrics.}
Scene quality is measured by request matching (MATCH), environment-side task readiness (READY), geometric validity (GEO), physical stability (PHYS), and visual quality (VIS): MATCH checks requested objects, counts, and attributes; READY assesses workspace clearances, reachability, and operable mechanisms; GEO checks interpenetration and placement bounds; PHYS evaluates settling and support stability under simulation stepping; and VIS rates visual realism and layout coherence. The first four use a $[0,100]$ scale and VIS averages $1$--$5$ ratings; all are higher-is-better, and means weight the 50 requests equally. Delivered scenes undergo unified conversion into a common evaluation representation that preserves original geometry, native interaction definitions, and measured scene properties without repair. Metric criteria and evaluation protocols are detailed in the supplementary material.

\paragraph{Implementation details.}
All six methods use \texttt{gpt-6.1-sol} with \texttt{high} reasoning effort. EmbodiedStudio runs the native Codex CLI with Isaac Sim~6.0.0.1 and resolves assets via tiered priority across simulation-ready libraries (ArtVIP, RoboDojo, RoboCasa365, NVIDIA SimReady), general repositories (Objathor/Objaverse, YCB, ambientCG), online retrieval, and procedural asset programs. Hardware, baseline configurations, and training hyperparameters for the learning studies are listed in the supplementary material.

\begin{table*}[!t]
\centering\small
\caption{\textbf{Scene quality and mechanism ablations across 50 common requests.} Means weight tasks equally ($\uparrow$ higher is better; MATCH/READY/GEO/PHYS in $[0,100]$, VIS in $1$--$5$).}
\label{tab:scene}
\begin{tabular*}{\textwidth}{@{\extracolsep{\fill}}lrrrrr@{}}
\toprule
Method / Variant & MATCH & READY & GEO & PHYS & VIS\\
\midrule
Holodeck & 23.19 & 11.38 & 77.36 & 60.76 & 1.68\\
LayoutVLM & 51.93 & 22.88 & 73.74 & 43.60 & 2.35\\
SAGE & 59.55 & 44.80 & 86.08 & 62.68 & 2.59\\
EmbodiedGen V2 & 26.66 & 4.23 & 80.34 & 72.71 & 2.31\\
SceneSmith & 74.26 & 69.25 & 90.86 & 71.19 & 2.77\\
\midrule
EmbodiedStudio (Full) & 97.63 & \textbf{98.36} & 99.07 & \textbf{79.84} & \textbf{2.95}\\
\quad w/o simulation-feedback revision & 96.90 & 97.39 & \textbf{99.10} & 60.38 & 2.81\\
\quad w/o independent review & \textbf{98.04} & 98.00 & 97.64 & 59.35 & 2.91\\
\bottomrule
\end{tabular*}
\vspace{-4pt}
\end{table*}

\begin{figure*}[!t]
\centering
\includegraphics[width=\textwidth]{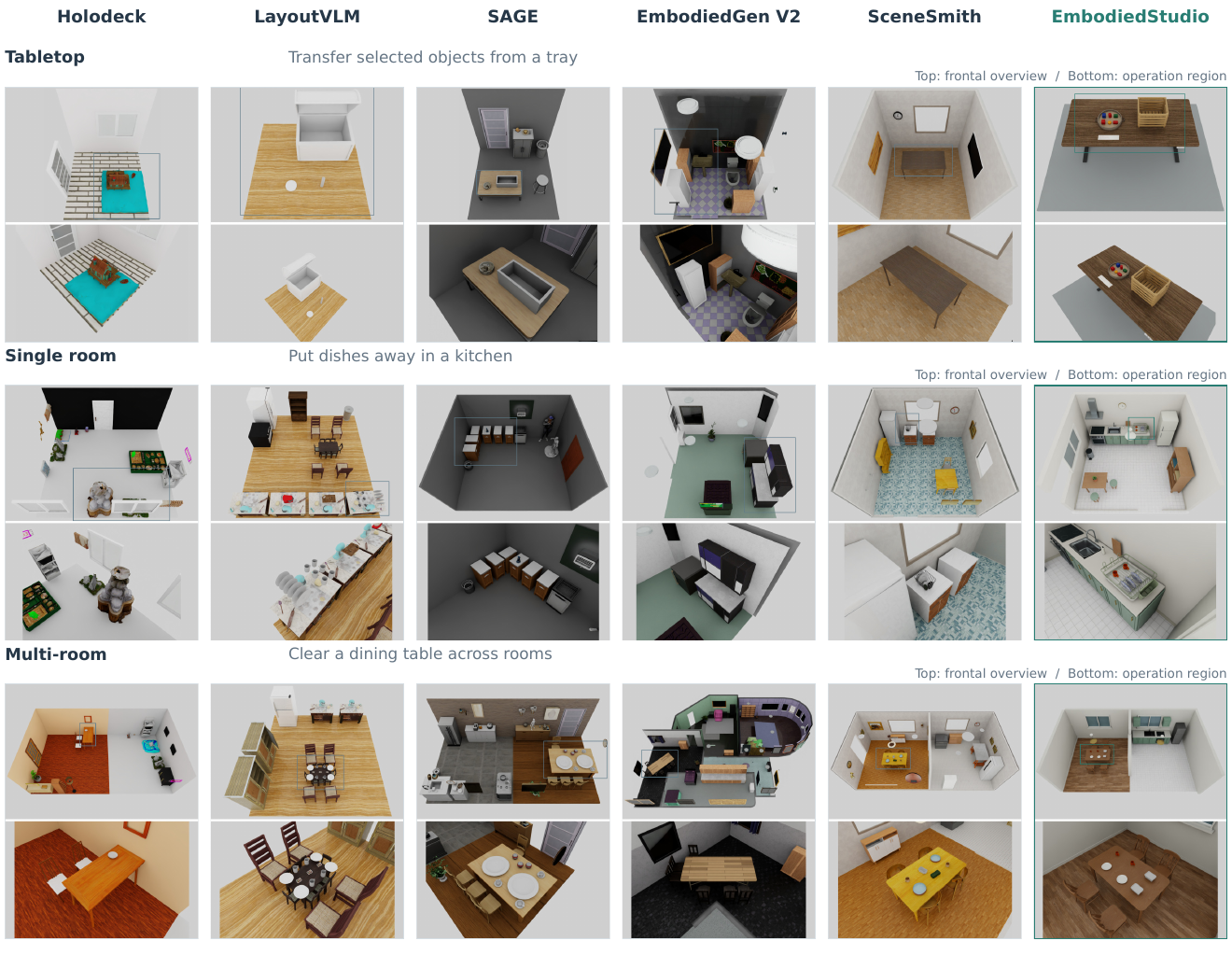}
\caption{\textbf{Qualitative comparison across three scales.} Each row pair shows a whole-scene overview (top) and its outlined operation detail (bottom) for tabletop object selection, single-room kitchen storage, and multi-room table clearing.}
\label{fig:scenes}
\vspace{-4pt}
\end{figure*}

\subsection{Scene fidelity and task readiness}
\label{sec:scene-eval}
Table~\ref{tab:scene} reports aggregate MATCH, READY, GEO, PHYS, and VIS across the 50-request suite. EmbodiedStudio leads all five scene-generation baselines across every metric, reaching 97.63 MATCH, 98.36 READY, and 99.07 GEO, compared with 74.26 MATCH and 69.25 READY for the strongest baseline, SceneSmith (gaps of 23.36 and 29.11 points, respectively). Beyond aggregate averages, the baseline scores reveal a sharp disconnect between local geometry or stability and task readiness: EmbodiedGen V2 attains higher physical stability (72.71 PHYS) than LayoutVLM (43.60) and SAGE (62.68), yet scores only 26.66 MATCH and 4.23 READY; even SceneSmith achieves 90.86 GEO while trailing by nearly 30 points in READY. Thus, locally plausible geometry or static stability does not automatically yield task-ready environments.

Figure~\ref{fig:scenes} illustrates these differences across tabletop, single-room, and multi-room requests. In the selected EmbodiedStudio outputs, target objects and distractors sit cleanly in a tray and receiving box (top), dishes are staged in a rack on the kitchen counter beside reachable storage cabinets (middle), and dishes and food containers span a dining table connected to a kitchen (bottom), preserving clear workspace margins for manipulation.

\begin{figure*}[!t]
\centering
\includegraphics[width=\textwidth]{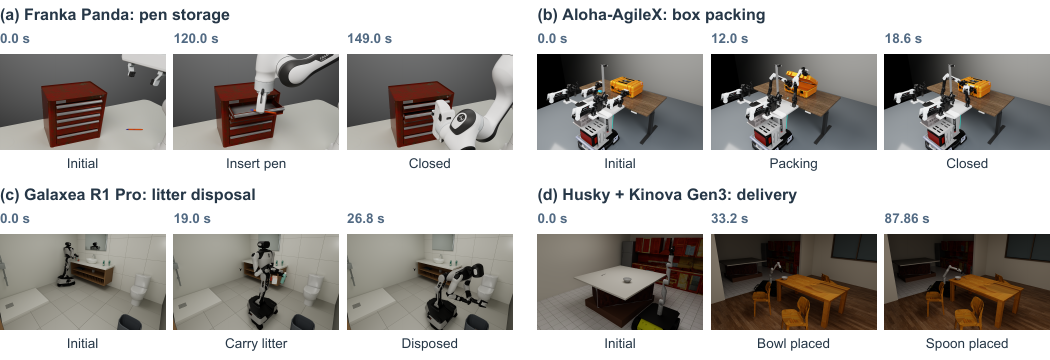}
\vspace{-4pt}
\caption{\textbf{Successful recorded interactions across four robot platforms.} Three chronological states per episode: (a) Franka Panda stores a pen in a drawer; (b) Aloha-AgileX packs and closes a box; (c) Galaxea R1 Pro disposes of litter; (d) Husky A200 with Kinova Gen3 delivers a bowl followed by a spoon. Original episode records report success in all four cases.}
\label{fig:robot-keyframes}
\vspace{-6pt}
\end{figure*}

\subsection{Robot interaction examples}
\label{sec:interactions}
Figure~\ref{fig:robot-keyframes} illustrates successful recorded episodes across four platforms and manipulation demands: Franka Panda opens a drawer, inserts a pen and closes it; Aloha-AgileX packs designated objects and closes a storage box; Galaxea R1 Pro removes two empty packages while preserving the toiletries; and Husky A200 with Kinova Gen3 delivers a bowl before a spoon. Keyframes are rendered from committed states, and success comes from the original frozen episode rubrics. The complete records retain earlier attempts, including those preceding the drawer episode's completion at 149.55\,s. Together, these examples illustrate multi-stage manipulation and cross-room delivery across diverse robot morphologies.

\begin{table}[!tb]
\centering\footnotesize
\caption{\textbf{Shared-scene rubric evaluation on 20 tasks} (\%). BA/Acc./FPR/FNR use valid replay outputs; LBC covers all layout probes.}
\label{tab:rubrics}
\setlength{\tabcolsep}{2pt}
\begin{tabular*}{\columnwidth}{@{\extracolsep{\fill}}lrrrrrrr@{}}\toprule
 & \multicolumn{2}{c}{Source review} & \multicolumn{4}{c}{State replay} & Layout\\
\cmidrule(lr){2-3}\cmidrule(lr){4-7}\cmidrule(l){8-8}
Method & Cov.$\uparrow$ & Expr.$\uparrow$ & BA$\uparrow$ & Acc.$\uparrow$ & FPR$\downarrow$ & FNR$\downarrow$ & LBC$\uparrow$\\\midrule
GenSim &71.97&38.01&73.53&75.49&20.59&\textbf{32.35}&52.50\\
RoboGen &93.67&62.65&50.00&66.67&\textbf{0.00}&100.00&27.50\\
RoboLab &94.25&56.85&73.13&80.00&6.25&47.50&60.00\\
EmbodiedStudio &\textbf{96.25}&\textbf{92.39}&\textbf{82.50}&\textbf{88.24}&\textbf{0.00}&35.00&\textbf{83.75}\\\bottomrule
\end{tabular*}
\vspace{-6pt}
\end{table}

\subsection{Shared-scene rubric study}
\label{sec:rubric-eval}
To isolate rubric quality from scene differences, GenSim, RoboGen, and RoboLab generate scoring programs for the same 20 frozen tabletop environments and original requests. Each baseline's scoring logic is retained, with joint conventions aligned to the shared scene; EmbodiedStudio is evaluated on its delivered programs from the full workflow.

The 20 requests decompose into 65 necessary requirements. Each of the 80 artifacts receives three independent reviews under the same fixed \texttt{gpt-6.1-sol} configuration with access to original requests, scene facts, active scoring sources, bindings, and relevant dependencies. For task $i$ with $n_i$ requirements, coverage $c_{ij}\in\{0,1\}$ records whether a predicate attempts requirement $j$, and correct expression $q_{ij}\in\{0,\frac12,1\}$ records incorrect/missing, partial/uncertain, or correct implementation, giving $C_i=\frac{1}{n_i}\sum_jc_{ij}$ and $Q_i=\frac{1}{n_i}\sum_jq_{ij}$ averaged over three reviews per task and weighted equally across tasks.

Table~\ref{tab:rubrics} shows that EmbodiedStudio achieves 96.25\% coverage (Cov.) and 92.39\% correct expression (Expr.), exceeding the strongest baseline by 29.74 points on expression.

We additionally replay 120 simulator-state sequences or controlled fragments (two successful and four unsuccessful instances per task) through the four delivered scoring programs using independently measured task labels, reporting balanced accuracy (BA), accuracy (Acc.), false-positive rate (FPR), and false-negative rate (FNR) over valid binary outputs (details in supplementary). EmbodiedStudio achieves the highest BA (82.50\%) and Acc. (88.24\%), with 0.00\% FPR across 79 valid negative instances and 35.00\% FNR. On 80 counterfactual layout probes with unchanged scoring sources and updated external task parameters, EmbodiedStudio reaches 83.75\% layout balanced correct-output rate (LBC) versus 27.50\%--60.00\% for the baselines, validating our decoupled task binding and stateful rubric design.

\begin{figure*}[!t]
\centering
{\small
\setlength{\tabcolsep}{2pt}
\begin{tabular}{@{}cccc@{}}
\multicolumn{2}{c}{\textbf{(a) Bowl-stacking area}} &
\multicolumn{2}{c@{}}{\textbf{(b) Study corner}} \\[1pt]
Before & After & Before & After \\[1pt]
\includegraphics[width=.184\textwidth]{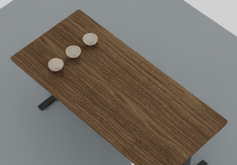} &
\includegraphics[width=.184\textwidth]{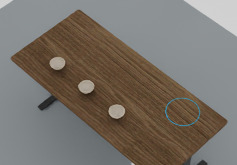} &
\includegraphics[width=.184\textwidth]{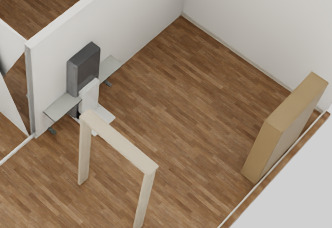} &
\includegraphics[width=.184\textwidth]{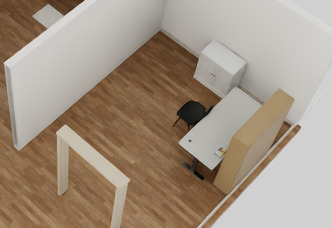} \\[2pt]
\multicolumn{2}{c}{\textbf{(c) Sofa and coffee table}} &
\multicolumn{2}{c@{}}{\textbf{(d) Wardrobe placement}} \\[1pt]
Before & After & Before & After \\[1pt]
\includegraphics[width=.184\textwidth]{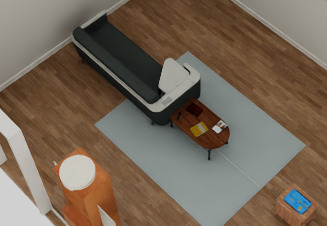} &
\includegraphics[width=.184\textwidth]{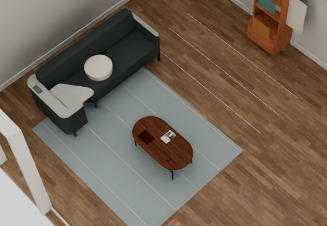} &
\includegraphics[width=.184\textwidth]{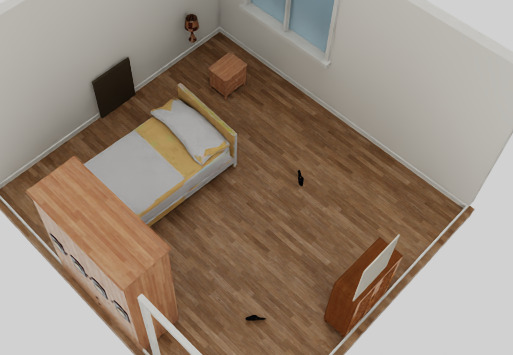} &
\includegraphics[width=.184\textwidth]{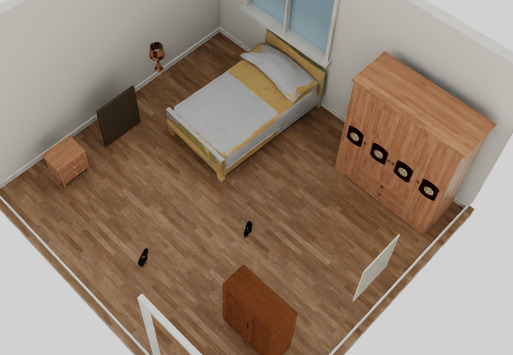}
\end{tabular}}
\vspace{-6pt}
\caption{\textbf{Layout details before and after fine-tuning on construction trajectories.} Each pair shows matched crops under shared lighting using the base checkpoint (left) and the fine-tuned checkpoint (right).}
\label{fig:training-scenes}
\vspace{-10pt}
\end{figure*}

\subsection{Scene-generation mechanisms}
\label{sec:ablations}
The lower block of Table~\ref{tab:scene} isolates the two closed-loop mechanisms from Section~\ref{sec:method} under identical requests, model configuration, asset access, and budgets: \textit{w/o simulation-feedback revision} removes post-build rendered images and physical/execution measurements from authoring and review while keeping static source checks, whereas \textit{w/o independent review} retains simulator feedback for author self-revision while removing the separate reviewer.

Across all three variants, MATCH and READY remain within 1.2 points of one another (96.90--98.04 MATCH and 97.39--98.36 READY), indicating that open-loop coding or author self-iteration already places requested objects in plausible task regions. In contrast, PHYS drops sharply from 79.84 in the full system to 60.38 without simulation feedback (-19.46) and 59.35 without independent review (-20.49). Thus, achieving physical stability requires both closed-loop simulator evidence and an independent reviewer beyond the author's own iteration.

\subsection{Supervising object placement from construction histories}
\label{sec:learning}
To test whether the leak-free forward construction histories $\mathcal H_x$ (Section~\ref{sec:method}) provide usable supervision for scene authoring, we fine-tune a 27B Qwen model on object-placement examples extracted from the construction trajectories exported by EmbodiedStudio. At inference, both the base and fine-tuned checkpoints receive the same request, fixed room structure, and asset catalog, and predict object placements to build the scene. Figure~\ref{fig:training-scenes} compares initial scene renders on four scenes: after fine-tuning, the marked bowl-stacking area is clearly formed (Figure~\ref{fig:training-scenes}a), the desk and chair form a coherent study corner (Figure~\ref{fig:training-scenes}b), the sofa and coffee table are separated cleanly (Figure~\ref{fig:training-scenes}c), and the wardrobe sits flush against the far wall (Figure~\ref{fig:training-scenes}d).

\subsection{Policy adaptation from execution experience}
\label{sec:policy-learning}
We next verify whether the recorded interaction datasets $\mathcal D_x$ provide valid visuomotor supervision for task-level policy adaptation. Starting from the base $\pi_{0.5}$ checkpoint~\cite{pi05}, we fine-tune two task-specific policies with LoRA on 30 successful simulated demonstrations each: moving an apple into an empty bowl while preserving other objects (\textbf{Apple}), and building a stable two-level block structure (\textbf{Blocks}).

\begin{figure}[!tb]
\centering
\includegraphics[width=.56\columnwidth]{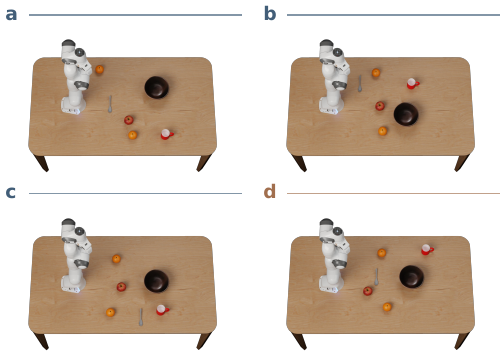}
\vspace{-4pt}
\caption{\textbf{Layout seed variation for the Apple task.} Training layouts (a--c) and held-out test layout (d) compiled from the same scene program with fixed assets and task semantics.}
\label{fig:seed-variation}
\vspace{-8pt}
\end{figure}

\begin{table}[!tb]
\centering\small
\caption{\textbf{$\pi_{0.5}$ adaptation on a held-out layout per task.} Mean rubric scores over five rollouts per task, scaled to $[0,100]$.}
\label{tab:pi05}
\setlength{\tabcolsep}{3pt}
\begin{tabularx}{\columnwidth}{@{}Yrrr@{}}
\toprule
Policy checkpoint & Apple & Blocks & Mean\\
\midrule
Before fine-tuning & 0.00 & 0.00 & 0.00\\
After fine-tuning & 32.00 & 35.09 & 33.55\\
\bottomrule
\end{tabularx}
\end{table}

\begin{figure}[!tb]
\centering
\includegraphics[width=.44\columnwidth]{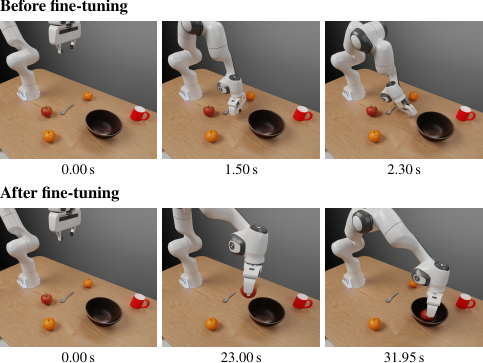}
\vspace{-4pt}
\caption{\textbf{Apple placement before and after fine-tuning.} Path-traced replay of recorded states.}
\label{fig:pi05-apple-comparison}
\vspace{-6pt}
\end{figure}

To separate training and evaluation layouts prior to data collection, the scene program generates explicit frozen revisions for training and held-out testing by varying object poses and distractor layouts while keeping task semantics, assets, robot configuration, and scoring logic fixed (Figure~\ref{fig:seed-variation}). For evaluation, we fix one held-out layout per task and run five rollouts per checkpoint under matched inference-noise seeds, camera views, instructions, and control interfaces. As reported in Table~\ref{tab:pi05}, the base $\pi_{0.5}$ checkpoint scores 0.00 on both tasks, whereas LoRA adaptation raises mean rubric scores to 32.00 on Apple and 35.09 on Blocks (33.55 overall); each fine-tuned policy fully completes one of five rollouts. Figure~\ref{fig:pi05-apple-comparison} compares both checkpoints on Apple: the base policy fails early by 2.30\,s, whereas the fine-tuned policy grasps and places the apple by 31.95\,s.

\section{Conclusion}
We presented EmbodiedStudio, a task-driven simulation platform uniting scene authoring, robot execution, and stateful rubrics under frozen revisions. Joint authoring aligns scenes with task requirements, while simulator feedback and independent review improve stability and capture construction histories and demonstrations. EmbodiedStudio improves readiness, stability, and rubric faithfulness while providing supervision for layout and policy adaptation.
\section*{LLM Usage Statement}
Large language models assisted with system and plotting code development, literature retrieval, and text polishing. Scene generation and evaluation in our experiments used \texttt{gpt-6.1-sol} as specified in Section~4. The authors reviewed the manuscript, experimental results, and references, and take full responsibility for the submission.

\clearpage
\bibliographystyle{ieeenat_fullname}
\bibliography{references}

\begin{thebibliography}{41}
\providecommand{\natexlab}[1]{#1}
\providecommand{\url}[1]{\texttt{#1}}
\expandafter\ifx\csname urlstyle\endcsname\relax
  \providecommand{\doi}[1]{doi: #1}\else
  \providecommand{\doi}{doi: \begingroup \urlstyle{rm}\Url}\fi

\bibitem[Black et~al.(2025)Black, Brown, Darpinian, Dhabalia, Driess, Esmail,
  Equi, Finn, Fusai, Galliker, Ghosh, Groom, Hausman, Ichter, Jakubczak, Jones,
  Ke, LeBlanc, Levine, Li-Bell, Mothukuri, Nair, Pertsch, Ren, Shi, Smith,
  Springenberg, Stachowicz, Tanner, Vuong, Walke, Walling, Wang, Yu, and
  Zhilinsky]{pi05}
Kevin Black, Noah Brown, James Darpinian, Karan Dhabalia, Danny Driess, Adnan
  Esmail, Michael~Robert Equi, Chelsea Finn, Niccolo Fusai, Manuel~Y. Galliker,
  Dibya Ghosh, Lachy Groom, Karol Hausman, Brian Ichter, Szymon Jakubczak, Tim
  Jones, Liyiming Ke, Devin LeBlanc, Sergey Levine, Adrian Li-Bell, Mohith
  Mothukuri, Suraj Nair, Karl Pertsch, Allen~Z. Ren, Lucy~Xiaoyang Shi, Laura
  Smith, Jost~Tobias Springenberg, Kyle Stachowicz, James Tanner, Quan Vuong,
  Homer Walke, Anna Walling, Haohuan Wang, Lili Yu, and Ury Zhilinsky.
\newblock {$\pi_{0.5}$: a Vision-Language-Action Model with Open-World
  Generalization}.
\newblock In \emph{Proceedings of The 9th Conference on Robot Learning}, pages
  17--40. PMLR, 2025.

\bibitem[Chen et~al.(2026{\natexlab{a}})Chen, Chen, Chen, Cai, Liu, Li, Liang,
  Lin, Ge, Gu, Deng, Guo, Nian, Xie, Chen, Su, Xu, Liu, Hu, Gao, Wang, Liang,
  Qin, Yang, Luo, and Mu]{robotwin2}
Tianxing Chen, Zanxin Chen, Baijun Chen, Zijian Cai, Yibin Liu, Zixuan Li,
  Qiwei Liang, Xianliang Lin, Yiheng Ge, Zhenyu Gu, Weiliang Deng, Yubin Guo,
  Tian Nian, Xuanbing Xie, Qiangyu Chen, Kailun Su, Tianling Xu, Guodong Liu,
  Mengkang Hu, Huan-ang Gao, Kaixuan Wang, Zhixuan Liang, Yusen Qin, Xiaokang
  Yang, Ping Luo, and Yao Mu.
\newblock {RoboTwin 2.0: A Scalable Data Generator and Benchmark with Strong
  Domain Randomization for Robust Bimanual Robotic Manipulation}.
\newblock In \emph{Proceedings of the 43rd International Conference on Machine
  Learning}, pages 13673--13699. PMLR, 2026{\natexlab{a}}.

\bibitem[Chen et~al.(2026{\natexlab{b}})Chen, Feng, Liu, Zhang, Wen, Liao, Qiu,
  Li, and Lin]{physcene3d}
Weixing Chen, Zhuoqian Feng, Yang Liu, Yexin Zhang, Yifan Wen, Yinghong Liao,
  Weichao Qiu, Guanbin Li, and Liang Lin.
\newblock {PhyScene3D: Physically Consistent Interactive 3D Tabletop Scene
  Generation}.
\newblock In \emph{Proceedings of the 43rd International Conference on Machine
  Learning}, pages 14170--14192. PMLR, 2026{\natexlab{b}}.

\bibitem[Deitke et~al.(2022)Deitke, VanderBilt, Herrasti, Weihs, Ehsani,
  Salvador, Han, Kolve, Kembhavi, and Mottaghi]{procthor}
Matt Deitke, Eli VanderBilt, Alvaro Herrasti, Luca Weihs, Kiana Ehsani, Jordi
  Salvador, Winson Han, Eric Kolve, Aniruddha Kembhavi, and Roozbeh Mottaghi.
\newblock {ProcTHOR: Large-Scale Embodied AI Using Procedural Generation}.
\newblock In \emph{Advances in Neural Information Processing Systems}, pages
  5982--5994, 2022.

\bibitem[Feng et~al.(2023)Feng, Zhu, Fu, Jampani, Akula, He, Basu, Wang, and
  Wang]{layoutgpt}
Weixi Feng, Wanrong Zhu, Tsu-jui Fu, Varun Jampani, Arjun Akula, Xuehai He,
  Sugato Basu, Xin~Eric Wang, and William~Yang Wang.
\newblock {LayoutGPT: Compositional Visual Planning and Generation with Large
  Language Models}.
\newblock In \emph{Advances in Neural Information Processing Systems}, pages
  18225--18250, 2023.

\bibitem[Ha et~al.(2023)Ha, Florence, and Song]{scalingup}
Huy Ha, Pete Florence, and Shuran Song.
\newblock {Scaling Up and Distilling Down: Language-Guided Robot Skill
  Acquisition}.
\newblock In \emph{Proceedings of The 7th Conference on Robot Learning}, pages
  3766--3777. PMLR, 2023.

\bibitem[Hua et~al.(2025)Hua, Liu, Macaluso, Lin, Zhang, Xu, and Wang]{gensim2}
Pu Hua, Minghuan Liu, Annabella Macaluso, Yunfeng Lin, Weinan Zhang, Huazhe Xu,
  and Lirui Wang.
\newblock {GenSim2: Scaling Robot Data Generation with Multi-modal and
  Reasoning LLMs}.
\newblock In \emph{Proceedings of The 8th Conference on Robot Learning}, pages
  5030--5066. PMLR, 2025.

\bibitem[Huang et~al.(2023)Huang, Wang, Zhang, Li, Wu, and Fei-Fei]{voxposer}
Wenlong Huang, Chen Wang, Ruohan Zhang, Yunzhu Li, Jiajun Wu, and Li Fei-Fei.
\newblock {VoxPoser: Composable 3D Value Maps for Robotic Manipulation with
  Language Models}.
\newblock In \emph{Proceedings of The 7th Conference on Robot Learning}, pages
  540--562. PMLR, 2023.

\bibitem[Huang et~al.(2025)Huang, Wang, Li, Zhang, and Fei-Fei]{rekep}
Wenlong Huang, Chen Wang, Yunzhu Li, Ruohan Zhang, and Li Fei-Fei.
\newblock {ReKep: Spatio-Temporal Reasoning of Relational Keypoint Constraints
  for Robotic Manipulation}.
\newblock In \emph{Proceedings of The 8th Conference on Robot Learning}, pages
  4573--4602. PMLR, 2025.

\bibitem[James et~al.(2020)James, Ma, Arrojo, and Davison]{rlbench}
Stephen James, Zicong Ma, David~Rovick Arrojo, and Andrew~J. Davison.
\newblock {RLBench: The Robot Learning Benchmark \& Learning Environment}.
\newblock \emph{IEEE Robotics and Automation Letters}, 5\penalty0 (2):\penalty0
  3019--3026, 2020.

\bibitem[Jiang et~al.(2026)Jiang, Zhang, Weng, Song, and Duh]{hoglayout}
Haiyan Jiang, Deyu Zhang, Dongdong Weng, Weitao Song, and Henry Been-Lirn Duh.
\newblock {HOG-Layout: Hierarchical 3D Scene Generation, Optimization and
  Editing via Vision-Language Models}.
\newblock In \emph{Proceedings of the IEEE/CVF Conference on Computer Vision
  and Pattern Recognition (CVPR)}, pages 31586--31596, 2026.

\bibitem[Li et~al.(2024)Li, Zhang, Wong, Gokmen, Srivastava, Martín-Martín,
  Wang, Levine, Ai, Martinez, Yin, Lingelbach, Hwang, Hiranaka, Garlanka,
  Aydin, Lee, Sun, Anvari, Sharma, Bansal, Hunter, Kim, Lou, Matthews,
  Villa-Renteria, Tang, Tang, Xia, Li, Savarese, Gweon, Liu, Wu, and
  Fei-Fei]{behavior1k}
Chengshu Li, Ruohan Zhang, Josiah Wong, Cem Gokmen, Sanjana Srivastava, Roberto
  Martín-Martín, Chen Wang, Gabrael Levine, Wensi Ai, Benjamin Martinez, Hang
  Yin, Michael Lingelbach, Minjune Hwang, Ayano Hiranaka, Sujay Garlanka, Arman
  Aydin, Sharon Lee, Jiankai Sun, Mona Anvari, Manasi Sharma, Dhruva Bansal,
  Samuel Hunter, Kyu-Young Kim, Alan Lou, Caleb~R Matthews, Ivan
  Villa-Renteria, Jerry~Huayang Tang, Claire Tang, Fei Xia, Yunzhu Li, Silvio
  Savarese, Hyowon Gweon, C.~Karen Liu, Jiajun Wu, and Li Fei-Fei.
\newblock {BEHAVIOR-1K: A Human-Centered, Embodied AI Benchmark with 1,000
  Everyday Activities and Realistic Simulation}.
\newblock \href{https://arxiv.org/abs/2403.09227}{arXiv preprint
  arXiv:2403.09227}, 2024.

\bibitem[Liang et~al.(2023)Liang, Huang, Xia, Xu, Hausman, Ichter, Florence,
  and Zeng]{codeaspolicies}
Jacky Liang, Wenlong Huang, Fei Xia, Peng Xu, Karol Hausman, Brian Ichter, Pete
  Florence, and Andy Zeng.
\newblock {Code as Policies: Language Model Programs for Embodied Control}.
\newblock In \emph{Proceedings of the IEEE International Conference on Robotics
  and Automation (ICRA)}, pages 9493--9500, 2023.

\bibitem[Lin and Mu(2024)]{instructscene}
Chenguo Lin and Yadong Mu.
\newblock {InstructScene: Instruction-Driven 3D Indoor Scene Synthesis with
  Semantic Graph Prior}.
\newblock In \emph{International Conference on Learning Representations}, pages
  25687--25718, 2024.

\bibitem[Liu et~al.(2023)Liu, Zhu, Gao, Feng, Liu, Zhu, and Stone]{libero}
Bo Liu, Yifeng Zhu, Chongkai Gao, Yihao Feng, Qiang Liu, Yuke Zhu, and Peter
  Stone.
\newblock {LIBERO: Benchmarking Knowledge Transfer for Lifelong Robot
  Learning}.
\newblock In \emph{Advances in Neural Information Processing Systems}, pages
  44776--44791, 2023.

\bibitem[Ma et~al.(2024)Ma, Liang, Wang, Huang, Bastani, Jayaraman, Zhu, Fan,
  and Anandkumar]{eureka}
Yecheng~Jason Ma, William Liang, Guanzhi Wang, De-An Huang, Osbert Bastani,
  Dinesh Jayaraman, Yuke Zhu, Jim Fan, and Anima Anandkumar.
\newblock {Eureka: Human-Level Reward Design via Coding Large Language Models}.
\newblock In \emph{International Conference on Learning Representations}, pages
  26516--26560, 2024.

\bibitem[Mandlekar et~al.(2023)Mandlekar, Nasiriany, Wen, Akinola, Narang, Fan,
  Zhu, and Fox]{mimicgen}
Ajay Mandlekar, Soroush Nasiriany, Bowen Wen, Iretiayo Akinola, Yashraj Narang,
  Linxi Fan, Yuke Zhu, and Dieter Fox.
\newblock {MimicGen: A Data Generation System for Scalable Robot Learning using
  Human Demonstrations}.
\newblock In \emph{Proceedings of The 7th Conference on Robot Learning}, pages
  1820--1864. PMLR, 2023.

\bibitem[Mao et~al.(2026)Mao, Huang, Lin, Zhao, Wang, Zhang, Wang, Zhou, Zhang,
  Zhong, and Yang]{voxscene}
Haotian Mao, Yuhan Huang, Jiatao Lin, Yang Zhao, Hui Wang, Yiheng Zhang, Yuwang
  Wang, Chenliang Zhou, Yan Zhang, Fangcheng Zhong, and Xubo Yang.
\newblock {VoxScene: Anchor-Conditioned Voxel Diffusion for Indoor Scene
  Arrangement}.
\newblock \href{https://arxiv.org/abs/2605.17102}{arXiv preprint
  arXiv:2605.17102}, 2026.

\bibitem[Nasiriany et~al.(2024)Nasiriany, Maddukuri, Zhang, Parikh, Lo, Joshi,
  Mandlekar, and Zhu]{robocasa}
Soroush Nasiriany, Abhiram Maddukuri, Lance Zhang, Adeet Parikh, Aaron Lo,
  Abhishek Joshi, Ajay Mandlekar, and Yuke Zhu.
\newblock {RoboCasa: Large-Scale Simulation of Household Tasks for Generalist
  Robots}.
\newblock In \emph{Proceedings of Robotics: Science and Systems}, 2024.

\bibitem[Paschalidou et~al.(2021)Paschalidou, Kar, Shugrina, Kreis, Geiger, and
  Fidler]{atiss}
Despoina Paschalidou, Amlan Kar, Maria Shugrina, Karsten Kreis, Andreas Geiger,
  and Sanja Fidler.
\newblock {ATISS: Autoregressive Transformers for Indoor Scene Synthesis}.
\newblock In \emph{Advances in Neural Information Processing Systems}, pages
  12013--12026, 2021.

\bibitem[Pfaff et~al.(2026)Pfaff, Cohn, Zakharov, Cory, and
  Tedrake]{scenesmith}
Nicholas Pfaff, Thomas Cohn, Sergey Zakharov, Rick Cory, and Russ Tedrake.
\newblock {SceneSmith: Agentic Generation of Simulation-Ready Indoor Scenes}.
\newblock In \emph{Proceedings of the 43rd International Conference on Machine
  Learning}, pages 98499--98567. PMLR, 2026.

\bibitem[Qin et~al.(2026)Qin, Deng, Liang, Song, Lin, Jiang, Fu, Sun, Lei,
  Gong, Chen, Zhao, Bai, Ding, Wang, and Li]{embodiedsmith}
Yikai Qin, Yifei Deng, Mingjian Liang, Wenxuan Song, Zepeng Lin, Zhiyi Jiang,
  Jiajun Fu, Qiao Sun, Huashuo Lei, Xicheng Gong, Jiayi Chen, Han Zhao,
  Shuanghao Bai, Pengxiang Ding, Pengwei Wang, and Haoang Li.
\newblock {EmbodiedSmith: Scaling Embodied Data through Recursive
  Self-Improvement Flywheel in Simulation}.
\newblock \href{https://arxiv.org/abs/2610.07969}{arXiv preprint
  arXiv:2610.07969}, 2026.

\bibitem[Raistrick et~al.(2024)Raistrick, Mei, Kayan, Yan, Zuo, Han, Wen,
  Parakh, Alexandropoulos, Lipson, Ma, and Deng]{infinigenindoors}
Alexander Raistrick, Lingjie Mei, Karhan Kayan, David Yan, Yiming Zuo, Beining
  Han, Hongyu Wen, Meenal Parakh, Stamatis Alexandropoulos, Lahav Lipson, Zeyu
  Ma, and Jia Deng.
\newblock {Infinigen Indoors: Photorealistic Indoor Scenes using Procedural
  Generation}.
\newblock In \emph{Proceedings of the IEEE/CVF Conference on Computer Vision
  and Pattern Recognition (CVPR)}, pages 21783--21794, 2024.

\bibitem[Ranawaka et~al.(2026)Ranawaka, Wong, Pai, Chu, Dai, Moghani, Yin,
  Jiang, Durbano, Huynh, Fang, Xu, Zhang, Fei-Fei, Fan, Wen, Mandlekar, and
  Zhu]{simfoundry}
Nadun Ranawaka, Josiah Wong, Wei-Lin Pai, Wei-Teng Chu, Tianyuan Dai, Masoud
  Moghani, Hang Yin, Yunfan Jiang, Wesley Durbano, Brandon Huynh, Yu Fang,
  Danfei Xu, Ruohan Zhang, Li Fei-Fei, Linxi Fan, Bowen Wen, Ajay Mandlekar,
  and Yuke Zhu.
\newblock {SimFoundry: Modular and Automated Scene Generation for Policy
  Learning and Evaluation}.
\newblock \href{https://arxiv.org/abs/2606.28276}{arXiv preprint
  arXiv:2606.28276}, 2026.

\bibitem[Shen et~al.(2026)Shen, You, Liu, Zheng, Zha, Yamazaki, Zhang, Huang,
  Sun, Chen, He, Liu, Chang, Fragkiadaki, Shah, Schwager, Henderson, Abraham,
  and Xu]{embodiedswe}
Zeyu Shen, Haoxiang You, Yilang Liu, Zhicheng Zheng, Lihan Zha, Kashu Yamazaki,
  Mingtong Zhang, Suning Huang, Jiankai Sun, Qianzhong Chen, Lucy He, Kaiyuan
  Liu, Haoran Chang, Katerina Fragkiadaki, Dhruv Shah, Mac Schwager, Peter
  Henderson, Ian Abraham, and Canwen Xu.
\newblock {EmbodiedSWE: Coding Agents for Long Horizon Dexterous Robotics}.
\newblock \href{https://arxiv.org/abs/2609.27308}{arXiv preprint
  arXiv:2609.27308}, 2026.

\bibitem[Sun et~al.(2024)Sun, Harini, Yi, Zhou, Zook, Tremblay, Cross, Wu, and
  Haber]{factorsim}
Fan-Yun Sun, S.~I. Harini, Angela Yi, Yihan Zhou, Alex Zook, Jonathan Tremblay,
  Logan Cross, Jiajun Wu, and Nick Haber.
\newblock {FactorSim: Generative Simulation via Factorized Representation}.
\newblock In \emph{Advances in Neural Information Processing Systems}, pages
  87438--87472, 2024.

\bibitem[Sun et~al.(2025)Sun, Liu, Gu, Lim, Bhat, Tombari, Li, Haber, and
  Wu]{layoutvlm}
Fan-Yun Sun, Weiyu Liu, Siyi Gu, Dylan Lim, Goutam Bhat, Federico Tombari,
  Manling Li, Nick Haber, and Jiajun Wu.
\newblock {LayoutVLM: Differentiable Optimization of 3D Layout via
  Vision-Language Models}.
\newblock In \emph{Proceedings of the IEEE/CVF Conference on Computer Vision
  and Pattern Recognition (CVPR)}, pages 29469--29478, 2025.

\bibitem[Tam et~al.(2026)Tam, Pun, Wang, Chang, and Savva]{sceneeval}
Hou In~Ivan Tam, Hou In~Derek Pun, Austin~T. Wang, Angel~X. Chang, and Manolis
  Savva.
\newblock {SceneEval: Evaluating Semantic Coherence in Text-Conditioned 3D
  Indoor Scene Synthesis}.
\newblock In \emph{Proceedings of the IEEE/CVF Winter Conference on
  Applications of Computer Vision (WACV)}, pages 7355--7365, 2026.

\bibitem[Tang et~al.(2024)Tang, Nie, Markhasin, Dai, Thies, and
  Nießner]{diffuscene}
Jiapeng Tang, Yinyu Nie, Lev Markhasin, Angela Dai, Justus Thies, and Matthias
  Nießner.
\newblock {DiffuScene: Denoising Diffusion Models for Generative Indoor Scene
  Synthesis}.
\newblock In \emph{Proceedings of the IEEE/CVF Conference on Computer Vision
  and Pattern Recognition (CVPR)}, pages 20507--20518, 2024.

\bibitem[Wang et~al.(2024{\natexlab{a}})Wang, Ling, Yuan, Shridhar, Bao, Qin,
  Wang, Xu, and Wang]{gensim}
Lirui Wang, Yiyang Ling, Zhecheng Yuan, Mohit Shridhar, Chen Bao, Yuzhe Qin,
  Bailin Wang, Huazhe Xu, and Xiaolong Wang.
\newblock {GenSim: Generating Robotic Simulation Tasks via Large Language
  Models}.
\newblock In \emph{International Conference on Learning Representations}, pages
  4890--4924, 2024{\natexlab{a}}.

\bibitem[Wang et~al.(2026)Wang, Liu, Ding, Choi, Huang, Zhao, Li, Jiang, Yu,
  Liu, Xu, and Su]{embodiedgenv2}
Xinjie Wang, Liu Liu, Taojun Ding, Andrew Choi, Chaodong Huang, Mengao Zhao,
  Ziang Li, Jackson Jiang, Chunlei Yu, Shengxiang Liu, Wei Xu, and Zhizhong Su.
\newblock {EmbodiedGen V2: An Agentic, Simulation-Ready 3D World Engine for
  Embodied AI}.
\newblock \href{https://arxiv.org/abs/2607.07459}{arXiv preprint
  arXiv:2607.07459}, 2026.

\bibitem[Wang et~al.(2024{\natexlab{b}})Wang, Xian, Chen, Wang, Wang,
  Fragkiadaki, Erickson, Held, and Gan]{robogen}
Yufei Wang, Zhou Xian, Feng Chen, Tsun-Hsuan Wang, Yian Wang, Katerina
  Fragkiadaki, Zackory Erickson, David Held, and Chuang Gan.
\newblock {RoboGen: Towards Unleashing Infinite Data for Automated Robot
  Learning via Generative Simulation}.
\newblock In \emph{Proceedings of the 41st International Conference on Machine
  Learning}, pages 51936--51983. PMLR, 2024{\natexlab{b}}.

\bibitem[Xia et~al.(2026)Xia, Li, Li, Ma, Xu, Liu, Cui, Lin, Ma, Wang, Song,
  and Wei]{sage}
Hongchi Xia, Xuan Li, Zhaoshuo Li, Qianli Ma, Jiashu Xu, Ming-Yu Liu, Yin Cui,
  Tsung-Yi Lin, Wei-Chiu Ma, Shenlong Wang, Shuran Song, and Fangyin Wei.
\newblock {SAGE: Scalable Agentic 3D Scene Generation for Embodied AI}.
\newblock In \emph{Proceedings of the IEEE/CVF Conference on Computer Vision
  and Pattern Recognition (CVPR)}, pages 22358--22368, 2026.

\bibitem[Xue et~al.(2025)Xue, Deng, Chen, Wang, Yuan, and Xu]{demogen}
Zhengrong Xue, Shuying Deng, Zhenyang Chen, Yixuan Wang, Zhecheng Yuan, and
  Huazhe Xu.
\newblock {DemoGen: Synthetic Demonstration Generation for Data-Efficient
  Visuomotor Policy Learning}.
\newblock In \emph{Proceedings of Robotics: Science and Systems}, 2025.

\bibitem[Yang et~al.(2026)Yang, Dagli, Zook, Hadfield, Goyal, Birchfield,
  Ramos, and Tremblay]{robolab}
Xuning Yang, Rishit Dagli, Alex Zook, Hugo Hadfield, Ankit Goyal, Stan
  Birchfield, Fabio Ramos, and Jonathan Tremblay.
\newblock {RoboLab: A High-Fidelity Simulation Benchmark for Analysis of Task
  Generalist Policies}.
\newblock In \emph{Proceedings of Robotics: Science and Systems}, 2026.

\bibitem[Yang et~al.(2024{\natexlab{a}})Yang, Jia, Zhi, and Huang]{physcene}
Yandan Yang, Baoxiong Jia, Peiyuan Zhi, and Siyuan Huang.
\newblock {PhyScene: Physically Interactable 3D Scene Synthesis for Embodied
  AI}.
\newblock In \emph{Proceedings of the IEEE/CVF Conference on Computer Vision
  and Pattern Recognition (CVPR)}, pages 16262--16272, 2024{\natexlab{a}}.

\bibitem[Yang et~al.(2024{\natexlab{b}})Yang, Sun, Weihs, VanderBilt, Herrasti,
  Han, Wu, Haber, Krishna, Liu, Callison-Burch, Yatskar, Kembhavi, and
  Clark]{holodeck}
Yue Yang, Fan-Yun Sun, Luca Weihs, Eli VanderBilt, Alvaro Herrasti, Winson Han,
  Jiajun Wu, Nick Haber, Ranjay Krishna, Lingjie Liu, Chris Callison-Burch,
  Mark Yatskar, Aniruddha Kembhavi, and Christopher Clark.
\newblock {Holodeck: Language Guided Generation of 3D Embodied AI
  Environments}.
\newblock In \emph{Proceedings of the IEEE/CVF Conference on Computer Vision
  and Pattern Recognition (CVPR)}, pages 16227--16237, 2024{\natexlab{b}}.

\bibitem[Yang et~al.(2025)Yang, Jia, Zhang, and Huang]{sceneweaver}
Yandan Yang, Baoxiong Jia, Shujie Zhang, and Siyuan Huang.
\newblock {SceneWeaver: All-in-One 3D Scene Synthesis with an Extensible and
  Self-Reflective Agent}.
\newblock In \emph{Advances in Neural Information Processing Systems
  (NeurIPS)}, 2025.

\bibitem[Zhang et~al.(2026)Zhang, Sun, Zhao, Liu, Gao, and Bian]{zonemaestro}
Meisheng Zhang, Shizhao Sun, Yang Zhao, Ziyuan Liu, Zhijun Gao, and Jiang Bian.
\newblock {Orchestrating Spatial Semantics via a Zone-Graph Paradigm for
  Intricate Indoor Scene Generation}.
\newblock \href{https://arxiv.org/abs/2605.02537}{arXiv preprint
  arXiv:2605.02537}, 2026.

\bibitem[Zhao et~al.(2026)Zhao, Sun, Zhang, Shi, Yang, and Bian]{scenerevis}
Yang Zhao, Shizhao Sun, Meisheng Zhang, Yingdong Shi, Xubo Yang, and Jiang
  Bian.
\newblock {SceneReVis: A Self-Reflective Vision-Grounded Framework for 3D
  Indoor Scene Synthesis via Multi-Turn RL}.
\newblock In \emph{Proceedings of the 34th ACM International Conference on
  Multimedia}, pages 15445--15453, 2026.

\bibitem[Çelen et~al.(2025)Çelen, Han, Schindler, Van~Gool, Armeni, Obukhov,
  and Wang]{idesign}
Ata Çelen, Guo Han, Konrad Schindler, Luc Van~Gool, Iro Armeni, Anton Obukhov,
  and Xi Wang.
\newblock {I-Design: Personalized LLM Interior Designer}.
\newblock In \emph{Computer Vision -- ECCV 2024 Workshops}, pages 217--234.
  Springer, 2025.

\end{thebibliography}
\clearpage
\appendix
\setcounter{figure}{0}
\setcounter{table}{0}
\renewcommand{\thefigure}{S\arabic{figure}}
\renewcommand{\thetable}{S\arabic{table}}
\renewcommand{\theHfigure}{appendix.\arabic{figure}}
\renewcommand{\theHtable}{appendix.\arabic{table}}
\twocolumn[{\noindent\sffamily\bfseries\LARGE\color{reportInk} Supplementary Material\par\vspace{0.9em}}]
\small
\setlength{\parskip}{0.33em plus 1pt minus 1pt}
\setlength{\emergencystretch}{3em}
\titleformat{\section}{\large\sffamily\bfseries\raggedright\color{reportInk}}{\thesection}{0.65em}{}
\titleformat{\subsection}{\normalsize\sffamily\bfseries\raggedright\color{reportInk}}{\thesubsection}{0.65em}{}
\makeatletter
\setlength{\@dblfpsep}{9pt plus 2pt minus 2pt}
\makeatother

\section{System and Tooling Details}
\label{sec:supp-system}

\subsection{Unified Scene Tree, Single World Authority, and Dynamic Robot Contracts}
EmbodiedStudio represents tabletop workspaces, single rooms, and multi-room environments in a single hierarchical scene tree rather than separate scale-specific planners. The runtime enforces cycle-free parent depth bounds and canonical orientation invariants while storing spatial ownership (\texttt{owner}), coordinate frames (\texttt{frame}), physical support/containment/joint relations, and task bindings (\texttt{bindings}) orthogonally: for example, a pen placed inside an articulated drawer belongs to the room's workspace hierarchy and is physically supported by the drawer bottom without becoming its rigid kinematic child, and child indices are derived programmatically rather than handwritten.

\paragraph{Single visual--physics world authority.}
Compiling $(S,T,R)$ produces one canonical frozen USD stage bound with semantic entity identifiers. Both Isaac Sim / PhysX physics simulation and web preview export derive directly from this compiled USD stage, eliminating representation drift between visual inspection and physical execution. Functional enclosure nodes (\texttt{role='enclosure'} for walls, floors, and ceilings) enable camera-ray cutaway visualization without disabling physical collision geometry.

\paragraph{Dynamic native I/O freezing and declarative robot setup.}
Rather than hardcoding a fixed robot embodiment or control frequency, environment compilation freezes observation and action conventions directly from native USD actuator, joint, sensor, and physics-clock bindings. Robots attach declaratively via versioned robot profiles and independent setup closures without modifying the base scene revision or the simulation kernel, allowing all 16 integrated robot platforms in the teaser and the four multi-platform rollouts in Figure~\ref{fig:robot-keyframes} of the main paper to share one loading, planning, and recording pipeline.

\subsection{Thin-Kernel Architecture, Multi-Frequency Freezing, and Agent Tools}
EmbodiedStudio enforces a strict boundary between a minimal task-agnostic runtime kernel and authored project sources. Rather than implementing a custom agent loop or task-specific motion/scoring heuristics, the host wraps the native Codex CLI solely for goal persistence, approval bridging, and session recovery, while interactive debugging and production execution share the same domain tools and execution kernel (exposing named clock transitions alongside task-agnostic numerical cuRobo motion planning and rigid-pose checks). Task semantics, scene logic (\texttt{scene.py}, \texttt{tasks.py}), expert behaviors (\texttt{expert.py}), task rubrics (\texttt{rubrics.py}), and format conversion scripts in \texttt{delivery/} remain inside versioned project sources.

\paragraph{Decoupled freeze layers and versioned runtimes.}
Artifacts freeze along their natural rate of change: (1) environment revisions $W_v$ lock geometry, task bindings, rubrics, and native I/O contracts (with multi-layout requests compiling parameterized seeds into explicit \texttt{train}/\texttt{validation}/\texttt{test} revisions); (2) robot setup and behavior closures ($B$) freeze independently against any compatible $W_v$; and (3) downstream \texttt{delivery/} exports freeze separately without triggering scene rebuilds or trajectory re-collection. The consumption runtime is version-pinned to the runtime source tree and interpreter environment, ensuring older environment revisions replay and rescore under their original kernel.

The native coding agent interacts with the workspace via standard file/shell operations and nine coarse-grained Model Context Protocol (MCP) tools:
\begin{enumerate}[leftmargin=*,nosep,topsep=2pt]
  \item \texttt{project\_inspect}: reads the full original request, requirement version, capability flags, and subtree context.
  \item \texttt{assets\_lookup}: queries curated simulation libraries with semantic, articulation, texture, and dimension filters.
  \item \texttt{assets\_prepare}: freezes library assets, self-contained online GLBs, or agent-authored procedural Python assets into immutable asset reports.
  \item \texttt{scene\_build}: compiles the locked $(S,T,R)$ source closure into an immutable revision $W_v$, outputting USD/GLB artifacts and build diagnostics.
  \item \texttt{scene\_observe}: renders multi-view RGB observations (supporting node focus and recorded episode frames).
  \item \texttt{task\_check}: verifies entity bindings, frozen rubric sources, and verification episodes against declared expectations.
  \item \texttt{robot\_run}: executes closed-loop expert or policy rollouts in Isaac Sim / PhysX and logs step-wise records.
  \item \texttt{job}: reconciles, polls, or cancels long-running simulation jobs on the GPU runner.
  \item \texttt{project\_deliver}: checks the independent reviewer's approval and runs downstream consumers to emit an immutable delivery manifest.
\end{enumerate}
Asset sourcing follows a four-tier priority order: (1) preferred simulation libraries (ArtVIP for articulated objects, RoboDojo, RoboCasa365, NVIDIA SimReady), (2) broad local repositories (Objathor/Objaverse, YCB, ambientCG), (3) online retrieval of self-contained assets, and (4) agent-authored procedural asset programs.

\subsection{Role Separation, Sandbox Isolation, and Three-Product Delivery}
Role separation between authoring, physical measurement, and acceptance verification is enforced through isolated processes and read-only mounts:
\begin{itemize}[leftmargin=*,nosep,topsep=2pt]
  \item \textbf{Authoring vs.\ review separation:} Before delivery, the authoring agent invokes an isolated read-only reviewer subagent that inspects the original request, multi-view renders, physical/execution measurements, and source bindings without editing files; delivery requires the independent reviewer's approval rather than author self-reports.
  \item \textbf{Execution and rubric isolation:} Controllers and procedural asset scripts execute in isolated sandboxes mounting only declared read-only inputs, while task rubrics run in a separate read-only process receiving only measured physical state JSON with zero simulator control handles and deterministic offline recomputation.
  \item \textbf{Acceptance by independent consumption:} Independent downstream consumers validate each requested product in fresh processes:
  \begin{itemize}[leftmargin=*,nosep,topsep=1pt]
    \item \emph{Construction experience}: verifies public trace events, trajectory-supervision JSONL records, source diffs, and linked visual/diagnostic consequences.
    \item \emph{Execution experience}: verifies paired LeRobot~v3 and robomimic HDF5 datasets, step-wise observation/action alignment, camera intrinsics/extrinsics, and rubric-verified demonstration eligibility.
    \item \emph{Evaluation environment}: reloads the frozen USD/robot/rubric closure in an isolated process and verifies reset, single-step control, and deterministic rubric recomputation from recorded state logs.
  \end{itemize}
\end{itemize}

\paragraph{Capability comparison criteria (Main Paper Table~\ref{tab:capability_comparison}).}
In Table~\ref{tab:capability_comparison} of the main paper, \emph{Tabletop}, \emph{Single Room}, and \emph{Multi Room} denote supported spatial scales; \emph{Sim-Refined Scene} indicates iterative scene revision verified by physical dynamics and contact stability in simulation; \emph{Task-Driven Scene} indicates constructing the scene around explicit task semantics; \emph{Behavior Program Generation ($B$)} indicates authoring executable robot manipulation or policy programs for the task; \emph{Stateful Rubric ($R$)} denotes evaluation programs that retain and use execution history to verify ordered stages and persistent conditions (distinguishing history-bearing stateful rubrics from instantaneous current-state boolean checks or reward shaping); \emph{Construction Exp.\ ($\mathcal H_x$)} denotes retaining multi-turn scene-authoring or editing trajectories for learning (covering both reverse-engineered editing trajectories in SceneReVis and forward simulator-guided debugging histories in EmbodiedStudio); and \emph{Execution Exp.\ ($\mathcal D_x$)} denotes automated method-authored generation of robot rollout datasets for downstream policy learning (as opposed to downstream human teleoperation or external-policy validation alone).

\section{Scene Evaluation Protocol and 50 Requests}
\label{sec:supp-scene-eval}

\subsection{Common Conversion and Evidence Collection}
All 50 requests (20 tabletop \texttt{T01--T20}, 20 single-room \texttt{R01--R20}, 10 multi-room \texttt{M01--M10}) are listed in Table~\ref{tab:requests-50}. The comparison covers five baselines, the full EmbodiedStudio system, and two ablations, yielding 400 evaluation positions. Request clauses and environment-side necessary conditions are fixed before scoring and shared across methods.

Delivered artifacts undergo unified conversion into a common observation and evaluation representation, preserving original geometry, placements, materials, and native interaction definitions via one reusable native-format route per method. Auxiliary geometric and interaction evidence—asset descriptions and entity bindings, collision and motion definitions, joint limits and control scripts, support/contact measurements, surface intersections, container openings and interior clearances, and placement/access-path samples—is extracted read-only from the delivered artifacts without modifying any scene.

\subsection{Five Scene-Quality Metrics}
\label{sec:supp-five-metrics}
The first four metrics use a $[0,100]$ scale; VIS uses $1$--$5$ ratings for delivered scenes. All are higher-is-better.

\paragraph{Request matching (MATCH).}
MATCH measures satisfaction of the requested pre-task scene: objects and counts, attributes, functional zones, and initial spatial relations. Each frozen clause receives 1 for full satisfaction, 0.5 for observable partial satisfaction or an ambiguous semantic arrangement, and 0 for an unmet or contradicted requirement (including missing objects). The score is the mean clause credit multiplied by 100.

\paragraph{Environment-side task readiness (READY).}
READY measures the fraction of equally weighted binary necessary conditions that pass, covering usable workspace, supporting surfaces, container capacity and openings, access paths, and required native motion or articulation. Free-object motion checks enabled motion, collision participation, and compatible constraints; moving a door or drawer relative to its frame requires a target-bound joint or documented control mechanism. Conditions are evaluated over the initial state and legal state changes supported by the delivered source; a missing mechanism fails only its corresponding condition. READY assesses environment-side prerequisites at finite spatial resolution independently of any bundled robot or policy.

\paragraph{Geometric validity (GEO).}
GEO is the percentage of semantic object/assembly groups with valid placement and support free of obvious unintended interpenetration, assessed over complete uncut geometry, measured surface contacts/intersections, and rendered views. Mesh/material subparts of one object are grouped together while independently movable objects remain separate. Legitimate support contact, wall-mounted or embedded fixtures, assembled parts, and objects inside hollow containers are distinguished from invalid crossings or unsupported placement.

\paragraph{Physical stability (PHYS).}
PHYS measures the percentage of tested free objects and freestanding furniture that remain within common displacement limits in a standardized gravity test. Building structure and fixed installations remain fixed, and non-task decorative rugs and base-backed thin floor finishes are excluded under shared rules. All parts of one physical object form one test body using source-derived collision proxies and mass properties (density $500\,\mathrm{kg/m^3}$, static/dynamic friction 0.5, restitution 0, gravity $9.81\,\mathrm{m/s^2}$, with mass properties held independent of collision approximation). Simulation runs for 5\,s at a 1\,ms time step (5,000 steps). For object $i$ with source bounding-box diagonal $d_i$ in metres, the recorded peak translation from the initial pose must satisfy
\begin{equation}
\Delta p_i \leq \max(0.01,\min(0.02,0.02d_i)),
\end{equation}
and peak rotation must not exceed $0.1$ radians (including normal settling). An empty free-object set scores 0.

\paragraph{Visual quality (VIS).}
VIS is a holistic 1--5 rating of geometry/material realism, style and scale consistency, scene detail, and layout coherence over the delivered area and operation close-ups: 1, severe visible defects; 2, substantial simplification or incoherence; 3, plausible appearance with noticeable limitations; 4, convincing appearance with minor flaws; and 5, consistently high realism and coherence globally and locally. Missing requested objects or rooms are accounted for in MATCH/READY rather than VIS, and missing source textures fall back to a shared neutral material.

\subsection{Common Observations, Scoring, and Aggregation}
\paragraph{Views and illumination.}
Evaluation uses a shared renderer at $1024\times768$ resolution, with a complete-scene overview, four views per task region, and two close-ups per task role. Camera placement and view budgets follow the same rules for each request across methods. Ceilings and camera-side occluding walls are hidden for observation, while furniture/object occlusions are retained. Full geometry remains available to geometric and physical tests. Primary VIS views use common reference illumination and fixed exposure, with source lamps disabled and source textures and material emission retained. Original scene lighting is supplementary and does not enter the primary VIS mean.

\paragraph{Scoring and retained evidence.}
Task/geometric judgments and visual ratings use anonymous scene evidence with a fixed \texttt{gpt-6.1-sol} evaluator configured at \texttt{high} reasoning effort. Each scene receives three task assessments and three visual ratings; MATCH, READY, and GEO are computed on their $[0,100]$ scales within each assessment and then averaged, while VIS averages the three integer ratings. PHYS is computed separately from simulated body motion. The evaluator sees task clauses, common views, and relevant source/measurement evidence without producer identity, and is instructed to assess existing artifacts without repairing them. This benchmark evaluator is separate from the internal generation reviewer.

Section~\ref{sec:supp-match-ready} illustrates the evidence behind selected MATCH and READY conditions for five requests across all six methods, and Section~\ref{sec:supp-cross-judge} validates evaluator consistency using an independent model family (Gemini~4~Neon~(High)) on a stratified 10-request subset.

\paragraph{Task weighting and output failures.}
Each method's reported score is the arithmetic mean over all 50 requests, weighting each request equally (20/20/10 across tabletop, single-room, and multi-room scales). Partially delivered scenes are scored as delivered. Confirmed native no-output failures contribute 0 to all five metrics: across all 400 positions ($8\text{ methods}\times 50\text{ requests}$), 10 no-output cases occur---8 in Holodeck (\texttt{T03}, \texttt{T05}, \texttt{M04--M07}, \texttt{M09}, \texttt{M10}; floor-plan parser failure) and 2 in LayoutVLM (\texttt{T08}, \texttt{T16}; layout solver exit without output). Delivered-only VIS means are $2.00$ for Holodeck ($n=42$) and $2.44$ for LayoutVLM ($n=48$). All other methods delivered 50/50 scenes.

\subsection{Baseline Configurations and Hardware}
All six scene-generation methods use \texttt{gpt-6.1-sol} with \texttt{high} reasoning effort:
\begin{itemize}[leftmargin=*,nosep,topsep=2pt]
  \item \textbf{Holodeck}~\cite{holodeck}: official native pipeline via headless adapter.
  \item \textbf{LayoutVLM}~\cite{layoutvlm}: published B.1 text frontend and native layout solver.
  \item \textbf{SAGE}~\cite{sage}: official native generator/critics with its released-client default limit of 15 tool calls.
  \item \textbf{EmbodiedGen V2}~\cite{embodiedgenv2}: official \texttt{room-cli} pipeline with Blender extension environment.
  \item \textbf{SceneSmith}~\cite{scenesmith}: official asset-retrieval configuration over its Objaverse and ArtVIP libraries.
\end{itemize}
EmbodiedStudio runs on Isaac Sim~6.0.0.1 using a workstation with an NVIDIA GeForce RTX~4090 (24\,GiB VRAM, driver 580.178.04, Vulkan), Intel Xeon Gold~6326 CPU (32 physical / 64 logical cores @ 2.90\,GHz), and 256\,GB RAM.

\subsection{Cross-Model Evaluator Validation}
\label{sec:supp-cross-judge}

To test whether using \texttt{gpt-6.1-sol} as both generator and benchmark evaluator introduces same-model preference bias, we conduct a double-blind cross-family evaluation using \textbf{Gemini~4~Neon~(High)} on a stratified 10-request subset spanning all three scales in the benchmark's exact 4:4:2 ratio: four tabletop tasks (\texttt{T01}, \texttt{T15}, \texttt{T17}, \texttt{T19}), four single-room tasks (\texttt{R06}, \texttt{R10}, \texttt{R11}, \texttt{R14}), and two multi-room tasks (\texttt{M03}, \texttt{M08}). To ensure clean apples-to-apples comparison without no-output zeros or asset-conversion artifacts, these ten tasks were selected from the pool of benchmark requests where all six primary methods (Holodeck, LayoutVLM, SAGE, EmbodiedGen~V2, SceneSmith, and EmbodiedStudio) delivered complete scenes with full five-metric measurements and clean format conversion (no missing textures or conversion errors), spanning diverse manipulation, room-organization, and multi-room task types ($60$ scene cases total).

\paragraph{Blind protocol and isolation.}
Each of the 60 cases is randomly assigned an anonymous identifier (\texttt{blind\_01}--\texttt{blind\_60}) with all producer identifiers and host paths redacted. Matching Section~\ref{sec:supp-scene-eval}, visual quality (\textsc{VIS}) and task/geometric assessment (\textsc{MATCH}, \textsc{READY}, \textsc{GEO}) run in strictly separated, context-free read-only subagent sessions ($120$ independent Gemini~4~Neon~(High) sessions: $60$ \textsc{VIS} and $60$ \textsc{TASK}) using the identical system prompt, scoring rubrics, JSON schemas, rendered views, and aggregation formulas as the primary evaluator. Each case receives one independent pass under Gemini~4~Neon~(High) and is compared against the three-repeat \texttt{gpt-6.1-sol} means.

\begin{table*}[!tb]
\centering\small
\caption{\textbf{Cross-model evaluator agreement} between blind Gemini~4~Neon~(High) and \texttt{gpt-6.1-sol} (3-repeat) across 60 cases (10 tasks $\times$ 6 methods).}
\label{tab:cross-judge-agreement}
\setlength{\tabcolsep}{7pt}
\begin{tabular}{@{}lccccccc@{}}
\toprule
Metric & Cases $n$ & Case $r$ & Case $\rho$ & Case MAE & Bias (Gem$-$GPT) & Method $r$ & Method $\rho$ \\
\midrule
\textsc{MATCH} & 60 & 0.9603 & 0.9608 & 6.27 & $-$2.83 & 0.9863 & 1.0000 \\
\textsc{READY} & 60 & 0.9728 & 0.9642 & 4.44 & $+$2.26 & 0.9997 & 1.0000 \\
\textsc{GEO}   & 60 & 0.9167 & 0.9010 & 3.47 & $-$2.12 & 0.9768 & 1.0000 \\
\textsc{VIS}   & 60 & 0.6535 & 0.6856 & 0.73 & $-$0.61 & 0.8943 & 0.8286 \\
\bottomrule
\end{tabular}
\end{table*}

\begin{table*}[!tb]
\centering\small
\caption{\textbf{Per-method comparison on the 10-request stratified subset (60 cases):} blind Gemini~4~Neon~(High) vs.\ \texttt{gpt-6.1-sol} (3-repeat). Each cell reports \texttt{Gemini / GPT} means across the 10 tasks.}
\label{tab:cross-judge-methods}
\setlength{\tabcolsep}{8pt}
\begin{tabular}{@{}lcccc@{}}
\toprule
Method & \textsc{MATCH} (Gem / GPT) & \textsc{READY} (Gem / GPT) & \textsc{GEO} (Gem / GPT) & \textsc{VIS} (Gem / GPT) \\
\midrule
Holodeck & 32.01 / 38.67 & 27.08 / 24.83 & 94.25 / 91.72 & 1.10 / 2.10 \\
LayoutVLM & 44.70 / 55.35 & 25.41 / 23.24 & 57.95 / 69.49 & 1.50 / 2.37 \\
SAGE & 58.52 / 62.44 & 44.73 / 43.20 & 79.85 / 82.69 & 1.60 / 2.60 \\
EmbodiedGen V2 & 28.09 / 28.14 & 12.60 / \phantom{0}8.53 & 76.87 / 77.12 & 1.80 / 2.27 \\
SceneSmith & 80.06 / 77.38 & 73.55 / 70.86 & 88.56 / 89.14 & 2.30 / 2.83 \\
\textbf{EmbodiedStudio} & \textbf{100.00} / 98.33 & \textbf{100.00} / 99.17 & \textbf{98.52} / 98.52 & \textbf{3.20} / 3.00 \\
\bottomrule
\end{tabular}
\end{table*}

\paragraph{Findings.}
Tables~\ref{tab:cross-judge-agreement} and~\ref{tab:cross-judge-methods} confirm strong cross-family agreement without evidence of same-model inflation by \texttt{gpt-6.1-sol}:
\begin{itemize}[leftmargin=*,nosep,topsep=2pt]
  \item \textbf{High structural agreement and identical rankings on \textsc{MATCH}, \textsc{READY}, and \textsc{GEO}:} Case-level Pearson correlations reach $0.9603$ (\textsc{MATCH}), $0.9728$ (\textsc{READY}), and $0.9167$ (\textsc{GEO}), while method-level correlations reach $0.9863$ (\textsc{MATCH}), $0.9997$ (\textsc{READY}), and $0.9768$ (\textsc{GEO}) with exact method-rank agreement ($\rho=1.0000$) across all three metrics. On \textsc{GEO}, EmbodiedStudio matches \texttt{gpt-6.1-sol} identically ($98.52$ vs.\ $98.52$), and its margins over the strongest baseline (SceneSmith) remain consistent under Gemini~4~Neon~(High): $+19.94$ \textsc{MATCH} ($100.00$ vs.\ $80.06$), $+26.45$ \textsc{READY} ($100.00$ vs.\ $73.55$), and $+9.96$ \textsc{GEO} ($98.52$ vs.\ $88.56$).
  \item \textbf{Stricter visual scoring widens EmbodiedStudio's \textsc{VIS} margin:} Gemini~4~Neon~(High) penalizes floating props, wall-mesh corruption, and missing enclosures more severely than \texttt{gpt-6.1-sol} (mean bias $-0.61$), lowering baseline \textsc{VIS} means by $0.47$--$1.00$ points while rating EmbodiedStudio higher ($3.20$ vs.\ $3.00$). EmbodiedStudio's \textsc{VIS} lead over SceneSmith expands from $+0.17$ under \texttt{gpt-6.1-sol} to $+0.90$ ($3.20$ vs.\ $2.30$), leading across all three scales (tabletop $3.50$ vs.\ $2.75$, single-room $3.25$ vs.\ $2.00$, multi-room $2.50$ vs.\ $2.00$).
\end{itemize}

\begin{table*}[!t]
\centering\scriptsize
\caption{\textbf{Complete 50-request benchmark suite} covering 20 tabletop (\texttt{T01--T20}), 20 single-room (\texttt{R01--R20}), and 10 multi-room (\texttt{M01--M10}) tasks.}
\label{tab:requests-50}
\renewcommand{\arraystretch}{0.97}
\setlength{\tabcolsep}{4pt}
\begin{tabular*}{\textwidth}{@{\extracolsep{\fill}}lll|lll@{}}
\toprule
ID & Scale & Task Summary & ID & Scale & Task Summary \\
\midrule
\texttt{T01} & tabletop & Place the target apple in a bowl & \texttt{R06} & single-room & Place a plate inside a microwave \\
\texttt{T02} & tabletop & Sort apples and oranges & \texttt{R07} & single-room & Put a pot on the stove and cover it \\
\texttt{T03} & tabletop & Set a place for one person & \texttt{R08} & single-room & Put away toys in a living room \\
\texttt{T04} & tabletop & Stack three bowls & \texttt{R09} & single-room & Restore a living-room reading area \\
\texttt{T05} & tabletop & Order blocks by size & \texttt{R10} & single-room & Clear a living-room passage \\
\texttt{T06} & tabletop & Hang a mug on a rack & \texttt{R11} & single-room & Move a potted plant to a window-side cabinet \\
\texttt{T07} & tabletop & Store stationery in a drawer & \texttt{R12} & single-room & Put slippers in a shoe cabinet \\
\texttt{T08} & tabletop & Retrieve a cup from a small cabinet & \texttt{R13} & single-room & Hang empty clothes hangers \\
\texttt{T09} & tabletop & Put the lid on a food container & \texttt{R14} & single-room & Organize bedside items \\
\texttt{T10} & tabletop & Stand a bottle upright & \texttt{R15} & single-room & Sort books onto shelves \\
\texttt{T11} & tabletop & Transfer selected objects from a tray & \texttt{R16} & single-room & Prepare a study workstation \\
\texttt{T12} & tabletop & Pour large wooden balls into a bowl & \texttt{R17} & single-room & Restock bathroom supplies \\
\texttt{T13} & tabletop & Transfer objects using kitchen tongs & \texttt{R18} & single-room & Remove bathroom counter litter \\
\texttt{T14} & tabletop & Sweep tabletop debris into a dustpan & \texttt{R19} & single-room & Store laundry-room supplies \\
\texttt{T15} & tabletop & Build a two-level block structure & \texttt{R20} & single-room & Pack tools in a storage room \\
\texttt{T16} & tabletop & Pack and close a rigid storage box & \texttt{M01} & multi-room & Set a dining table for two \\
\texttt{T17} & tabletop & Put pens in a pen holder & \texttt{M02} & multi-room & Clear the dining table \\
\texttt{T18} & tabletop & Place a phone in its stand & \texttt{M03} & multi-room & Deliver a snack tray to the living room \\
\texttt{T19} & tabletop & Press and release a physical button & \texttt{M04} & multi-room & Put away groceries from the entrance \\
\texttt{T20} & tabletop & Move a loaded tray across a tabletop & \texttt{M05} & multi-room & Collect recyclables from two rooms \\
\texttt{R01} & single-room & Put away groceries in a kitchen & \texttt{M06} & multi-room & Restock the bathroom from storage \\
\texttt{R02} & single-room & Put dishes away in a kitchen & \texttt{M07} & multi-room & Return books from two rooms to the study \\
\texttt{R03} & single-room & Load a dishwasher & \texttt{M08} & multi-room & Fetch tools, sweep, and return them \\
\texttt{R04} & single-room & Gather baking utensils & \texttt{M09} & multi-room & Pack items for leaving home \\
\texttt{R05} & single-room & Store condiment bottles on an open shelf & \texttt{M10} & multi-room & Tidy three rooms before bedtime \\
\bottomrule
\end{tabular*}
\end{table*}

\section{Shared-Scene Rubric Study Details}
\label{sec:supp-rubric}

The 20 tabletop tasks (\texttt{T01--T20}) decompose into 65 necessary requirements. Each of the 80 scoring artifacts ($4\text{ methods}\times 20\text{ tasks}$) receives three independent reviews using \texttt{gpt-6.1-sol} under a fixed model configuration (240 sessions, 780 clause judgments). For RoboGen, reversible joint conventions are mapped to the shared scene without altering native scoring logic, and nine RoboGen artifacts containing syntax errors are evaluated on their readable source semantics rather than being assigned zero. Table~\ref{tab:rubric-ci} reports means and 95\% bootstrap confidence intervals (10,000 task-level resamples) alongside paired differences against EmbodiedStudio. Consistent with the gap between requirement coverage and faithful implementation, RoboGen, RoboLab, and EmbodiedStudio all achieve high coverage ($93.67\%$--$96.25\%$), whereas correct expression separates them sharply ($+29.74$ to $+54.38$ points).

\begin{table*}[!tb]
\centering\scriptsize
\caption{\textbf{Shared-scene rubric semantics with 95\% bootstrap CIs} (10,000 task-level resamples) and paired differences against EmbodiedStudio.}
\label{tab:rubric-ci}
\setlength{\tabcolsep}{4pt}
\begin{tabular}{@{}lcccccc@{}}
\toprule
Method & Cov. (\%) & 95\% CI & Cov. diff. [95\% CI] & Corr. expr. (\%) & 95\% CI & Expr. diff. [95\% CI] \\
\midrule
GenSim & 71.97 & [60.31, 82.95] & +24.28 [14.00, 35.11] & 38.01 & [28.44, 47.53] & +54.38 [44.87, 64.00] \\
RoboGen & 93.67 & [89.25, 97.75] & +2.58 [$-$4.67, 8.50] & 62.65 & [56.51, 69.46] & +29.74 [18.65, 38.60] \\
RoboLab & 94.25 & [87.25, 100.00] & +2.00 [$-$5.50, 10.00] & 56.85 & [49.25, 64.74] & +35.54 [24.35, 45.31] \\
EmbodiedStudio & \textbf{96.25} & [88.75, 100.00] & --- & \textbf{92.39} & [83.64, 98.61] & --- \\
\bottomrule
\end{tabular}
\end{table*}

\paragraph{Actual-state execution validation.}
Complementing source review, we construct two successful instances and four targeted unsuccessful instances for each of the 20 tasks, yielding 120 simulator-state sequences or controlled fragments (40 positive, 80 negative) and 480 method-level scoring attempts. The same physical records are supplied to each method's delivered scoring programs, with separate private scoring states initialized from reset. Ground-truth labels follow the original task instructions, physical contacts, and geometry measurements.

\begin{table}[!tb]
\centering\footnotesize
\caption{\textbf{Execution denominators and raw outcomes.} A/E denotes abstentions/source errors. Errors and abstentions are not replaced with negative task predictions.}
\label{tab:rubric-execution-counts}
\setlength{\tabcolsep}{3pt}
\begin{tabular*}{\columnwidth}{@{\extracolsep{\fill}}lccc@{}}\toprule
Method & Valid / 120 & TP/TN/FP/FN & A/E\\\midrule
GenSim &102&23/54/14/11&0/18\\
RoboGen &66&0/44/0/22&0/54\\
RoboLab &120&21/75/5/19&0/0\\
EmbodiedStudio &119&26/79/0/14&1/0\\\bottomrule
\end{tabular*}
\end{table}

Acc., BA, FPR and FNR in the main-paper table are conditional on valid binary outputs ($\mathrm{Acc.}=(TP+TN)/N_{\mathrm{valid}}$, $\mathrm{FPR}=FP/(FP+TN)$, $\mathrm{FNR}=FN/(FN+TP)$, and $\mathrm{BA}=1-(\mathrm{FPR}+\mathrm{FNR})/2$), whose subsets differ across methods. For source review, readable source semantics are assessed even when a generated program contains syntax errors; runtime validity is measured separately in execution replay. Table~\ref{tab:rubric-execution-counts} gives their explicit denominators. Correct predictions divided by all 120 planned tests are 64.17\%, 36.67\%, 80.00\% and 87.50\% for GenSim, RoboGen, RoboLab and EmbodiedStudio, respectively; an abstention or source error contributes no correct prediction but is not counted as a task false negative or true negative. RoboGen has nine syntax-invalid task artifacts (54 case attempts), and GenSim has generated API/parameter errors on three tasks (18 attempts). RoboGen's 66 valid outputs are all negative: with twice as many negatives as positives, this gives 66.67\% Acc. despite 100\% FNR and only 50.00\% BA.

For EmbodiedStudio, the single abstention occurs on a \texttt{T06} negative instance where the mug handle is threaded onto the hook while still being held (\texttt{releaseVerified=False}, hook support ratio $\approx 0.22$ vs.\ required $0.75$--$1.25$); rather than guessing on an unverified release state, the rubric conservatively returns uncertain (not a runtime error). All 14 false negatives concentrate in 8 tasks and stem from three localized authoring patterns rather than structural defects of the stateful evaluation runtime: (i) overly tight continuous post-release velocity/settling thresholds (7/14 cases across \texttt{T06}, \texttt{T09}, \texttt{T18}, and \texttt{T20}, where spatial and contact conditions pass but strict angular-velocity caps of $0.10$--$0.15\,\mathrm{rad/s}$ over $0.4$--$0.8\,\mathrm{s}$ reject valid executions with slight residual swinging or settling micro-motion); (ii) narrow direct-contact predicates in multi-item tool tasks (4/14 cases in \texttt{T13} tongs transfer and \texttt{T14} dustpan sweeping, where the authored code requires direct tool contact with every individual piece rather than force transmitted through adjacent pieces); and (iii) conservative region/height bounds or an omitted release-free branch (3/14 cases in \texttt{T01} and \texttt{T03}). This conservative authoring bias is directly coupled with EmbodiedStudio's 0.00\% FPR across all 79 valid negatives; because rubrics live in editable versioned sources decoupled from the frozen scene, calibrating settling tolerances or generalizing contact-chain predicates requires only a localized source update without rebuilding the environment.

To test rubric scoring on tasks requiring fine tool--object contact (such as \texttt{T13} tongs transfer and \texttt{T14} dustpan sweeping), the controlled rubric-test fragments step the original scene geometry, masses, friction, and joint limits in physics simulation (480\,Hz integration, 20\,Hz state sampling) through the required pre-contact, transfer, and settled post-release stages without supplying private rubric state.

Each method retains its generated predicates, with read-only adapters binding entity identities, coordinates, and physical measurements without modifying missing conditions. RoboLab uses its subtask-state-machine completion signal, and RoboGen's scoring calls are evaluated on recorded postconditions.

\paragraph{Counterfactual layout robustness.}
For each task, we derive two task-preserving tabletop layouts and reuse one positive and one negative record, yielding 80 counterfactual inputs (40 positive, 40 negative) and 320 scoring attempts. Selected task objects and assemblies are relocated on unchanged support furniture; task-relevant relative geometry, event histories, clocks, and contact identities are preserved, with associated contact points translated consistently. Independent geometry and requirement checks verify label preservation. Scoring source remains unchanged, including embedded coordinate constants. Where the existing representation exposes external task parameters, their layout values are regenerated with the same \texttt{gpt-6.1-sol} (\texttt{high}) setup using only the original task, geometry, source, and new placements, without labels or predictions, while predicates, tolerances, durations, and stage order remain fixed. Accordingly, methods exposing external task bindings can adapt target locations to the new layout while keeping scoring code intact, whereas programs with hardcoded world coordinates cannot.

For task $i$, let $C_i^+$ and $C_i^-$ count correct positive and negative outputs among the two probes of each class. Layout balanced correct-output rate is $\mathrm{LBC}=\frac{100}{20}\sum_{i=1}^{20}\frac12(C_i^+/2+C_i^-/2)$. Valid outputs are 68/80, 44/80, 80/80, and 80/80 for GenSim, RoboGen, RoboLab, and EmbodiedStudio; the remaining 12 and 36 attempts retain original source errors. Table~\ref{tab:rubric-execution-counts} concerns the separate 120-record physical-state study; the LBC column uses the 80 counterfactual probes.

\section{Construction-Trajectory Learning Details}
\label{sec:supp-qwen}

\begin{figure*}[!t]
  \centering
  {\small
  \setlength{\tabcolsep}{2pt}
  \begin{tabular}{@{}cccc@{}}
  \multicolumn{2}{c}{\textbf{(a) Multi-room tidying (\texttt{M10})}} & \multicolumn{2}{c@{}}{\textbf{(b) Study workstation (\texttt{R16})}} \\[1pt]
  Before & After & Before & After \\[1pt]
  \includegraphics[width=.24\textwidth]{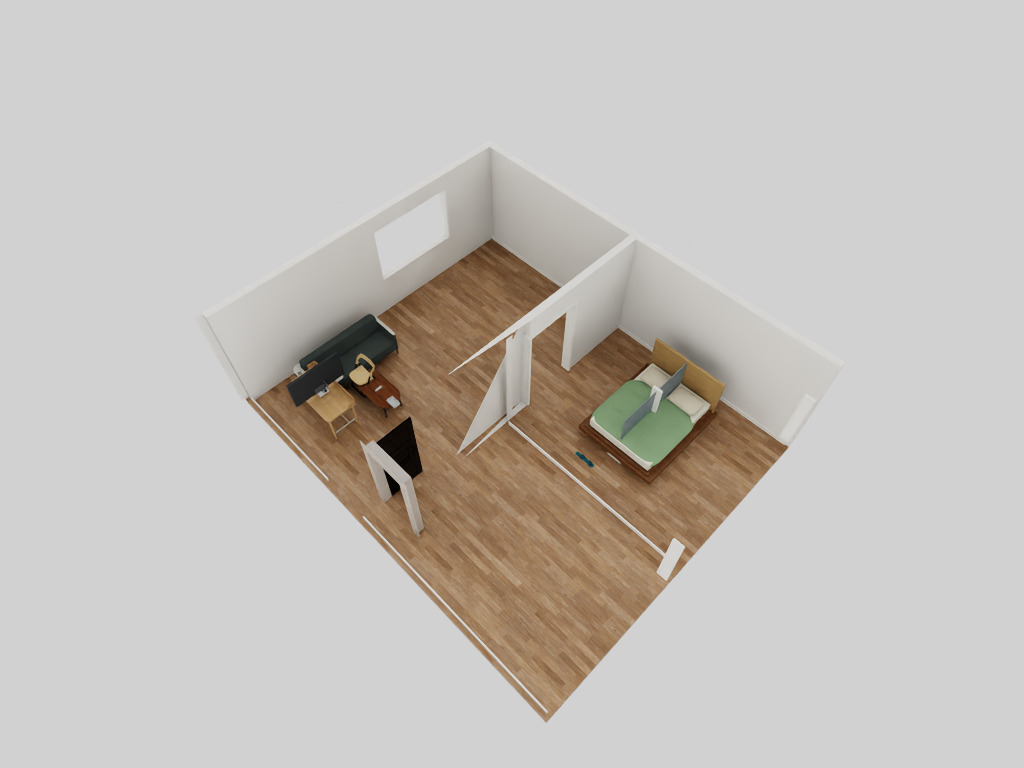} &
  \includegraphics[width=.24\textwidth]{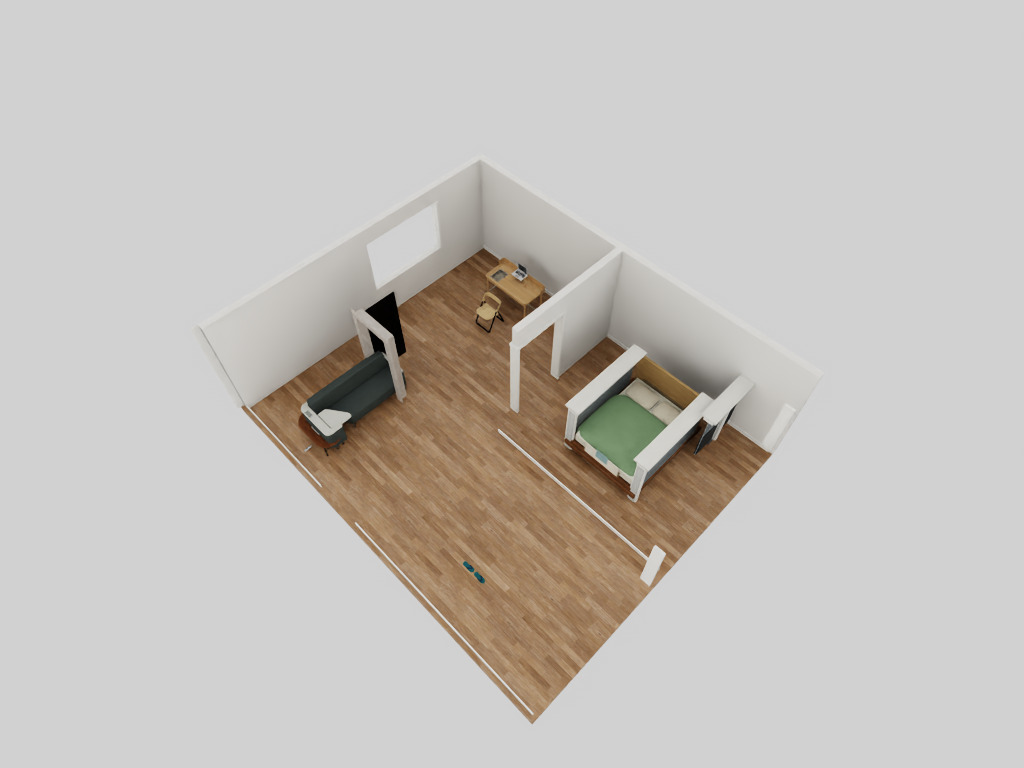} &
  \includegraphics[width=.24\textwidth]{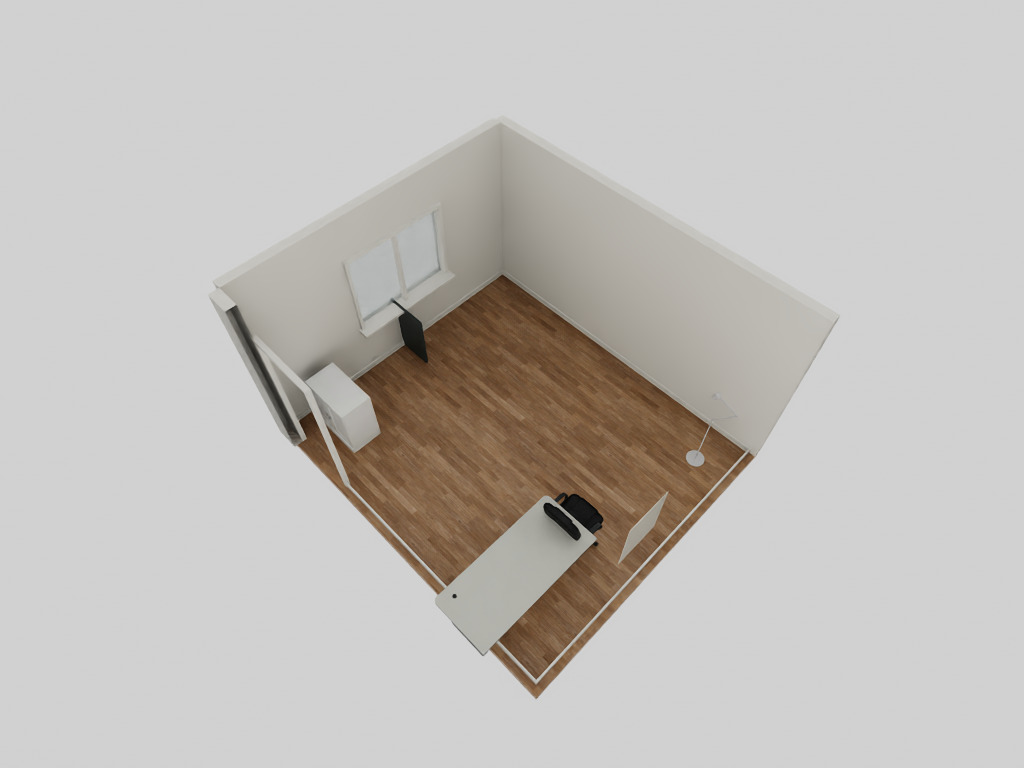} &
  \includegraphics[width=.24\textwidth]{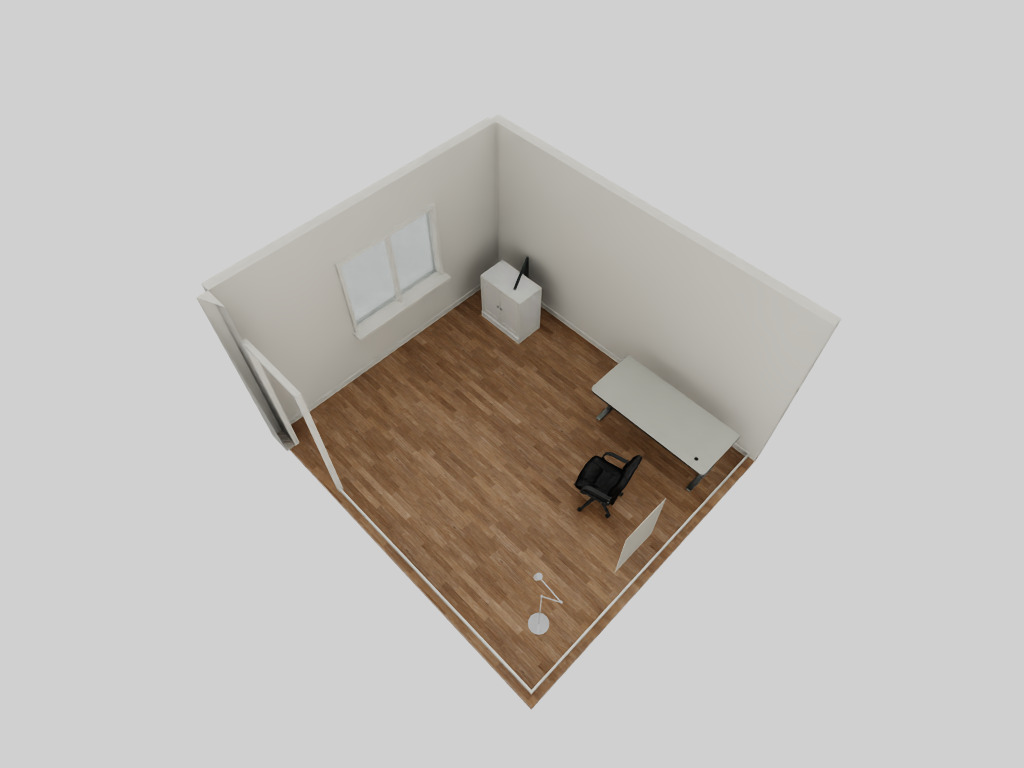}
  \end{tabular}}
  \caption{\textbf{Additional layout comparisons before and after fine-tuning on construction trajectories.} Matched renders using the base 27B checkpoint (left) and the fine-tuned checkpoint (right).}
  \label{fig:supp-traj-extra}
\end{figure*}

To evaluate whether forward construction histories $\mathcal{H}_x$ supervise scene authoring without future-information leakage, each training sample exposes only the prior request, current programs, and pre-action diagnostic context with assistant-only loss masks. We fine-tune \texttt{Qwen/Qwen3.8-27B} in BF16 precision using LoRA (rank $r=16$, $\alpha=32$), a peak learning rate of $5\times 10^{-5}$ (6 warmup steps, 0.1 cosine end ratio), a 32,768-token context window, per-GPU micro-batch size 1 with gradient accumulation 2 (global batch size 4 across two GPUs), and 112 optimizer updates on placement supervision extracted from EmbodiedStudio construction records. Training runs on two NVIDIA A100 80\,GB GPUs with DDP. At inference, both the base and fine-tuned checkpoints receive the same request, fixed room structure, and asset catalog; the main paper compares four scenes (\texttt{T04}, \texttt{M09}, \texttt{R09}, \texttt{R12}), and Figure~\ref{fig:supp-traj-extra} shows two additional comparisons (\texttt{M10} and \texttt{R16}).

\subsection{Quantitative Local Object Placement}
\label{sec:extended-local-placement}

To further validate a distinct quantitative facet of our forward construction experience data $\mathcal{H}_x$ beyond the qualitative full-scene layout study above, we train a separate focused LoRA probe on \texttt{Qwen/Qwen3.8-27B} using 44 recorded object translations from 12 training projects and compare it against the base checkpoint on placing an object onto a crowded work surface. Each prompt specifies the object's dimensions, the support bounds, nearby obstacle bounds, and a requested half of the surface (177--487 tokens per input). The model predicts the world coordinates of the object's bottom center at a fixed orientation using greedy decoding.

\begin{figure*}[!t]
  \centering
  \includegraphics[width=\textwidth]{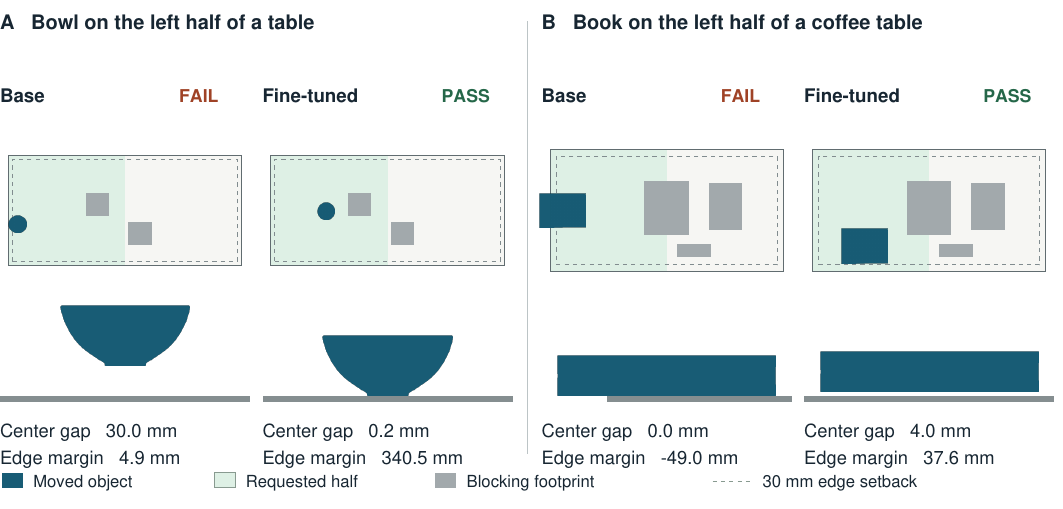}
  \caption{\textbf{Two held-out examples of improved local object placement.} Each pair shows a plan view of the support bounds and a side elevation centered horizontally on the target, with blue silhouettes projecting the visible target mesh from the reopened USD scene at matched scale. \textbf{A:} Tuning lowers the bowl onto the tabletop (center gap $30.0\,\mathrm{mm} \to 0.2\,\mathrm{mm}$) and clears the $30\,\mathrm{mm}$ edge setback. \textbf{B:} The base prediction overhangs the table by $49.0\,\mathrm{mm}$; after tuning, the book sits $37.6\,\mathrm{mm}$ inside the nearest edge and passes all five support checks.}
  \label{fig:local-placement-views}
\end{figure*}

\paragraph{Evaluation protocol and results.}
Each predicted pose is applied in a separate USD overlay and evaluated on the reopened scene across 21 held-out placements spanning eight projects from two task families (each verified feasible by deterministic grid search). A placement succeeds only when all three geometric conditions hold: (1) five-point footprint support (vertical gap $-2$ to $+5$\,mm), (2) full containment inside the requested half, and (3) at least $30$\,mm horizontal clearance from table edges and height-overlapping obstacles. As shown in Table~\ref{tab:local-placement}, fine-tuning raises overall pass rate from $3/21$ to $14/21$ ($+52.4$ percentage points), driven primarily by improved surface support ($4/21 \to 18/21$) and obstacle clearance ($7/21 \to 14/21$). In a post-hoc height-aligned control that sets both models' predicted heights to the tabletop surface, horizontal placements still improve from $7/21$ to $14/21$. Figure~\ref{fig:local-placement-views} illustrates two held-out examples.

\begin{table}[!tb]
\centering\scriptsize
\caption{\textbf{Quantitative local object placement on 21 held-out probes.} Base vs.\ fine-tuned \texttt{Qwen3.8-27B} across support, requested half, clearance, and combined pass rates.}
\label{tab:local-placement}
\setlength{\tabcolsep}{4pt}
\begin{tabular*}{\columnwidth}{@{\extracolsep{\fill}}lrrrr@{}}
\toprule
Model & Support & Requested half & Clearance & \textbf{All three} \\
\midrule
Base Qwen 27B & 4/21 & 21/21 & 7/21 & \textbf{3/21} \\
Fine-tuned (ours) & 18/21 & 20/21 & 14/21 & \textbf{14/21} \\
\bottomrule
\end{tabular*}
\end{table}

\paragraph{Training details.}
The local placement adapter starts from the base checkpoint and uses rank-8 LoRA for 44 updates on one NVIDIA A100 GPU ($117$\,s wall-clock time, $57.7$\,GiB peak reserved memory). This controlled probe isolates spatial placement supervision from recorded construction states.

\section{\texorpdfstring{$\pi_{0.5}$}{pi0.5} Policy Adaptation Details}
\label{sec:supp-pi05}

\begin{figure}[!tb]
  \centering
  \includegraphics[width=\columnwidth]{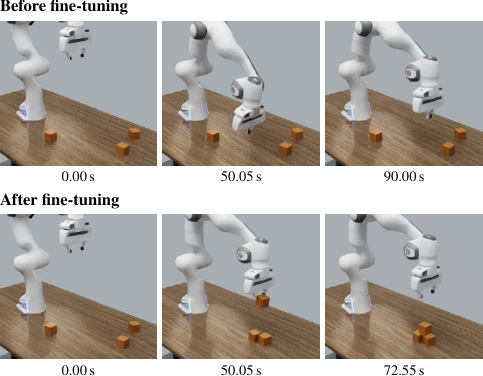}
  \caption{\textbf{Blocks task (\texttt{T15}) before and after $\pi_{0.5}$ LoRA fine-tuning} on held-out layout seed \texttt{9001}.}
  \label{fig:supp-pi05-blocks}
\end{figure}

Starting from the official OpenPI \texttt{pi05\_base} checkpoint (commit \texttt{215abfb2}), we fine-tune two task-specific LoRA policies on 30 verified simulated demonstrations each:
\begin{itemize}[leftmargin=*,nosep,topsep=2pt]
  \item \textbf{Apple (\texttt{T01})}: moving a designated apple into an empty bowl while leaving other objects undisturbed; evaluated with a staged rubric authored by our method (partial credit once the apple is lifted, full credit only after stable in-bowl support and verified release), scaled to $[0,100]$.
  \item \textbf{Blocks (\texttt{T15})}: building a stable two-level block structure; evaluated with a partial-credit rubric authored by our method tracking block placement and post-release stability, scaled to $[0,100]$.
\end{itemize}
Demonstrations are collected across 30 training layout seeds (\texttt{1001--1030}) and evaluated on held-out layout seed \texttt{9001} under five matched inference-noise seeds per checkpoint, with rollout horizons of 50\,s (Apple) and 90\,s (Blocks). Both tasks share a unified 9-DoF joint-position observation state (\texttt{panda\_joint1..7} in rad, \texttt{panda\_finger\_joint1..2} in m), an 8-DoF absolute position target (\texttt{panda\_joint1..7}, \texttt{panda\_finger\_joint1}), external and wrist RGB cameras at 20\,Hz, 20\,Hz control frequency, 120\,Hz physics stepping, and 50-step action chunks (2.5\,s). LoRA adaptation runs on a single NVIDIA A100 80\,GB GPU using batch size 16 for 20,000 updates with training-split-only normalization. Figure~\ref{fig:supp-pi05-blocks} compares the base and fine-tuned checkpoints on the Blocks task.

\section{Robot Interaction Records, Keyframes, and Models}
\label{sec:supp-robots}

The main-paper robot keyframes (Figure~\ref{fig:robot-keyframes}) select four successful recorded episodes: Franka Panda stores a pen in a drawer (\texttt{T07}; 2,991 control cycles, 149.55\,s), Aloha-AgileX packs and closes a box (\texttt{T16}; 558 cycles, 18.60\,s), Galaxea R1 Pro removes bathroom litter (\texttt{R18}; 268 cycles, 26.80\,s), and an onboarded Husky~A200 + Kinova~Gen3 delivers a bowl followed by a spoon (4,393 cycles, 87.86\,s). All four episodes satisfy their frozen task rubrics with a reward of 1.0, and the drawer record retains its earlier failed lip-grasp attempts prior to completion. A separate Franka blocks episode (\texttt{T15}; 798 cycles, 39.90\,s, 1.0\,s verified stability) is also included in the execution records. Table~\ref{tab:robots-16} lists the 16 locally pre-integrated robot models displayed in Figure~1 with their provenance. Beyond this pre-integrated local library, the system onboards external robot models on demand via declarative robot profiles during task generation---such as Husky~A200 + Kinova~Gen3 (main paper Figure~\ref{fig:robot-keyframes}), \texttt{I2RT YAM} (Section~\ref{sec:supp-trajectory-capture}), and \texttt{TIAGo++} (Section~\ref{sec:supp-rollouts}).

\subsection{Tabletop Trajectory Collection}
\label{sec:supp-trajectory-capture}

We collected rubric-verified demonstrations for 20 of the 20 tabletop tasks (100.0\%). Figure~\ref{fig:supp-tabletop-capture} shows one representative keyframe per collected task, including multi-robot coordination where helpful (e.g., two Fetch robots collaborating in \texttt{T08} so one holds the cabinet door open while the other retrieves the cup). Across all 20 collected tasks (302 physical rollout attempts), the agent averages 13.45 tested behavior revisions (median 11) and 15.1 physical rollout attempts (median 13) to reach the first rubric-verified completed demonstration (Table~\ref{tab:supp-capture-effort}). Figure~\ref{fig:supp-original-t01-embodiments} further compares four successful executions across distinct robot embodiments on the \texttt{T01} apple-placement scene.

\begin{table}[!tb]
\centering\small
\caption{\textbf{Tabletop collection coverage and iteration effort.} Effort is measured through the first rubric-verified completed demonstration in each recorded collection run. A rollout counts after committing a physical control cycle. A revision counts when the frozen controller or helper digest changes between consecutive rollout attempts.}
\label{tab:supp-capture-effort}
\begin{tabular}{@{}lrr@{}}
\toprule
Metric & Mean & Median \\
\midrule
Tested behavior revisions per task & 13.45 & 11 \\
Physical rollout attempts per task & 15.1 & 13 \\
\bottomrule
\end{tabular}

\smallskip
{\footnotesize Coverage: 20/20 (100.0\%). Effort: 20 tasks, 302 rollout attempts.}
\end{table}

\newcommand{\captureimage}[1]{\includegraphics[width=.24\textwidth]{figures/evidence/trajectory-capture-20261124/#1.jpg}}
\newcommand{\capturelabel}[2]{\scriptsize\texttt{#1}\enspace#2}

\begin{figure*}[p]
\centering
\setlength{\tabcolsep}{2pt}
\begin{tabular}{@{}cccc@{}}
\captureimage{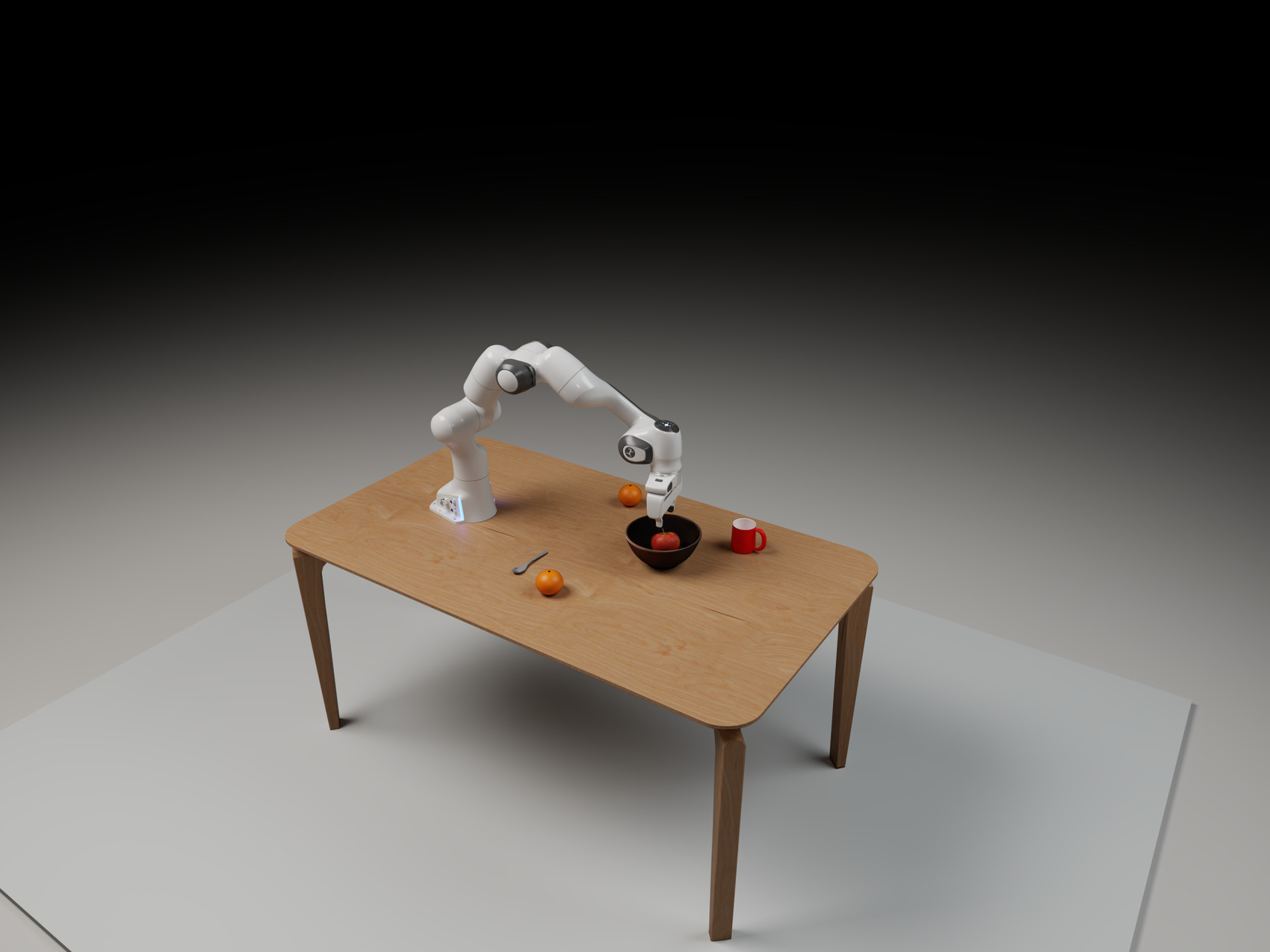} & \captureimage{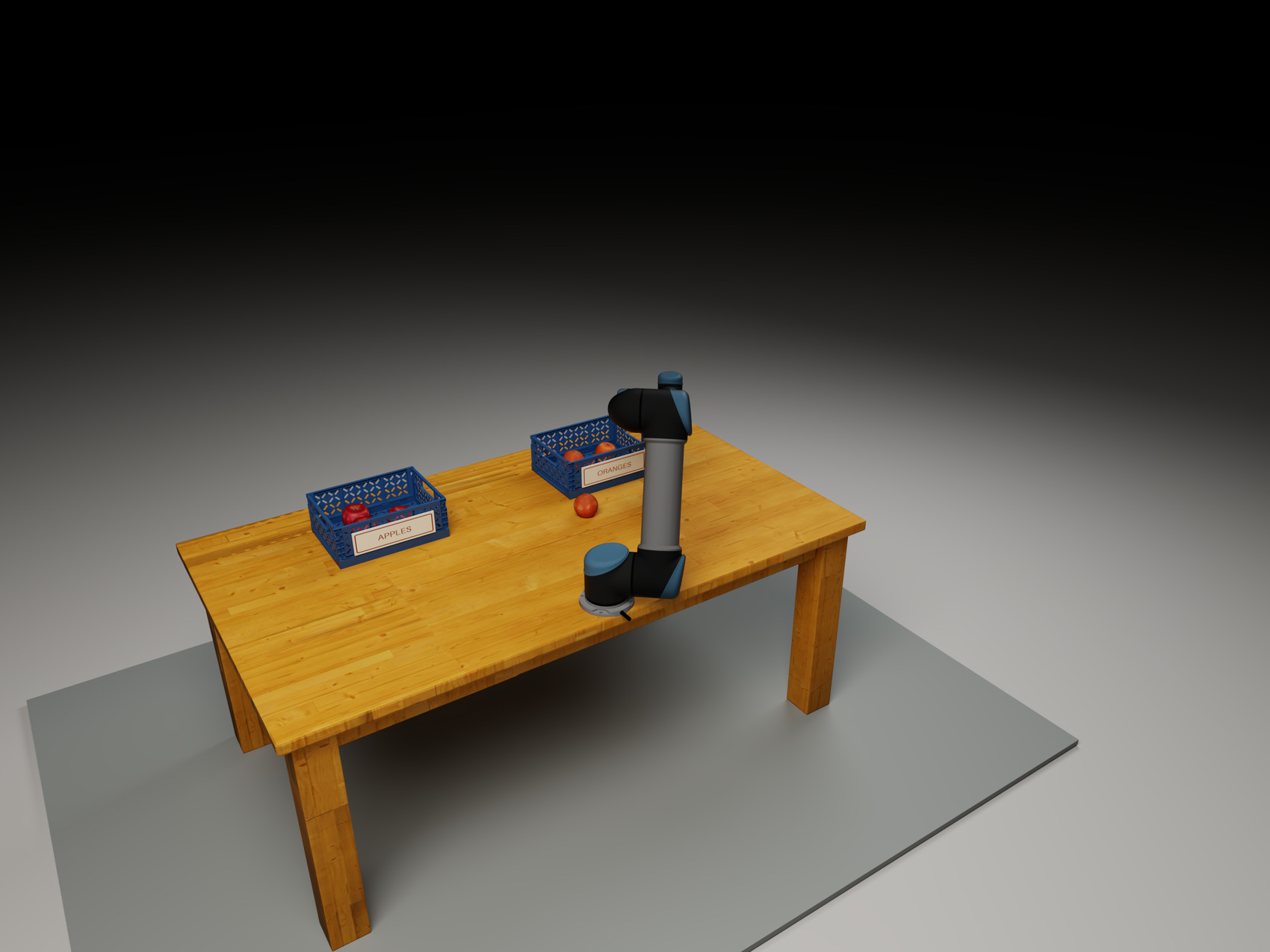} & \captureimage{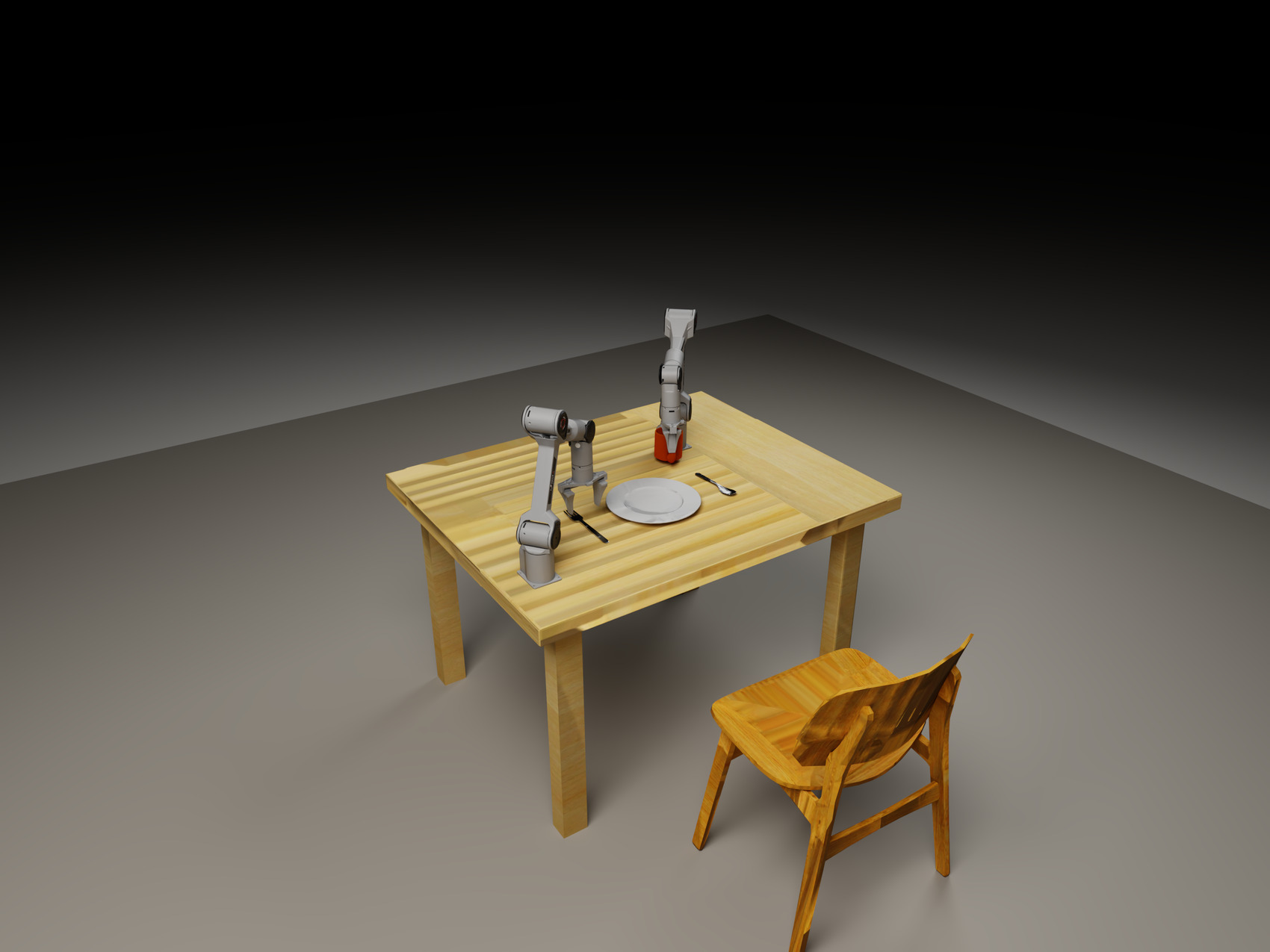} & \captureimage{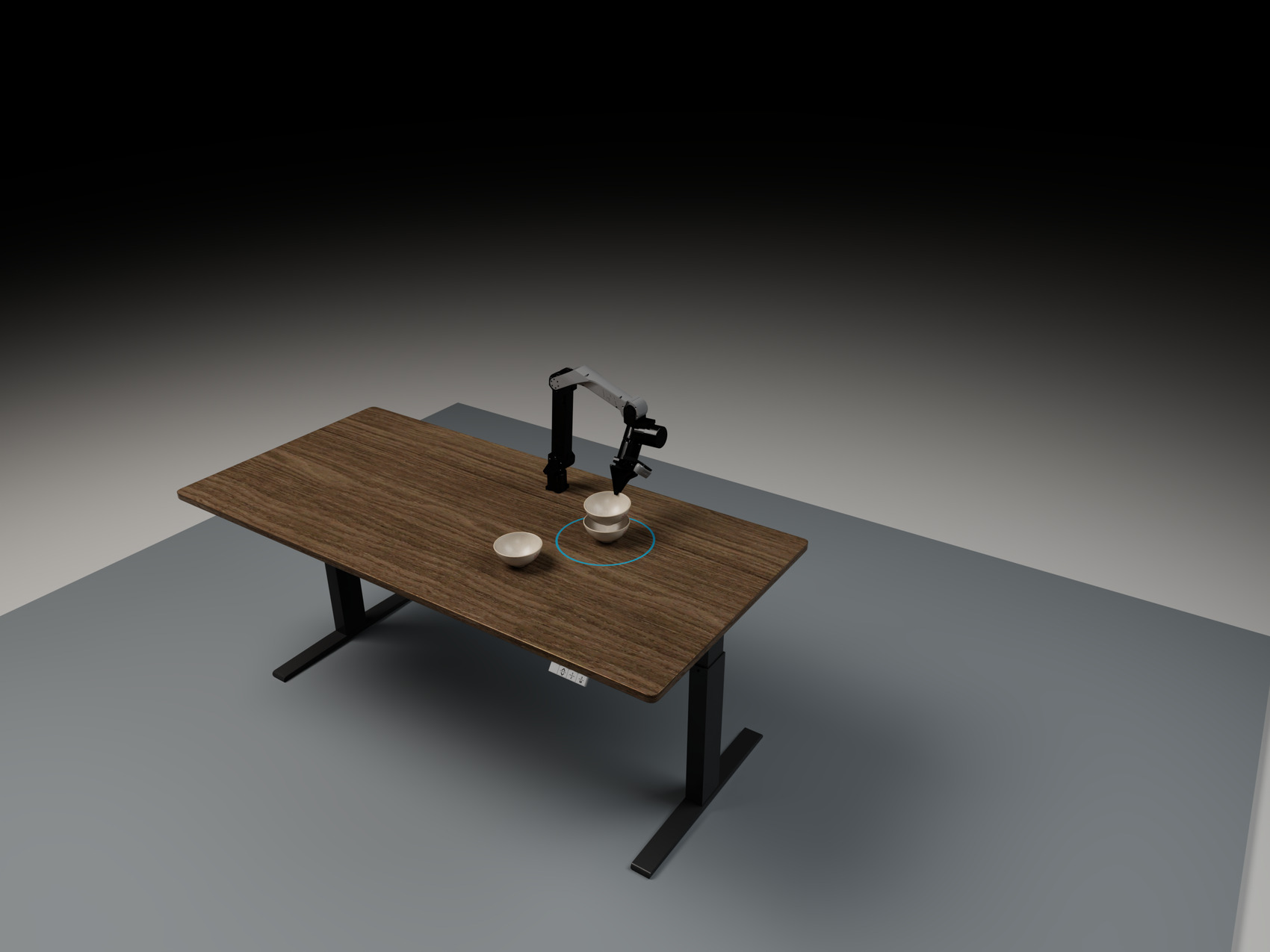} \\
\capturelabel{T01}{Place apple} & \capturelabel{T02}{Sort fruit} & \capturelabel{T03}{Set the table} & \capturelabel{T04}{Stack bowls} \\[5pt]
\captureimage{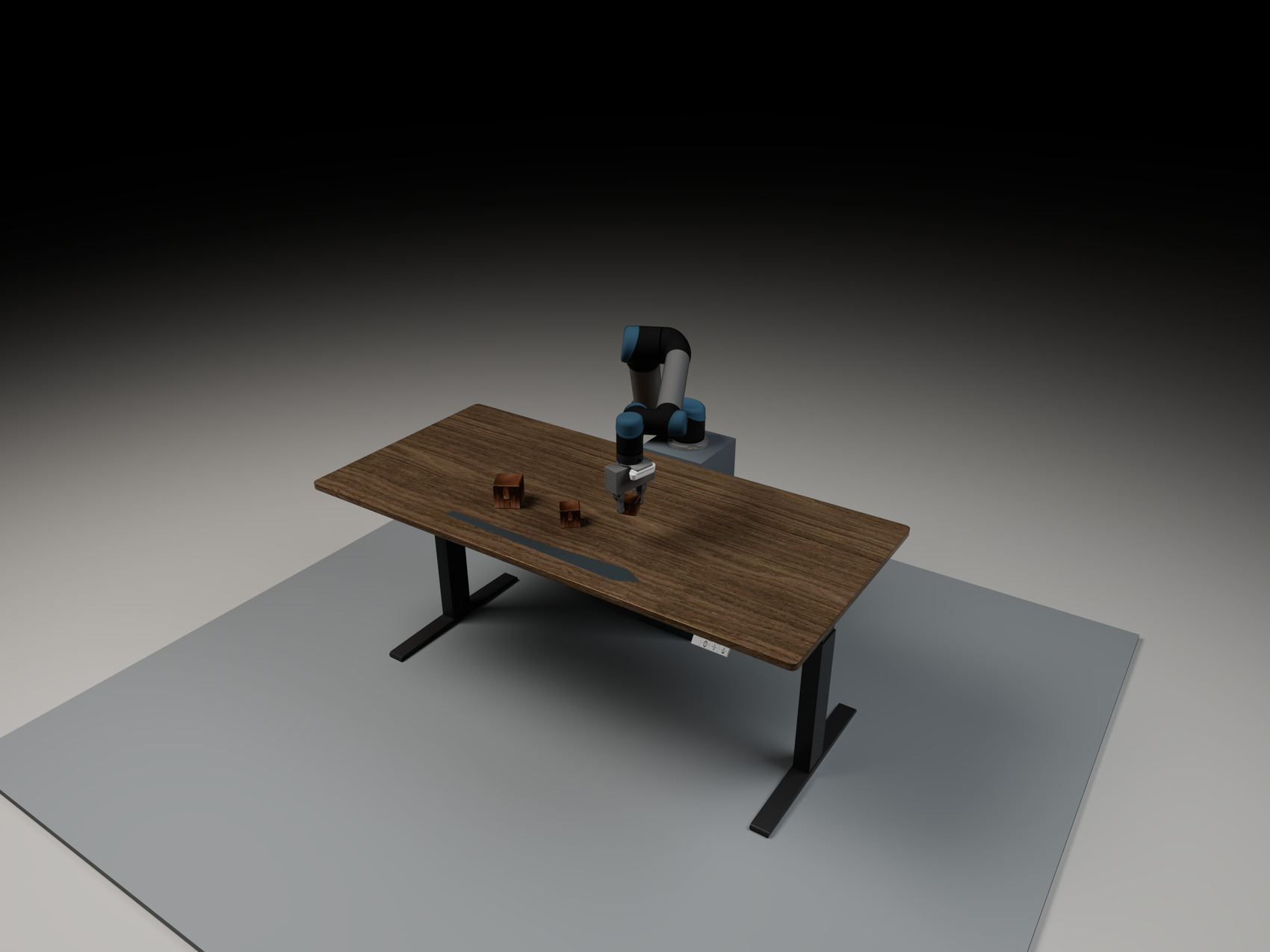} & \captureimage{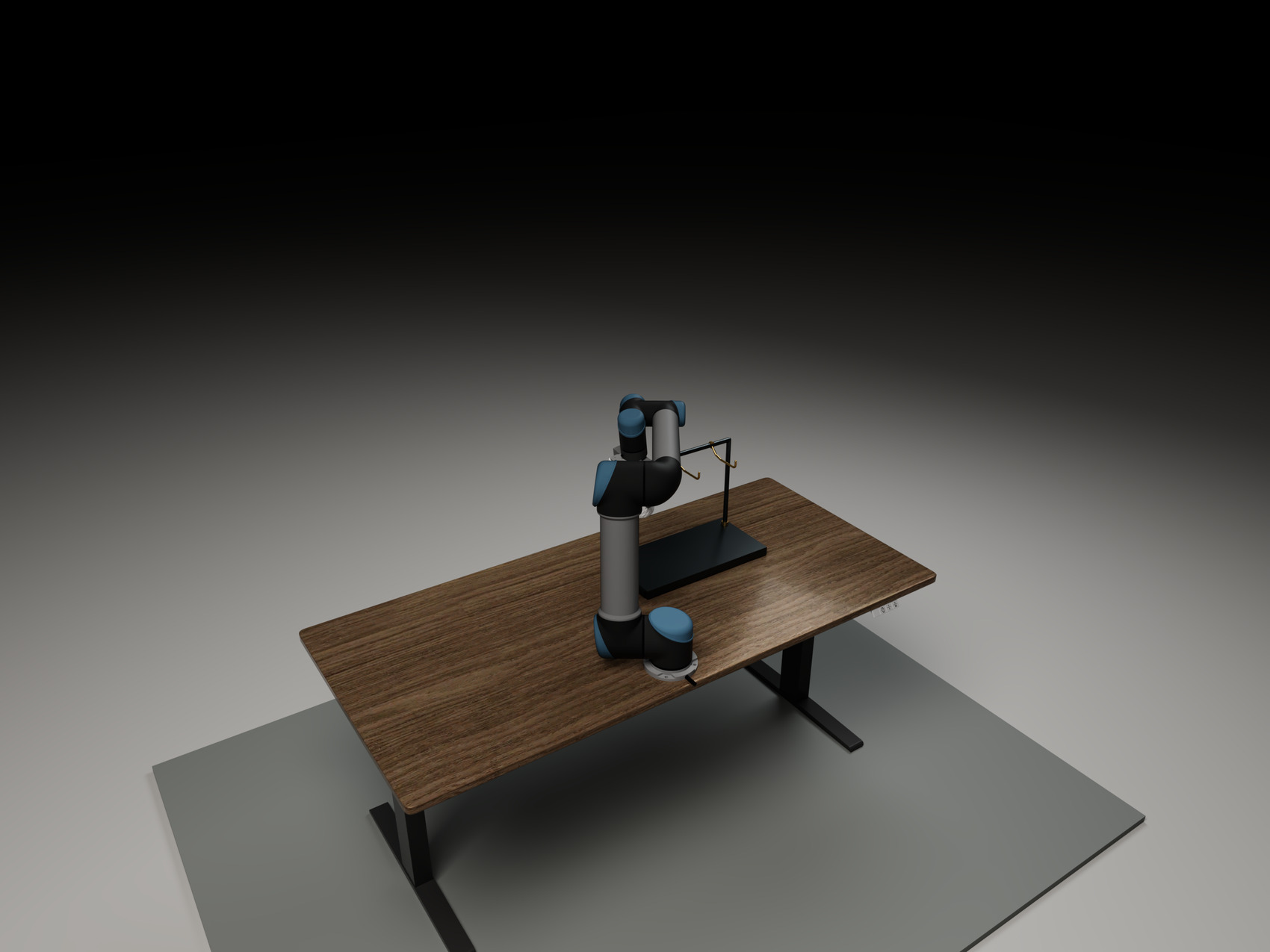} & \captureimage{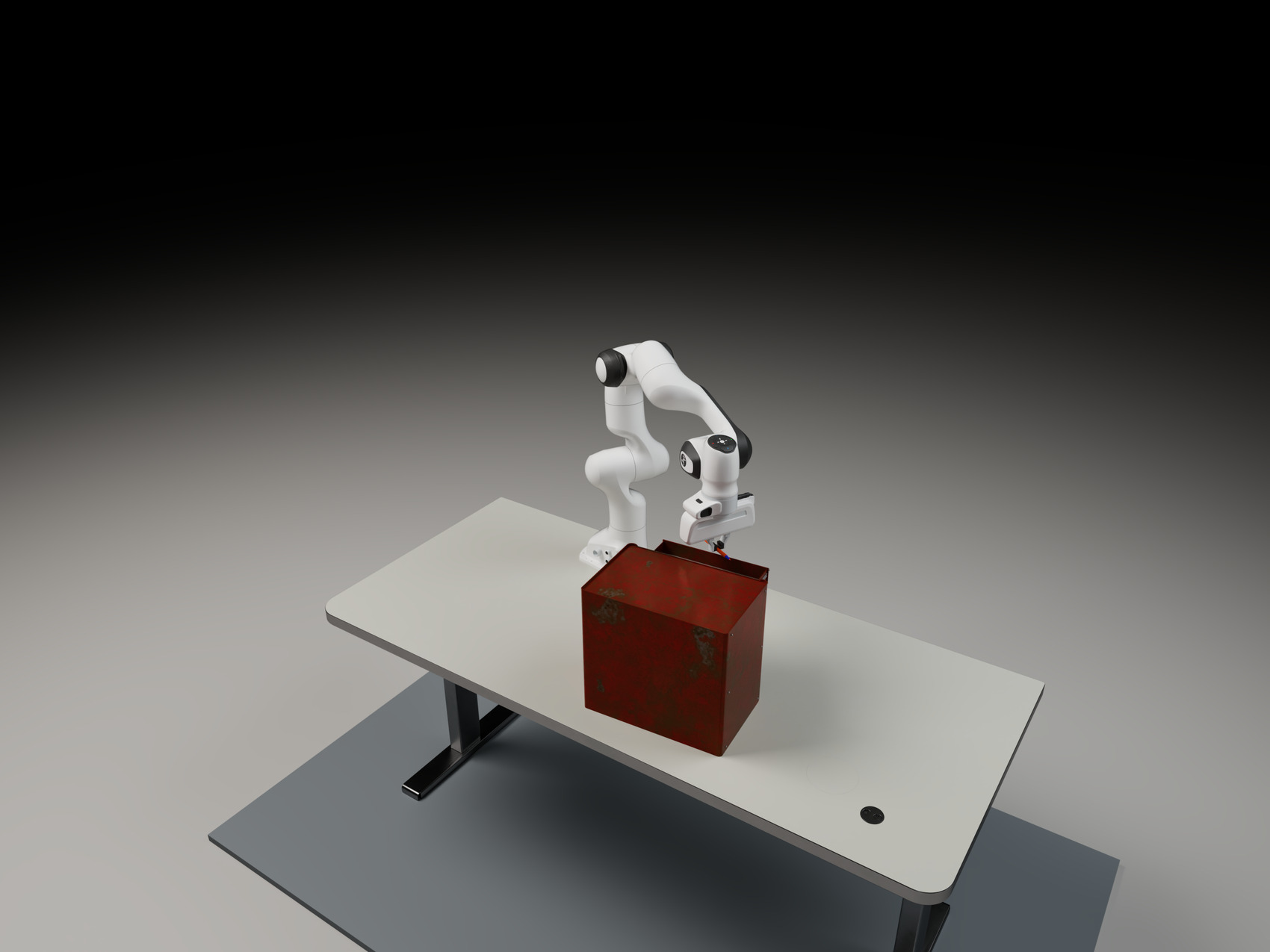} & \captureimage{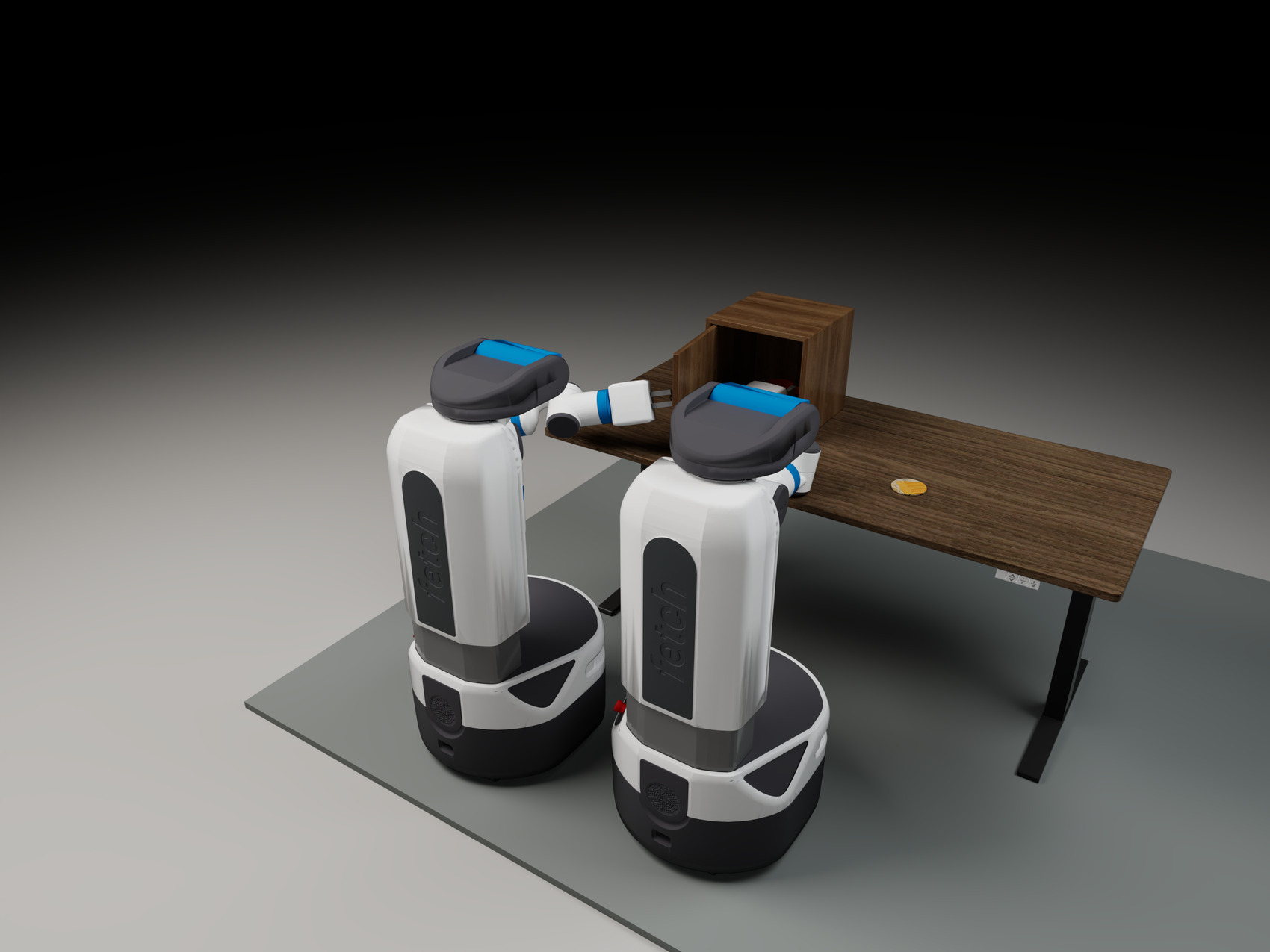} \\
\capturelabel{T05}{Order blocks} & \capturelabel{T06}{Hang mug} & \capturelabel{T07}{Store stationery} & \capturelabel{T08}{Retrieve cup} \\[5pt]
\captureimage{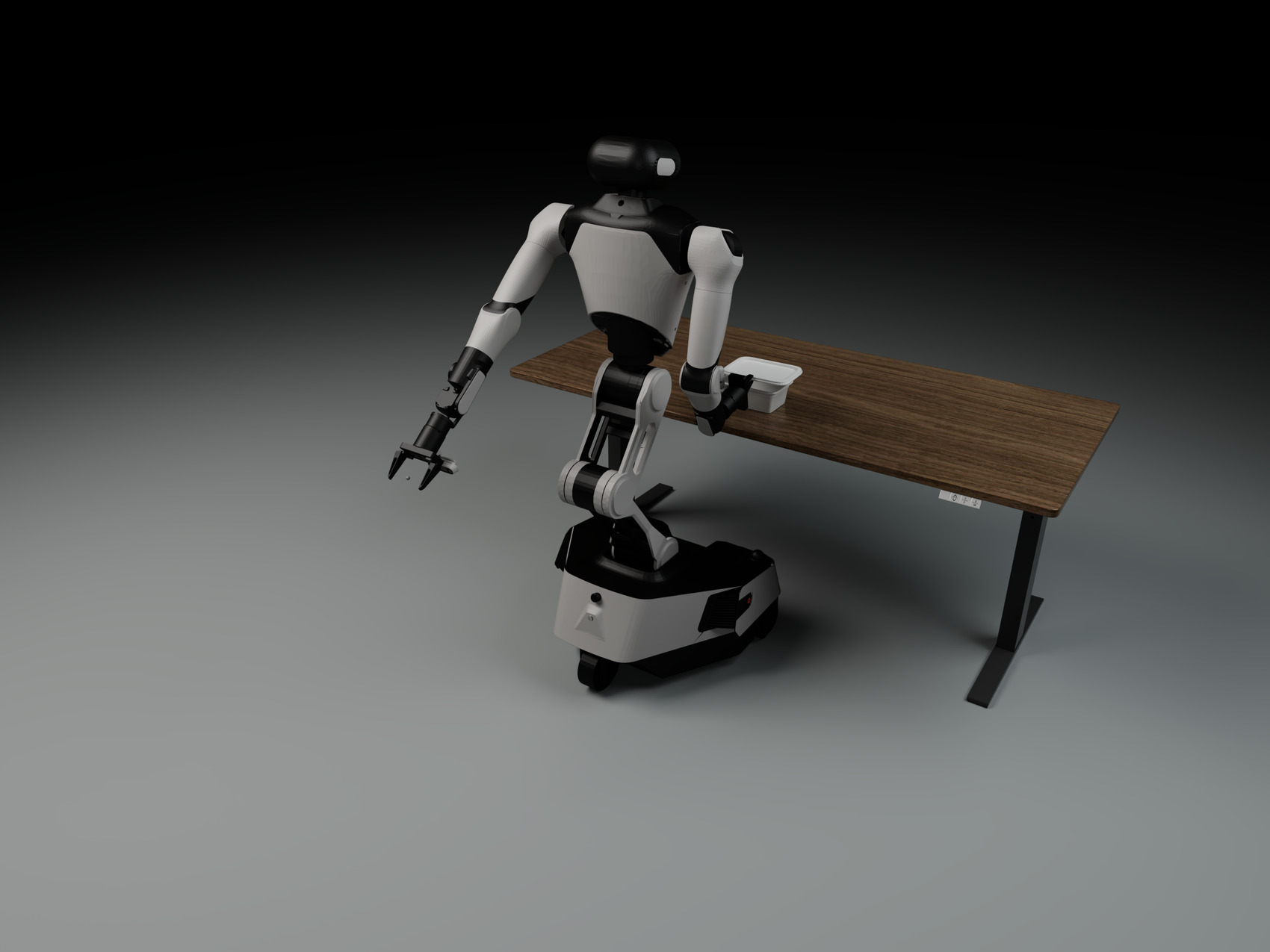} & \captureimage{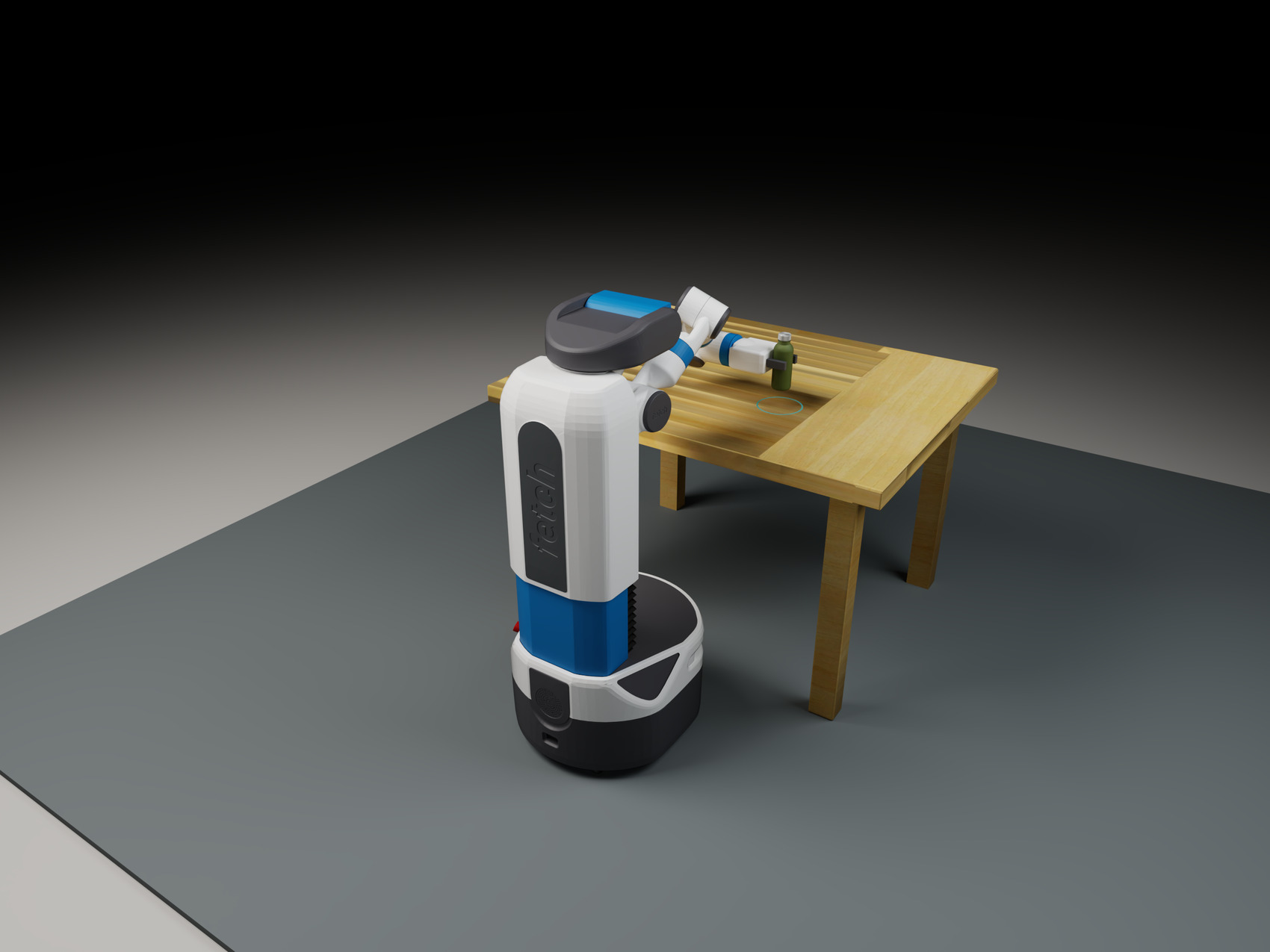} & \captureimage{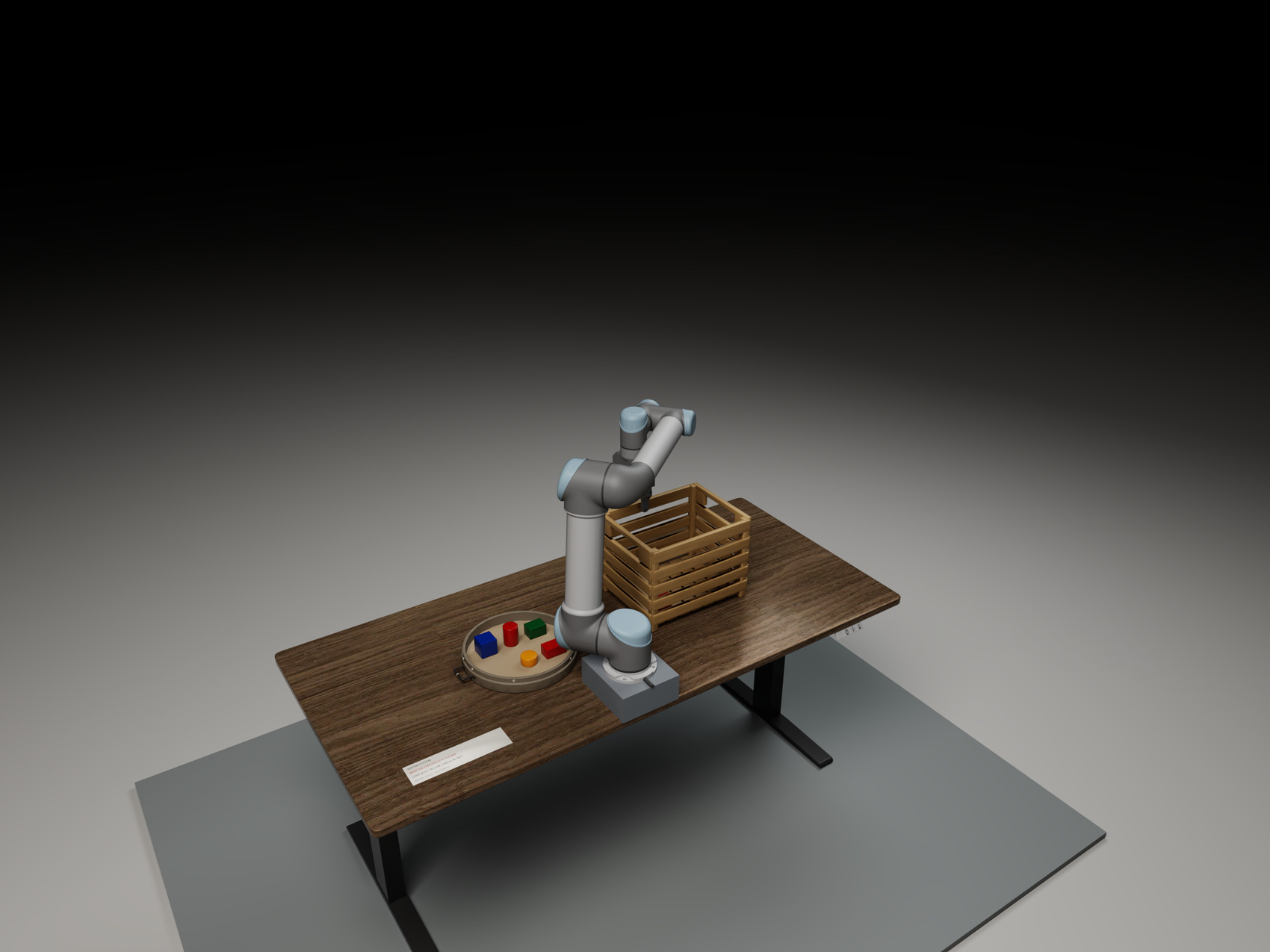} & \captureimage{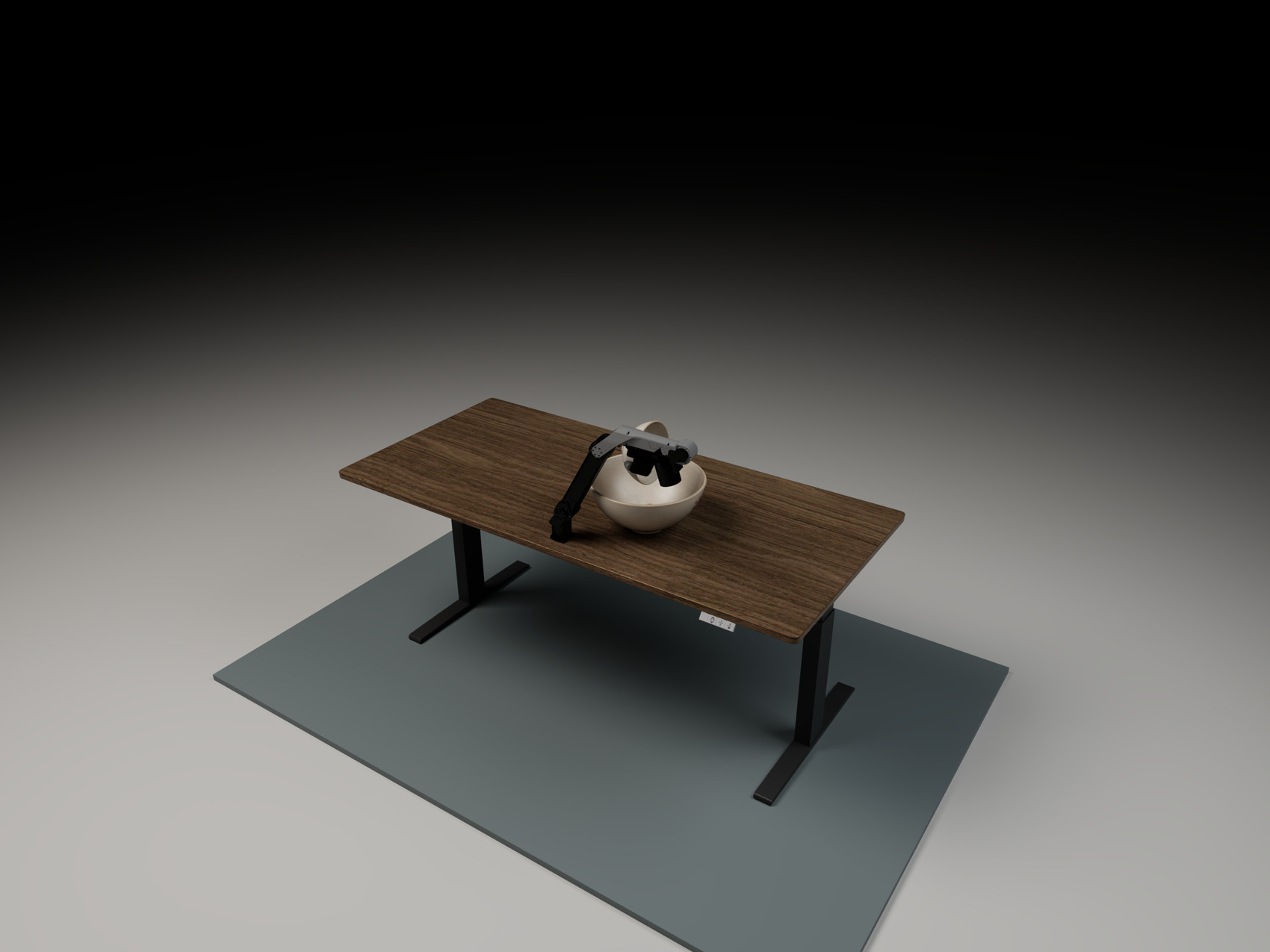} \\
\capturelabel{T09}{Close container} & \capturelabel{T10}{Stand bottle} & \capturelabel{T11}{Transfer from tray} & \capturelabel{T12}{Pour wooden balls} \\[5pt]
\captureimage{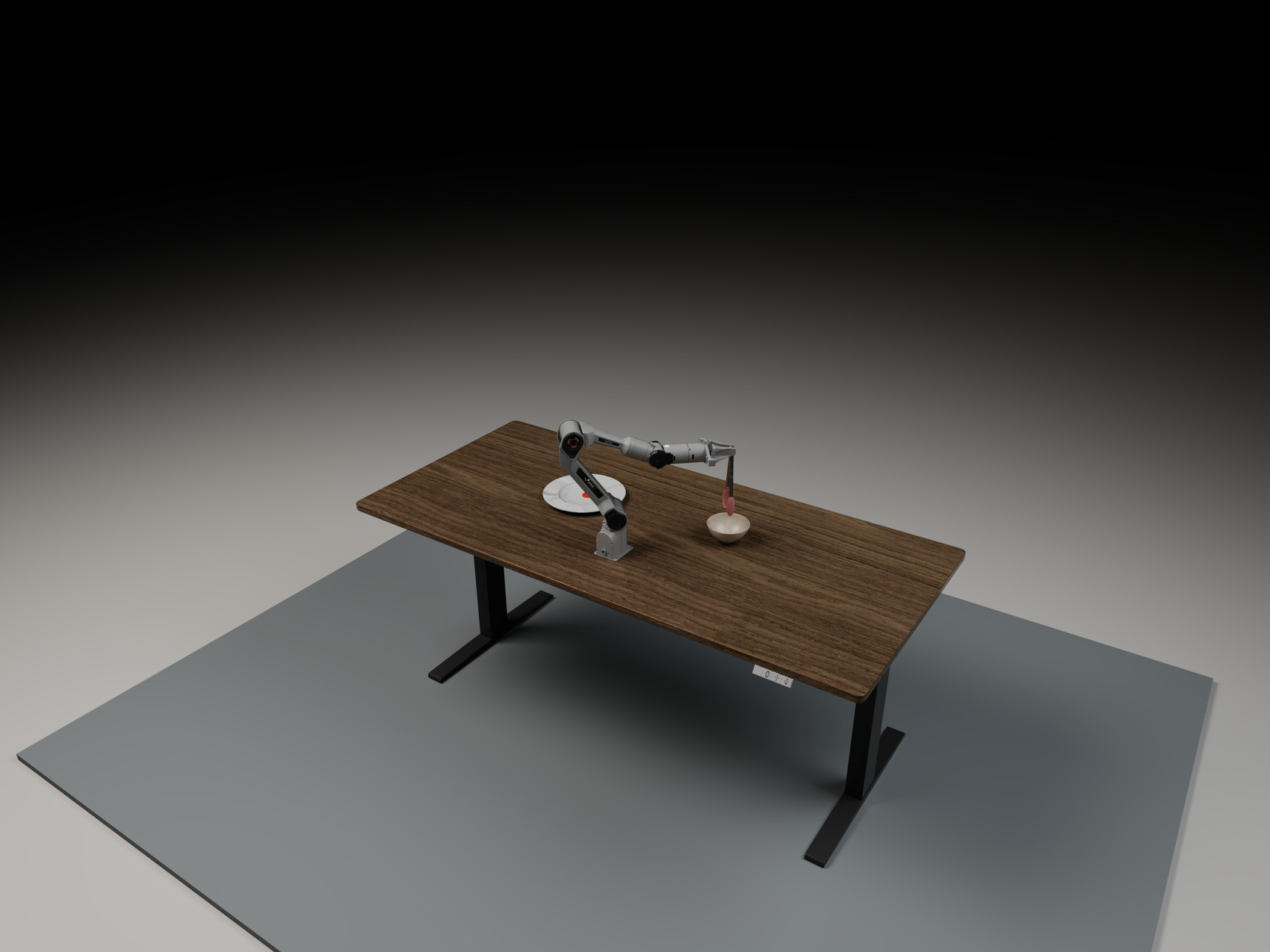} & \captureimage{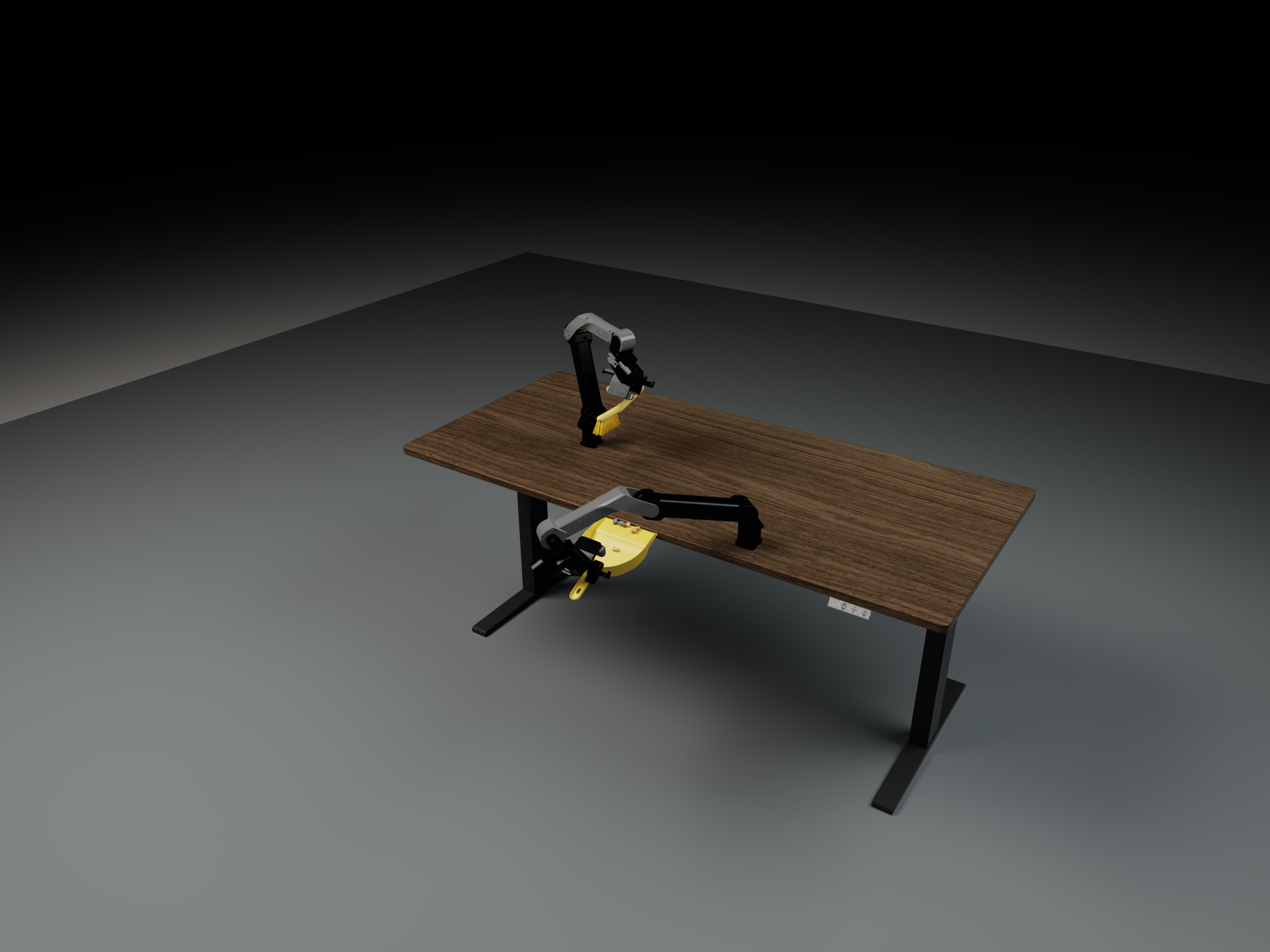} & \captureimage{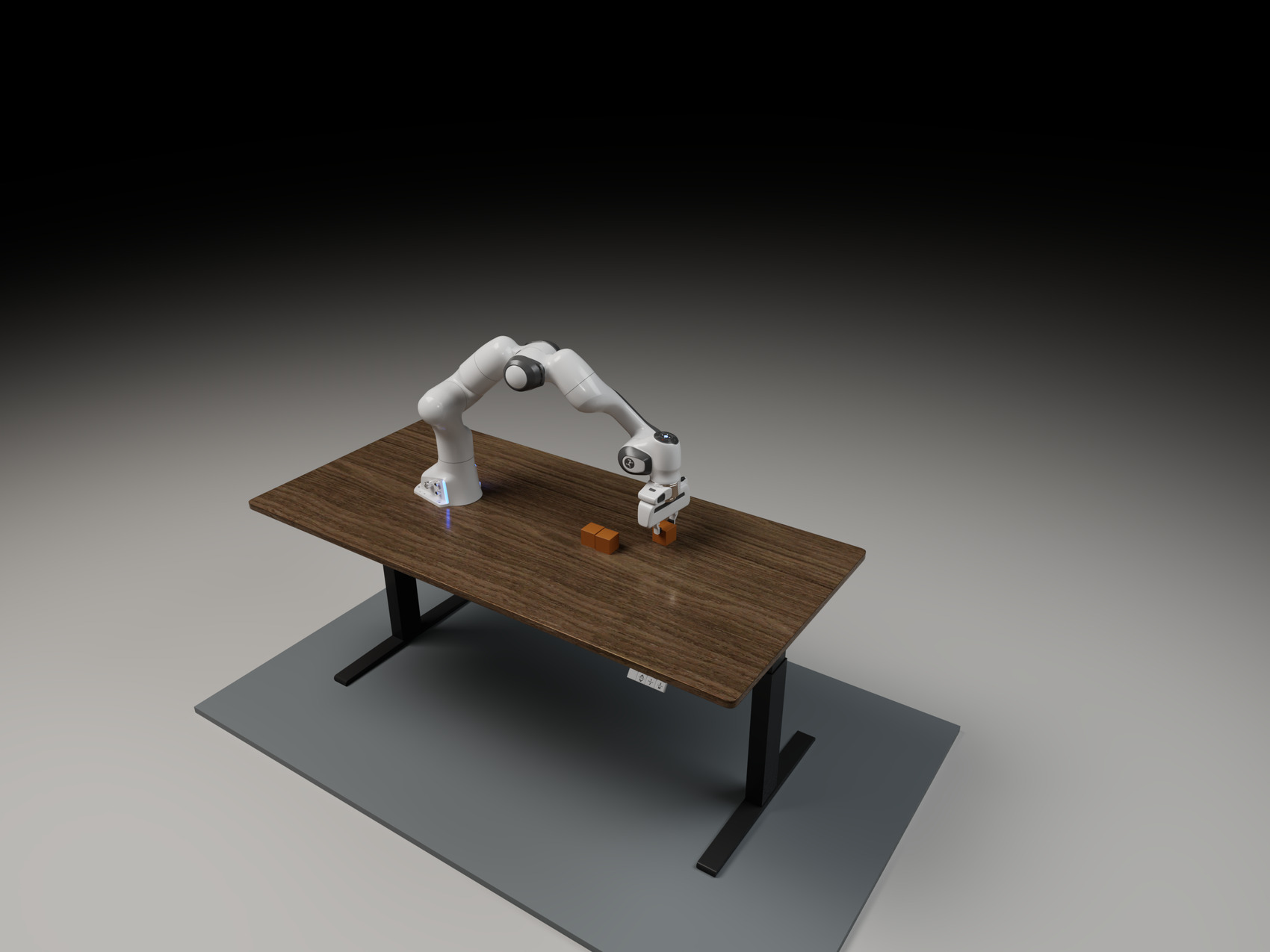} & \captureimage{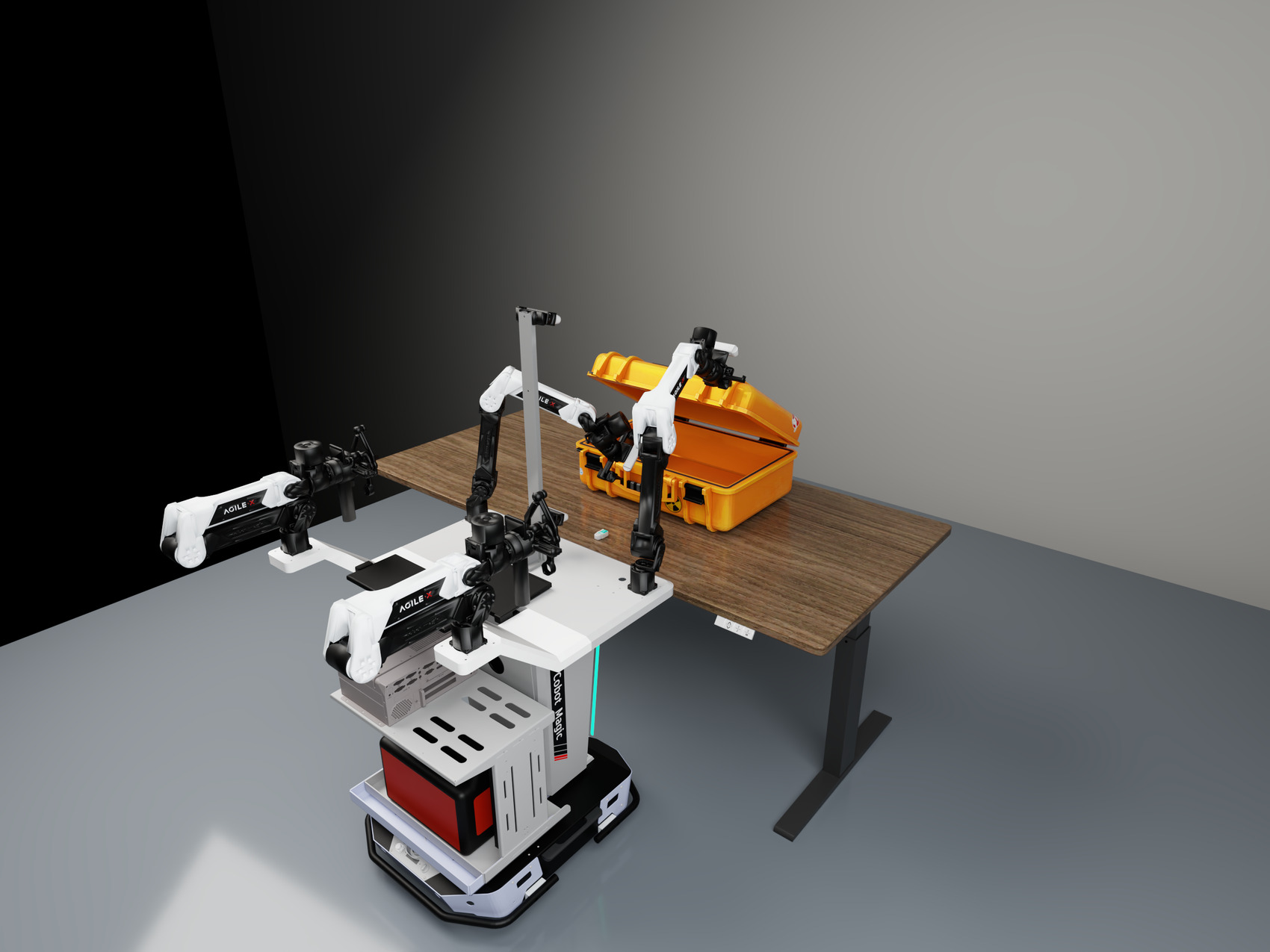} \\
\capturelabel{T13}{Use tongs} & \capturelabel{T14}{Sweep debris} & \capturelabel{T15}{Build block structure} & \capturelabel{T16}{Pack storage box} \\[5pt]
\captureimage{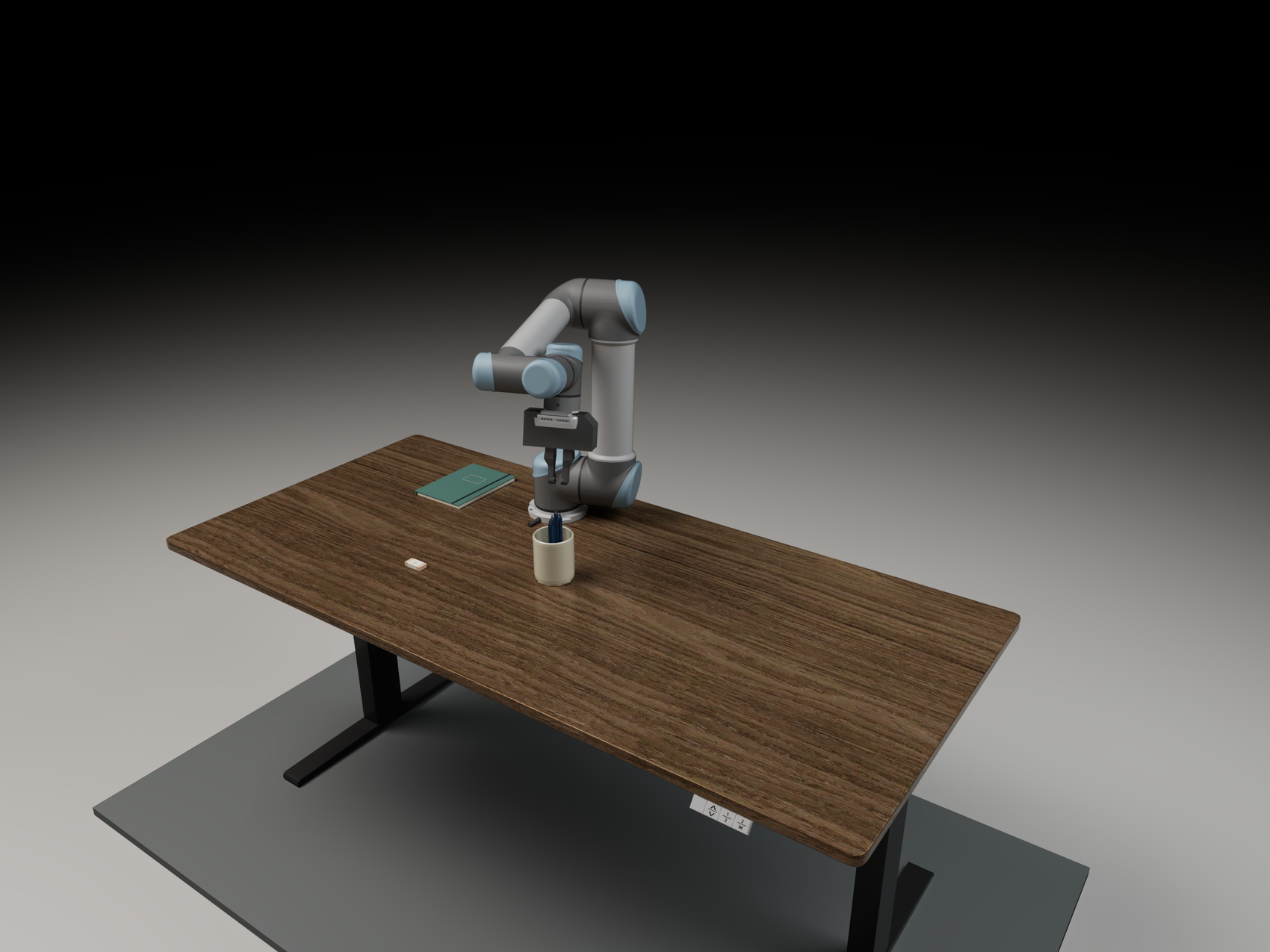} & \captureimage{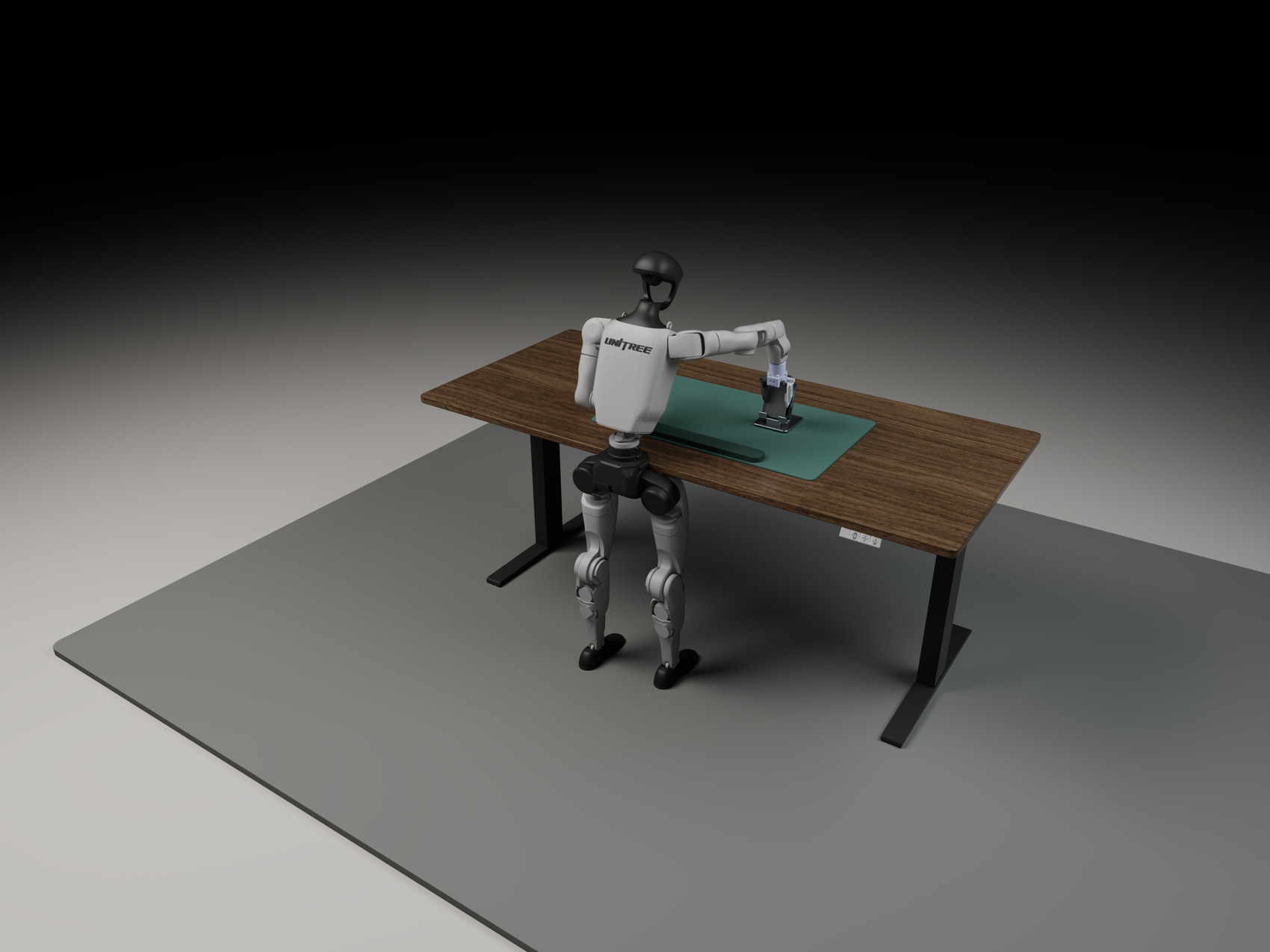} & \captureimage{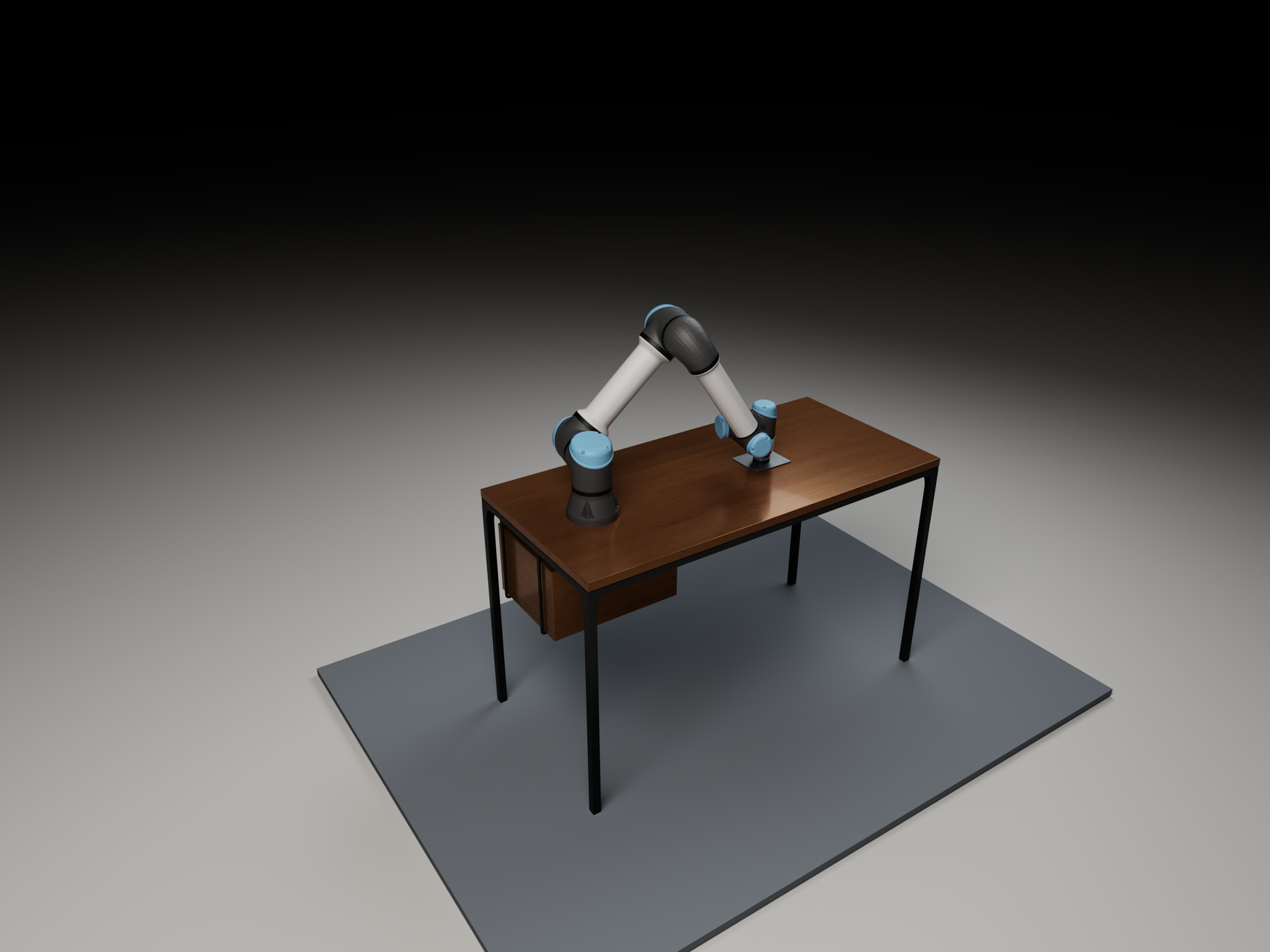} & \captureimage{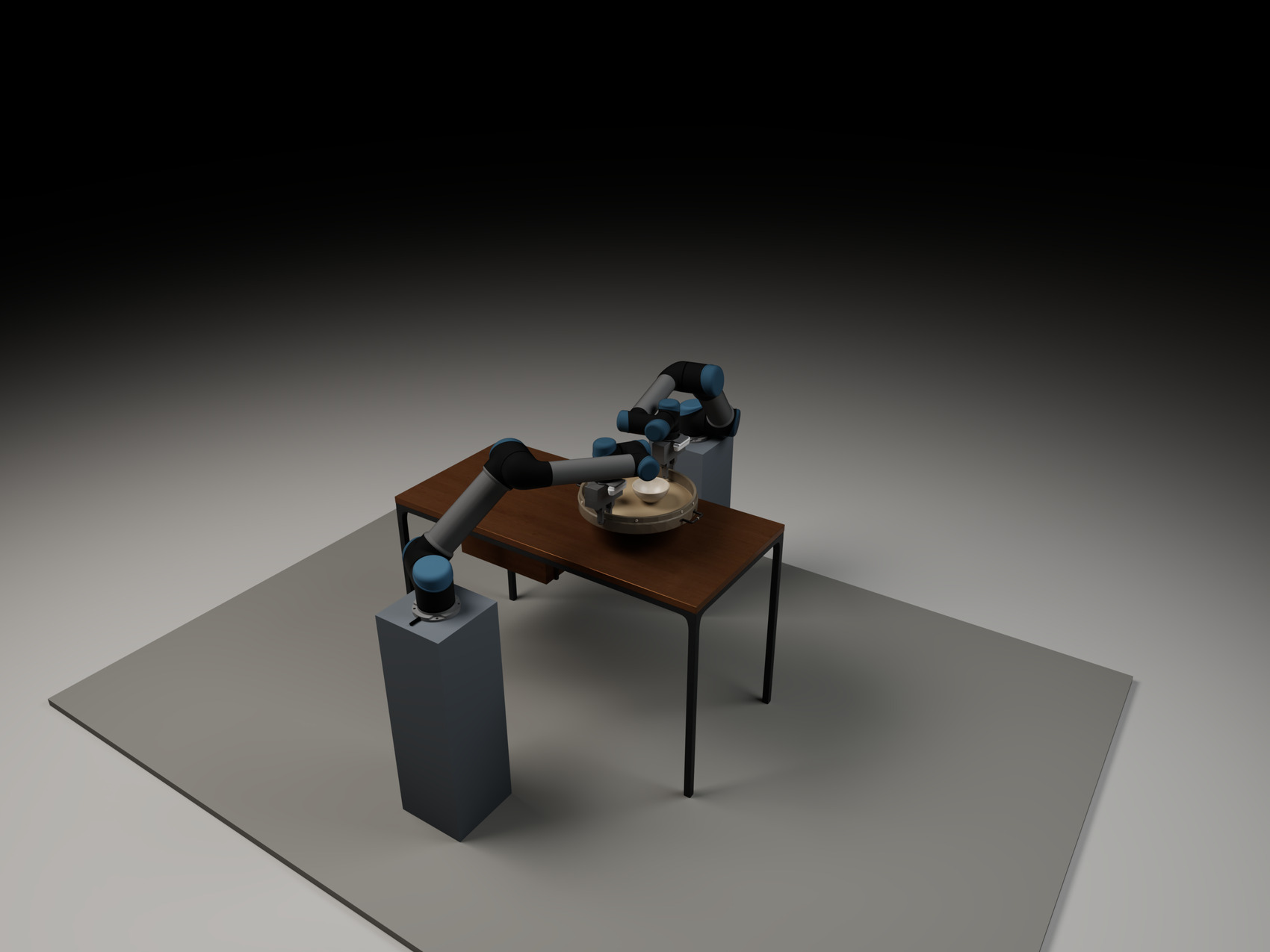} \\
\capturelabel{T17}{Organize pens} & \capturelabel{T18}{Place phone} & \capturelabel{T19}{Press button} & \capturelabel{T20}{Move loaded tray} \\
\end{tabular}
\caption{\textbf{Successful robot data collection across the tabletop suite.}
Each 4:3 panel renders a logged state from its demonstration at
$2560\times1920$ (1024 path-tracing samples per pixel, 12 diffuse bounces,
up to 16 specular/transmission bounces), covering 20 of 20 tasks (100.0\%).}
\label{fig:supp-tabletop-capture}
\end{figure*}

\begin{figure*}[p]
\centering
\newcommand{\tOneImage}[1]{\includegraphics[width=.225\textwidth]{figures/evidence/trajectory-capture-20261124/original-t01/#1.jpg}}
\setlength{\tabcolsep}{2pt}
\begin{tabular}{@{}cccc@{}}
\multicolumn{4}{@{}l@{}}{\small\textbf{Franka Panda} \enspace Single arm} \\[2pt]
\tOneImage{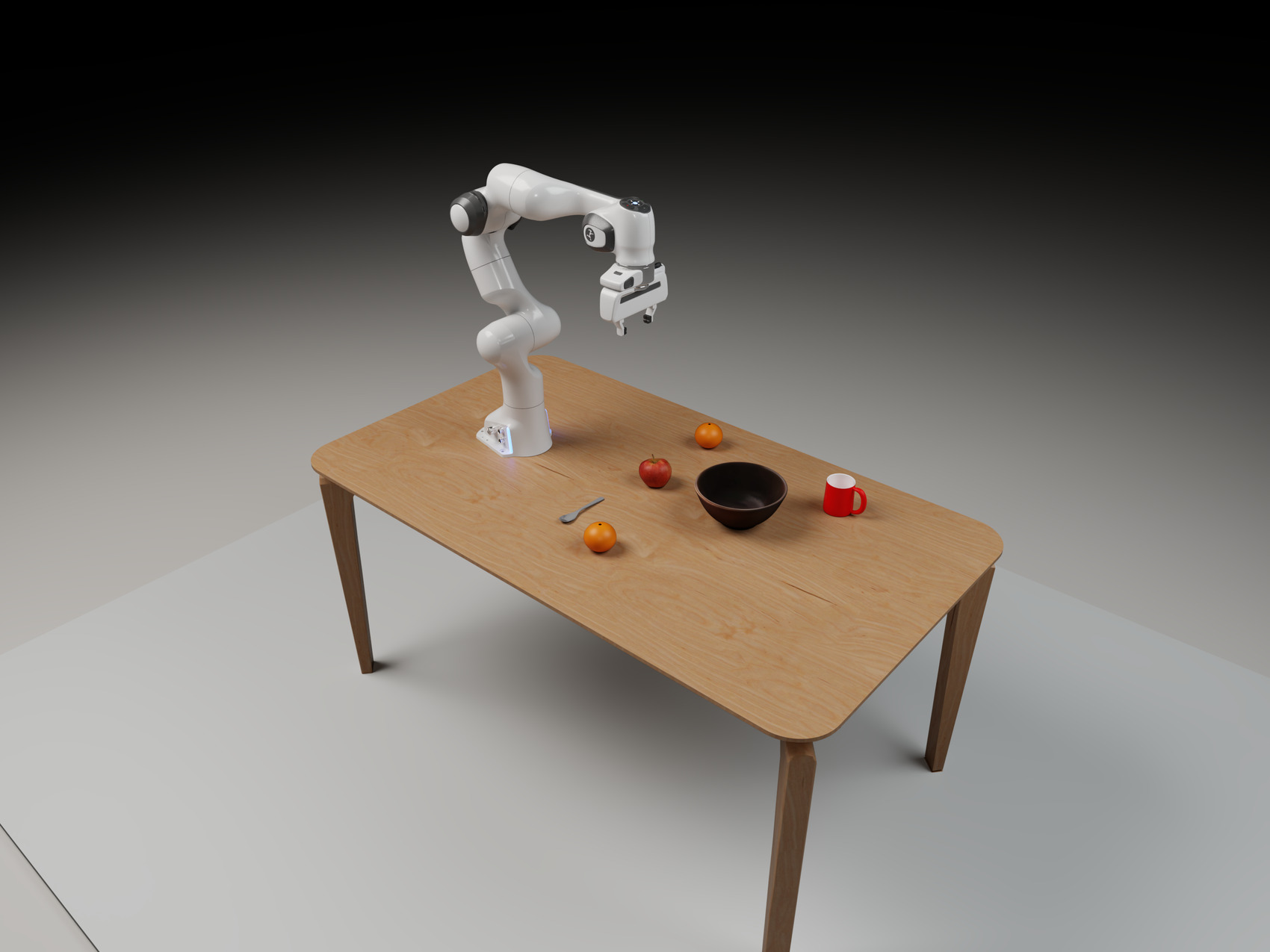} & \tOneImage{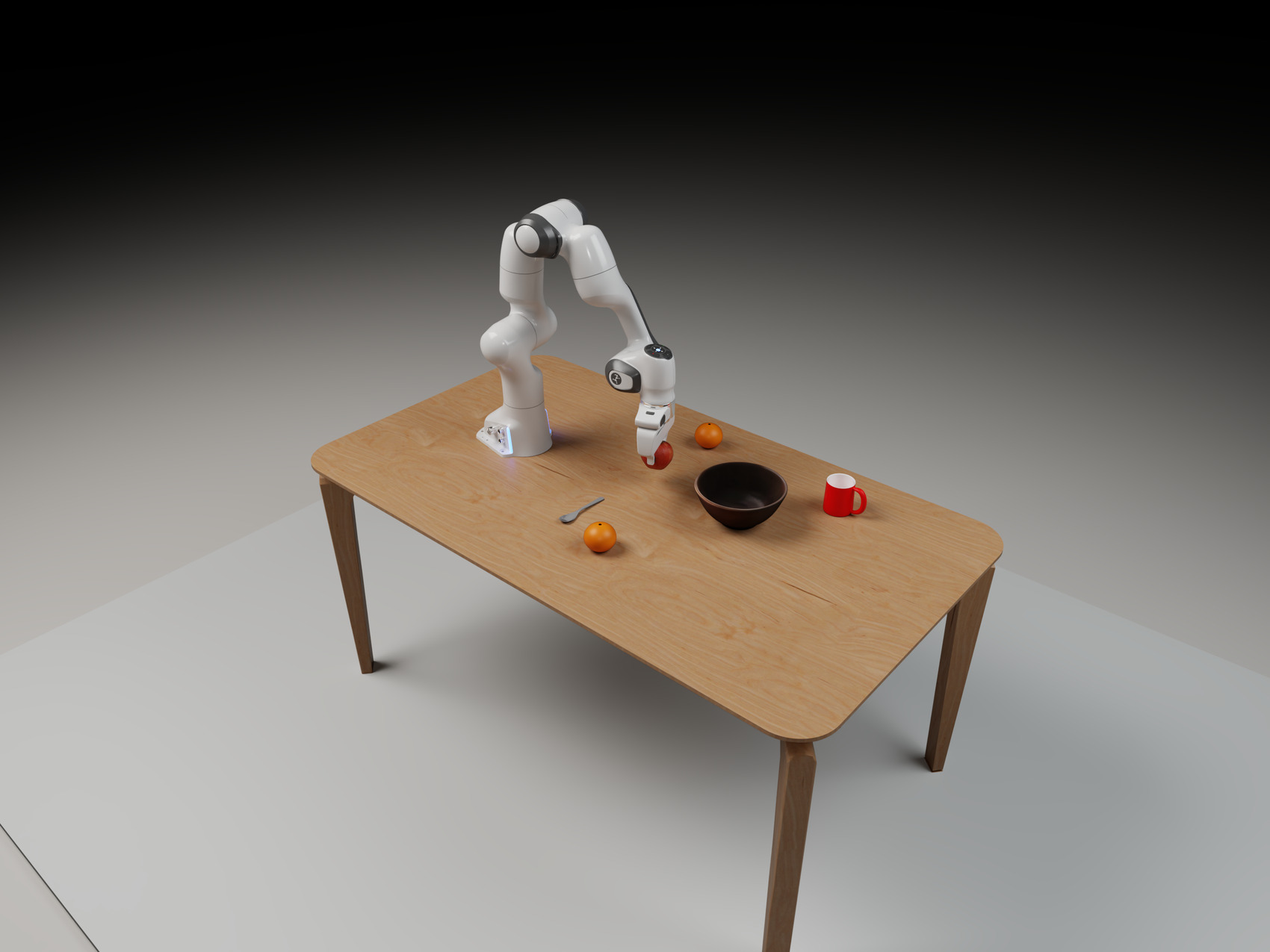} & \tOneImage{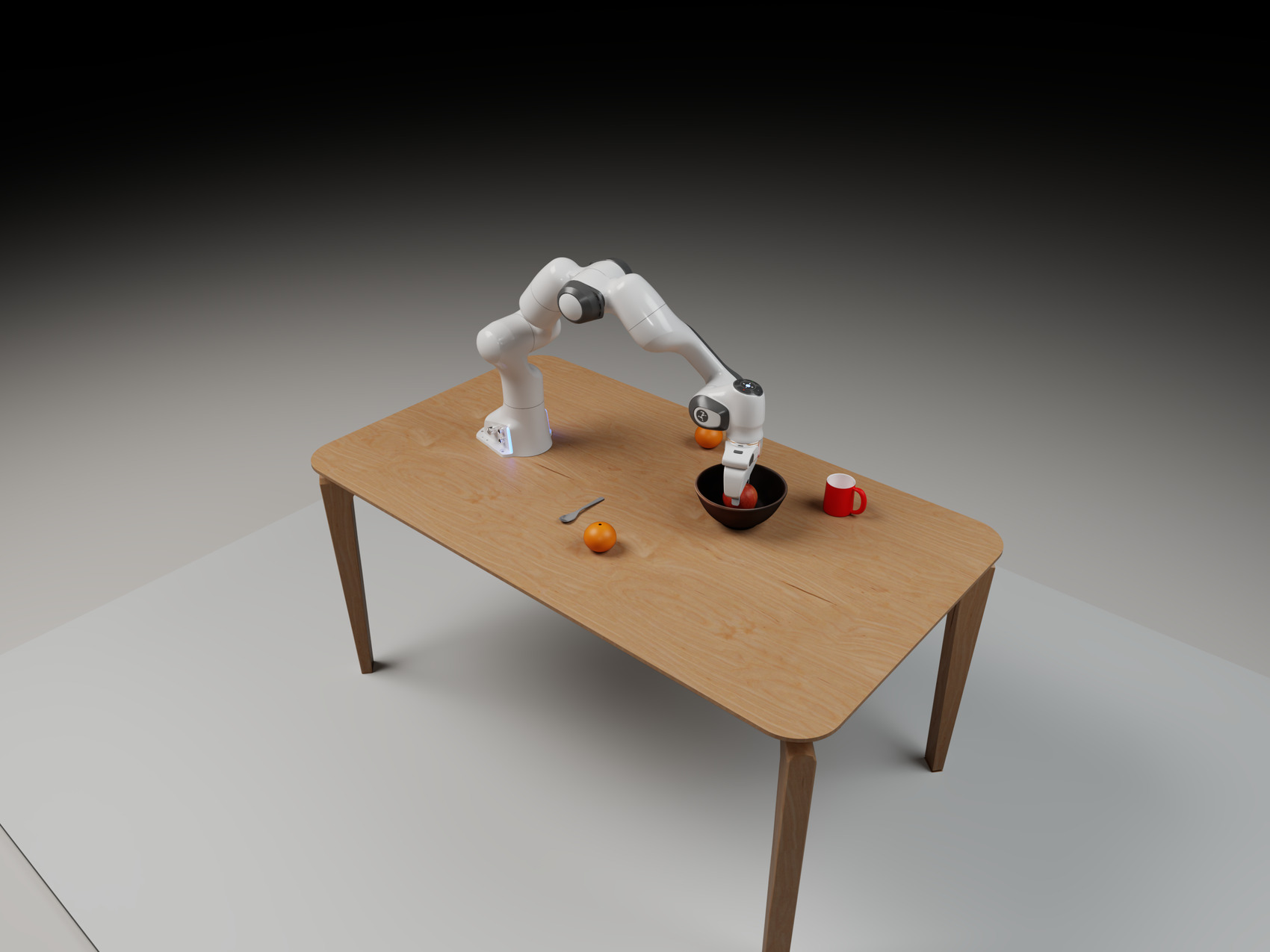} & \tOneImage{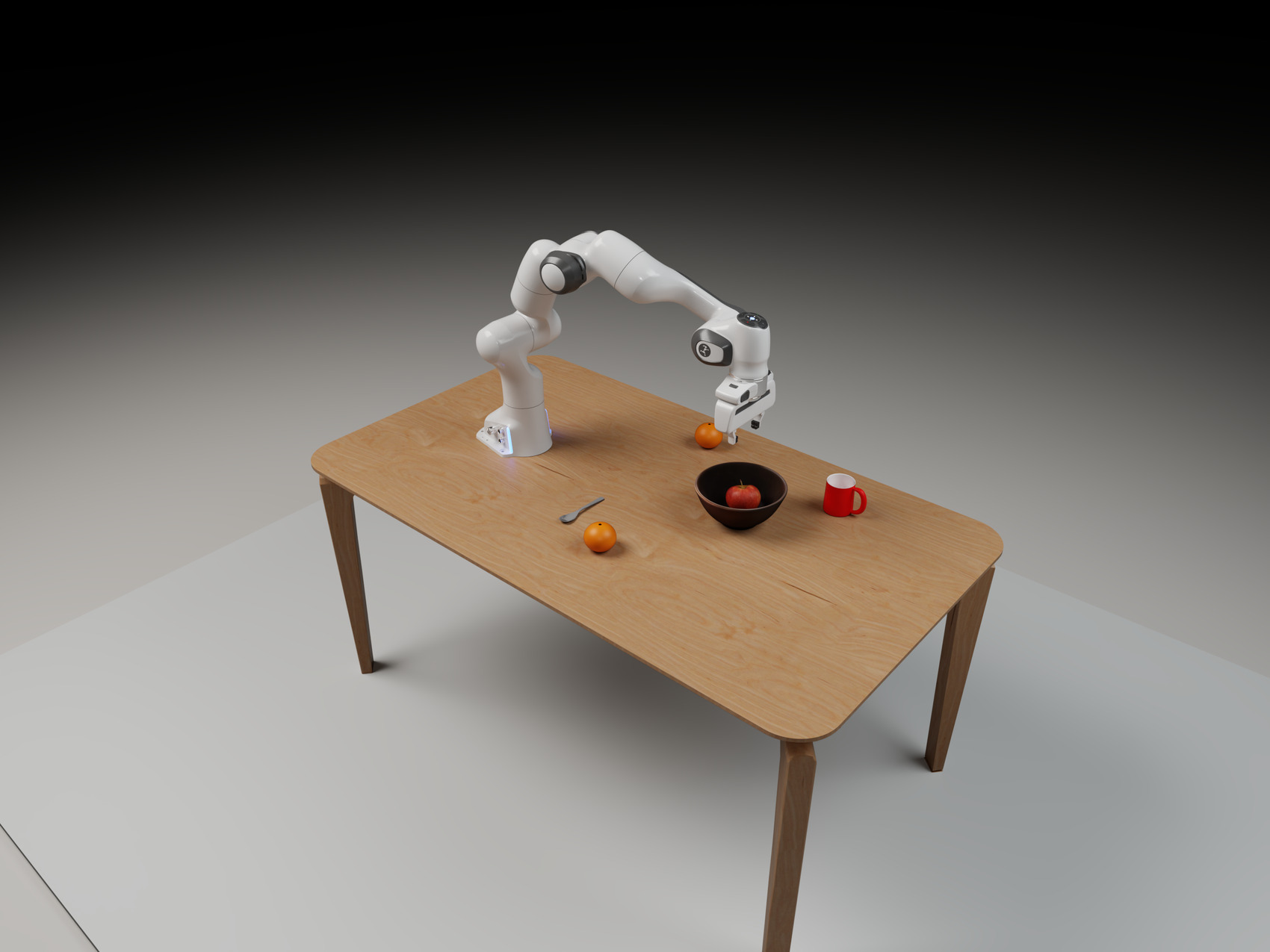} \\
\scriptsize 0.05\,s & \scriptsize 16.60\,s & \scriptsize 25.95\,s & \scriptsize 30.40\,s \\[5pt]
\multicolumn{4}{@{}l@{}}{\small\textbf{Fetch} \enspace Wheeled mobile manipulator} \\[2pt]
\tOneImage{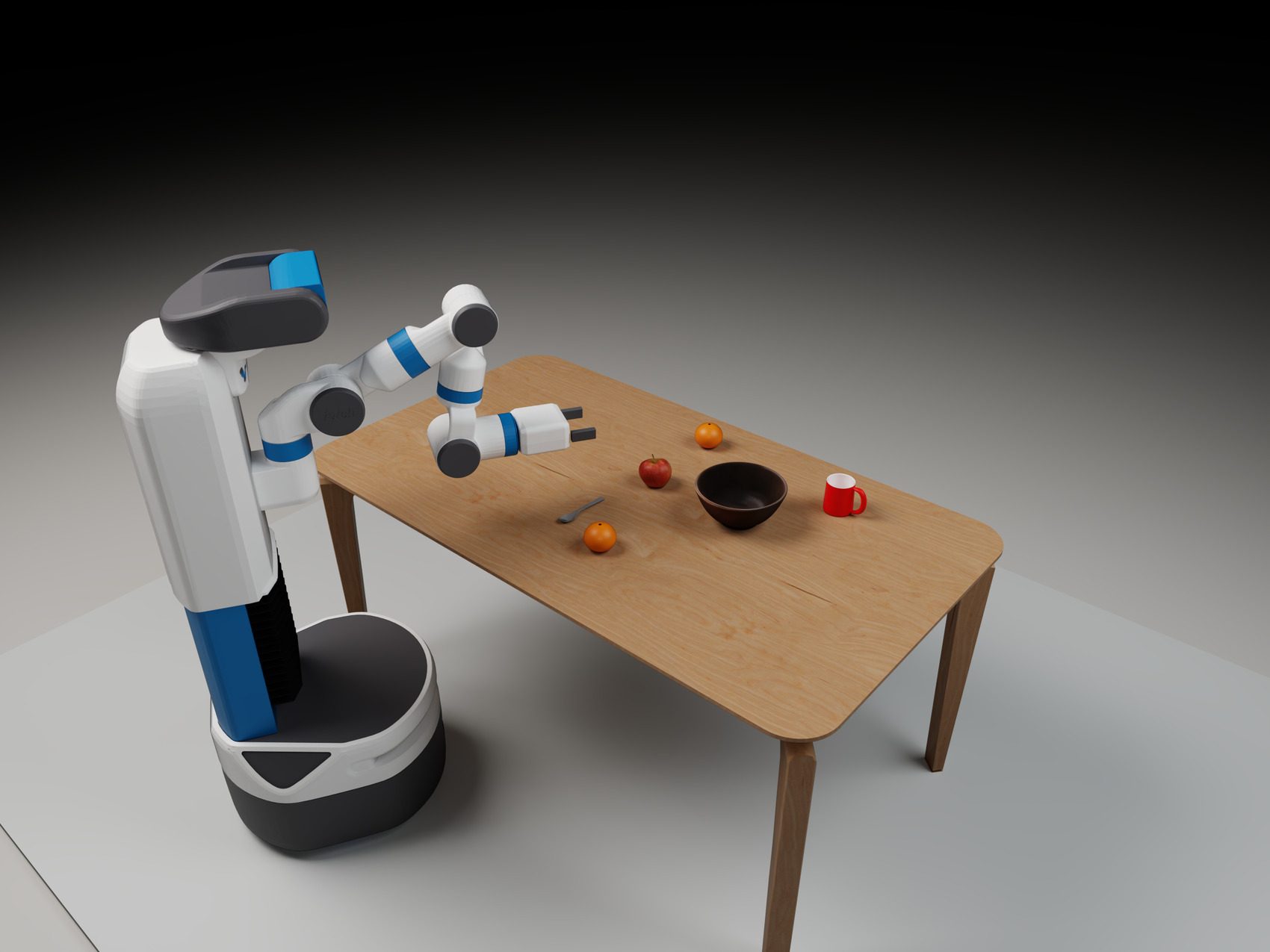} & \tOneImage{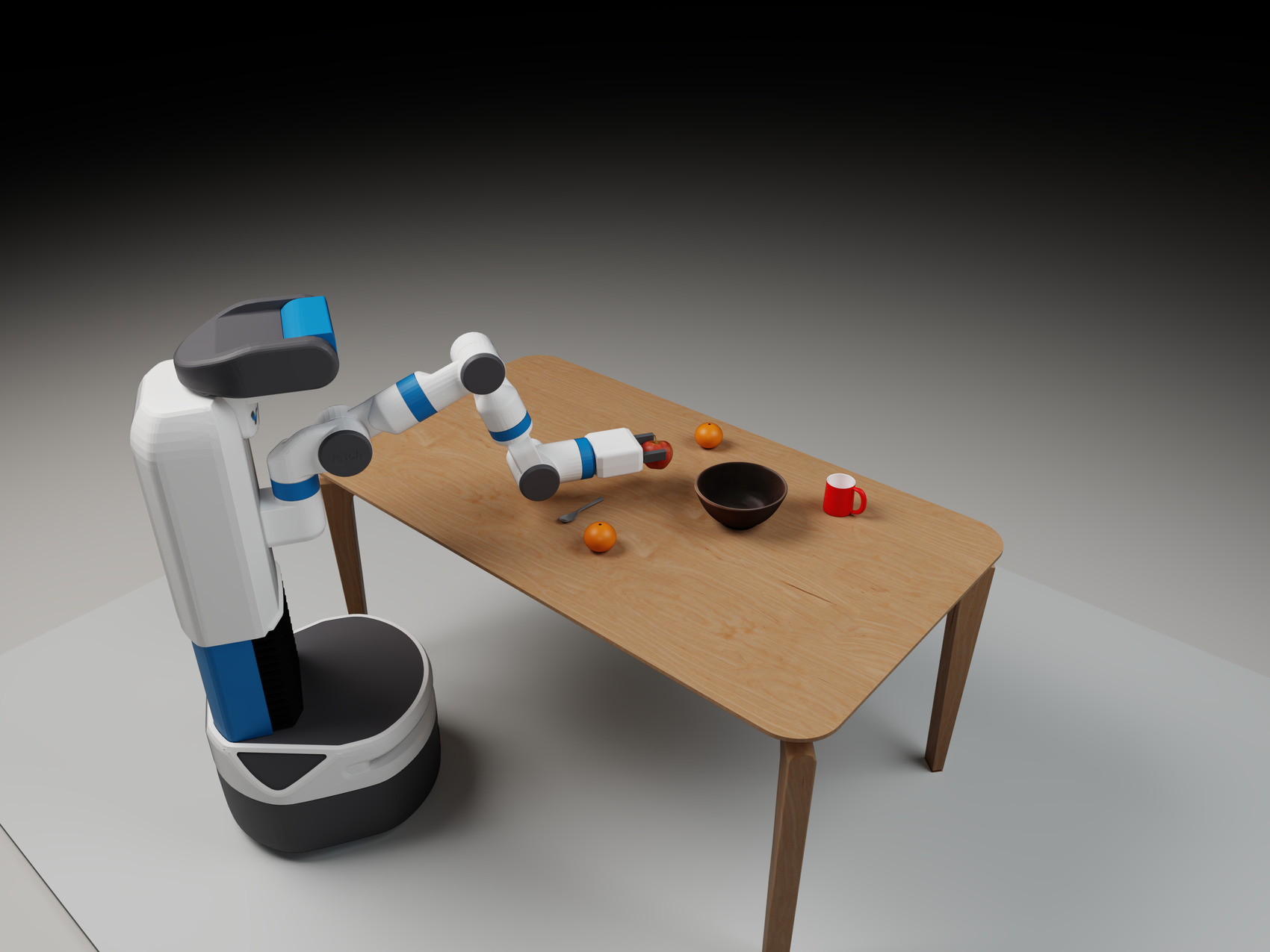} & \tOneImage{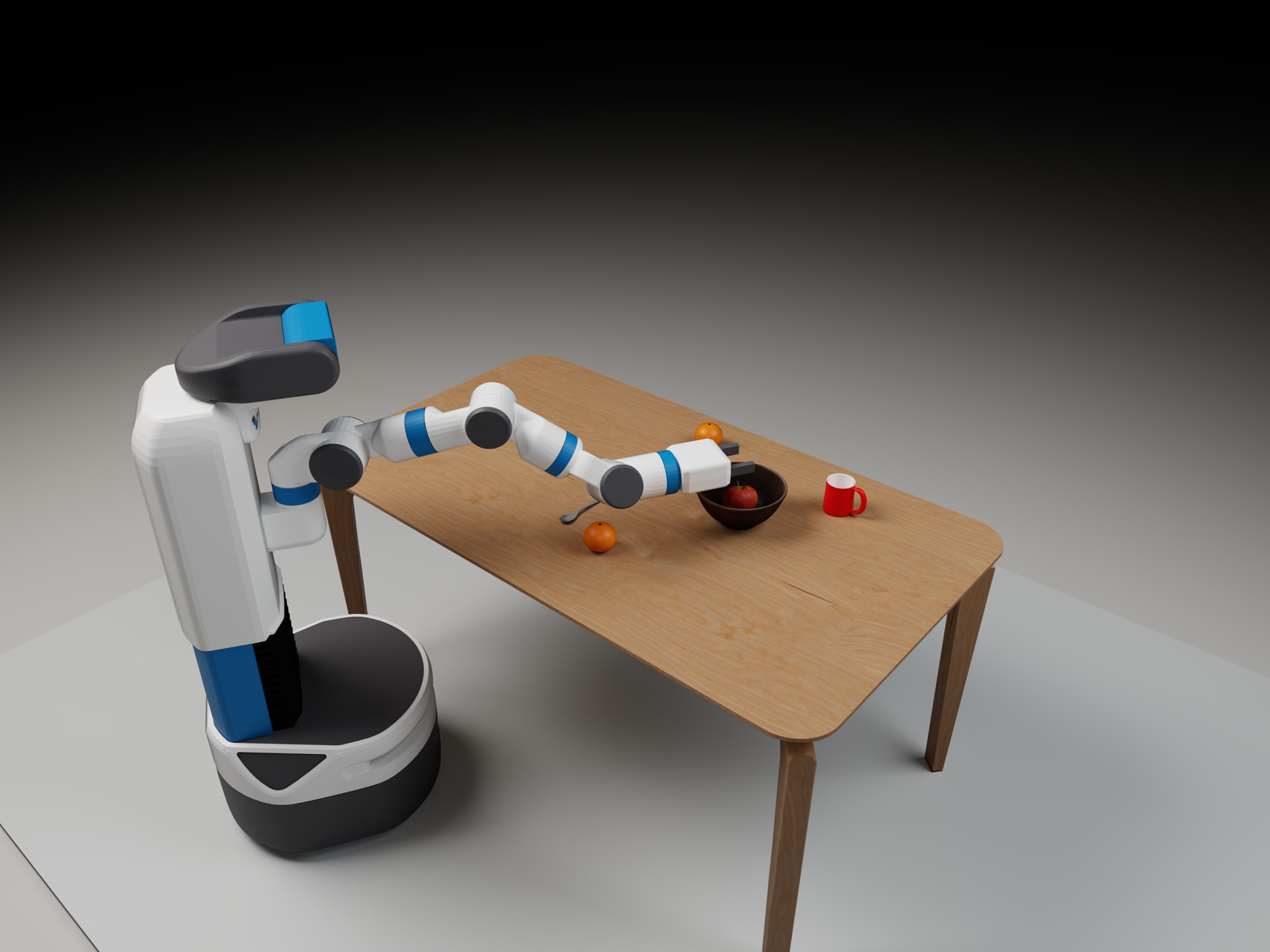} & \tOneImage{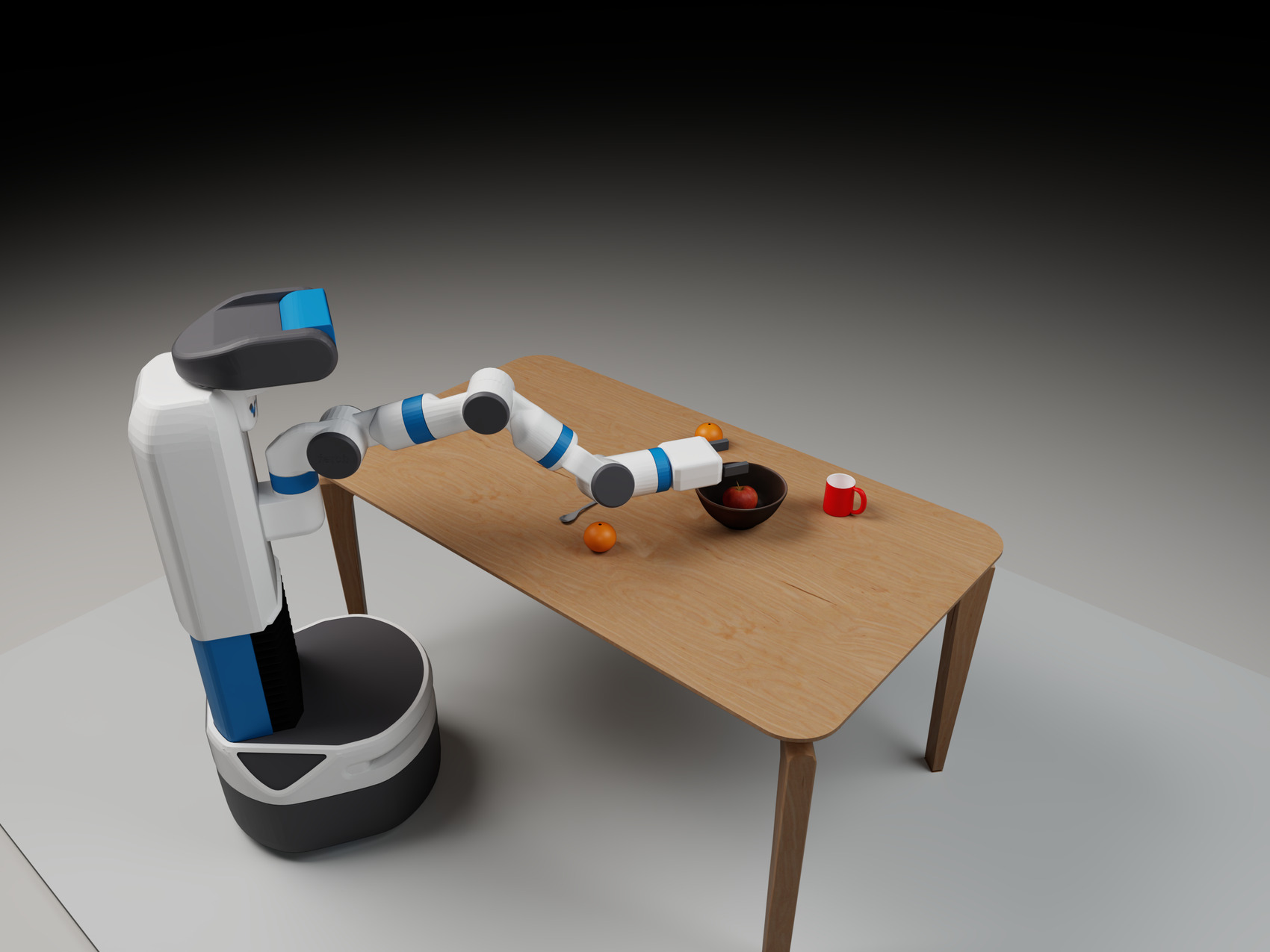} \\
\scriptsize 0.05\,s & \scriptsize 13.45\,s & \scriptsize 20.90\,s & \scriptsize 21.90\,s \\[5pt]
\multicolumn{4}{@{}l@{}}{\small\textbf{ARX-X5} \enspace Dual arm} \\[2pt]
\tOneImage{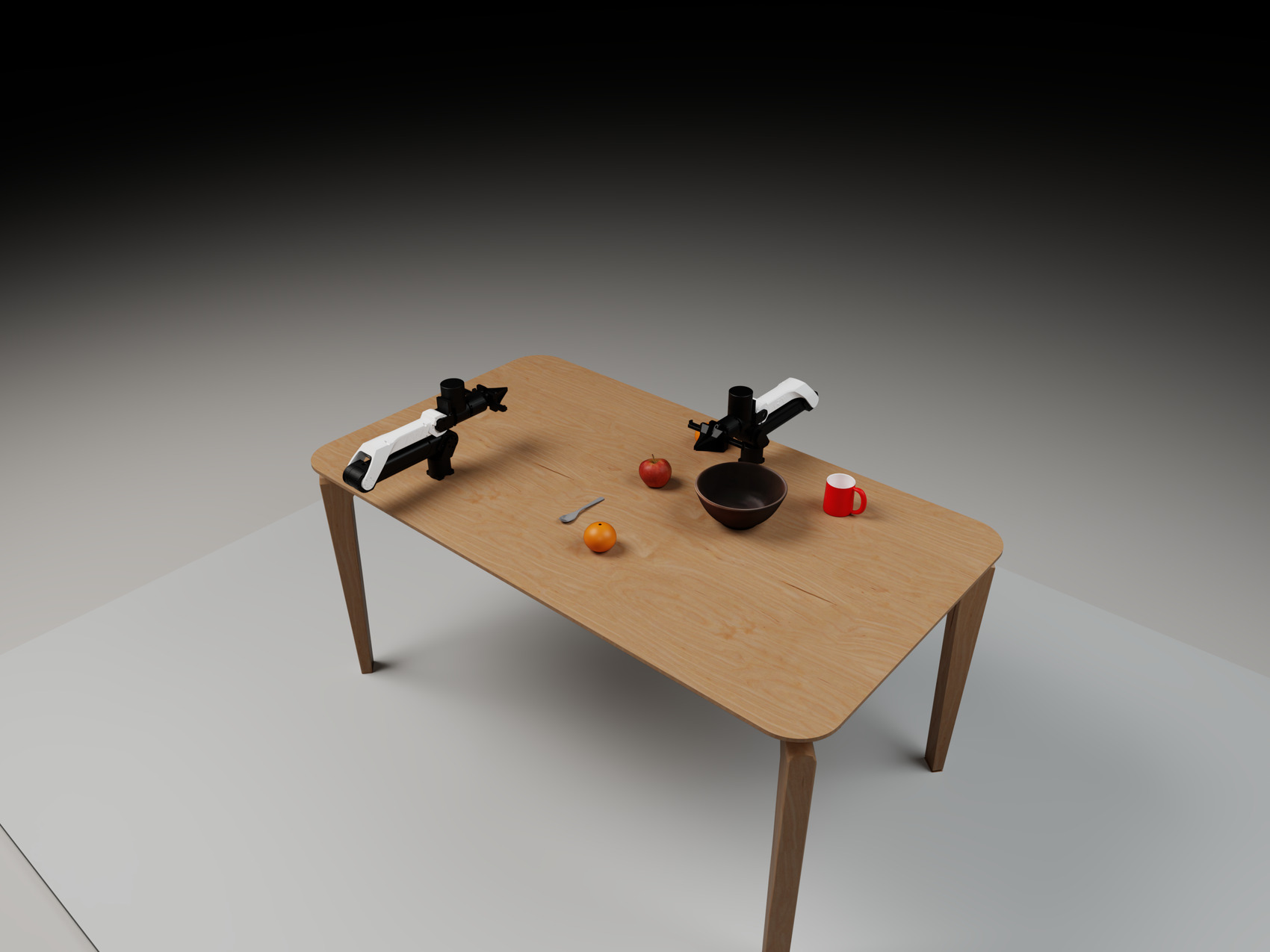} & \tOneImage{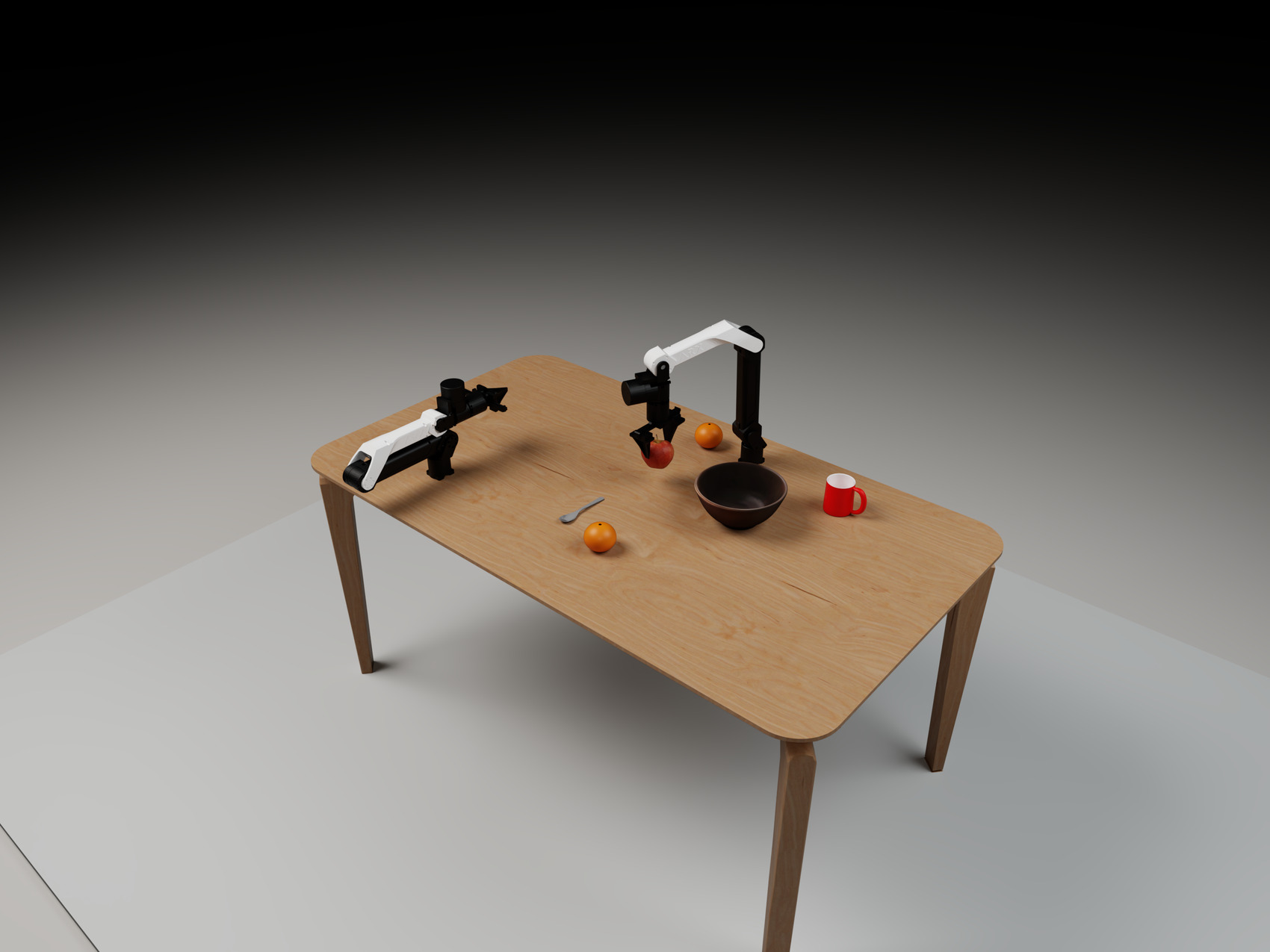} & \tOneImage{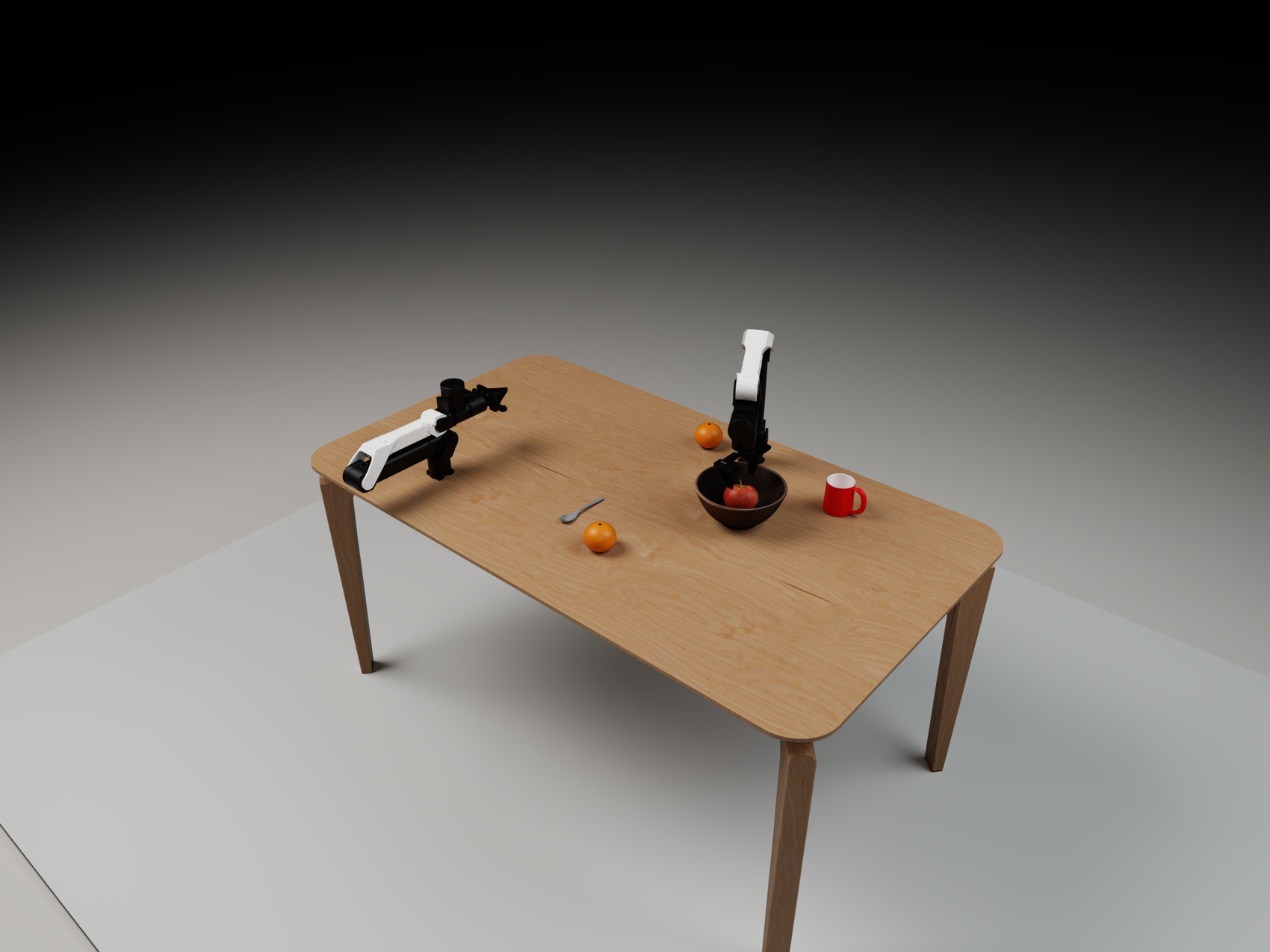} & \tOneImage{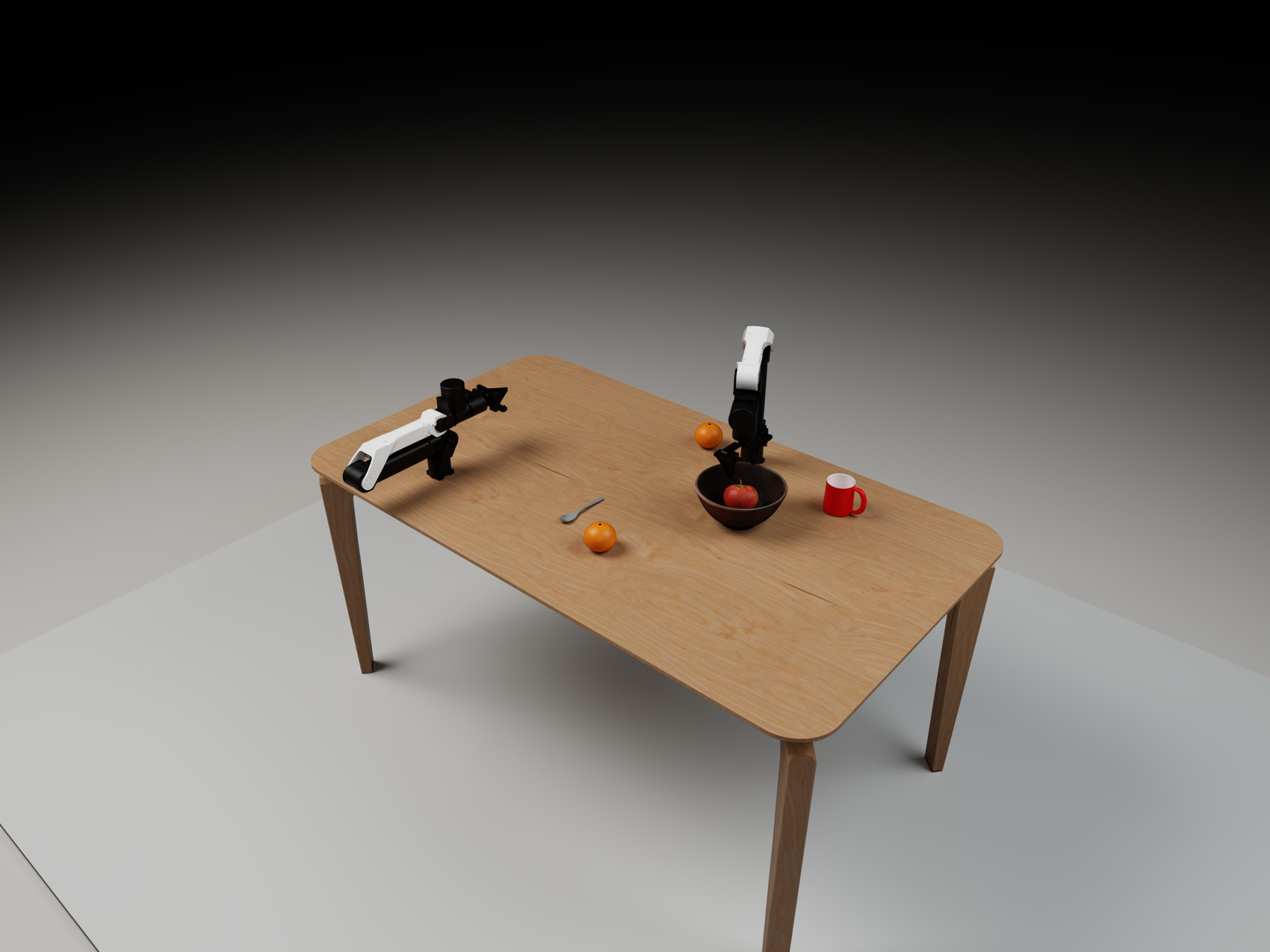} \\
\scriptsize 0.05\,s & \scriptsize 10.75\,s & \scriptsize 18.45\,s & \scriptsize 19.05\,s \\[5pt]
\multicolumn{4}{@{}l@{}}{\small\textbf{I2RT YAM} \enspace Dual arm (onboarded profile)} \\[2pt]
\tOneImage{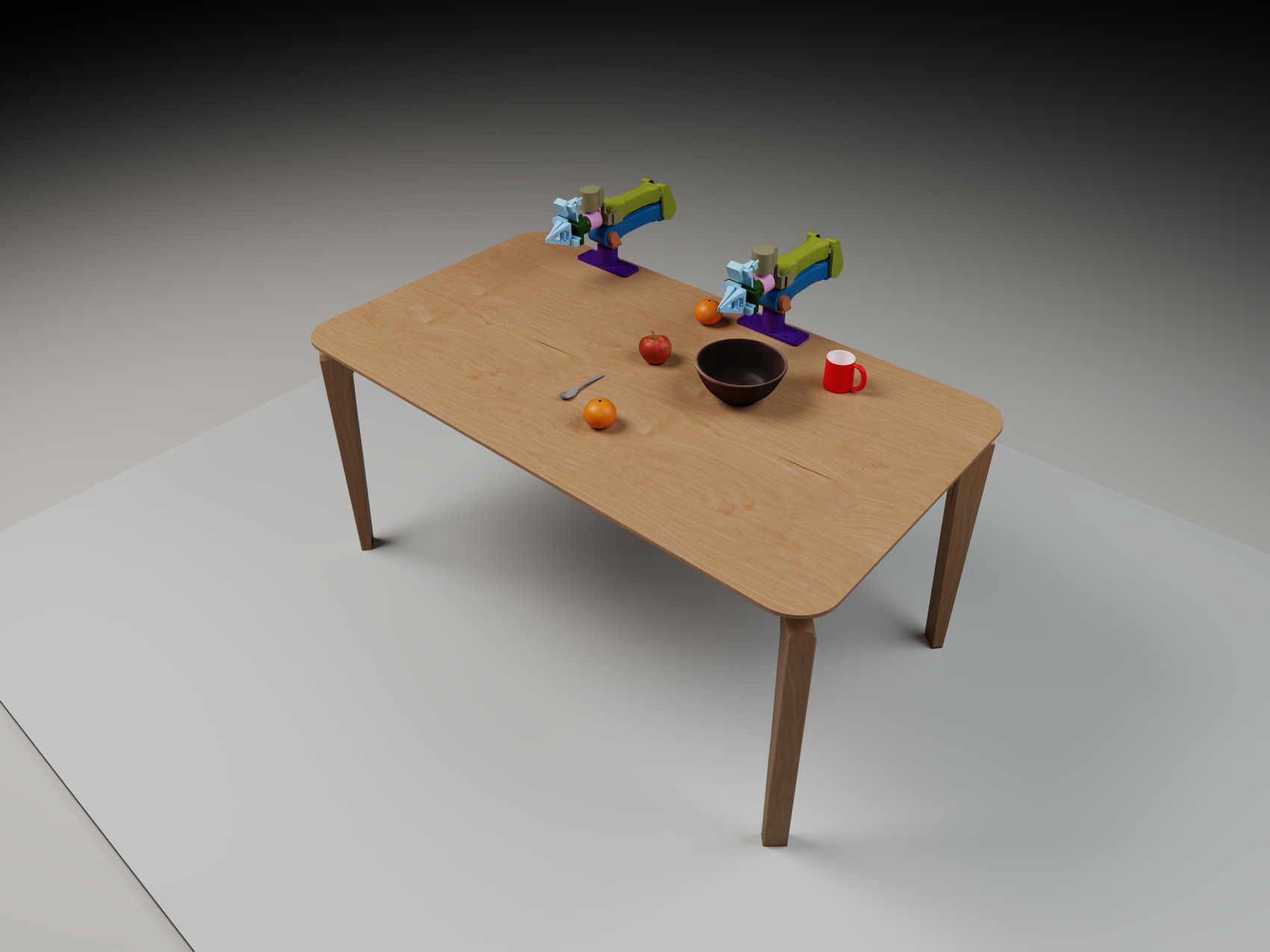} & \tOneImage{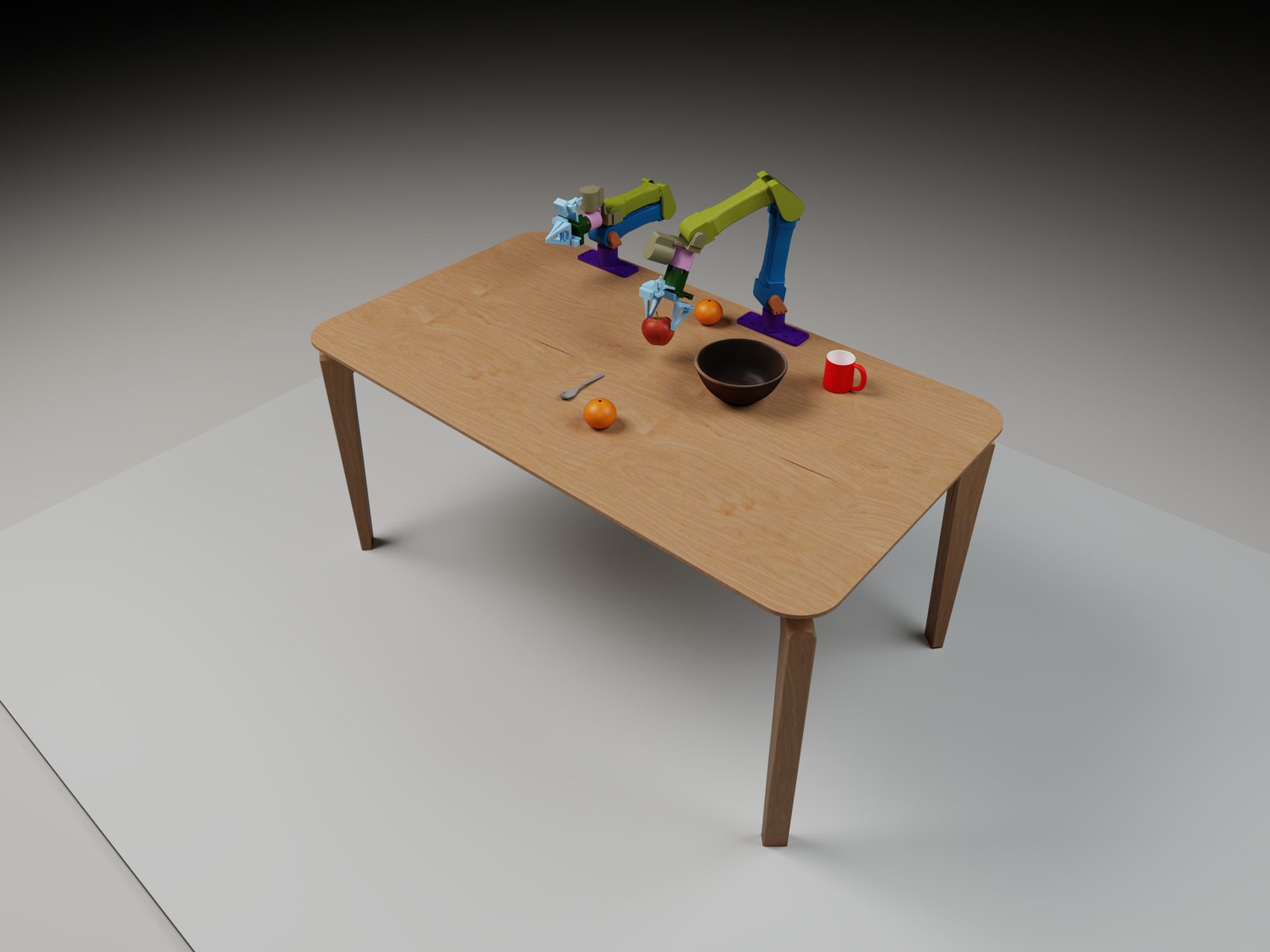} & \tOneImage{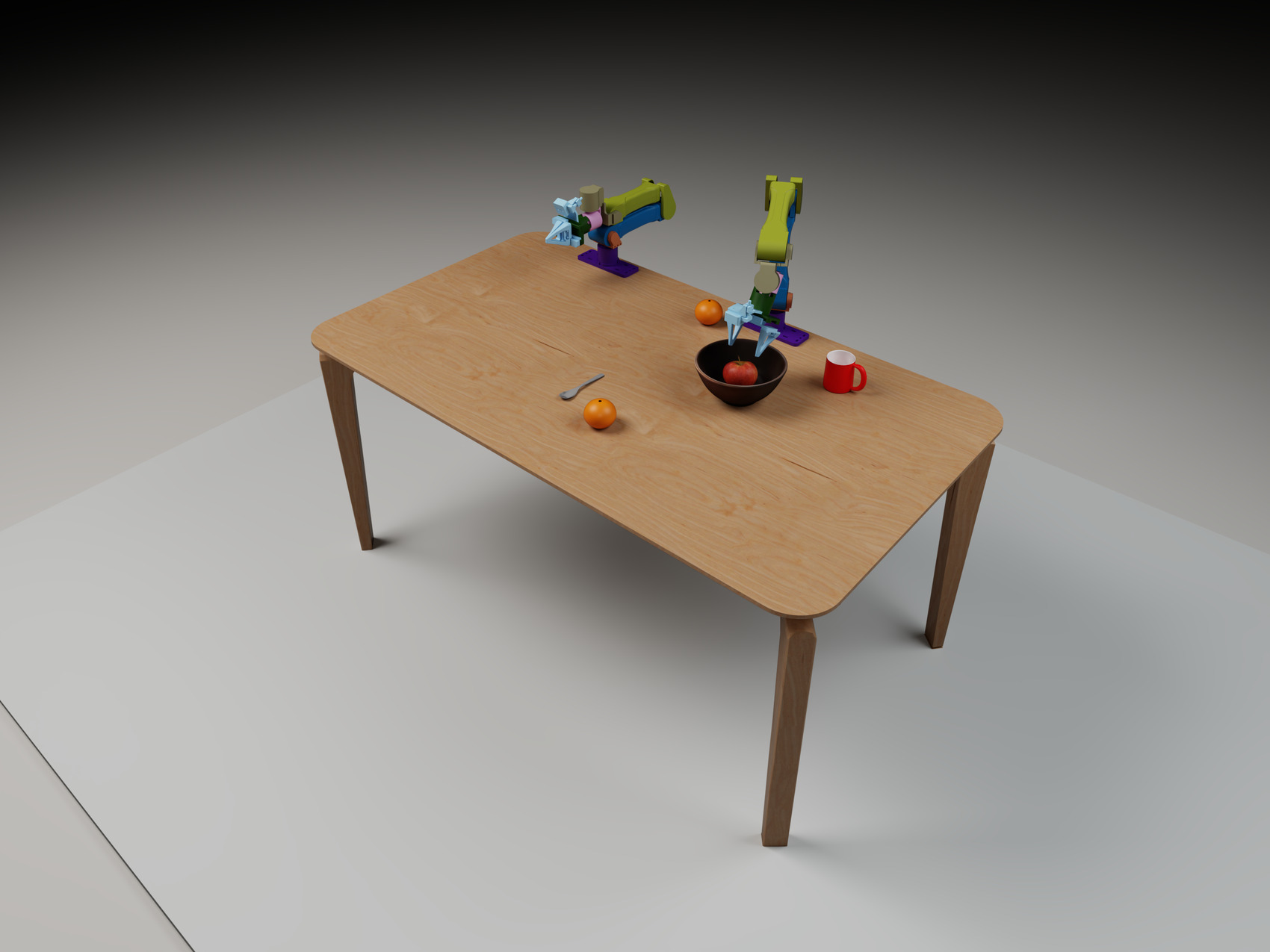} & \tOneImage{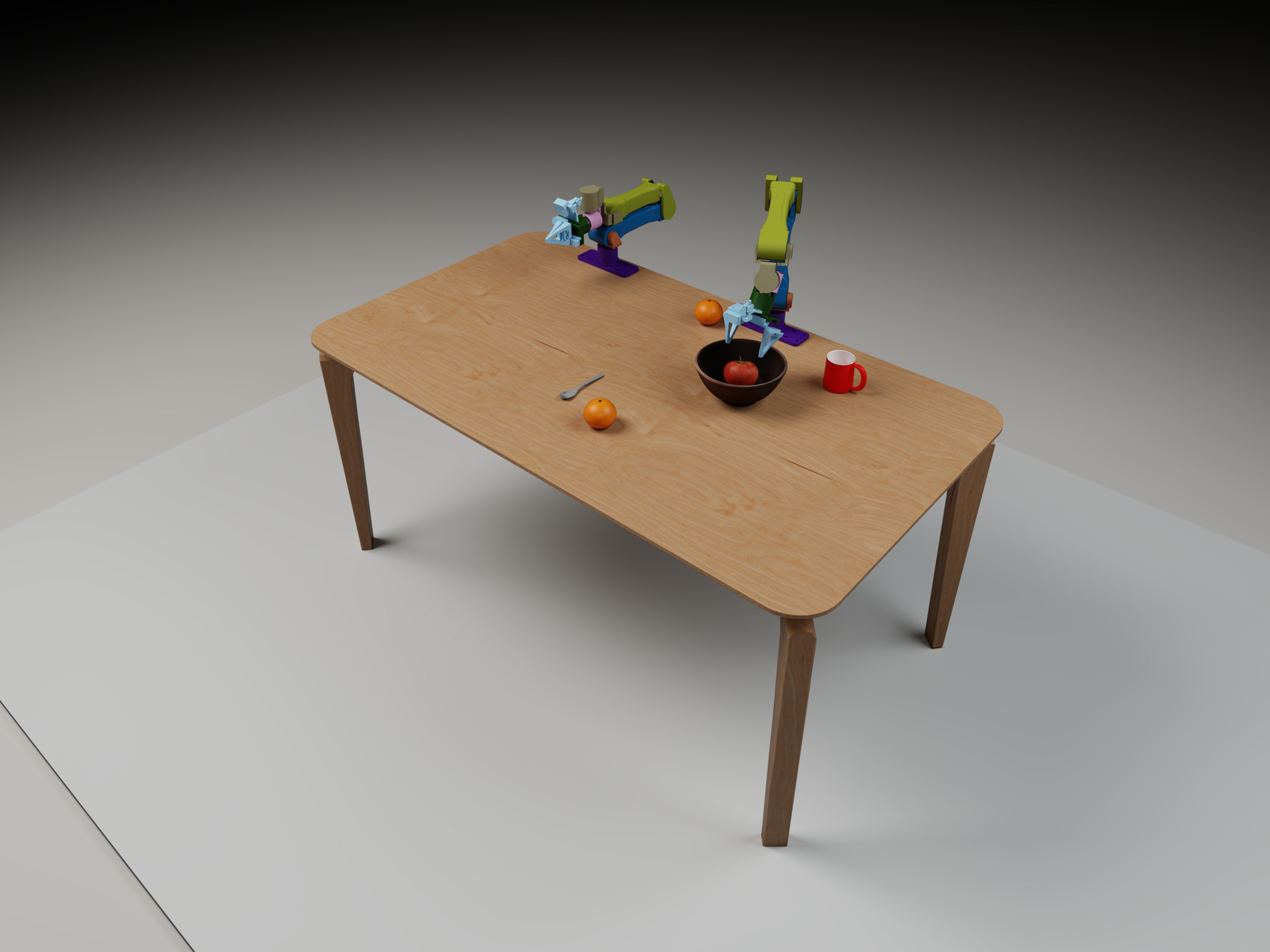} \\
\scriptsize 0.05\,s & \scriptsize 10.65\,s & \scriptsize 16.75\,s & \scriptsize 17.35\,s \\[5pt]
\end{tabular}
\caption{\textbf{Cross-embodiment apple placement on the shared \texttt{T01} tabletop scene.} The four columns show recorded states from the start, lift, bowl placement, and final release across four distinct robot embodiments (including the onboarded \texttt{I2RT YAM} profile). Each completed episode satisfies the frozen \texttt{T01} rubric for physical bowl support, stability, finger clearance, and preservation of the other objects. All illustrated states use the same viewpoint and $2560\times1920$ path tracing at 1024 samples per pixel.}
\label{fig:supp-original-t01-embodiments}
\end{figure*}

\subsection{Additional Non-Overlapping Rollouts}
\label{sec:supp-rollouts}
\begin{figure*}[p]
\centering
\setlength{\tabcolsep}{2pt}
\begin{tabular}{@{}cccc@{}}
\multicolumn{4}{@{}l@{}}{\small\textbf{Fetch: clear a living-room passage (\texttt{R10}); success, score 1.0}} \\[2pt]
\includegraphics[width=.24\textwidth]{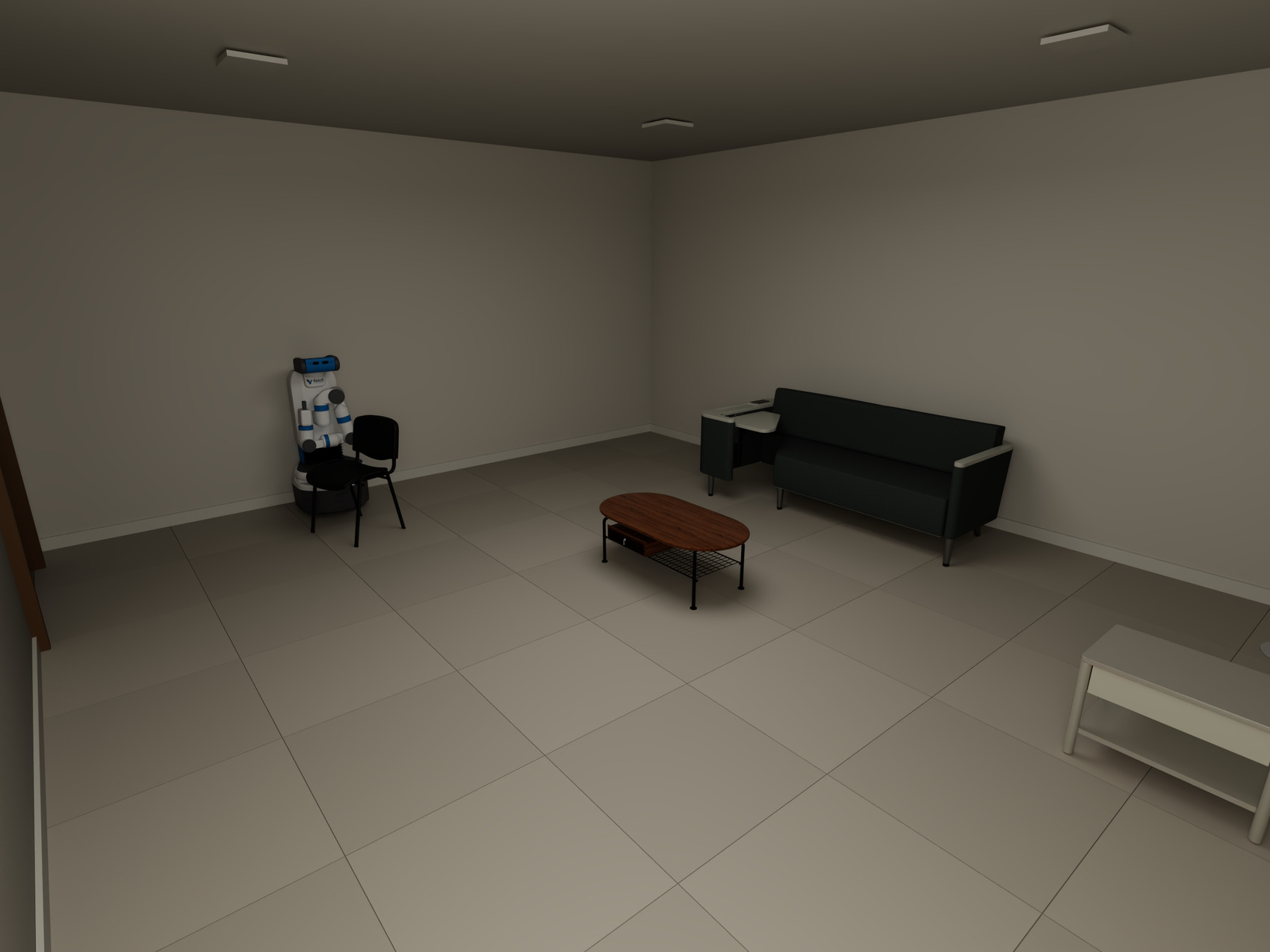} &
\includegraphics[width=.24\textwidth]{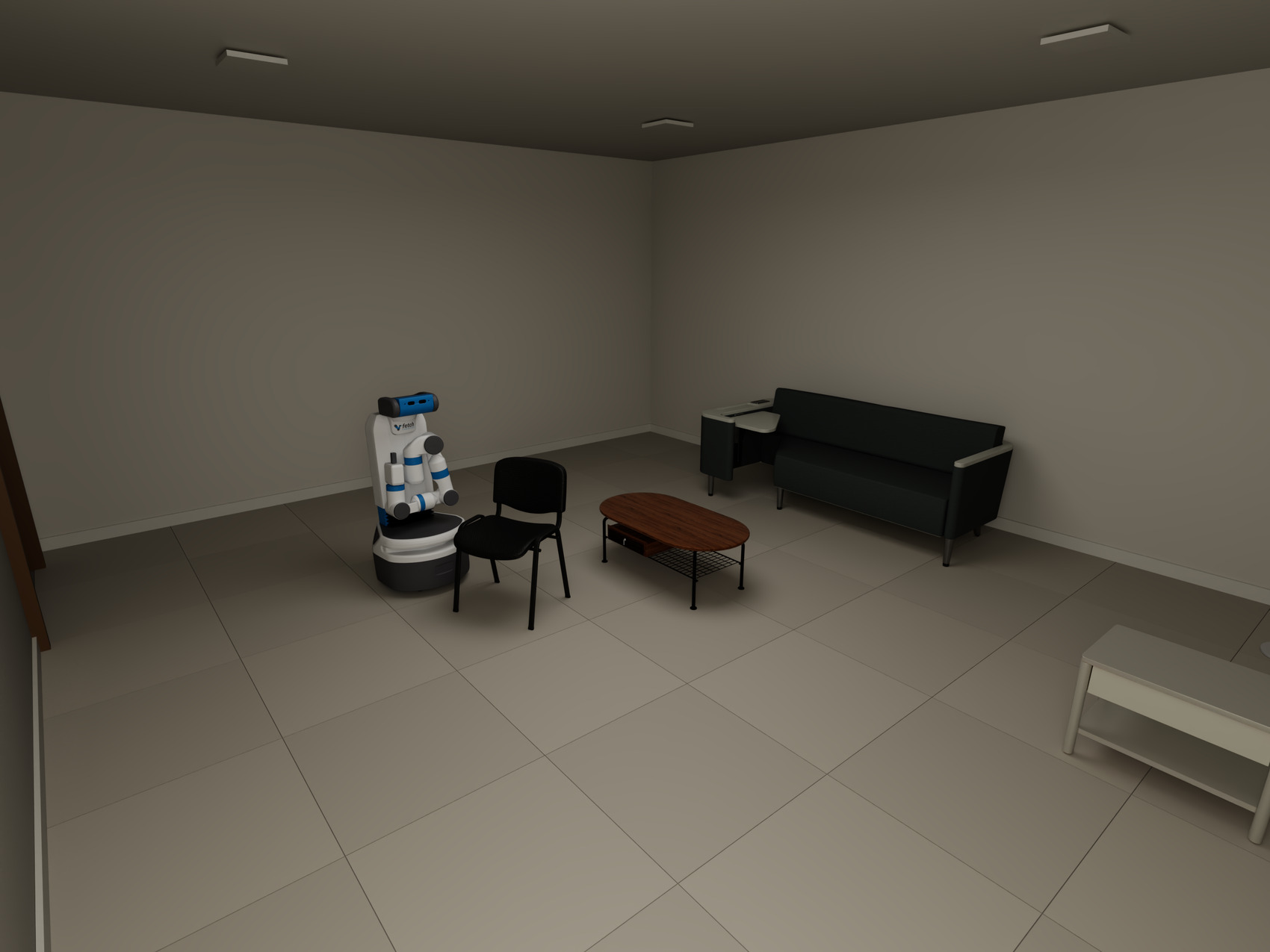} &
\includegraphics[width=.24\textwidth]{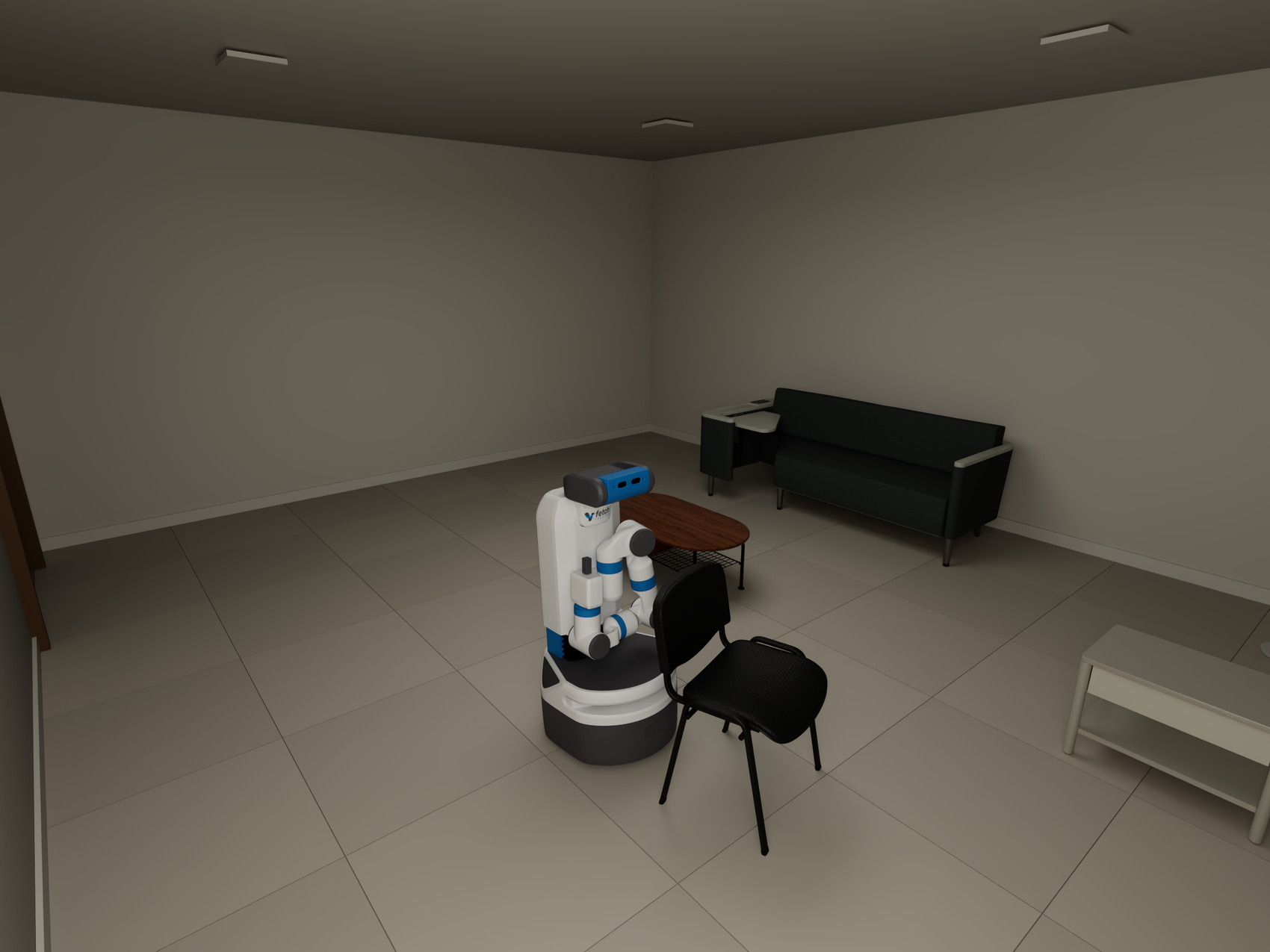} &
\includegraphics[width=.24\textwidth]{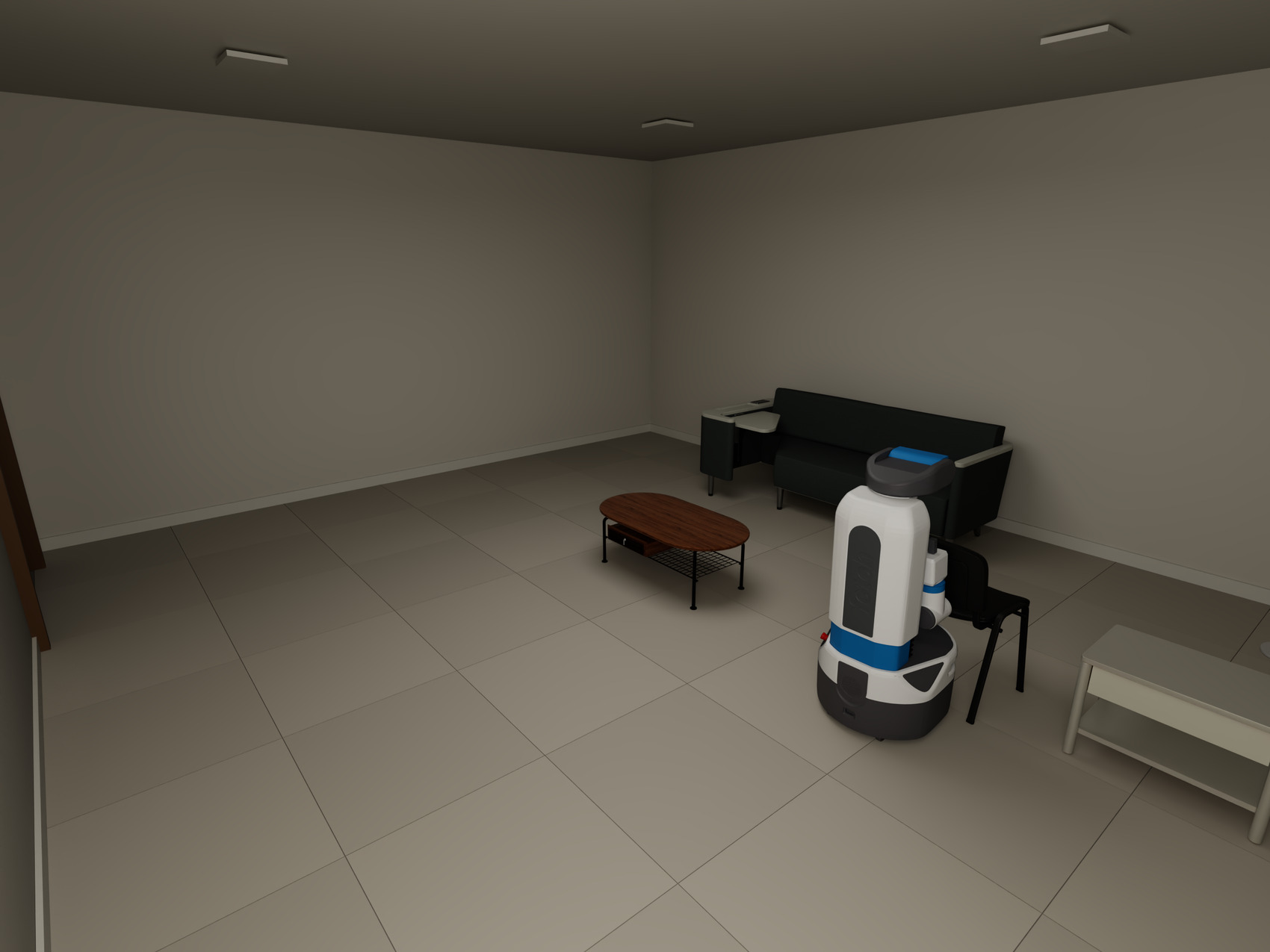} \\
\scriptsize $t=0.00$\,s & \scriptsize $t=25.00$\,s & \scriptsize $t=50.00$\,s & \scriptsize $t=72.30$\,s \\[8pt]
\multicolumn{4}{@{}l@{}}{\small\textbf{AGIBOT Genie G2: kitchen grocery storage (\texttt{R01}); incomplete, score 0.0}} \\[2pt]
\includegraphics[width=.24\textwidth]{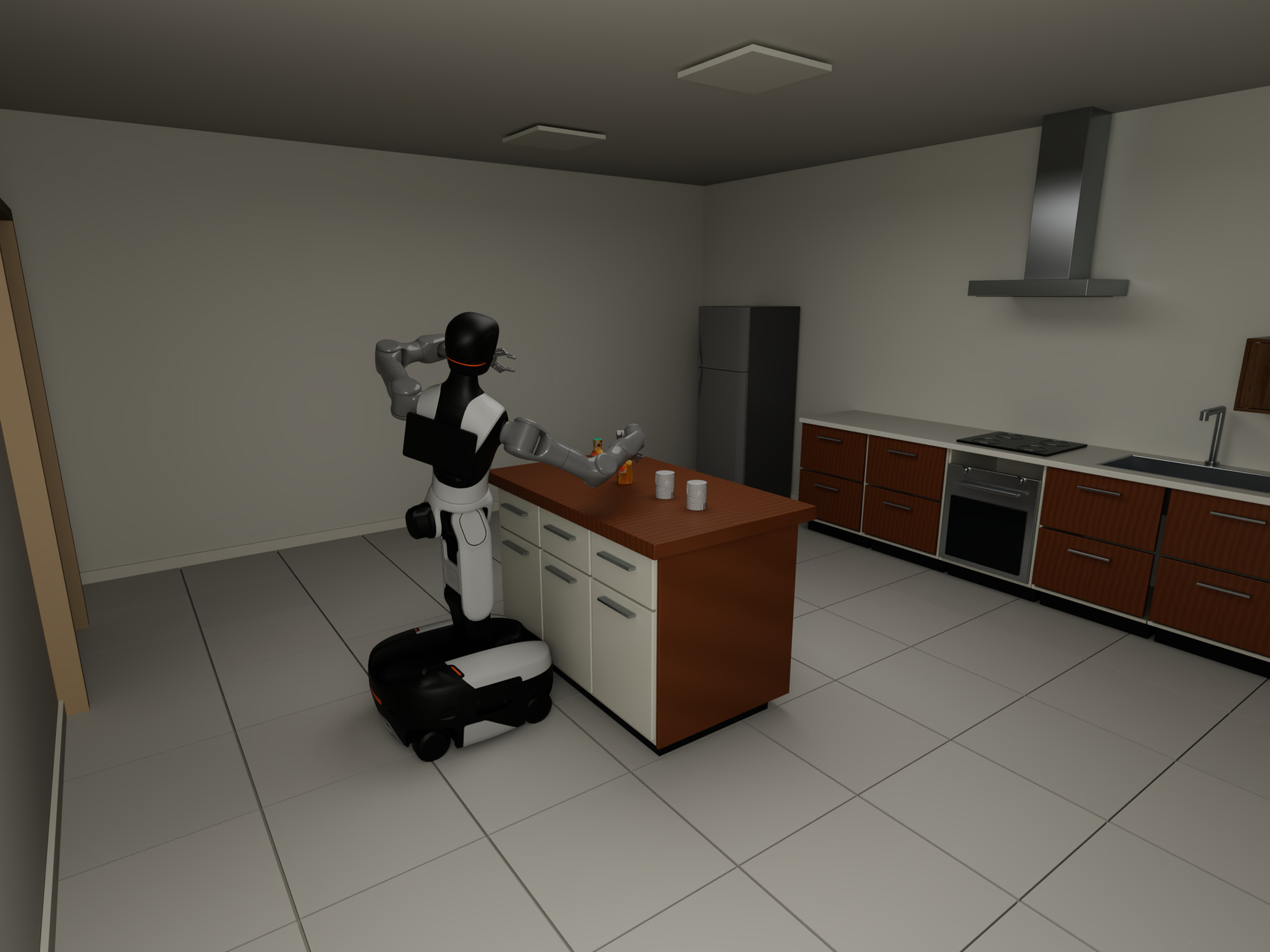} &
\includegraphics[width=.24\textwidth]{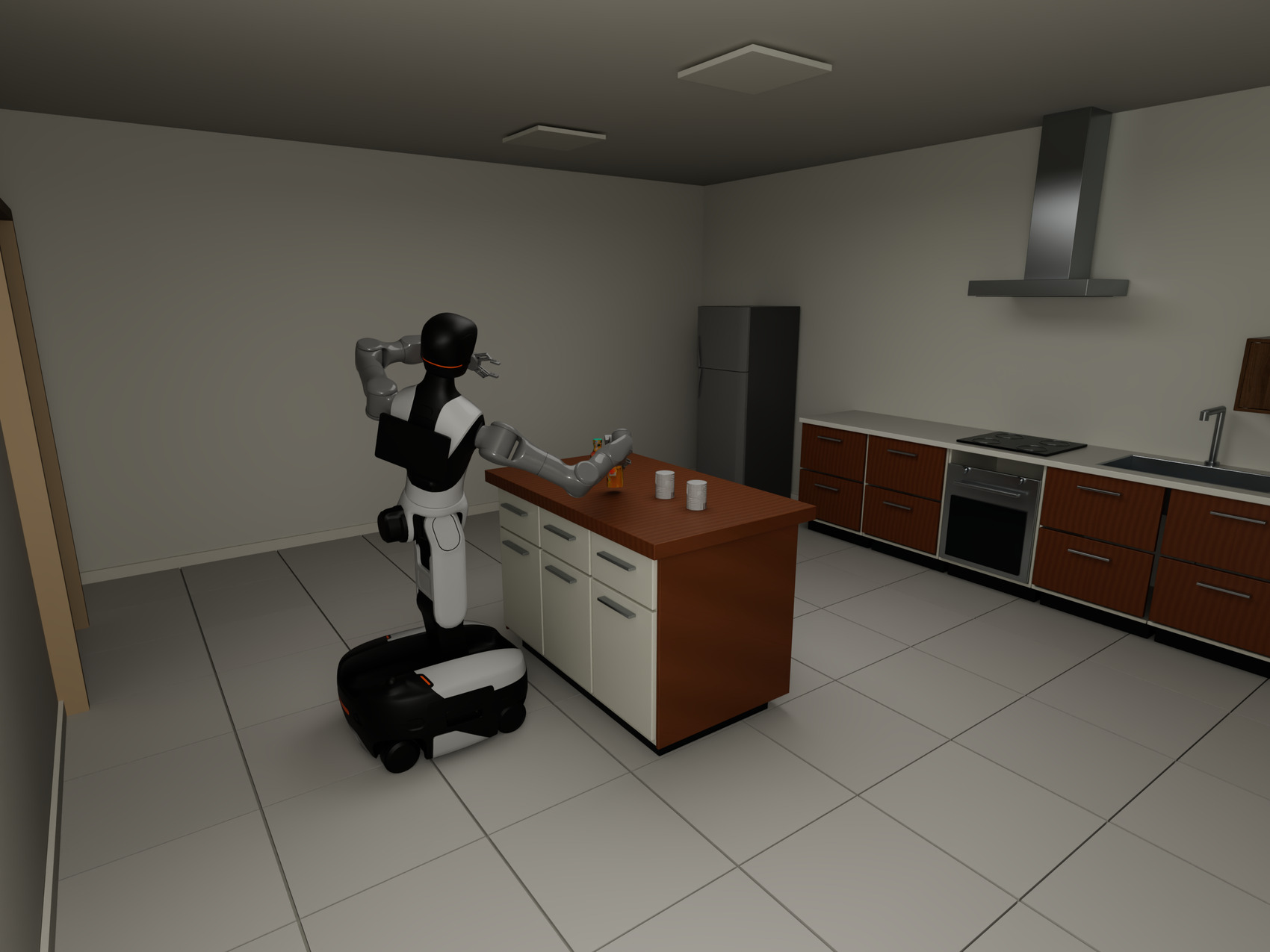} &
\includegraphics[width=.24\textwidth]{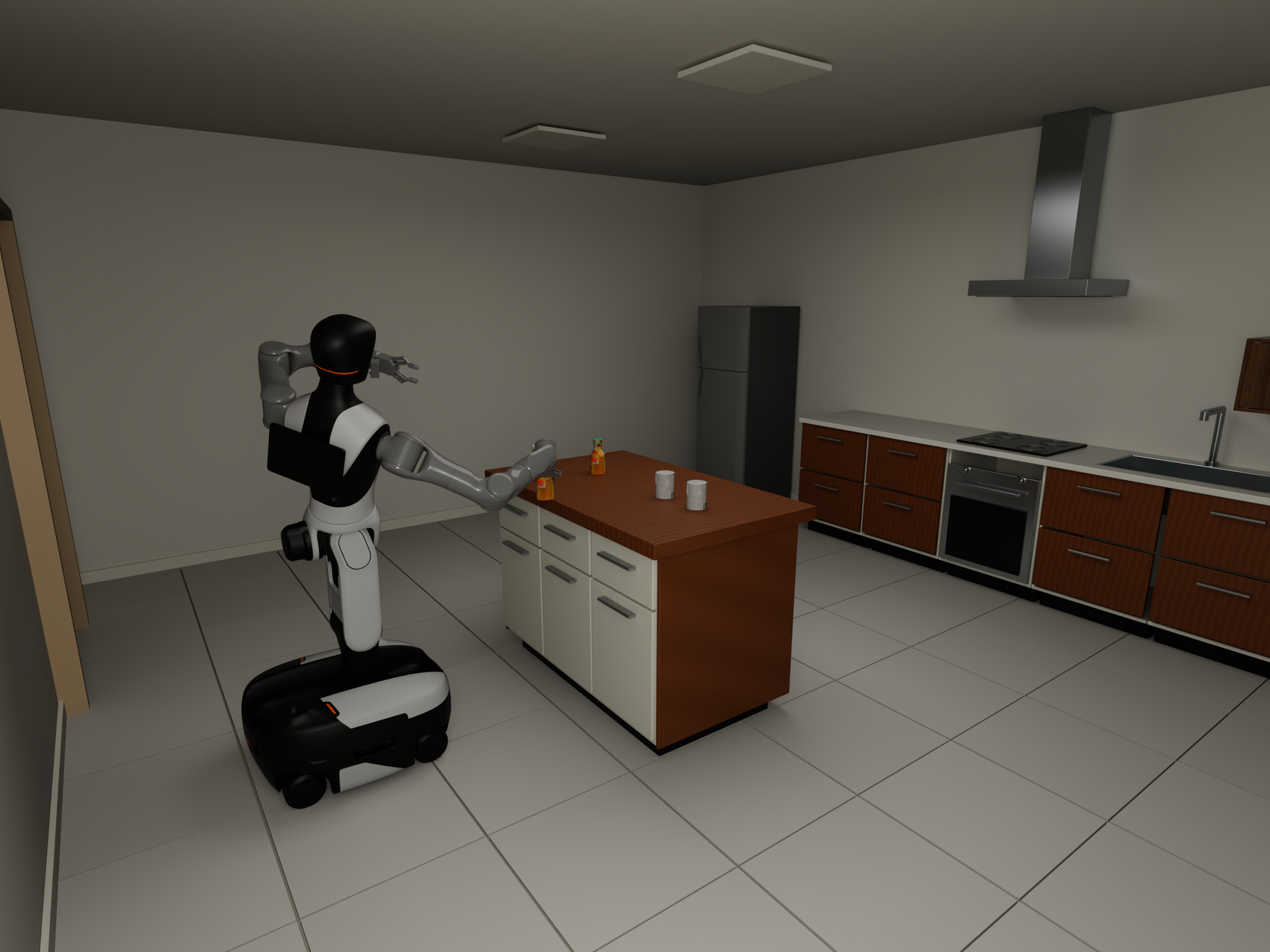} &
\includegraphics[width=.24\textwidth]{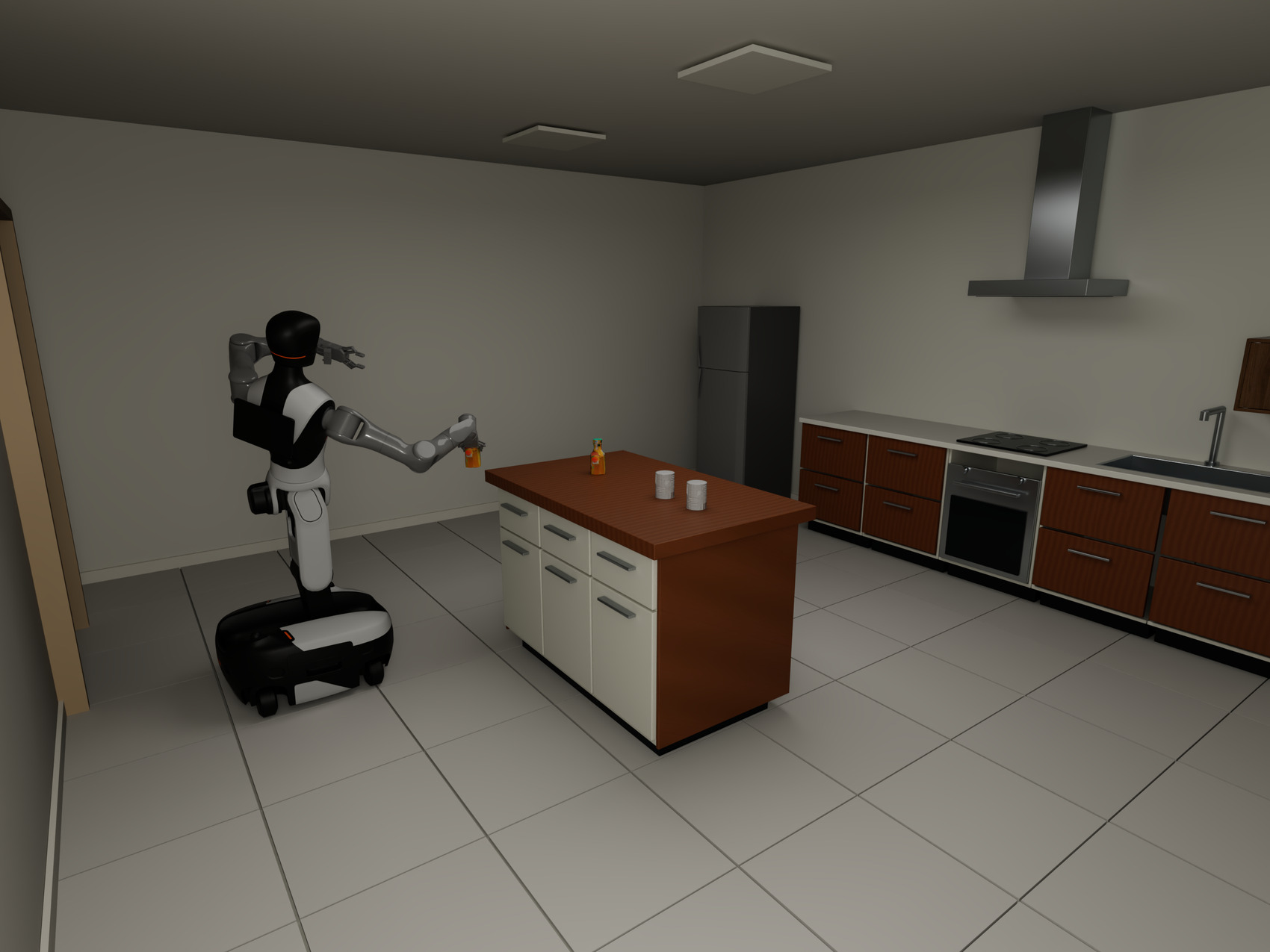} \\
\scriptsize $t=0.00$\,s & \scriptsize $t=15.00$\,s & \scriptsize $t=35.00$\,s & \scriptsize $t=68.80$\,s \\[8pt]
\multicolumn{4}{@{}l@{}}{\small\textbf{TIAGo++ (onboarded profile): cross-room recycling (\texttt{M05}); incomplete, score 0.5}} \\[2pt]
\includegraphics[width=.24\textwidth]{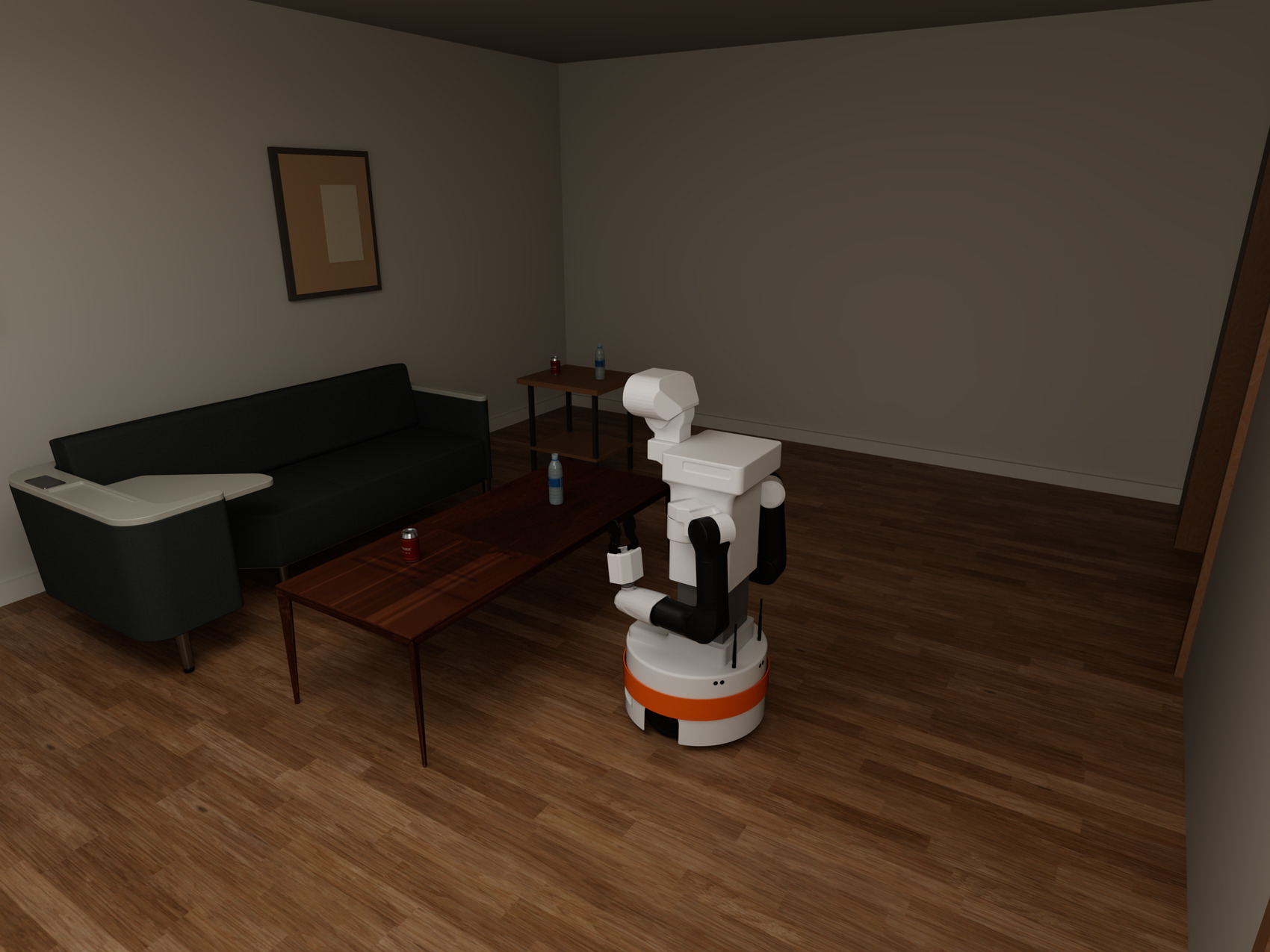} &
\includegraphics[width=.24\textwidth]{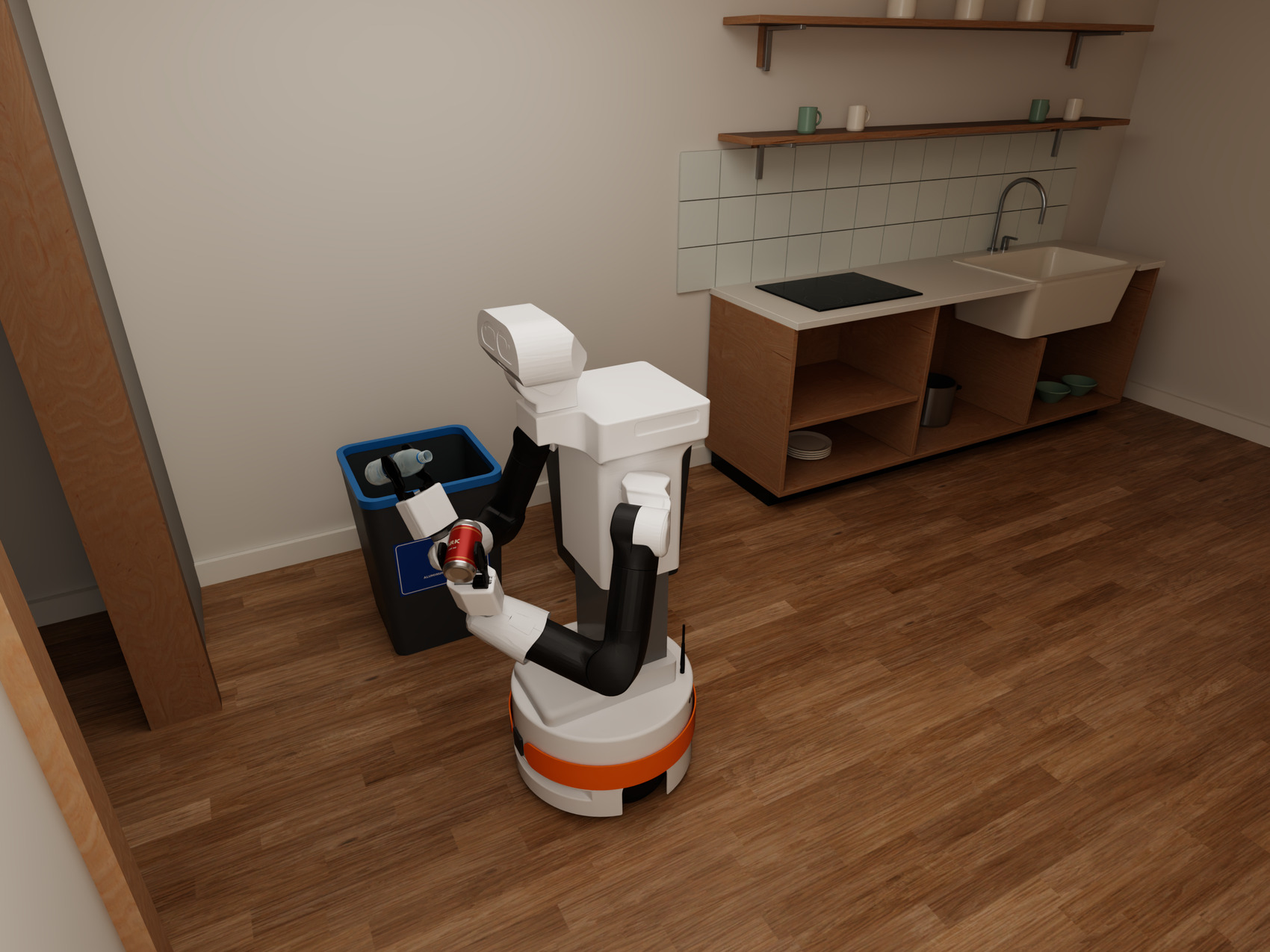} &
\includegraphics[width=.24\textwidth]{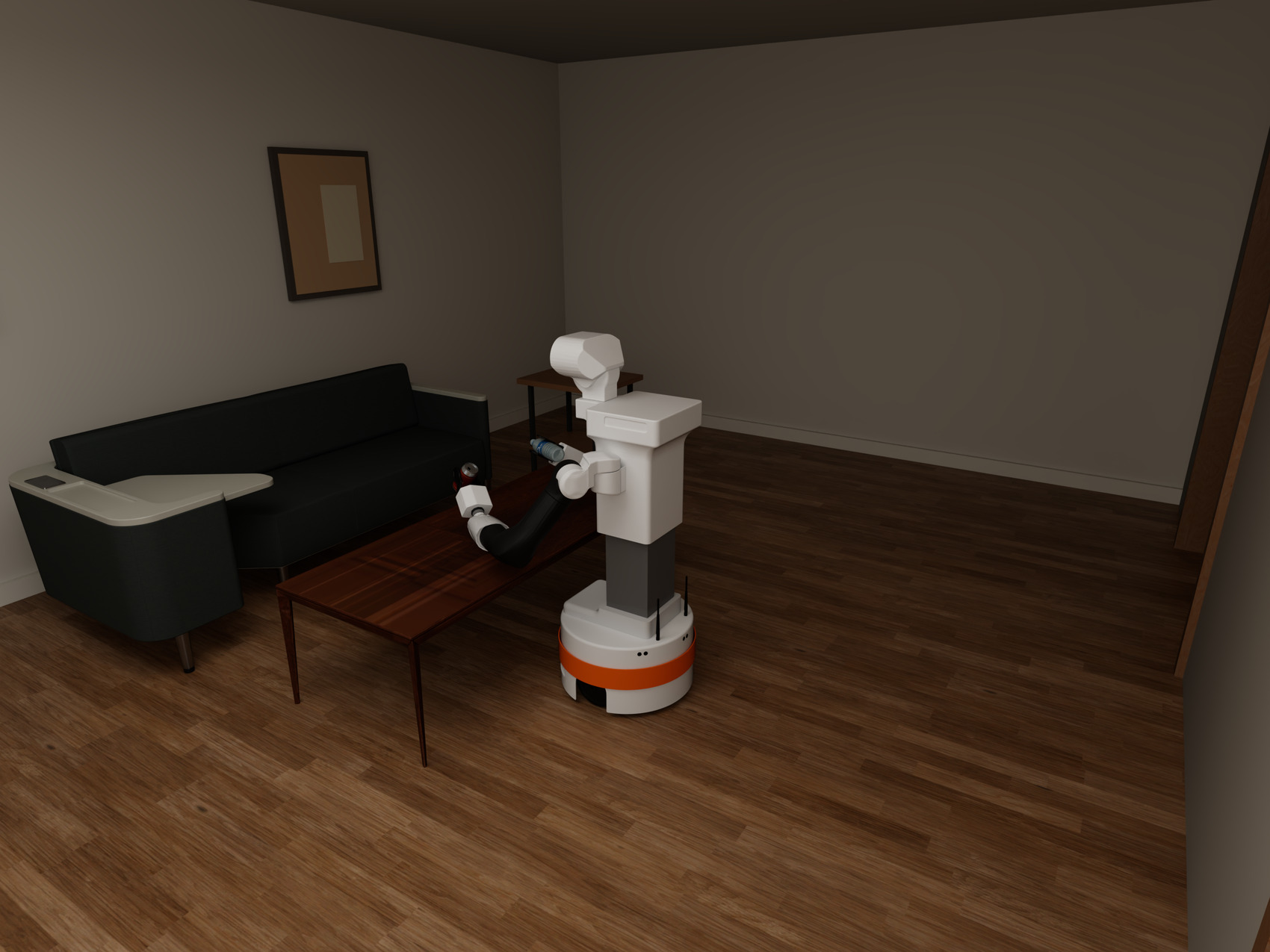} &
\includegraphics[width=.24\textwidth]{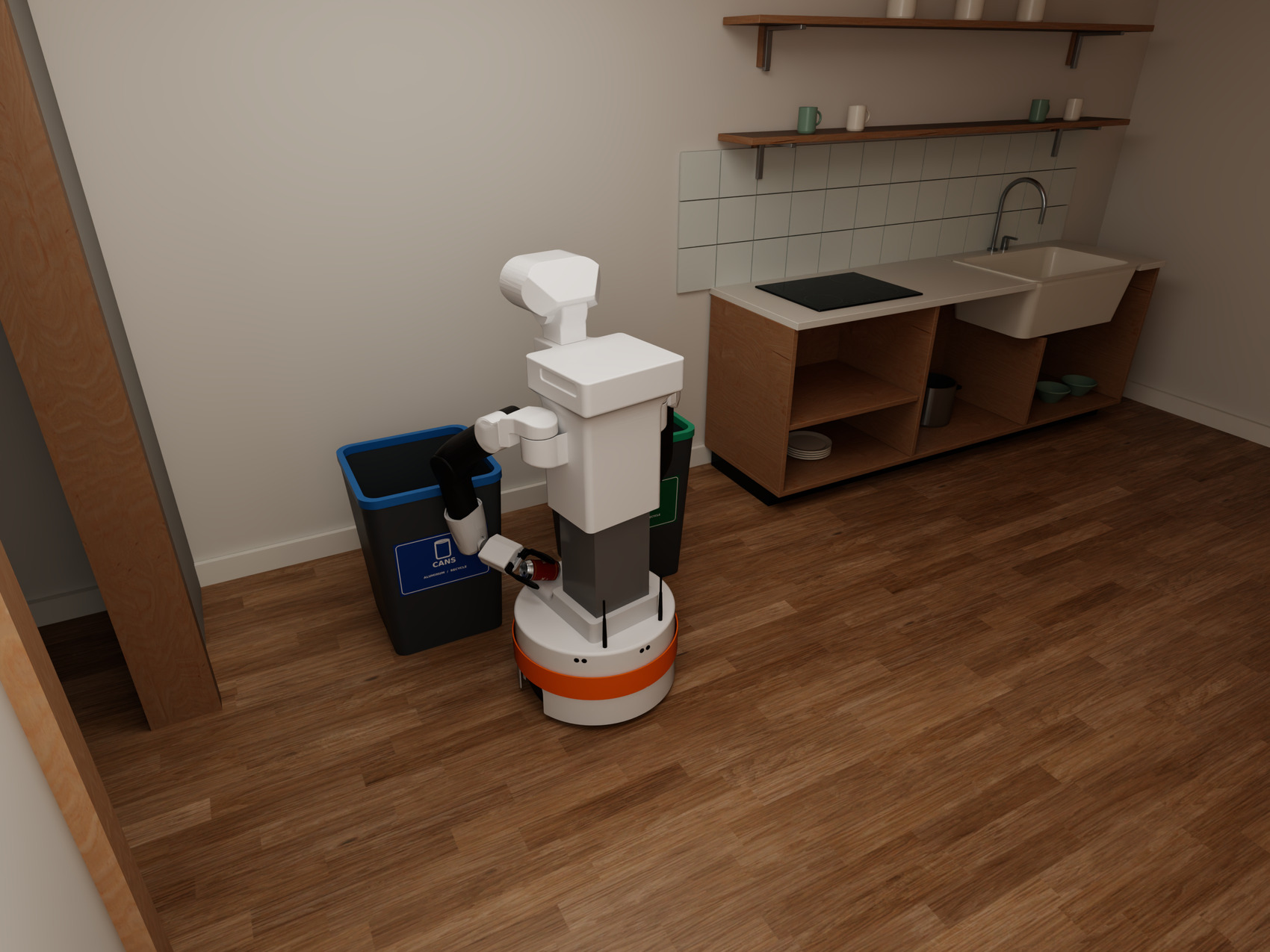} \\
\scriptsize $t=0.00$\,s & \scriptsize $t=25.00$\,s & \scriptsize $t=60.00$\,s & \scriptsize $t=119.65$\,s \\[6pt]
\end{tabular}
\caption{\textbf{Additional non-overlapping room-scale and multi-room robot rollouts.} \textbf{Top:} Fetch relocates the obstructing chair to clear the living-room passage (\texttt{R10}, score 1.0). \textbf{Middle:} AGIBOT Genie G2 grasps and carries a bottle away from the kitchen island, but the storage and door-opening conditions remain unsatisfied (\texttt{R01}, score 0.0). \textbf{Bottom:} An onboarded TIAGo++ deposits four of eight target recyclables across rooms before terminating (\texttt{M05}, score 0.5). All keyframes are path-traced from logged simulation states at $2560\times1920$ (1024 samples per pixel, 12 diffuse bounces, up to 16 specular/transmission bounces).}
\label{fig:supp-extra-rollouts}
\end{figure*}

To complement the four executions in Figure~\ref{fig:robot-keyframes} of the main paper and the tabletop suite in Figure~\ref{fig:supp-tabletop-capture} without repeating main-paper episodes, Figure~\ref{fig:supp-extra-rollouts} presents three additional room-scale and multi-room robot rollouts: one completed single-room episode (\texttt{Fetch} clearing a living-room passage on \texttt{R10}) and two incomplete episodes that illustrate partial progress and preserved failure logs (\texttt{AGIBOT Genie G2} on \texttt{R01} and an onboarded \texttt{TIAGo++} model on \texttt{M05}). All keyframes are path-traced directly from logged simulation states using fixed interior cameras and consistent lighting, with reported scores taken from each episode's frozen rubric.

\begin{table}[!tb]
\centering\scriptsize
\caption{\textbf{16 integrated robot models} across single-arm, dual-arm, mobile-manipulator, and humanoid categories.}
\label{tab:robots-16}
\setlength{\tabcolsep}{2pt}
\begin{tabular}{@{}lll@{}}
\toprule
Robot Model & Category & Provenance / Benchmark \\
\midrule
Franka Panda & Single-arm & Isaac Sim / LIBERO / RoboTwin \\
UR5e & Single-arm & UR SDK / OmniGibson \\
UR10e & Single-arm & UR SDK \\
UR5 + WSG & Single-arm & RoboTwin \\
Unitree Z1 & Single-arm & Unitree ROS \\
Piper & Single-arm & RoboTwin \\
ARX-X5 & Single/Dual-arm & RoboTwin / RoboDojo \\
Aloha-AgileX & Dual-arm & RoboTwin \\
AGIBOT Genie G1 & Dual-arm upper & GenieSim \\
AGIBOT Genie G2 & Dual-arm upper & GenieSim \\
Fetch & Mobile manip. & OmniGibson / BEHAVIOR \\
Dual-arm TIAGo & Mobile manip. & OmniGibson / BEHAVIOR \\
Galaxea R1Pro & Mobile dual-arm & OmniGibson / BEHAVIOR \\
Unitree G1 (23/29-DoF) & Humanoid & Unitree ROS \\
Unitree H1 & Humanoid & Unitree ROS \\
Unitree H1-2 & Humanoid & Unitree ROS \\
\bottomrule
\end{tabular}
\end{table}

\section{Extended Scene Comparisons and Close-Up Renders}
\label{sec:supp-qualitative}

\begin{figure*}[p]
\centering
\setlength{\tabcolsep}{1.5pt}
\begin{tabular}{@{}c*{6}{c}@{}}
 & \scriptsize Holodeck & \scriptsize LayoutVLM & \scriptsize SAGE & \scriptsize EmbodiedGen V2 & \scriptsize SceneSmith & \scriptsize EmbodiedStudio \\[2pt]
\rotatebox[origin=c]{90}{\scriptsize T01 / Common light} &
\includegraphics[width=.142\textwidth]{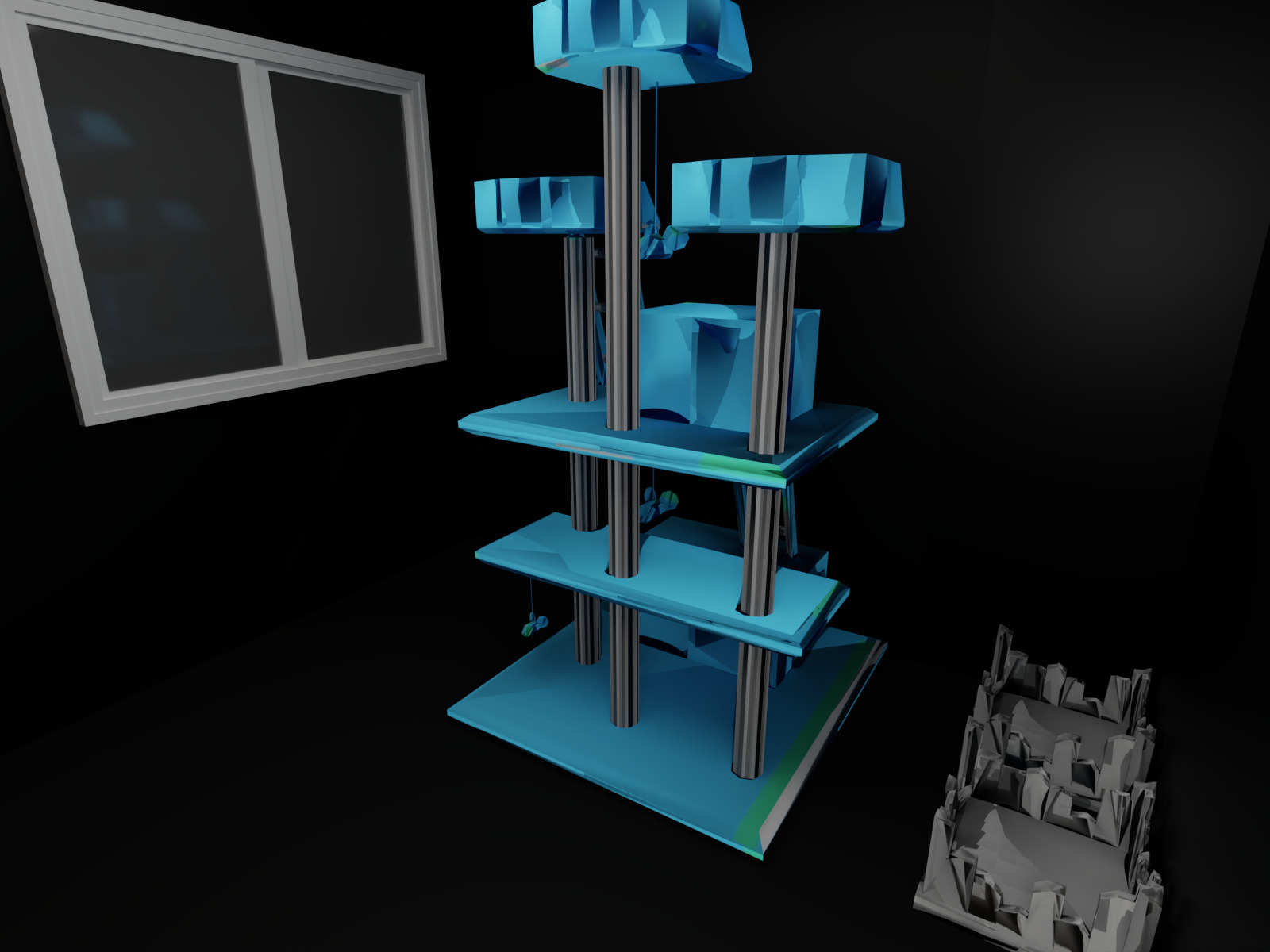} &
\includegraphics[width=.142\textwidth]{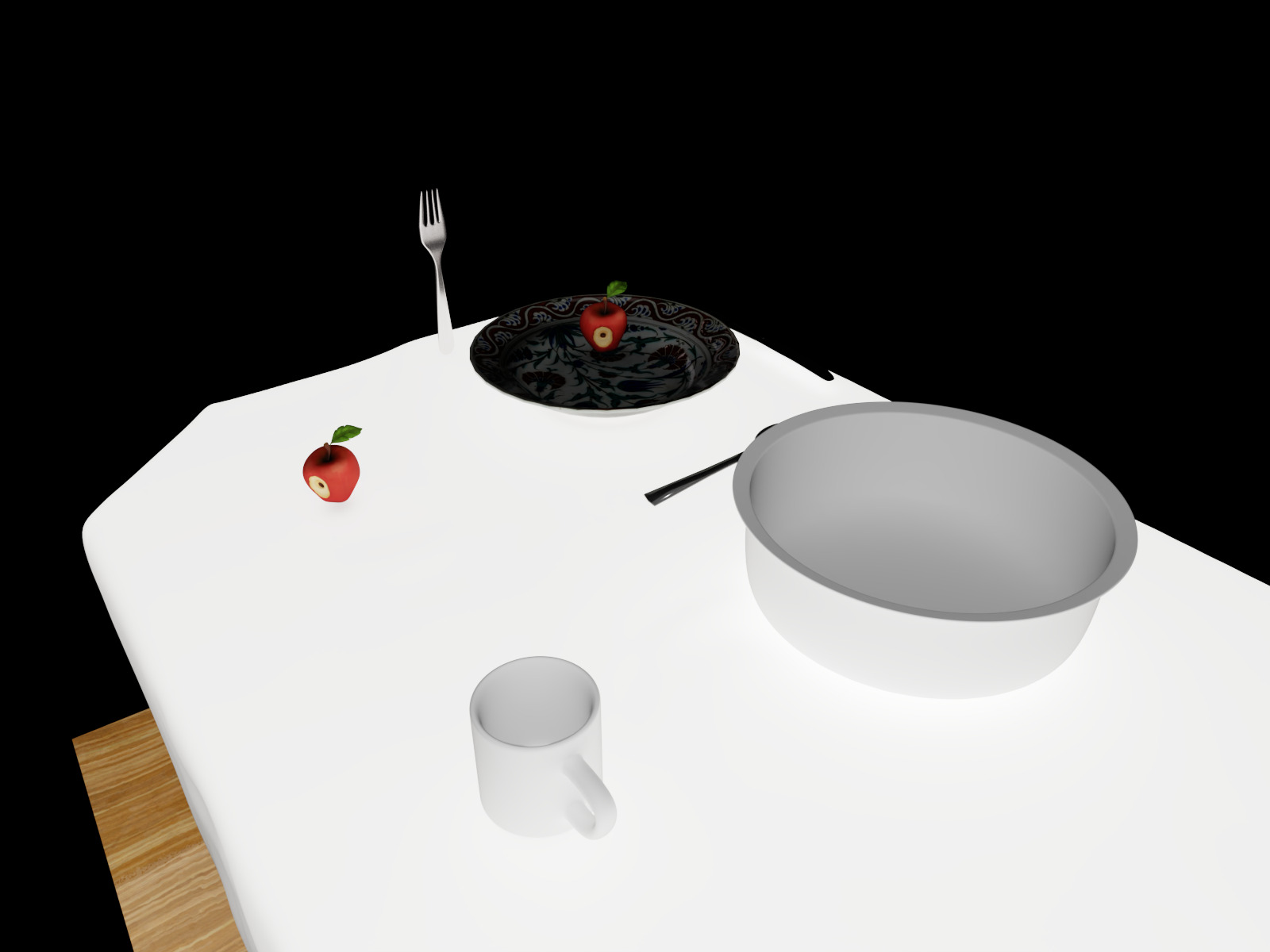} &
\includegraphics[width=.142\textwidth]{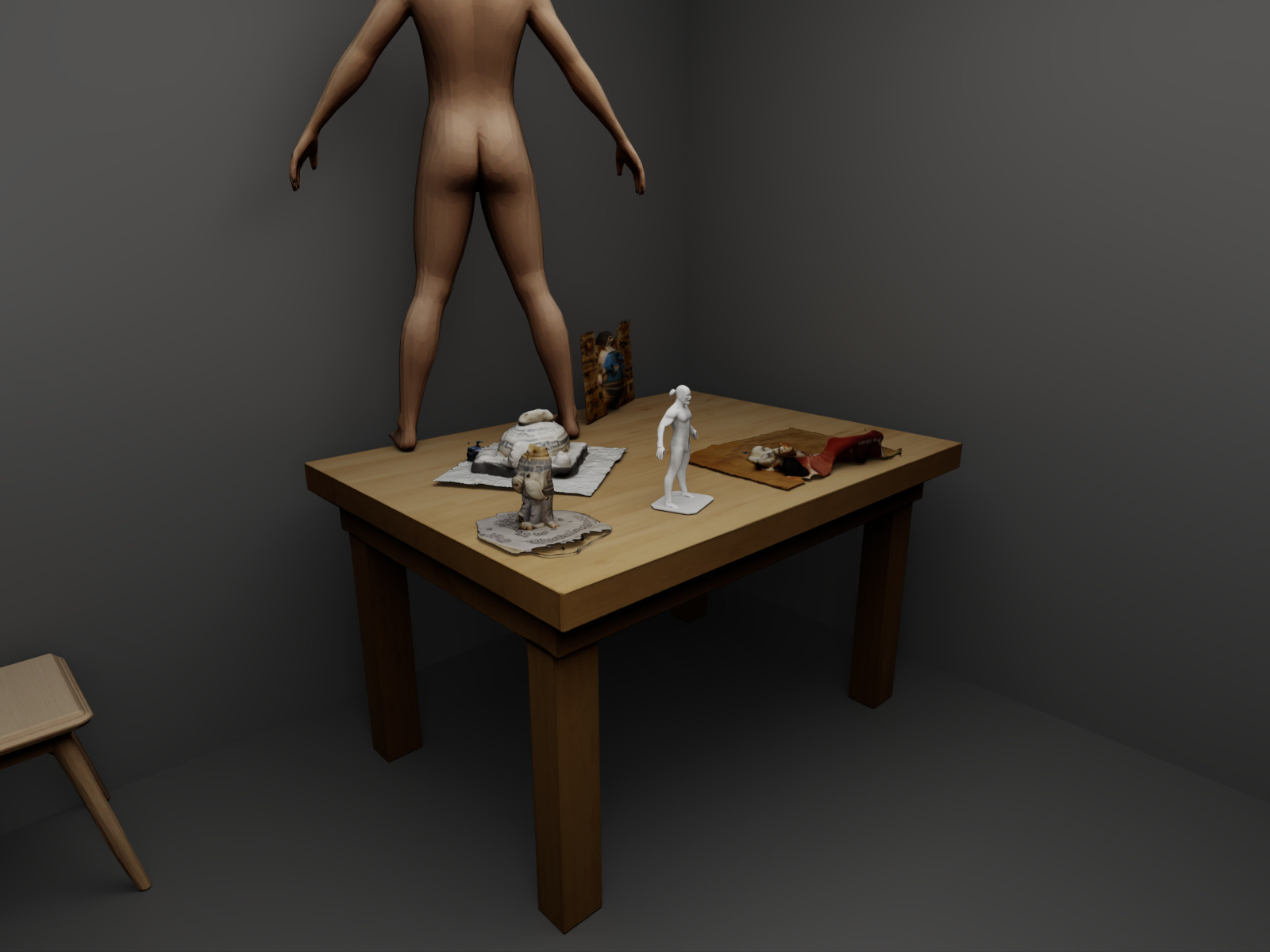} &
\includegraphics[width=.142\textwidth]{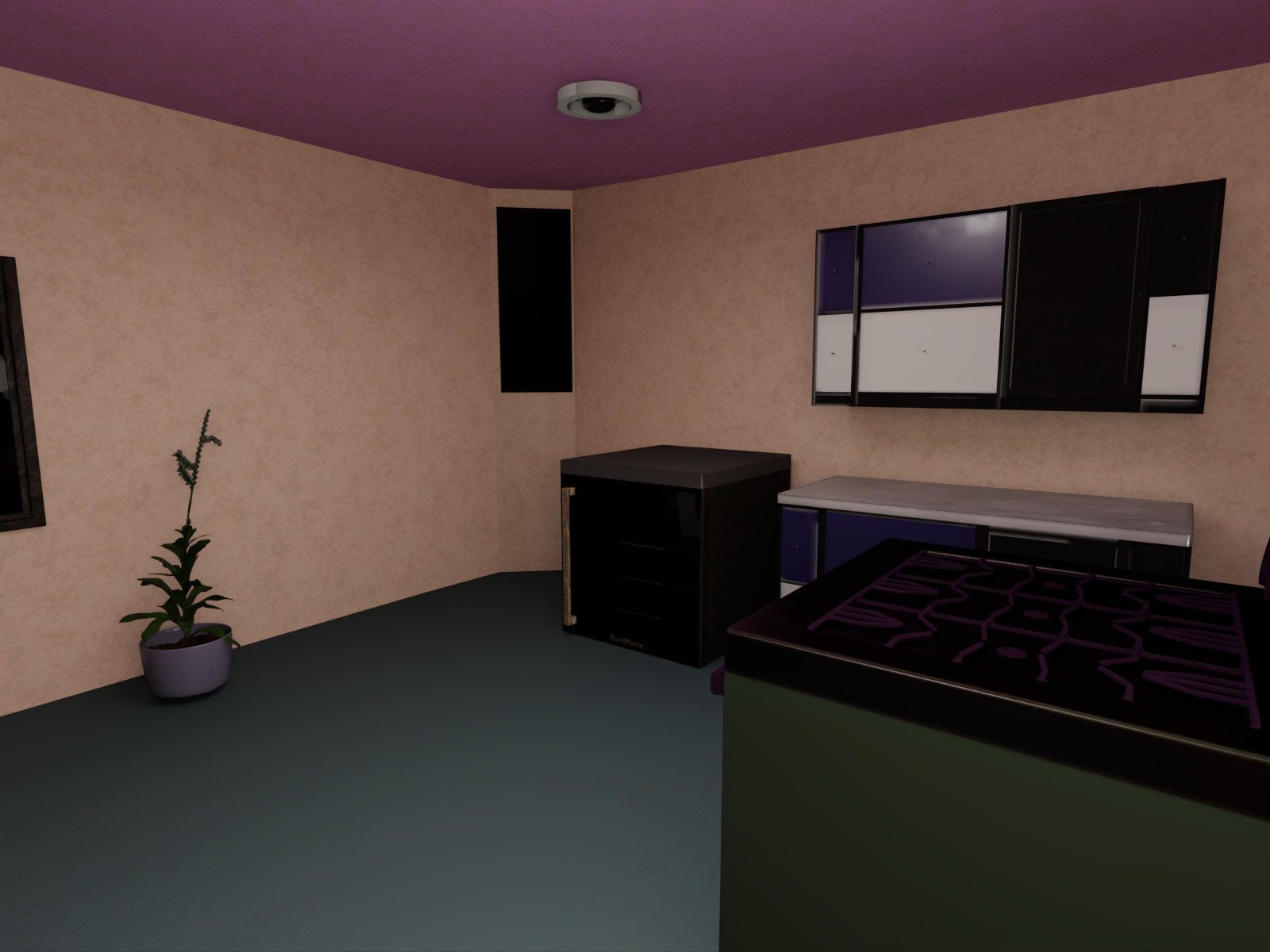} &
\includegraphics[width=.142\textwidth]{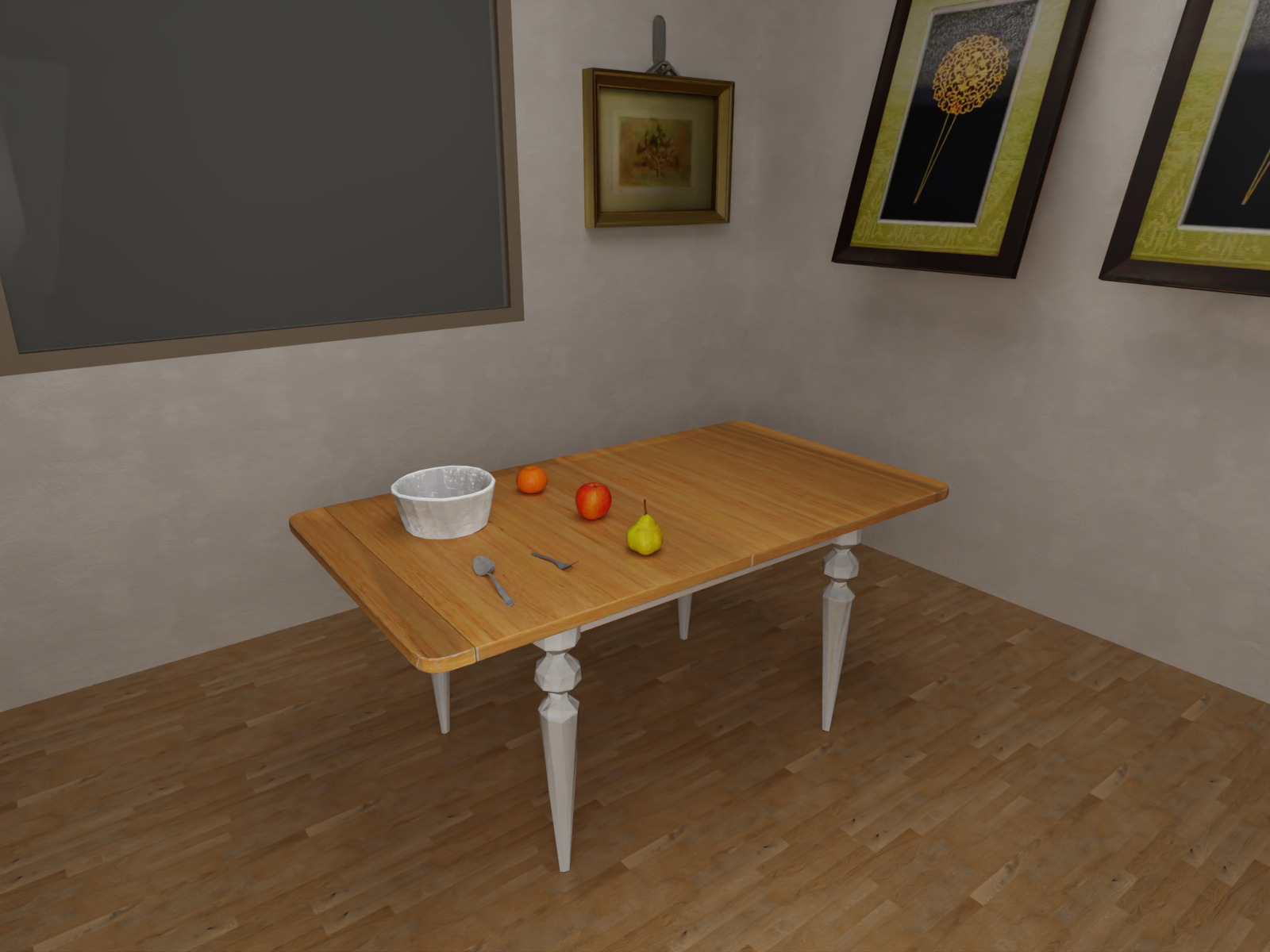} &
\includegraphics[width=.142\textwidth]{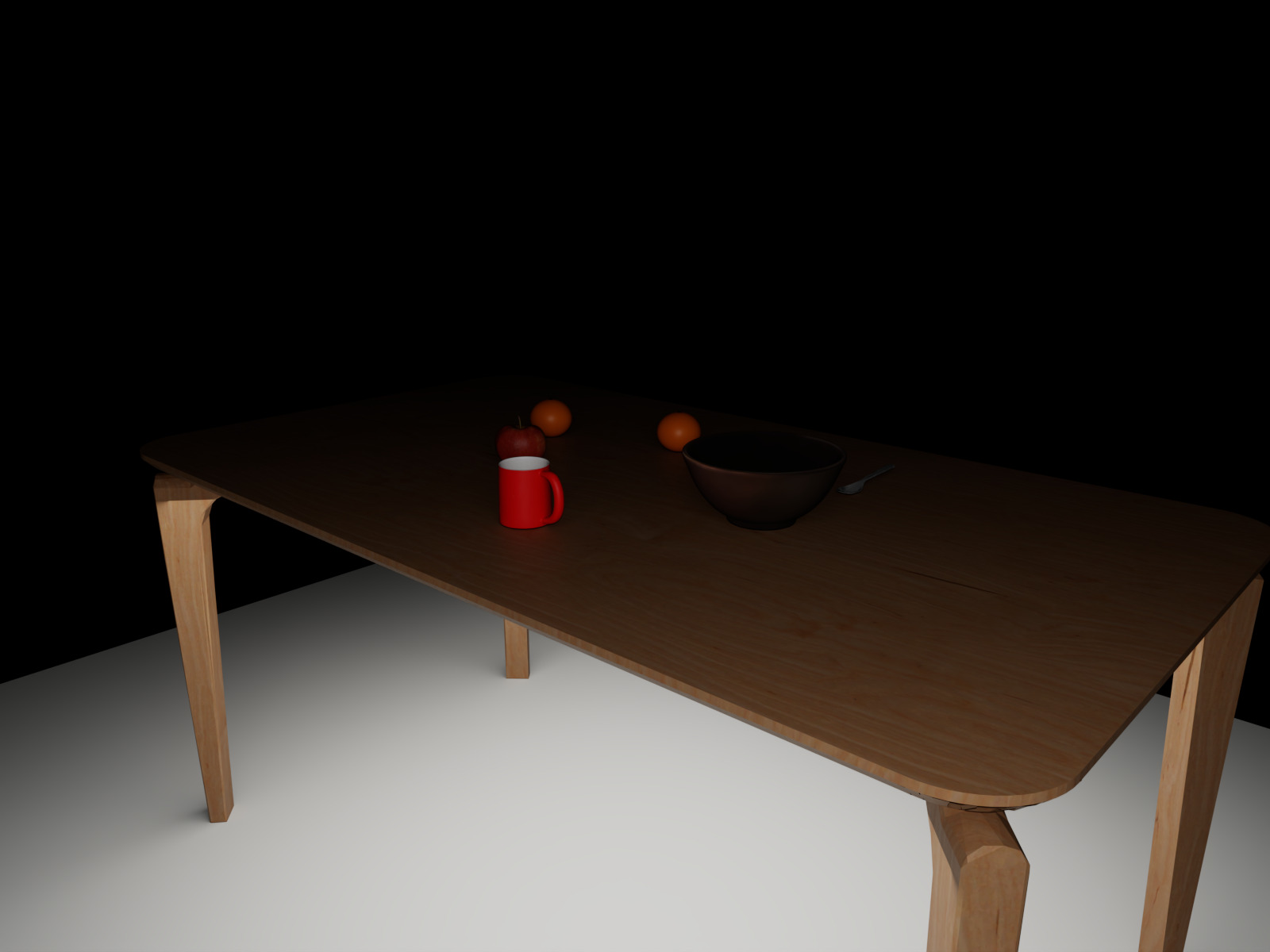} \\[1pt]
\rotatebox[origin=c]{90}{\scriptsize T01 / Native light} &
\includegraphics[width=.142\textwidth]{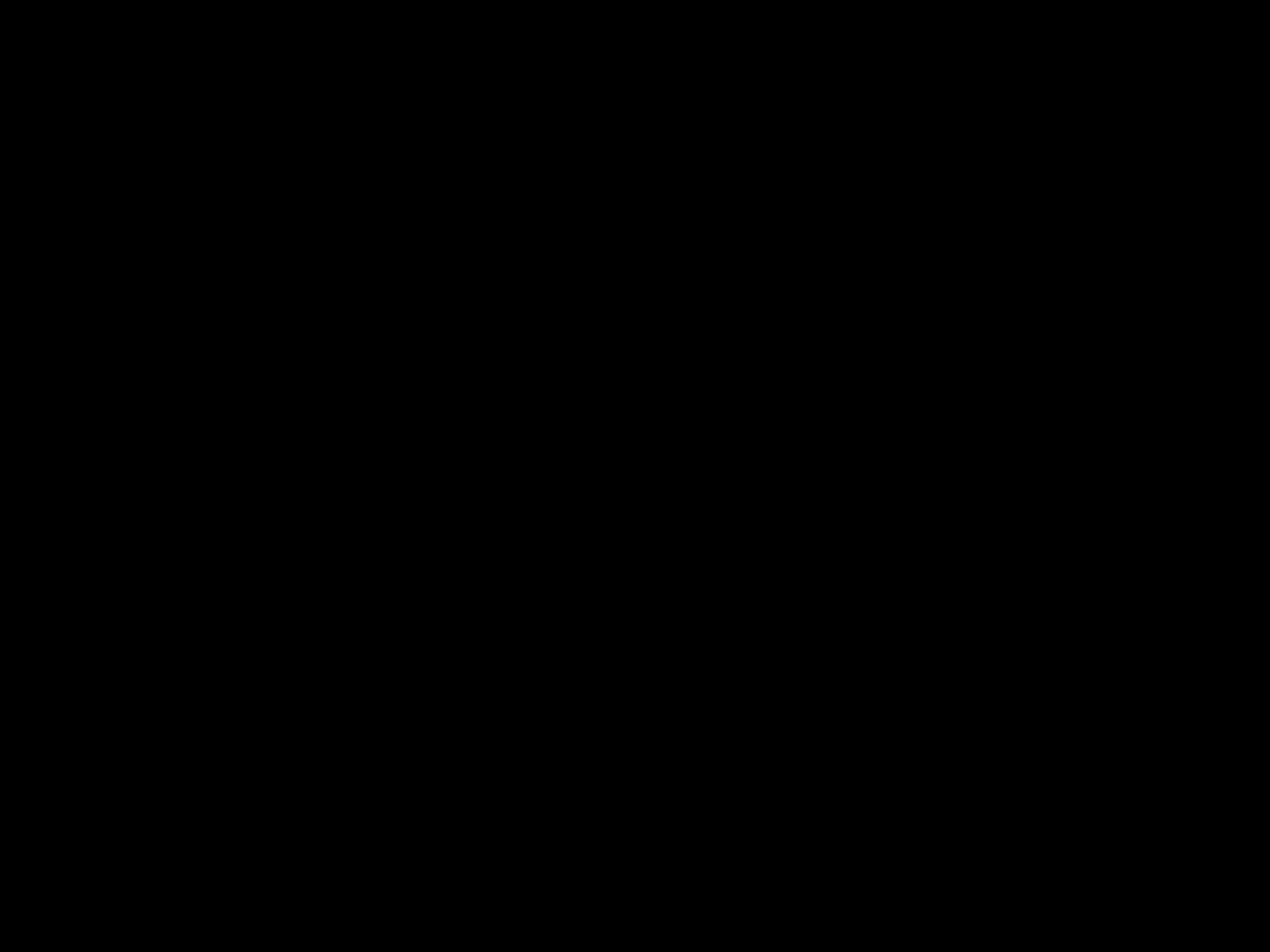} &
\includegraphics[width=.142\textwidth]{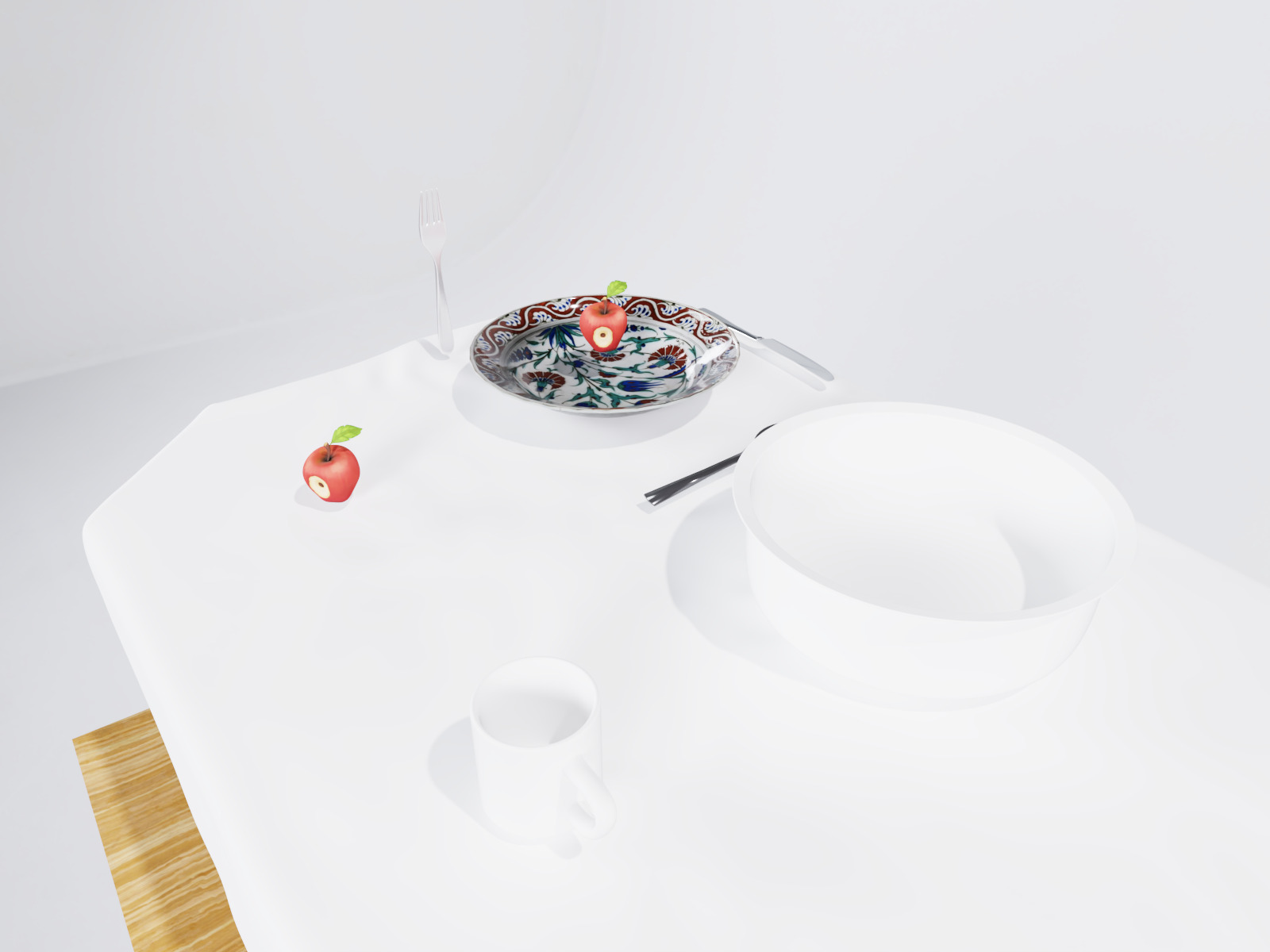} &
\includegraphics[width=.142\textwidth]{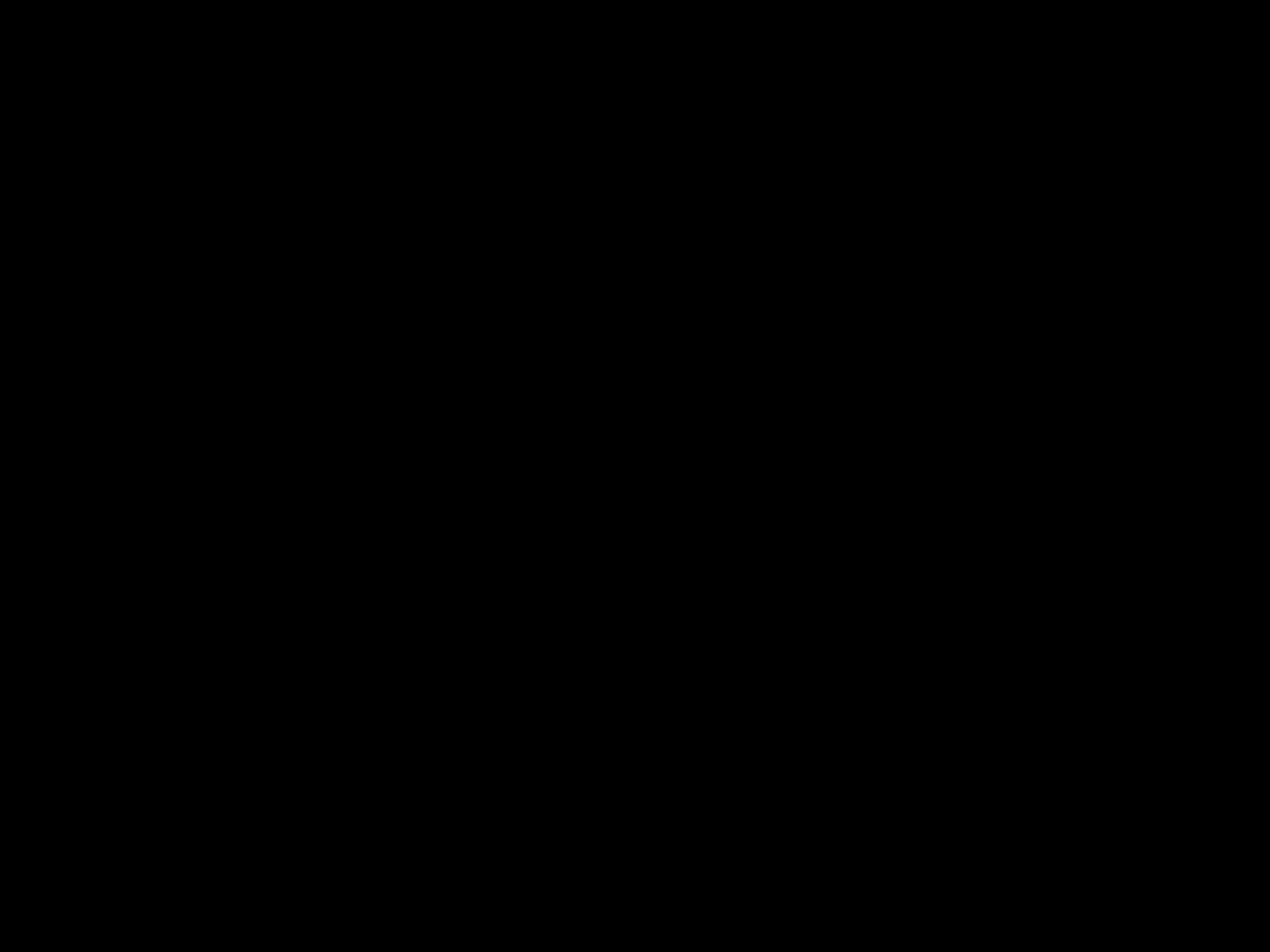} &
\includegraphics[width=.142\textwidth]{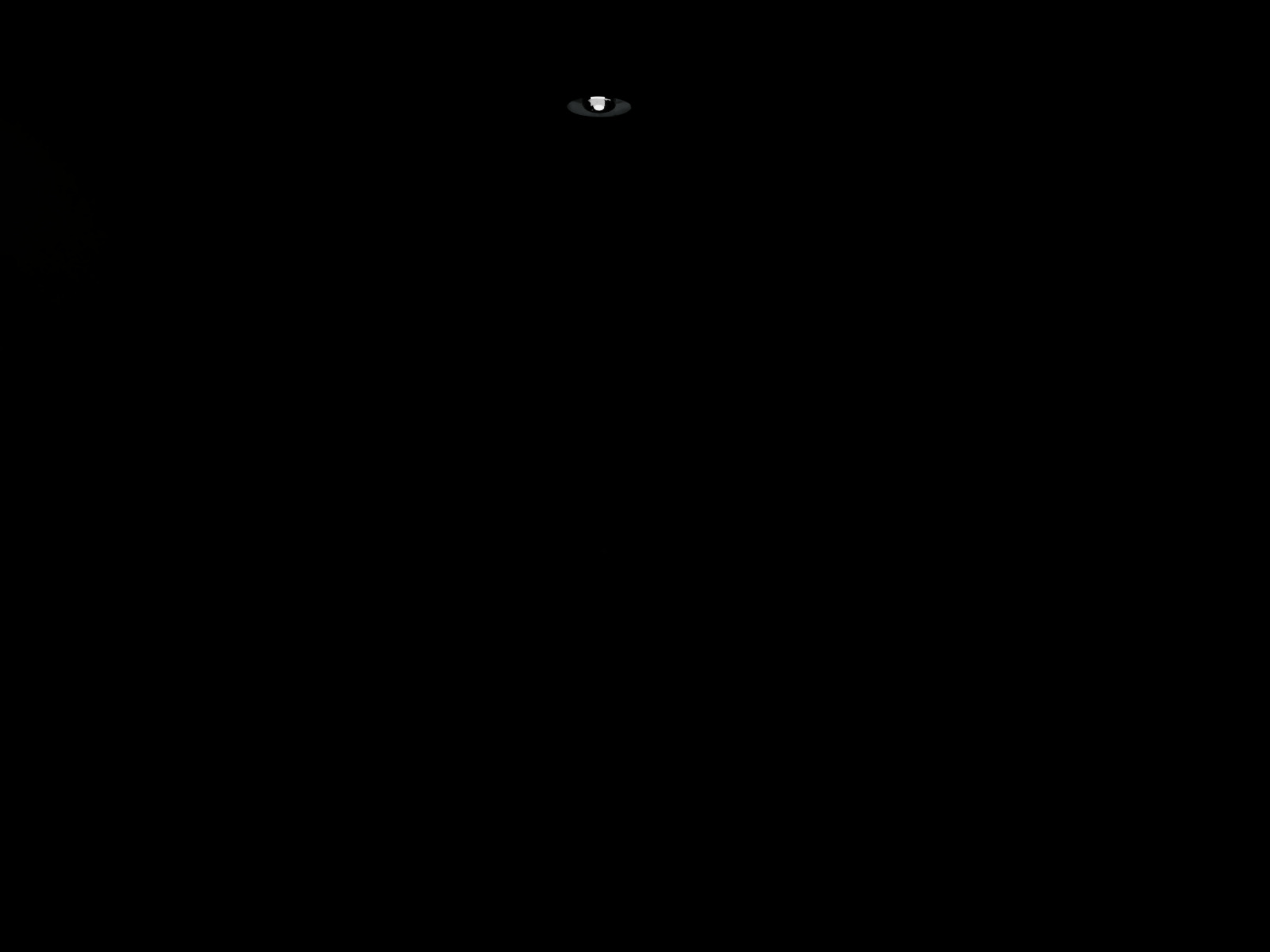} &
\includegraphics[width=.142\textwidth]{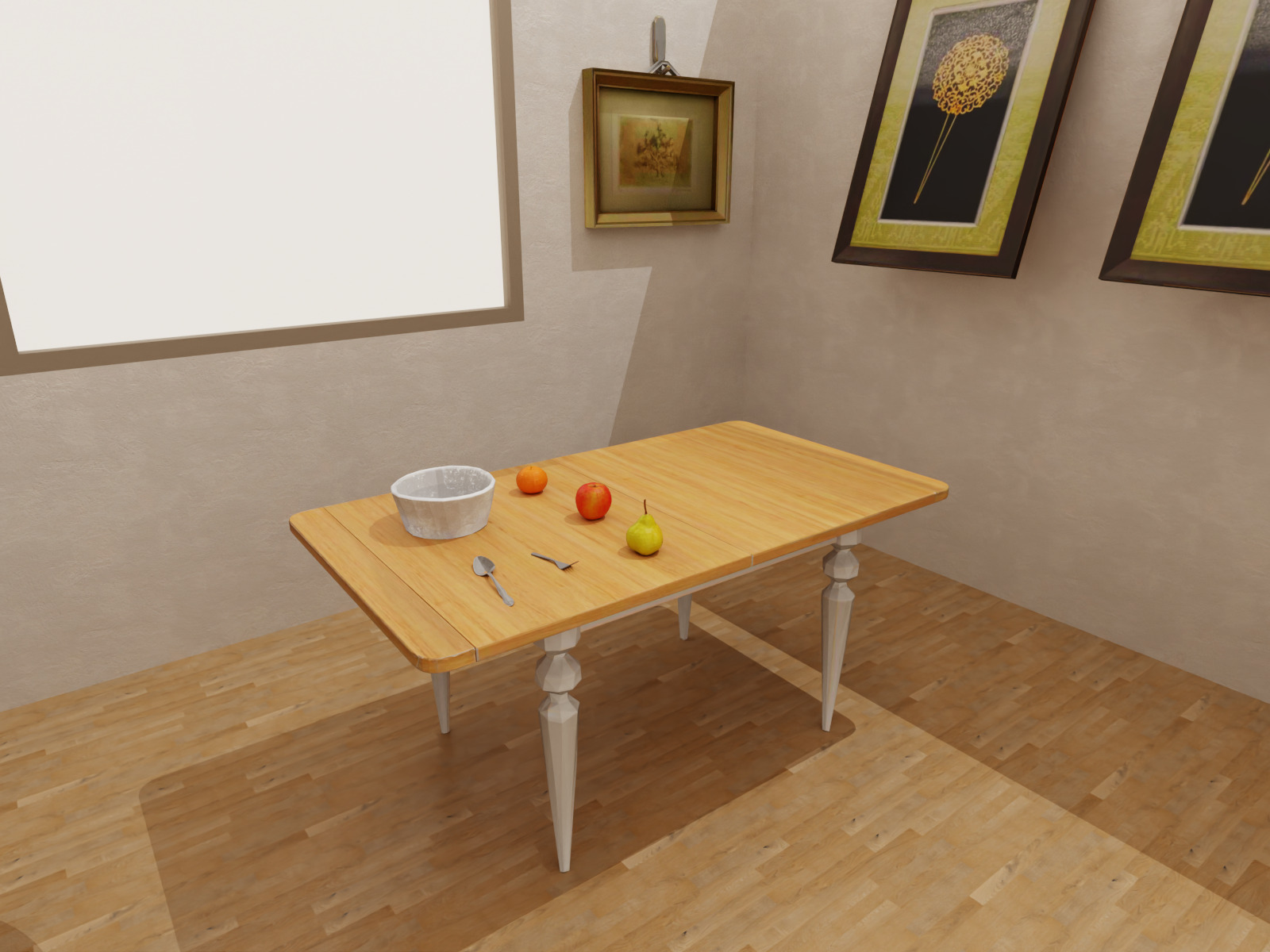} &
\includegraphics[width=.142\textwidth]{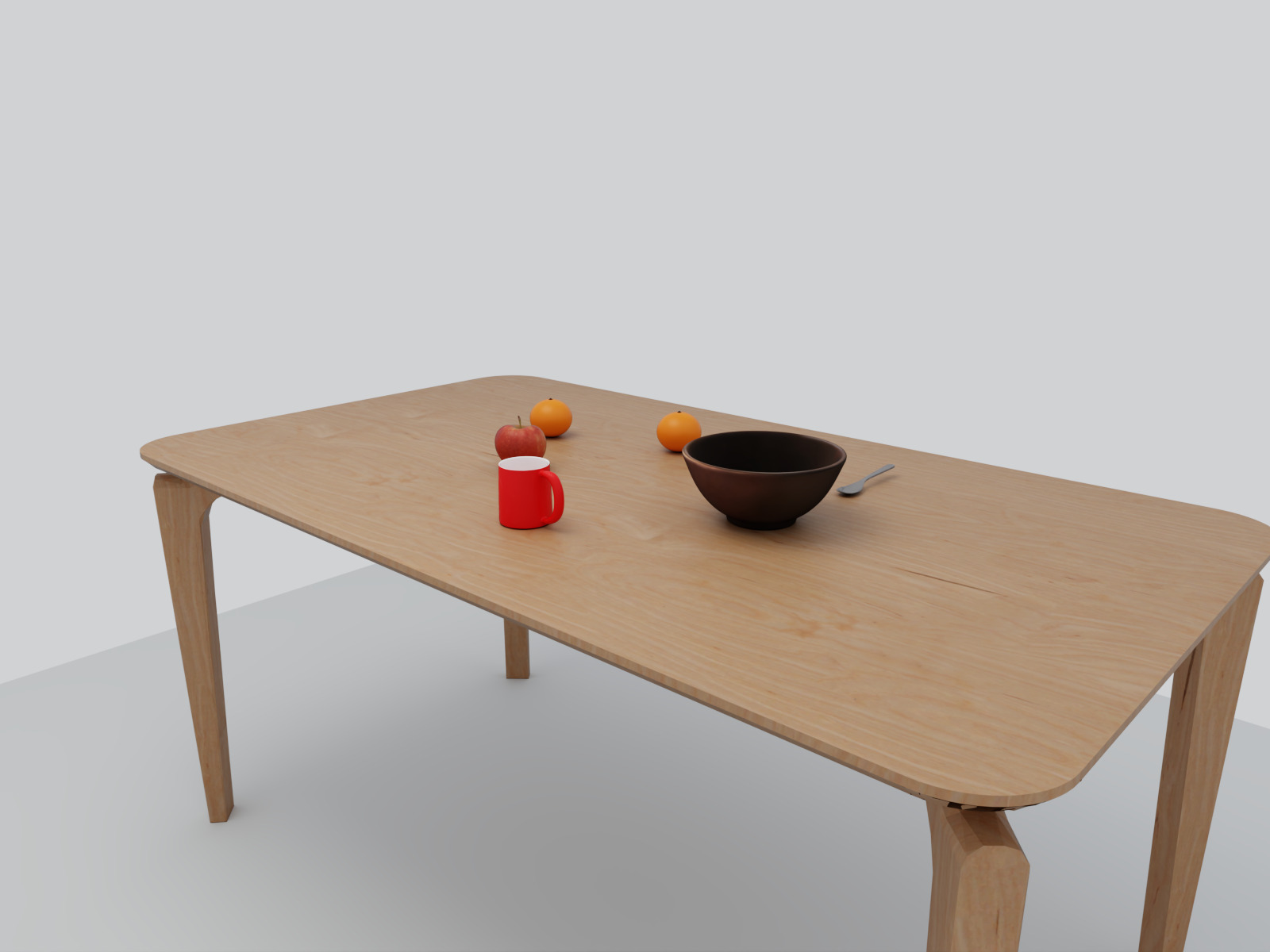} \\[1pt]
 & \tiny 3 native lights & \tiny 2 native lights & \tiny No exported lights & \tiny 7 native lights & \tiny 2 native lights & \tiny 4 native lights \\[4pt]
\rotatebox[origin=c]{90}{\scriptsize R14 / Common light} &
\includegraphics[width=.142\textwidth]{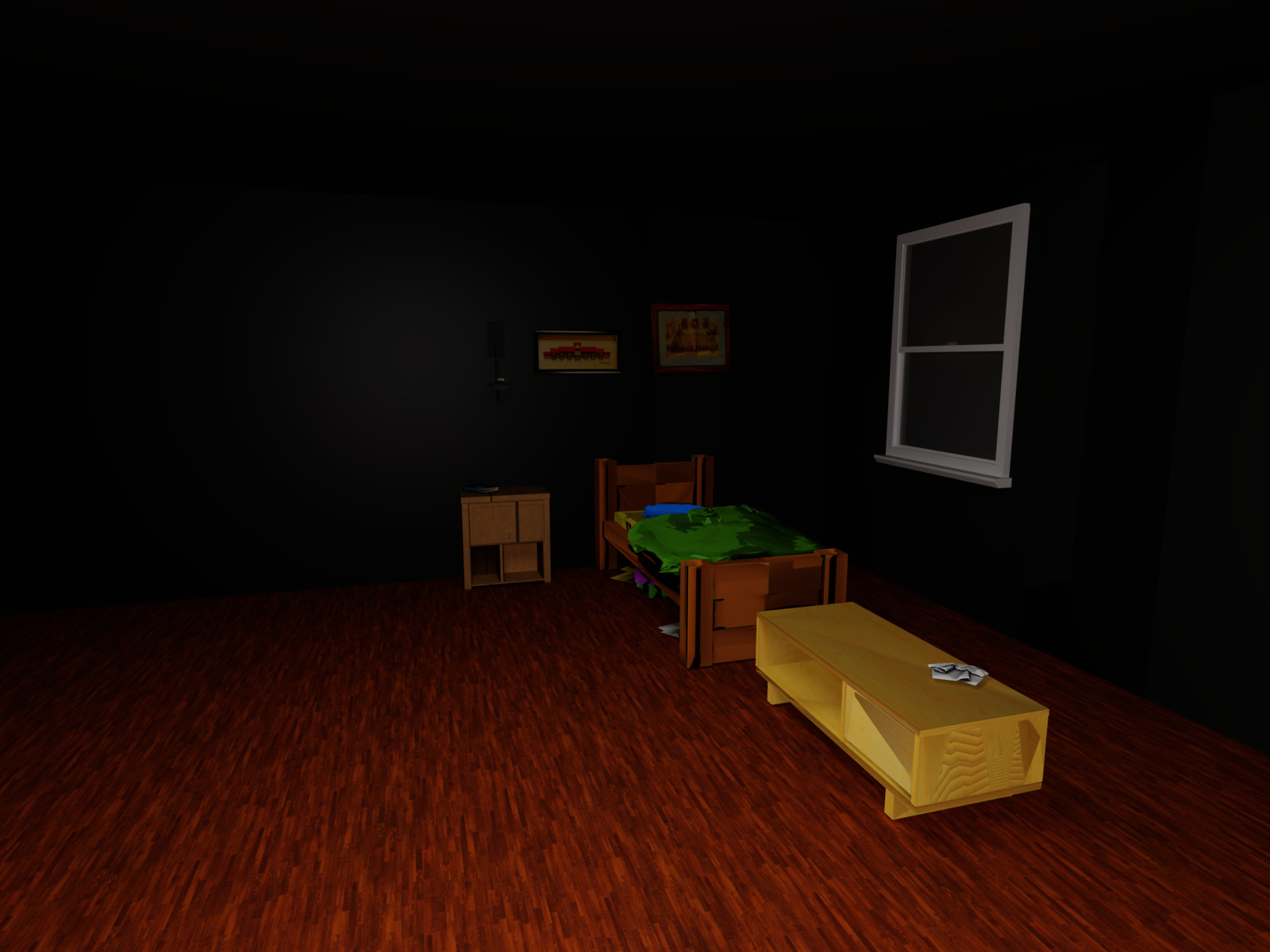} &
\includegraphics[width=.142\textwidth]{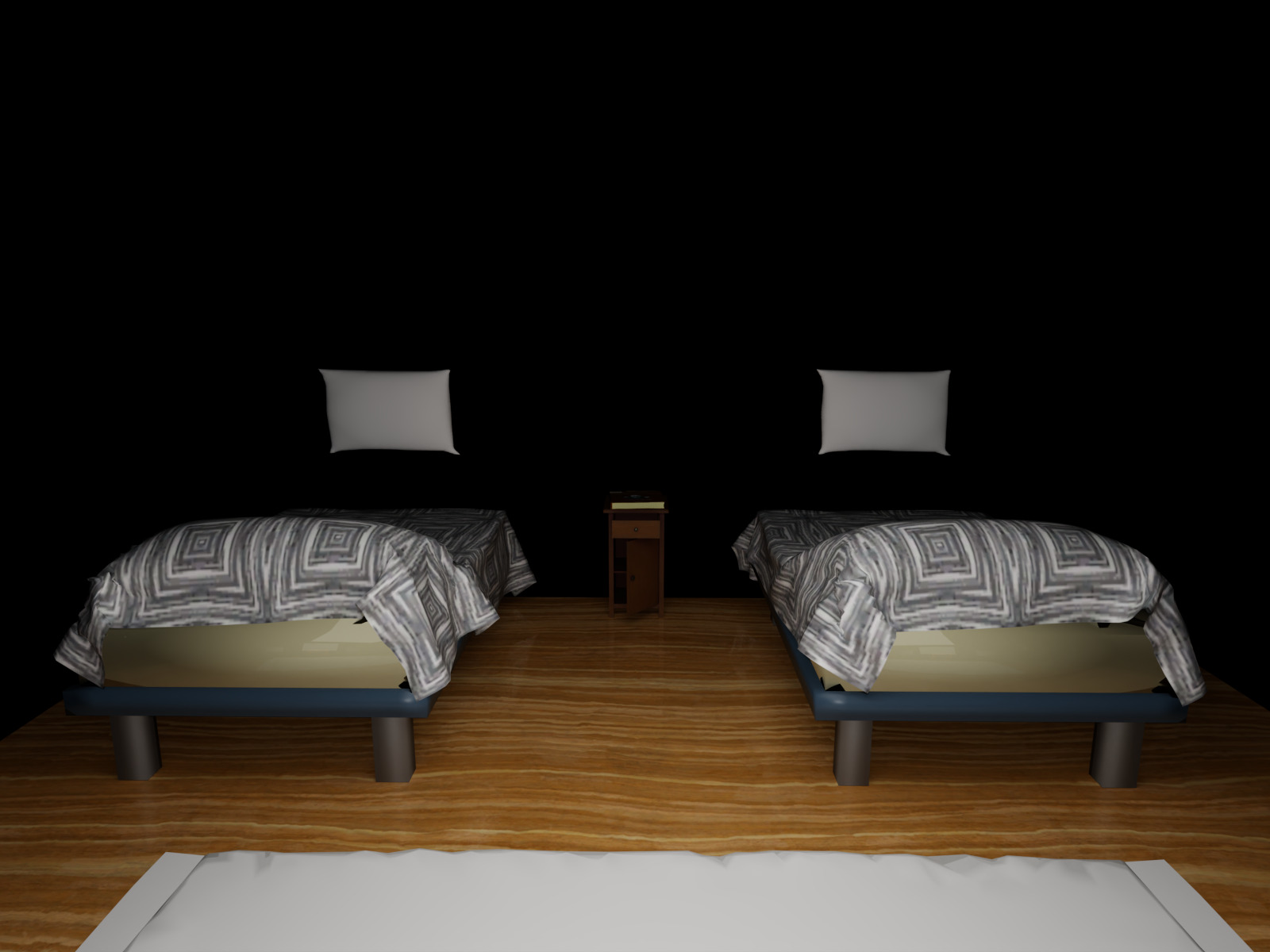} &
\includegraphics[width=.142\textwidth]{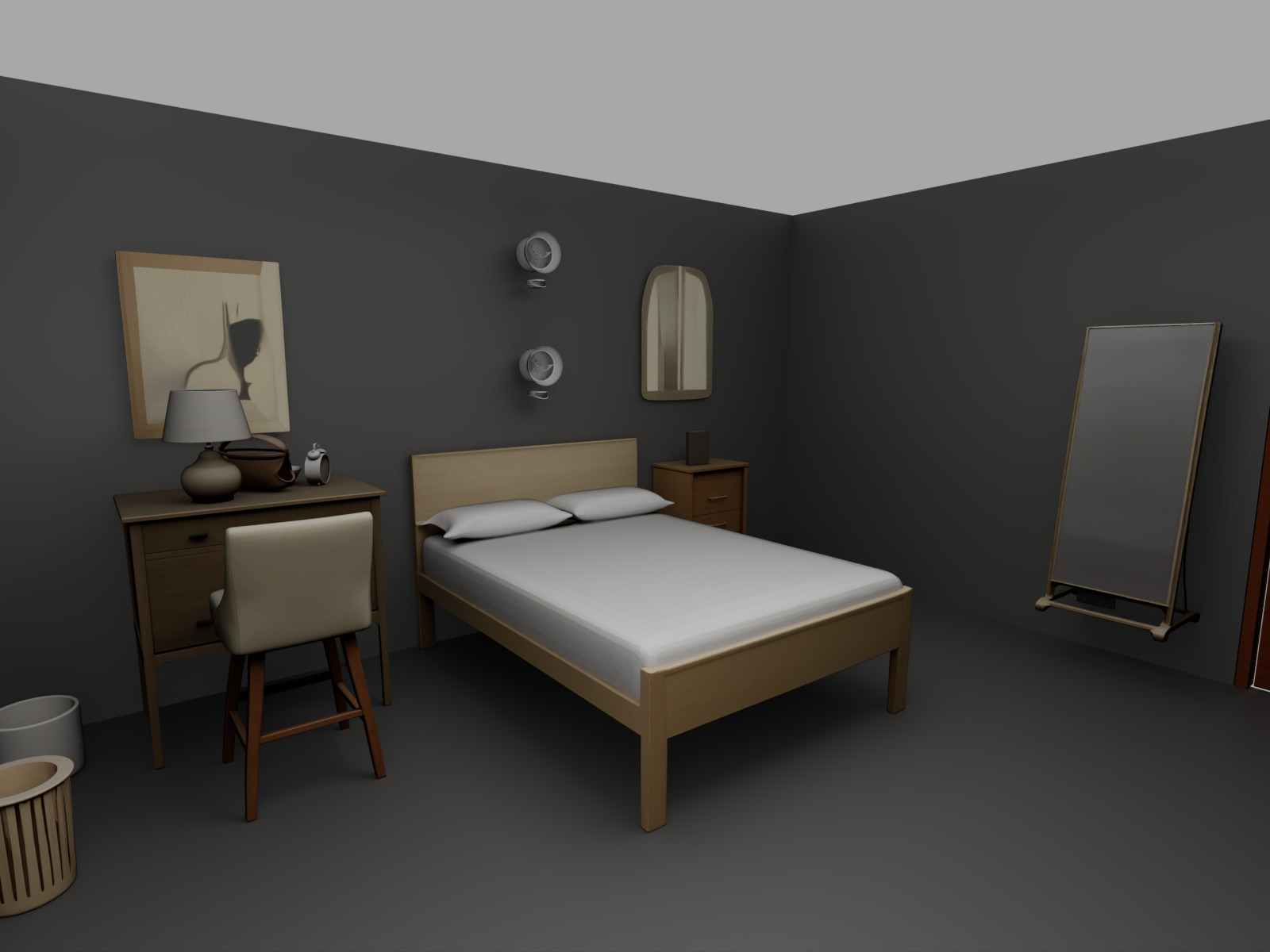} &
\includegraphics[width=.142\textwidth]{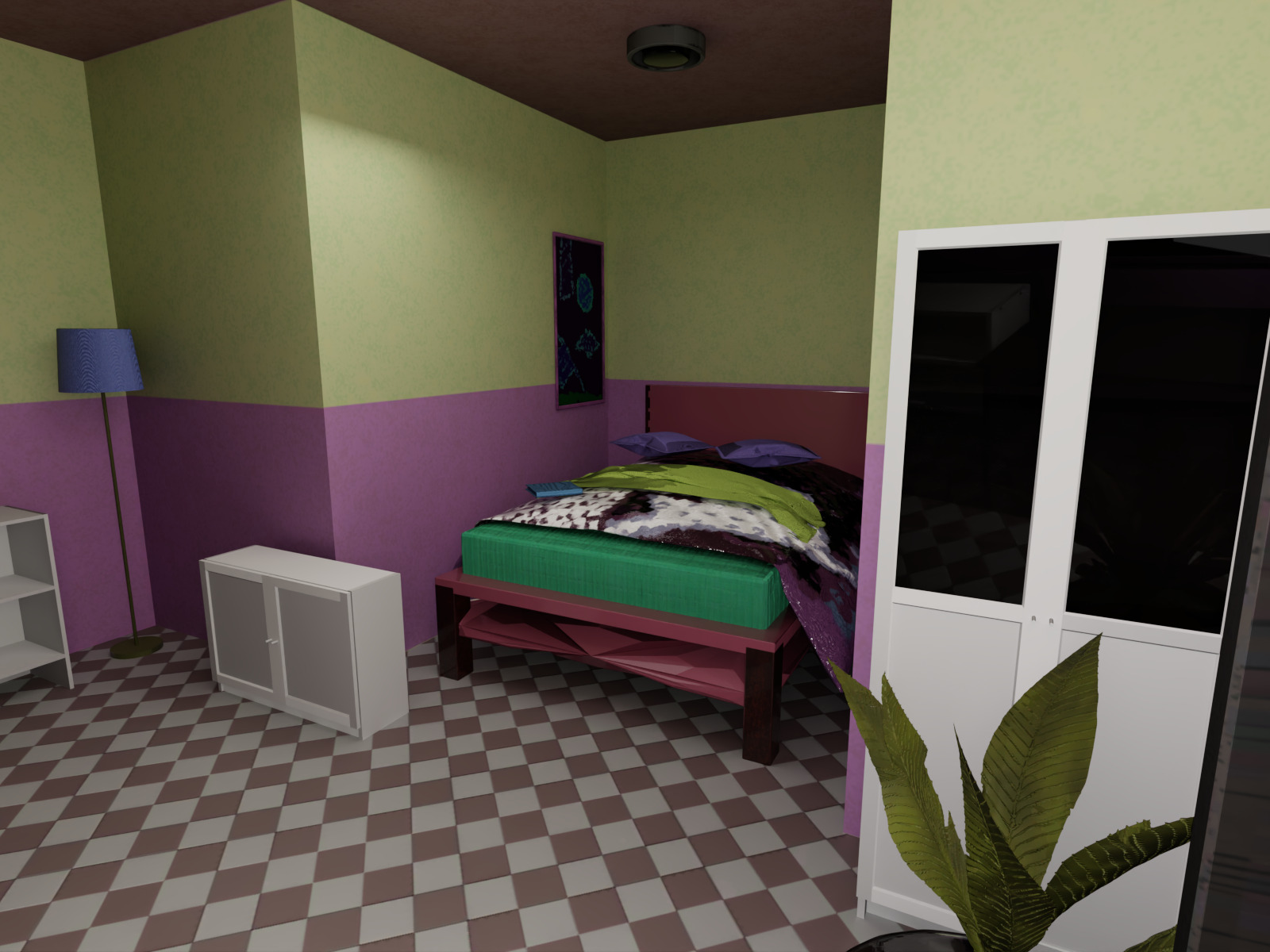} &
\includegraphics[width=.142\textwidth]{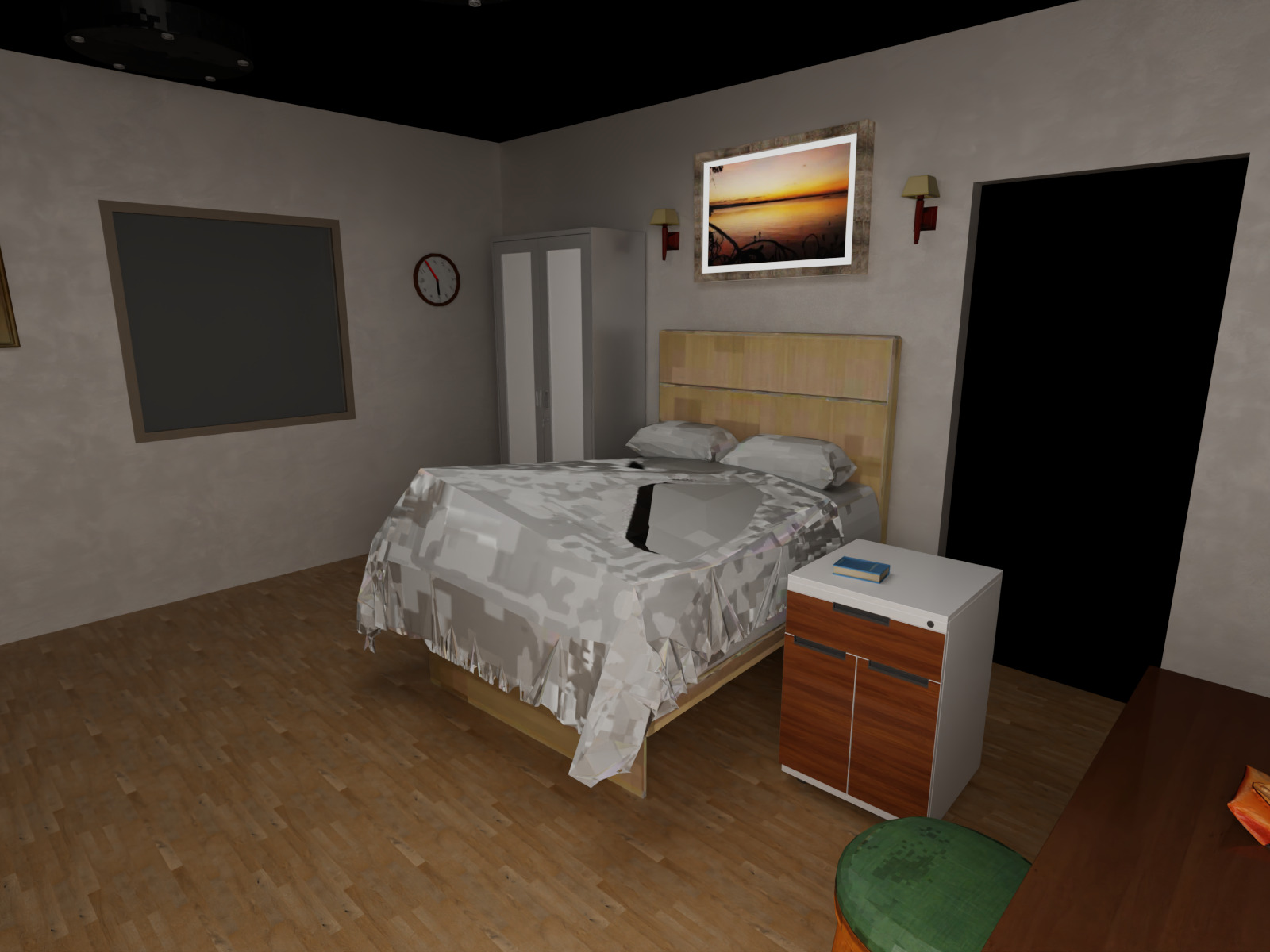} &
\includegraphics[width=.142\textwidth]{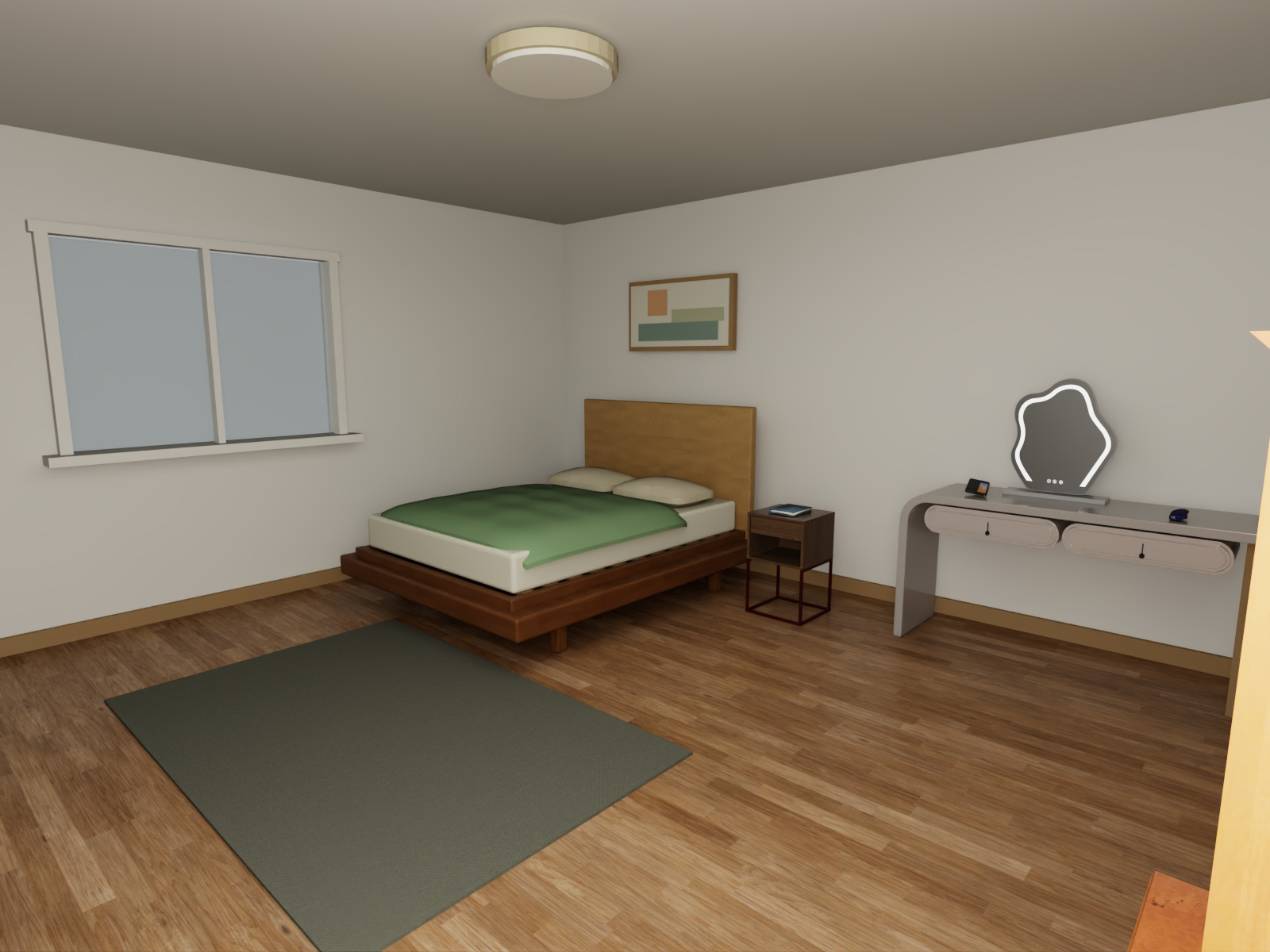} \\[1pt]
\rotatebox[origin=c]{90}{\scriptsize R14 / Native light} &
\includegraphics[width=.142\textwidth]{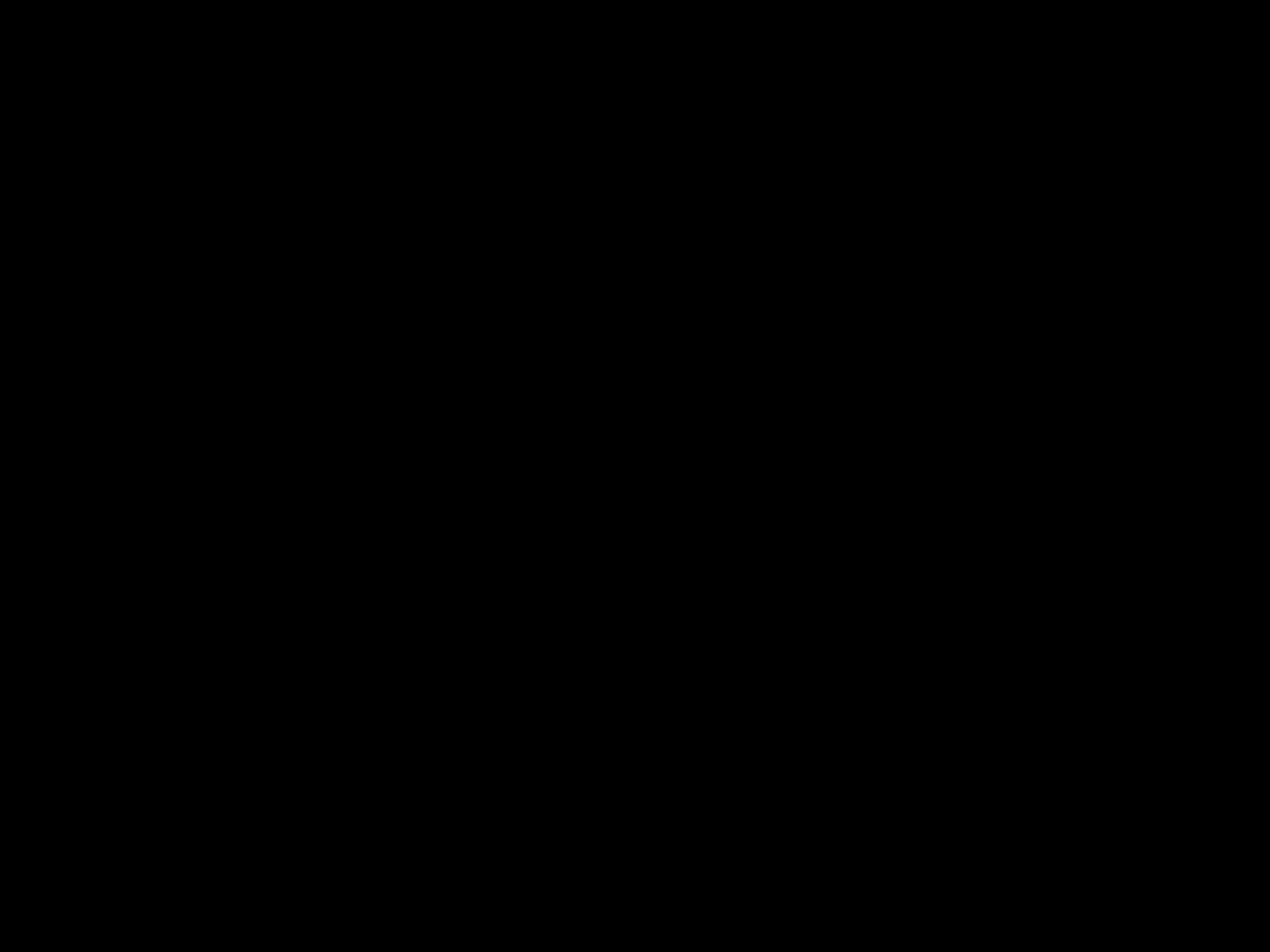} &
\includegraphics[width=.142\textwidth]{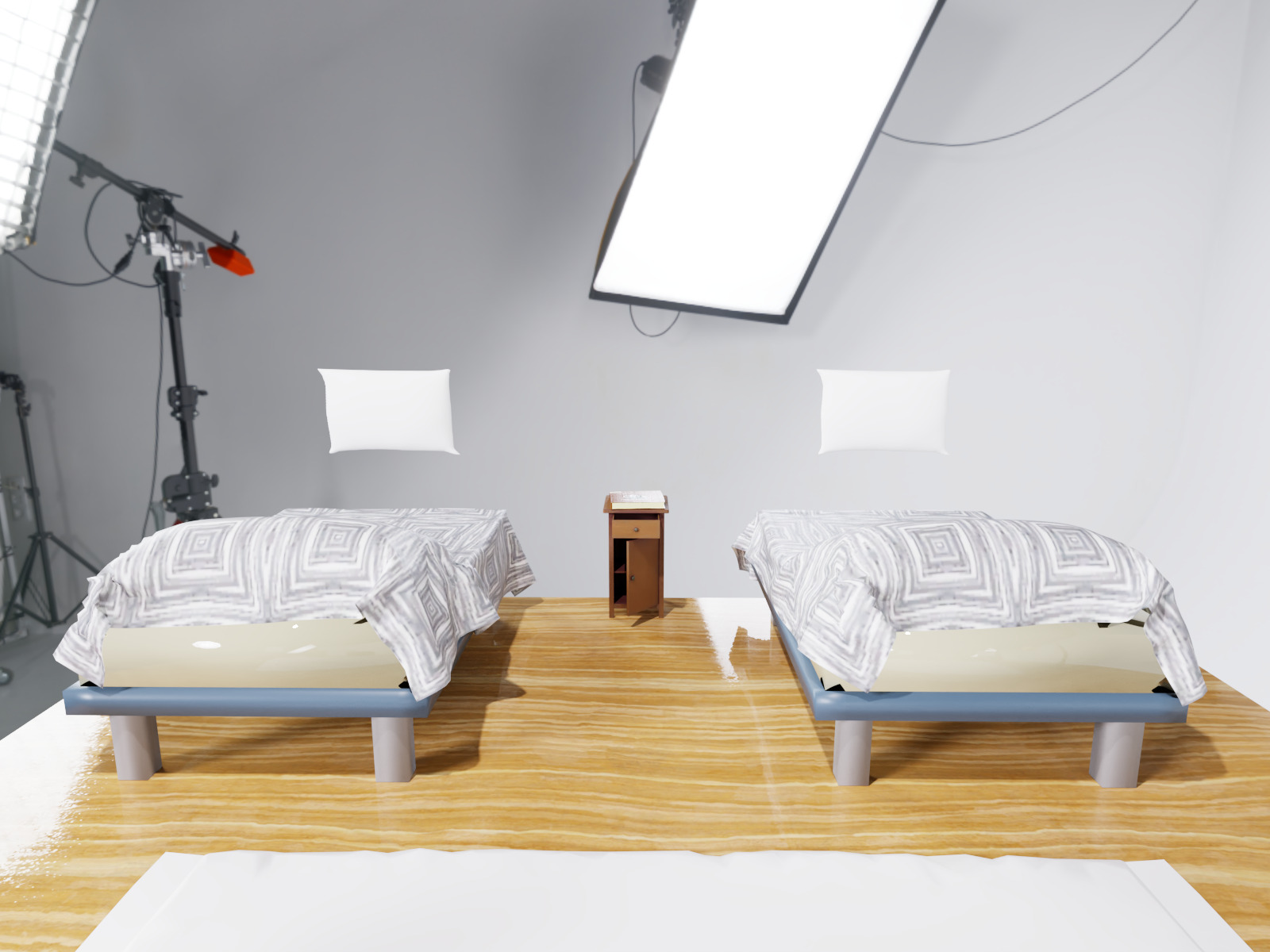} &
\includegraphics[width=.142\textwidth]{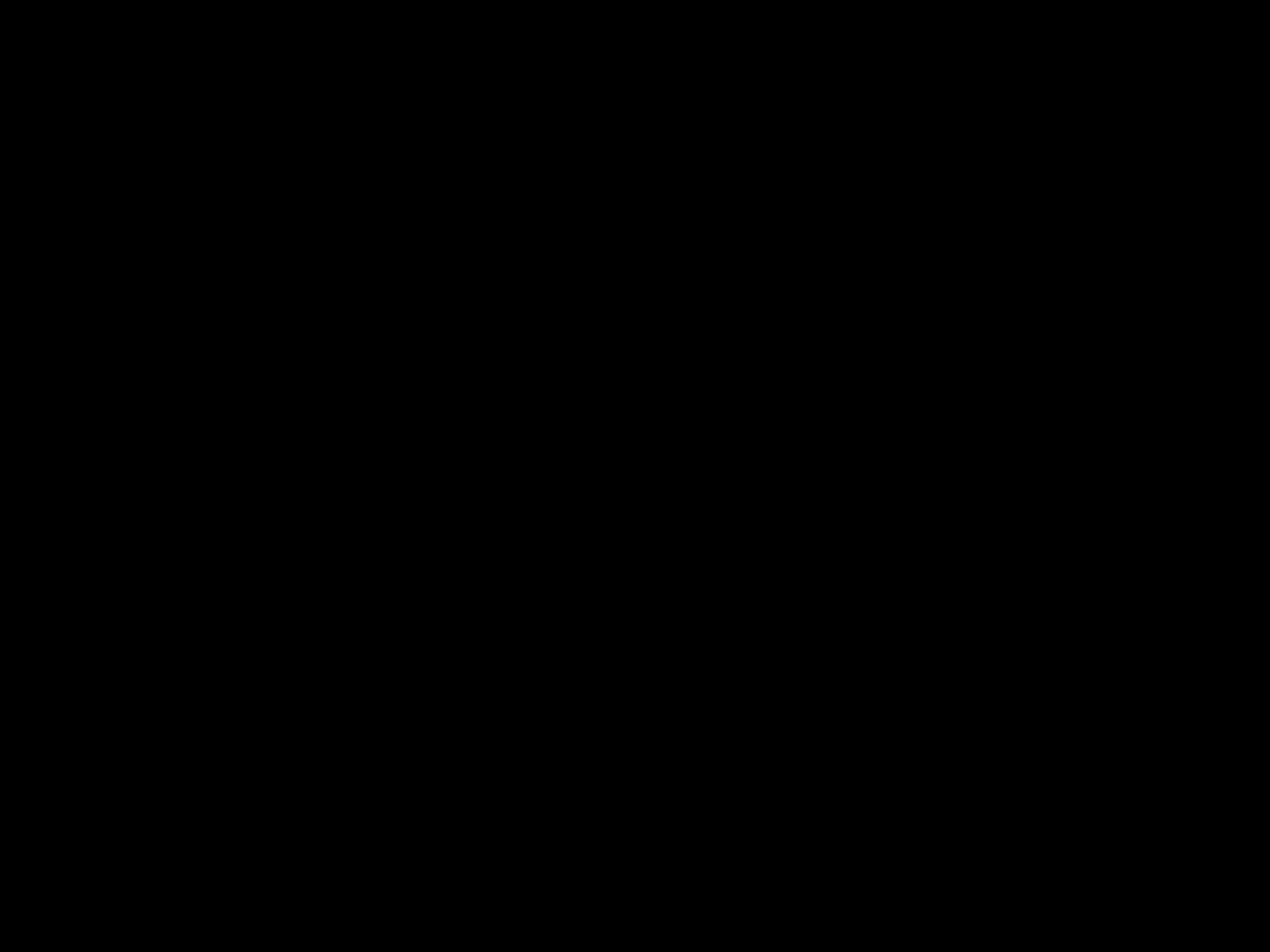} &
\includegraphics[width=.142\textwidth]{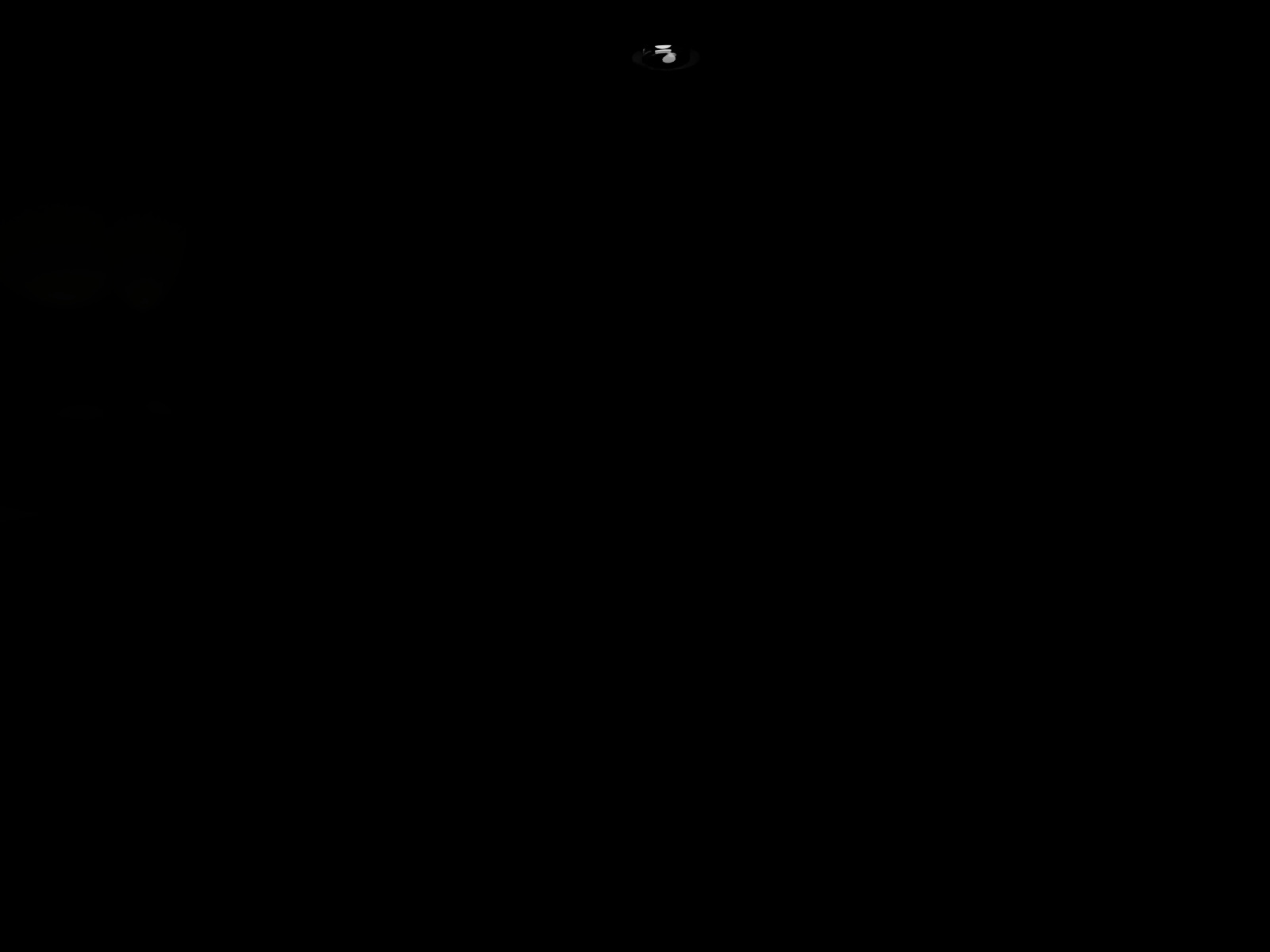} &
\includegraphics[width=.142\textwidth]{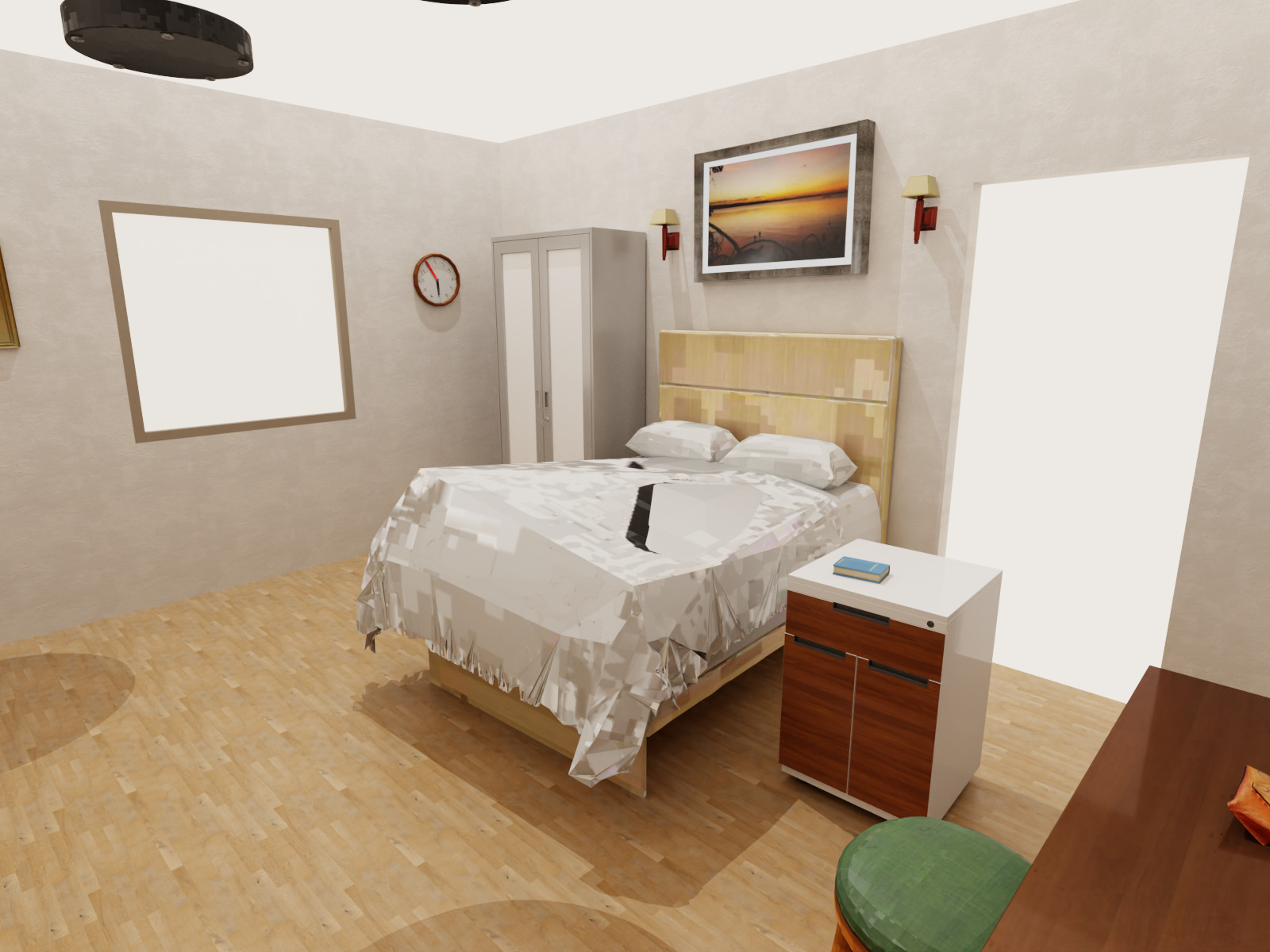} &
\includegraphics[width=.142\textwidth]{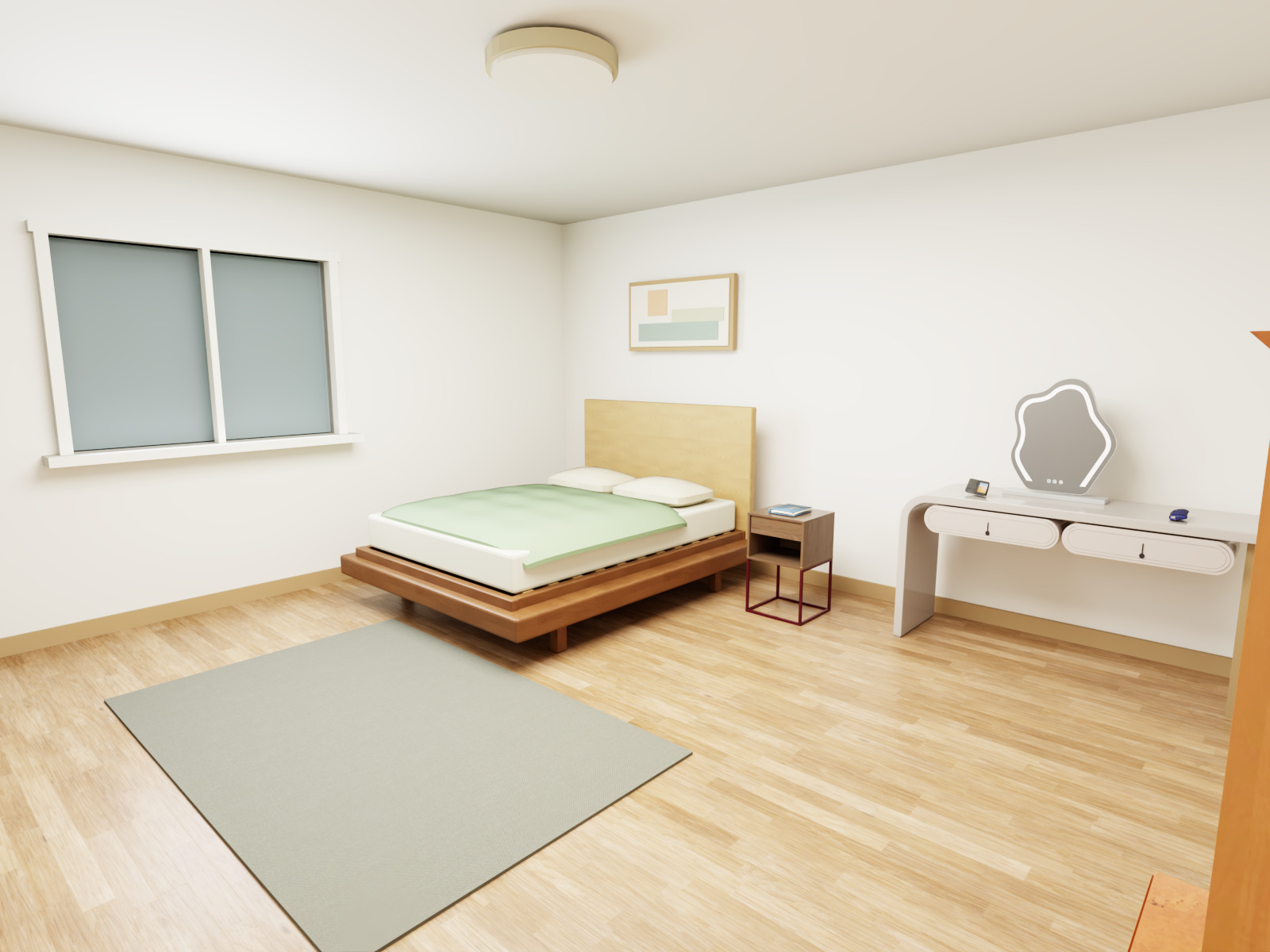} \\[1pt]
 & \tiny 3 native lights & \tiny 2 native lights & \tiny No exported lights & \tiny 7 native lights & \tiny 2 native lights & \tiny 3 native lights \\[4pt]
\rotatebox[origin=c]{90}{\scriptsize M01 / Common light} &
\includegraphics[width=.142\textwidth]{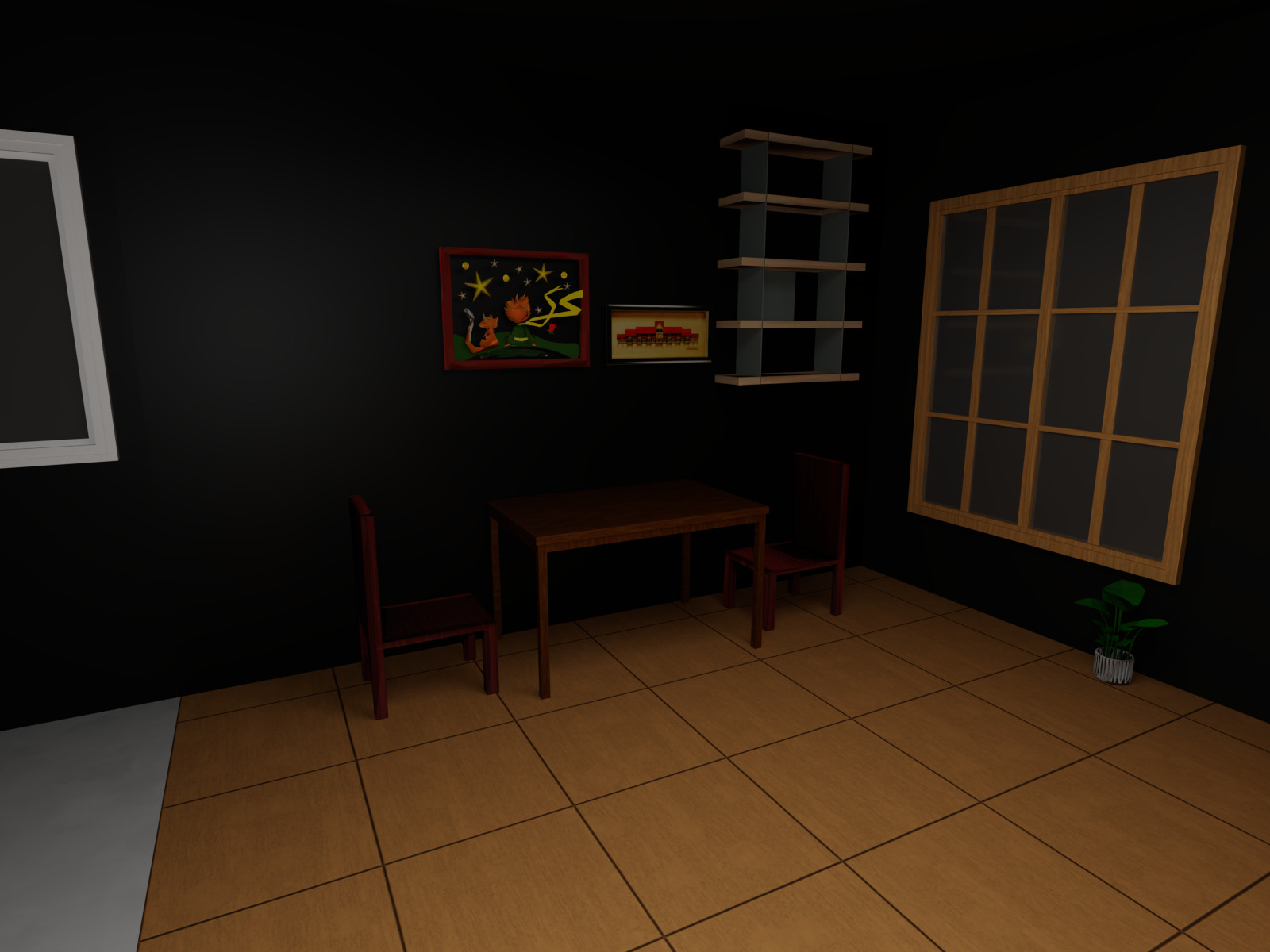} &
\includegraphics[width=.142\textwidth]{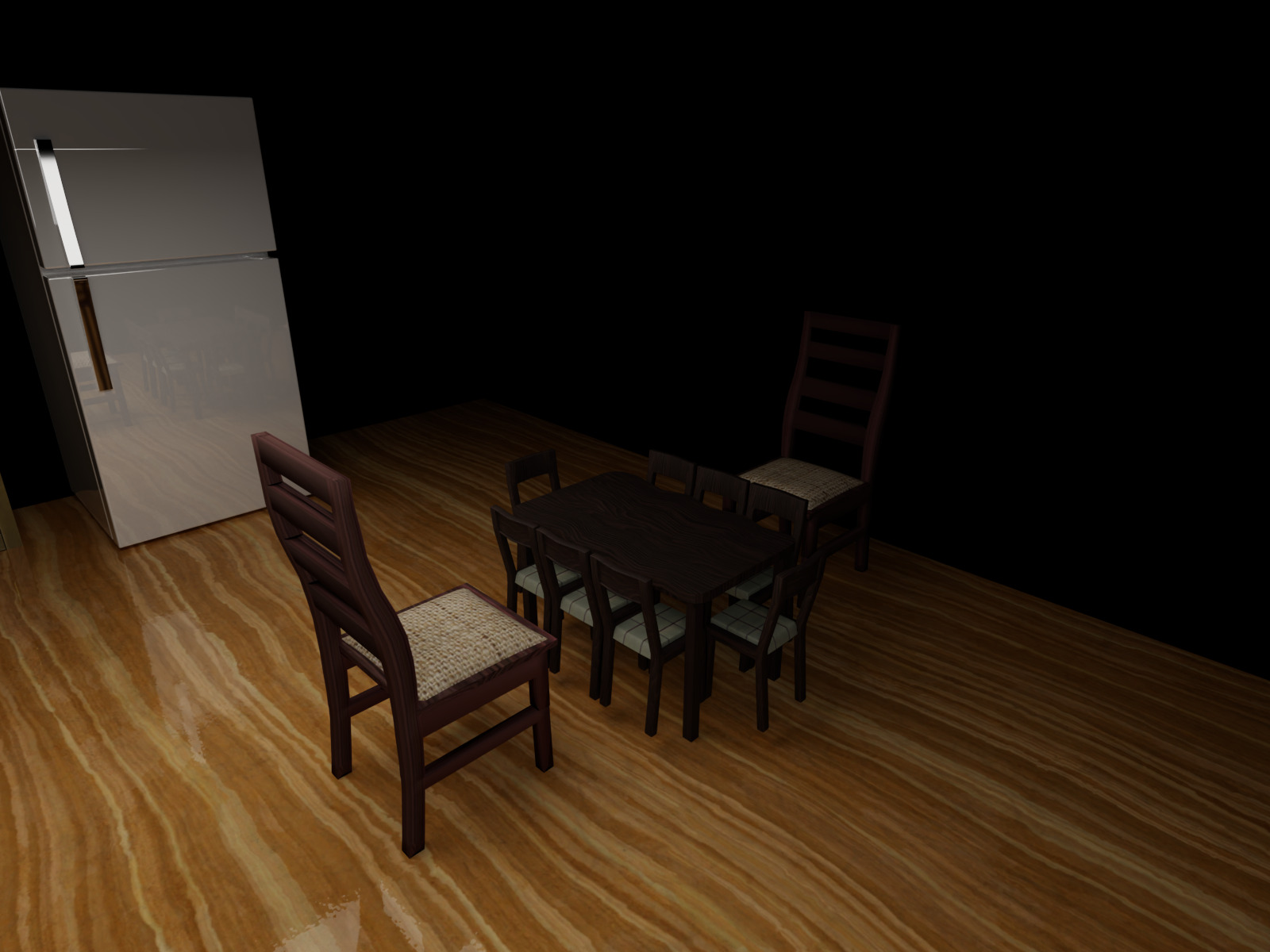} &
\includegraphics[width=.142\textwidth]{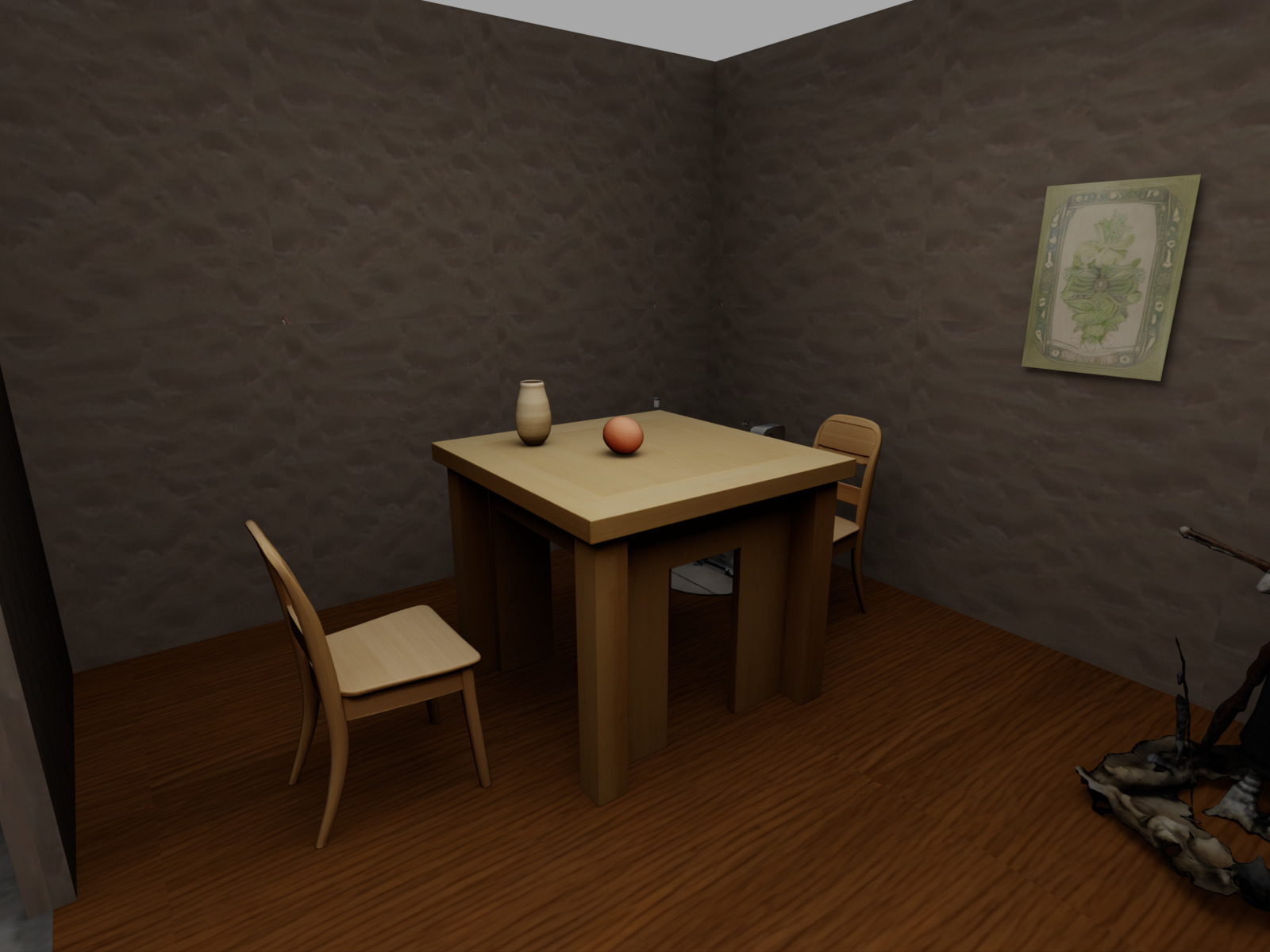} &
\includegraphics[width=.142\textwidth]{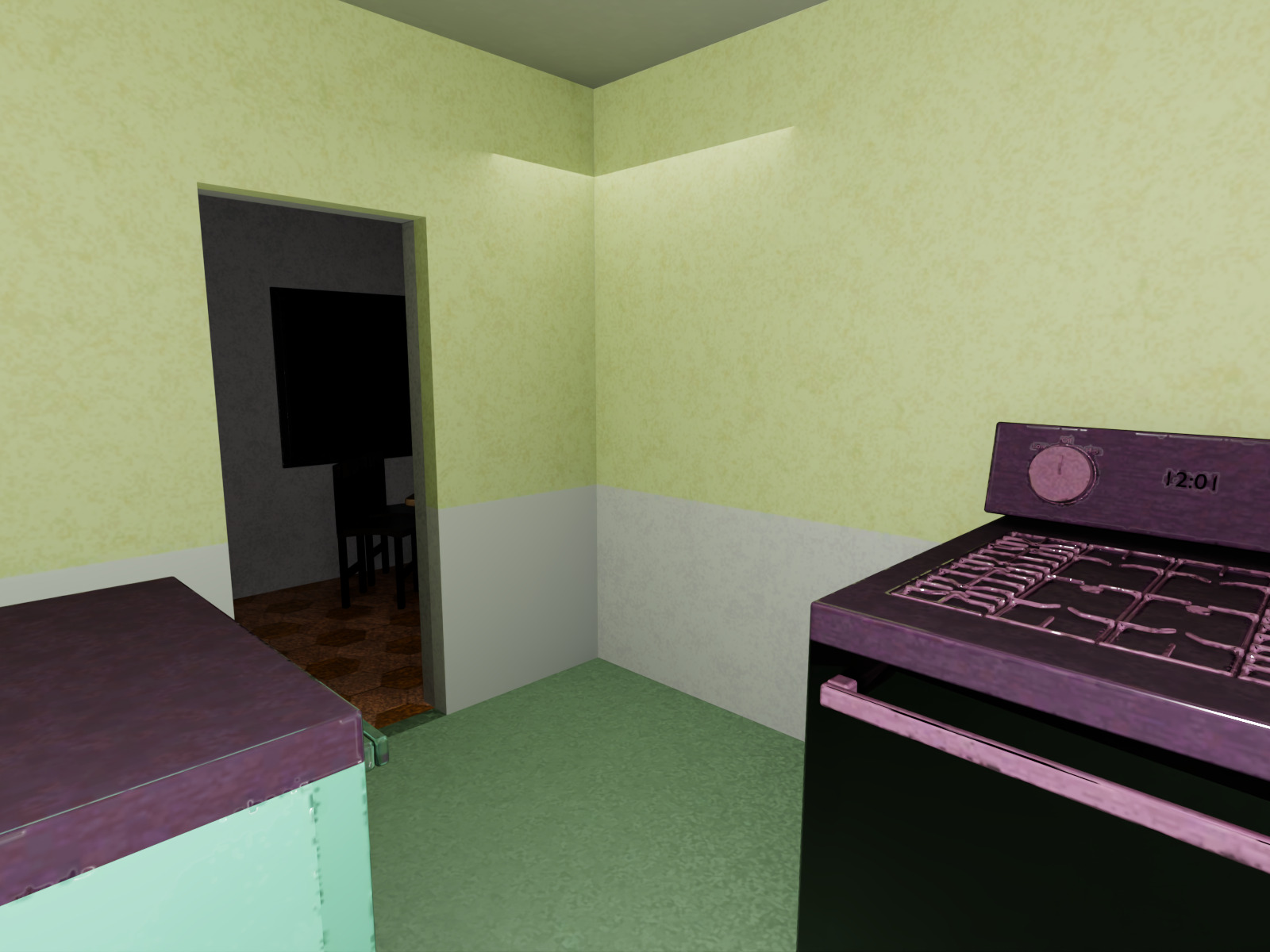} &
\includegraphics[width=.142\textwidth]{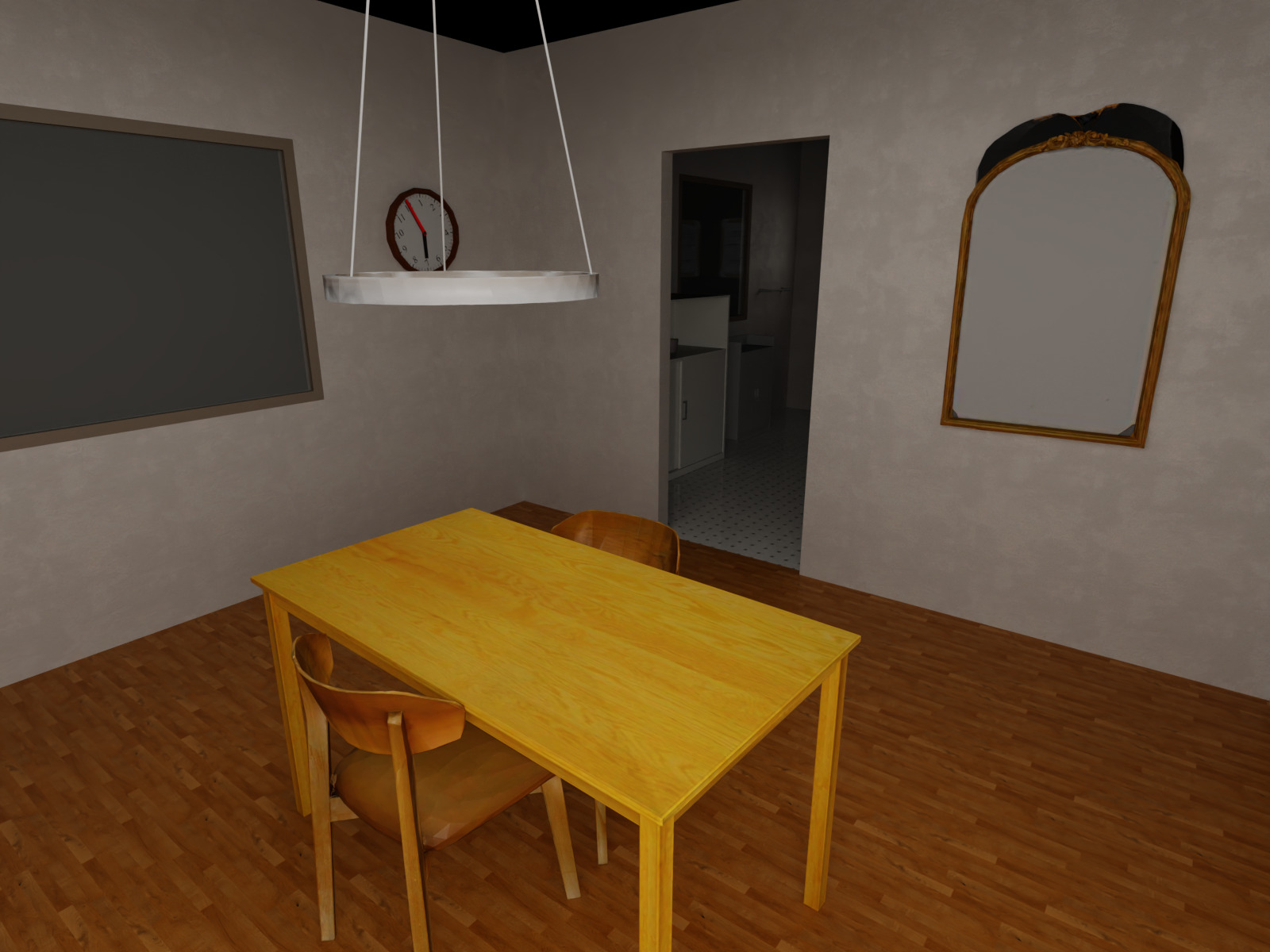} &
\includegraphics[width=.142\textwidth]{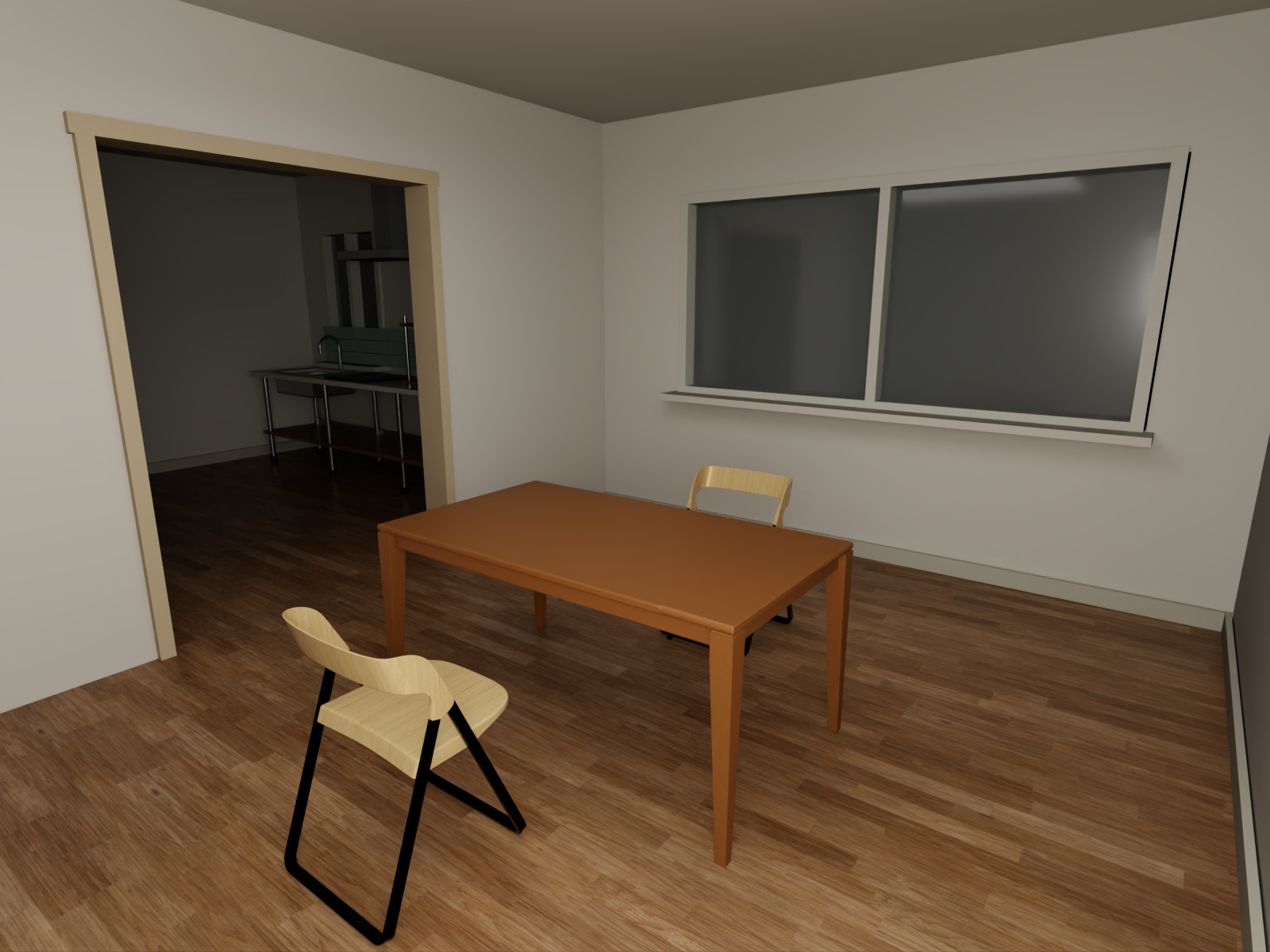} \\[1pt]
\rotatebox[origin=c]{90}{\scriptsize M01 / Native light} &
\includegraphics[width=.142\textwidth]{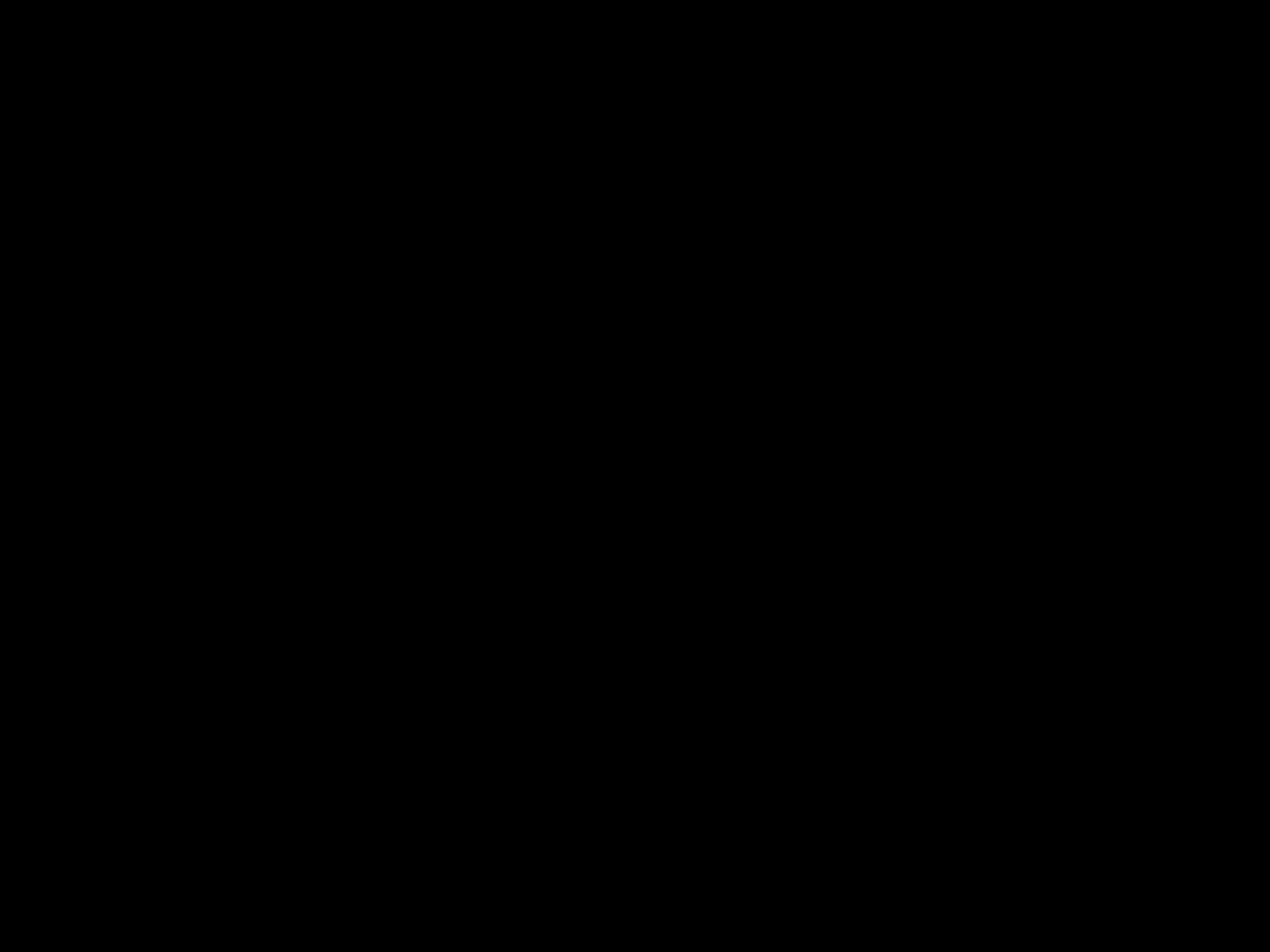} &
\includegraphics[width=.142\textwidth]{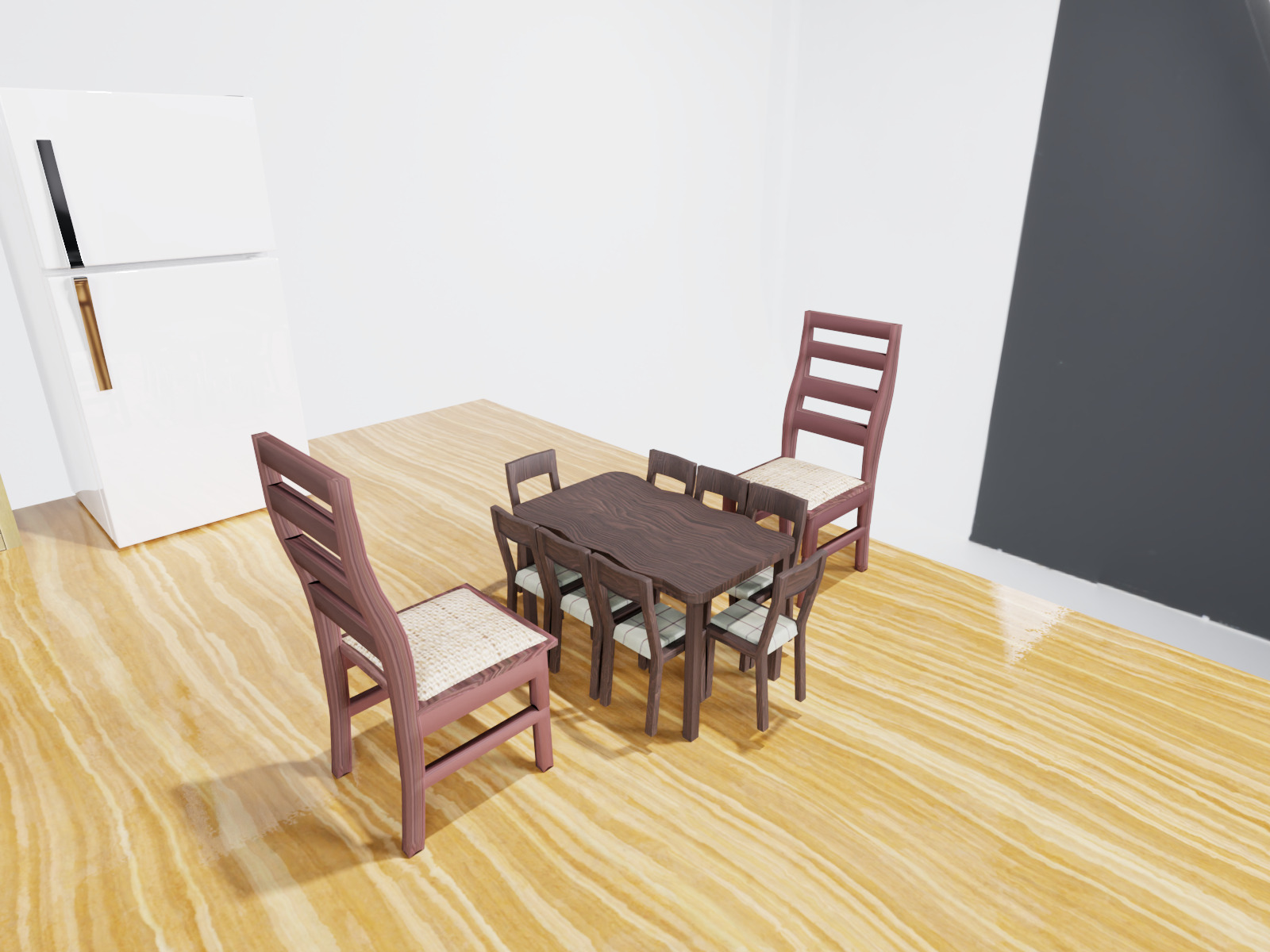} &
\includegraphics[width=.142\textwidth]{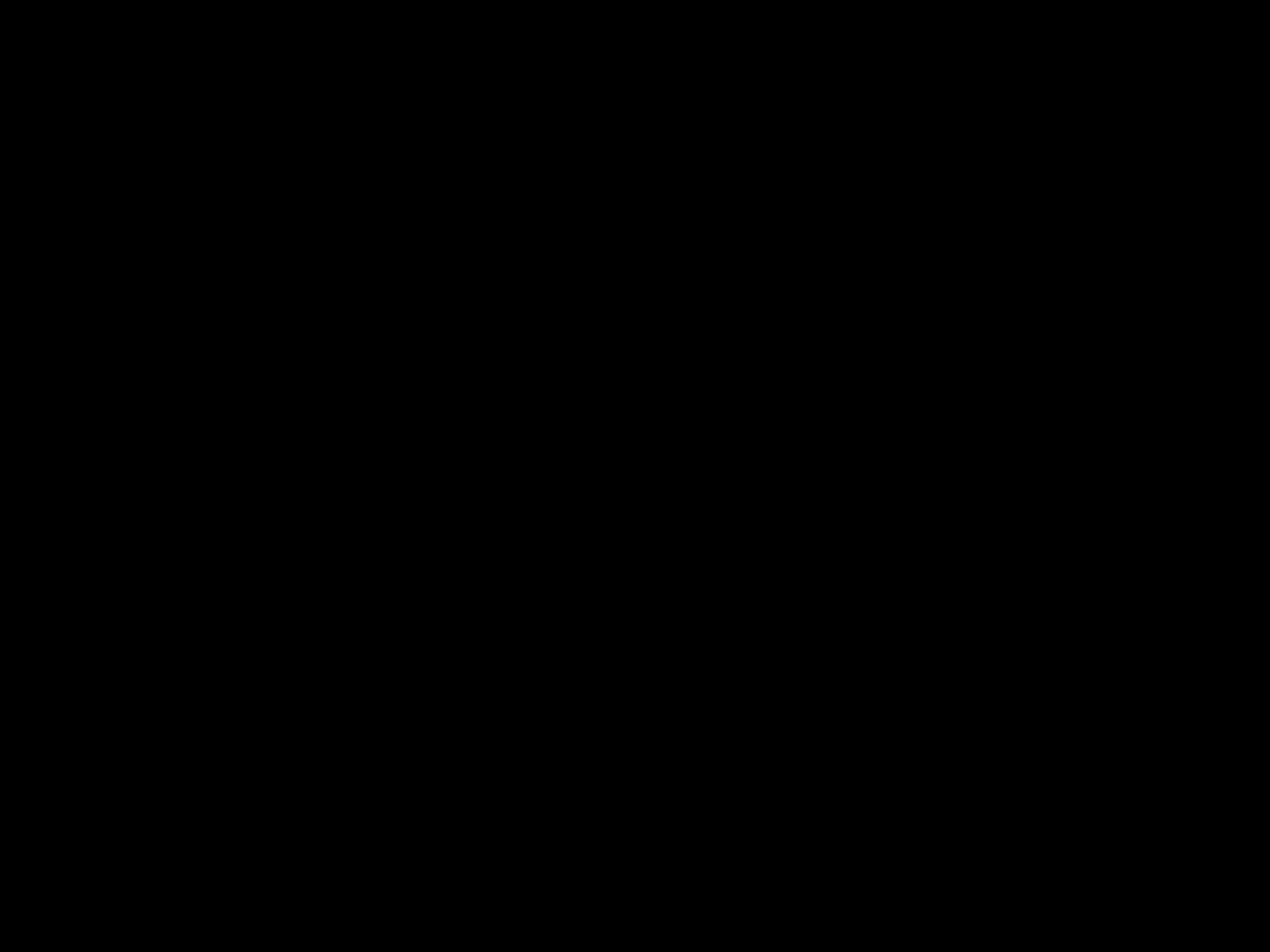} &
\includegraphics[width=.142\textwidth]{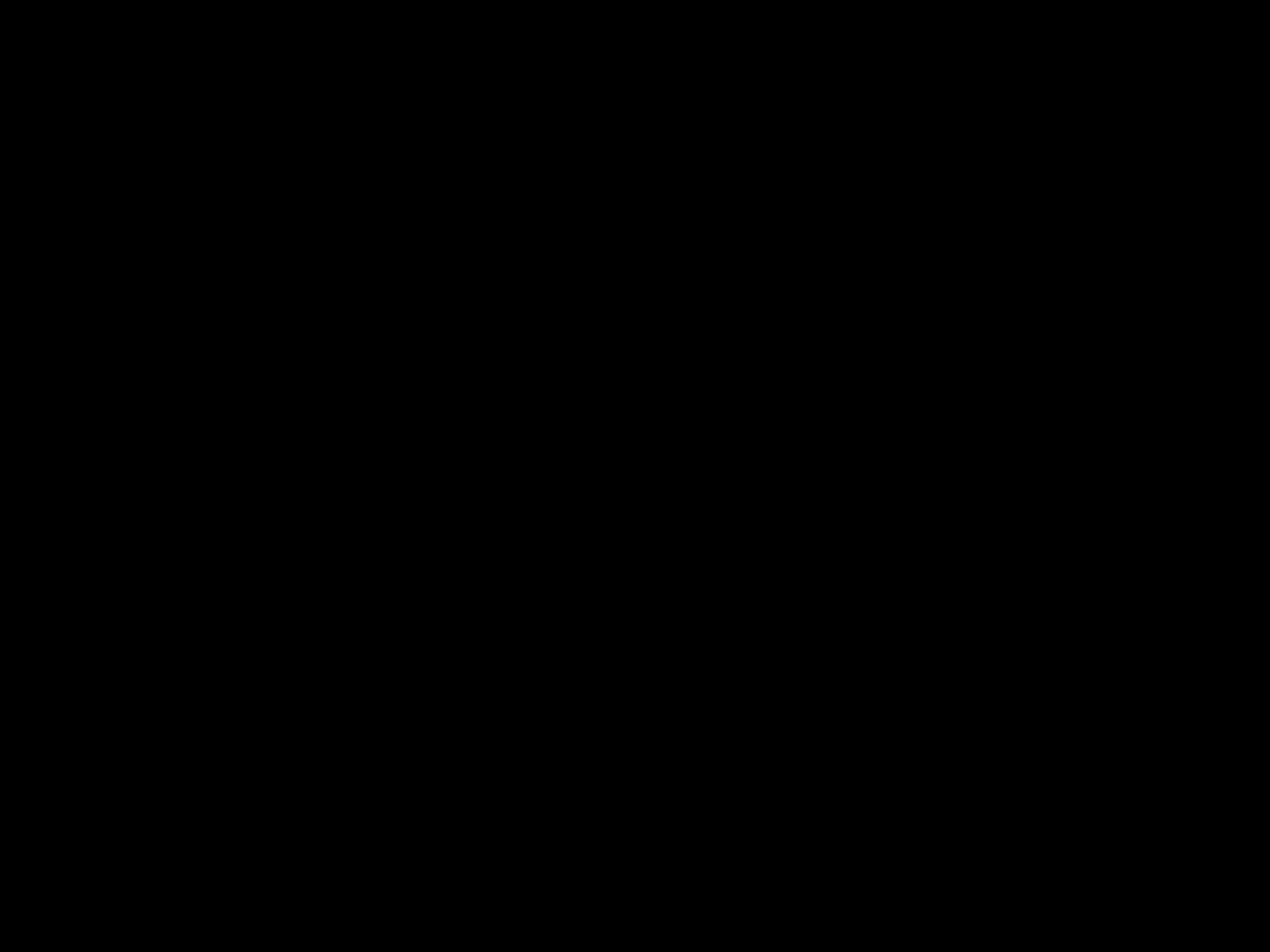} &
\includegraphics[width=.142\textwidth]{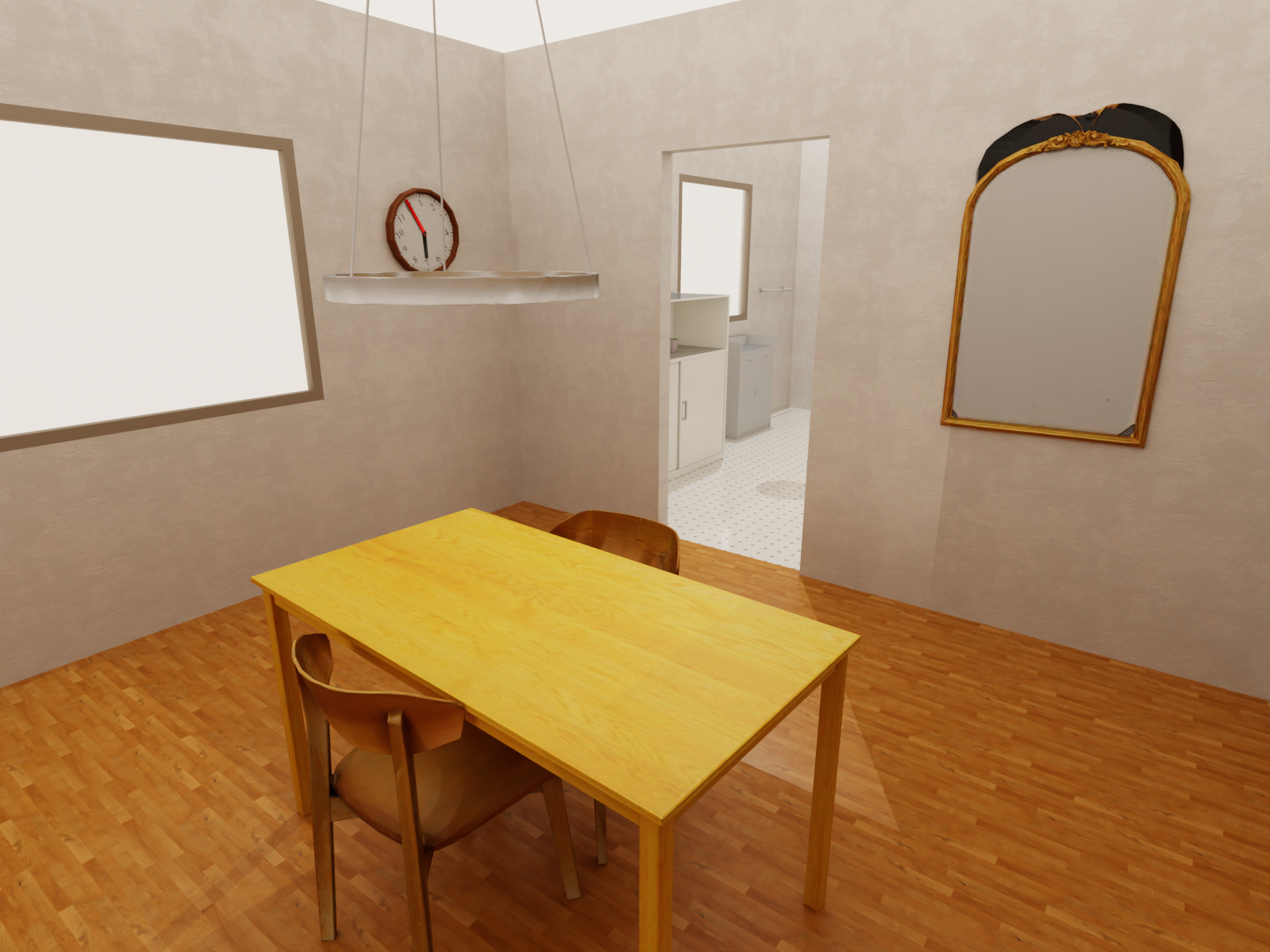} &
\includegraphics[width=.142\textwidth]{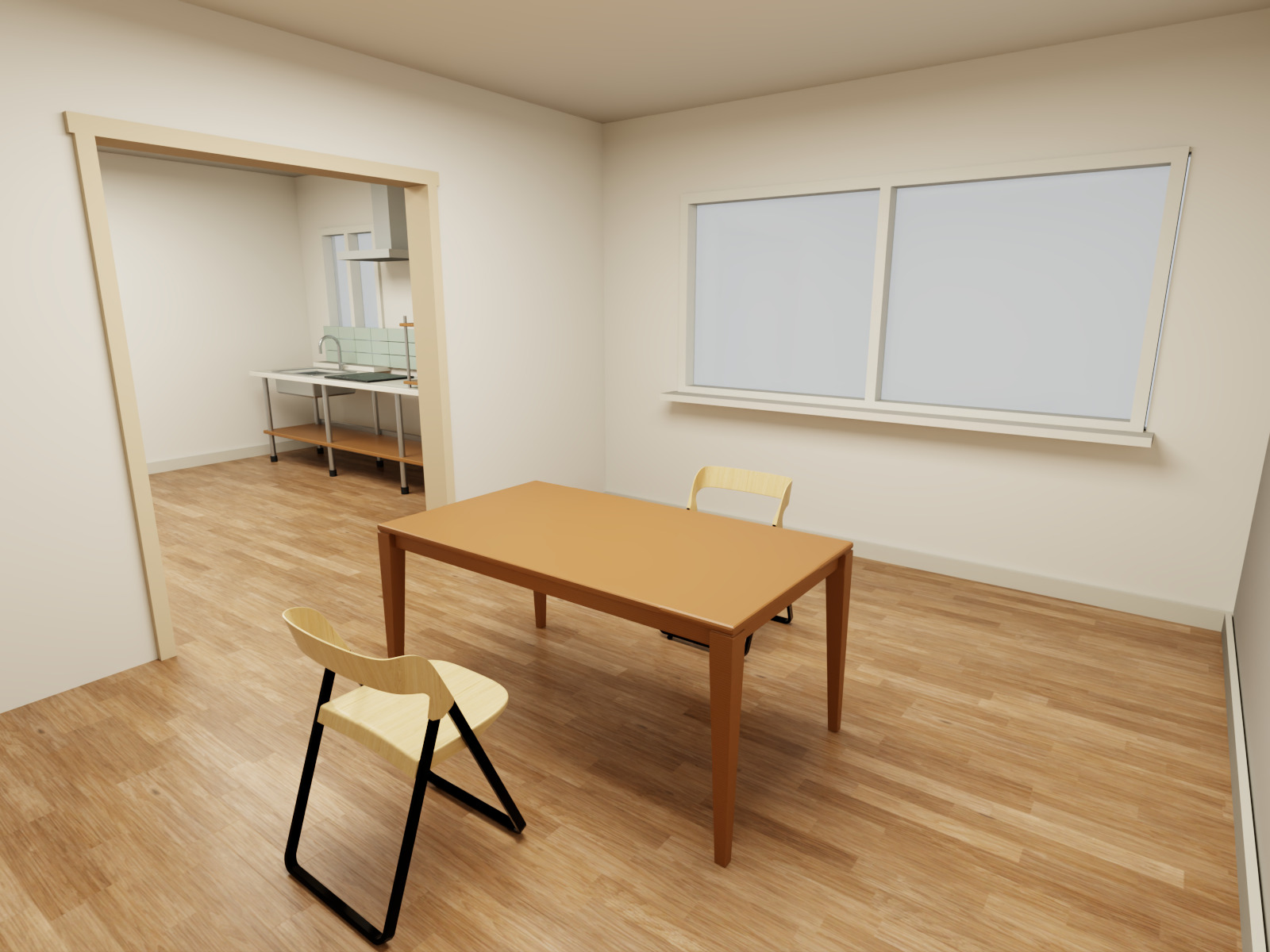} \\[1pt]
 & \tiny 4 native lights & \tiny 2 native lights & \tiny No exported lights & \tiny 12 native lights & \tiny 2 native lights & \tiny 3 native lights \\[4pt]
\end{tabular}
\caption{\textbf{Interior-view scene comparisons under two lighting conditions.} Each task shows common reference lighting (top) and source-native lighting (bottom) from identical cameras, geometry, materials, and exposure settings. Common lighting applies the same ceiling and camera-fill setup across all methods; native lighting renders each method's exported light definitions directly under fixed physical camera exposure without manual rescaling. Consequently, scenes without exported lights (SAGE) appear black, and scenes whose exported light intensities use non-USD/unscaled conventions (Holodeck's unitless Unity intensities of $0.75$--$1.0$ and EmbodiedGen~V2's raw Blender lamp values of $0.25$--$27$) render dark under physical path tracing; primary VIS scoring therefore uses common reference illumination. Tabletop requests use task-level perspective views and room requests use interior cameras without architectural cutaways (1024-sample path tracing at $1600\times1200$).}
\label{fig:supp-more-scenes}
\end{figure*}

\begin{figure*}[p]
\centering
\setlength{\tabcolsep}{3pt}
\begin{tabular}{@{}cc@{}}
\includegraphics[width=.465\textwidth]{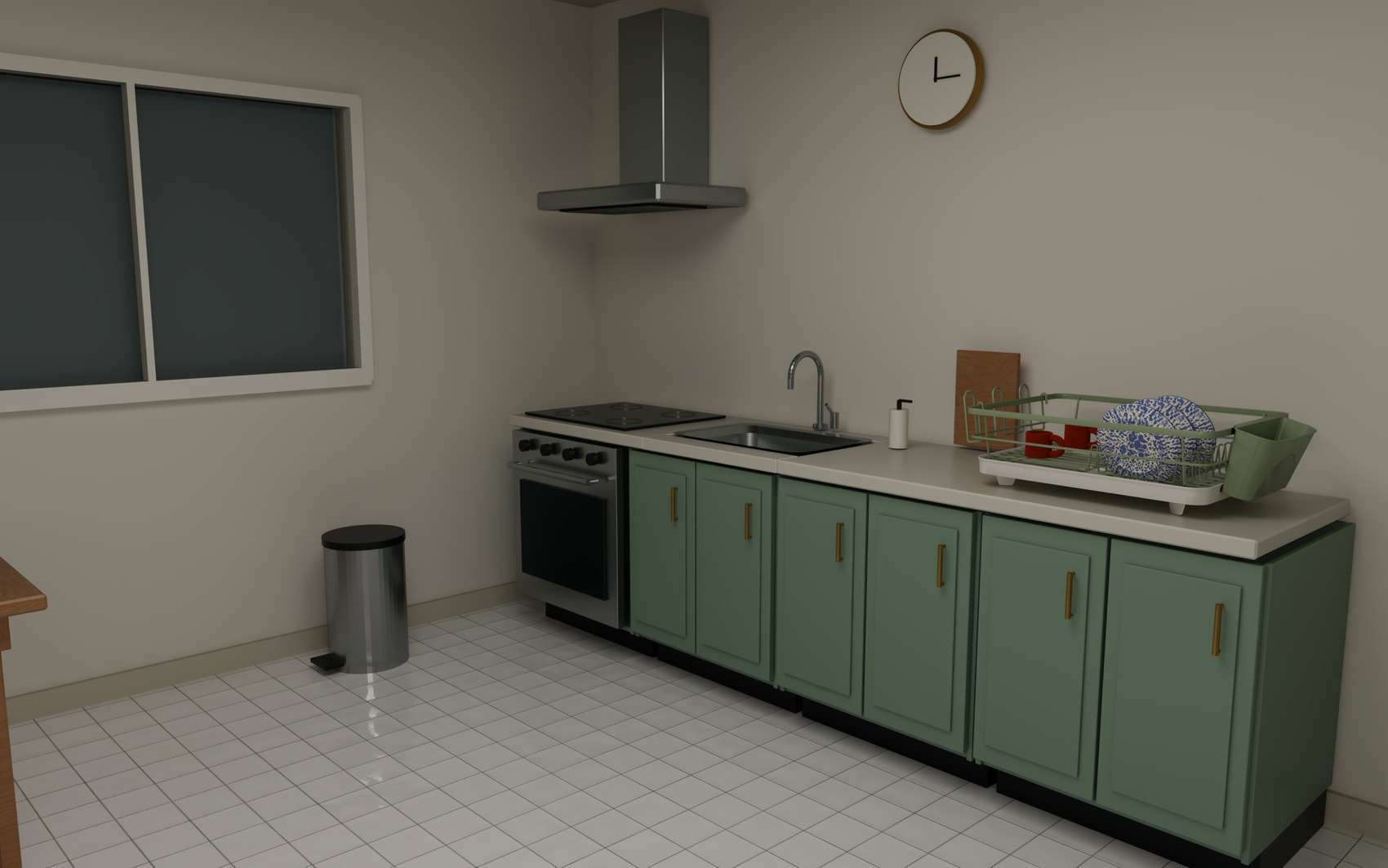} &
\includegraphics[width=.465\textwidth]{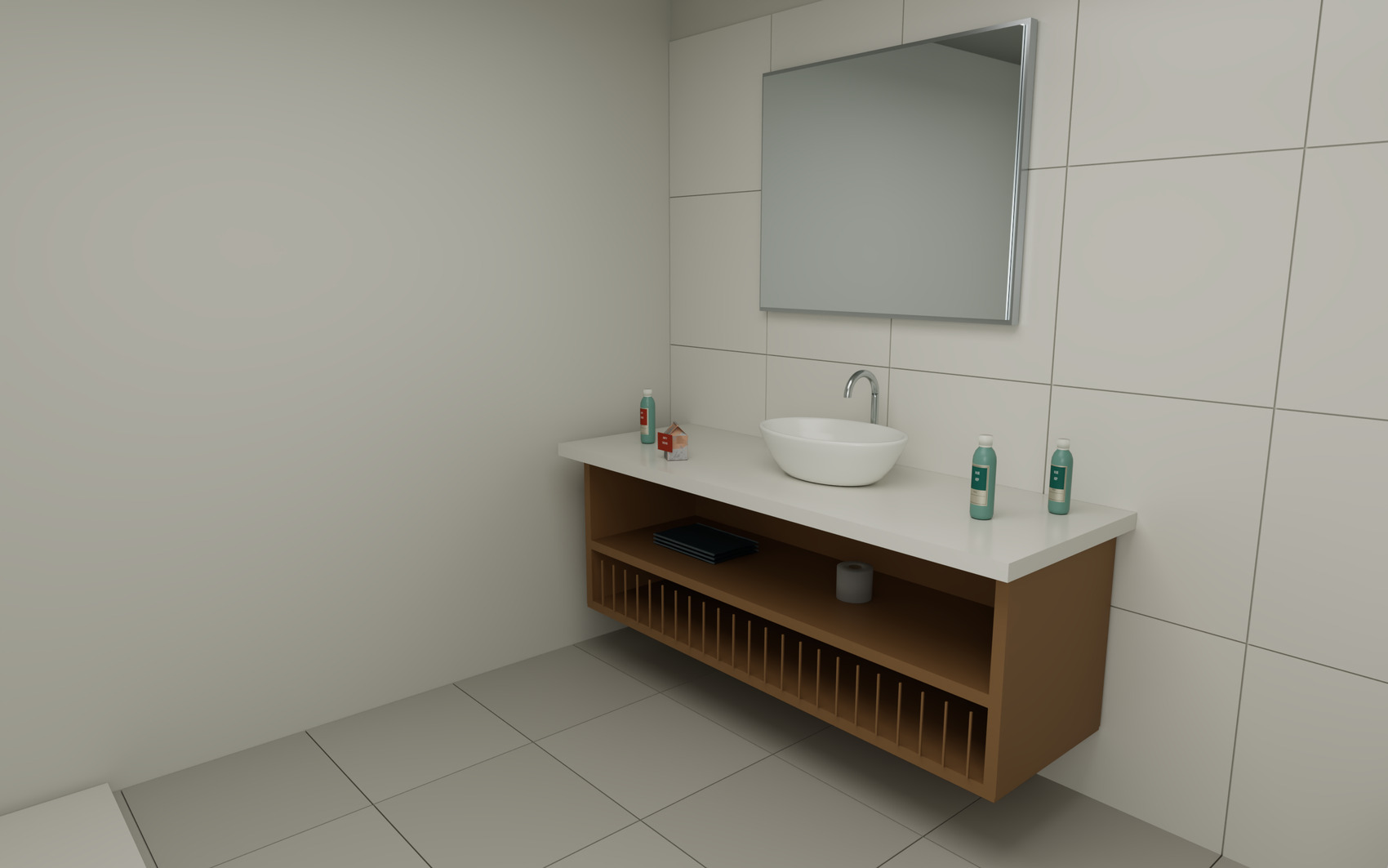} \\
\scriptsize (a) Kitchen: sink, dish rack and storage & \scriptsize (b) Bathroom: washbasin and storage \\[4pt]
\includegraphics[width=.465\textwidth]{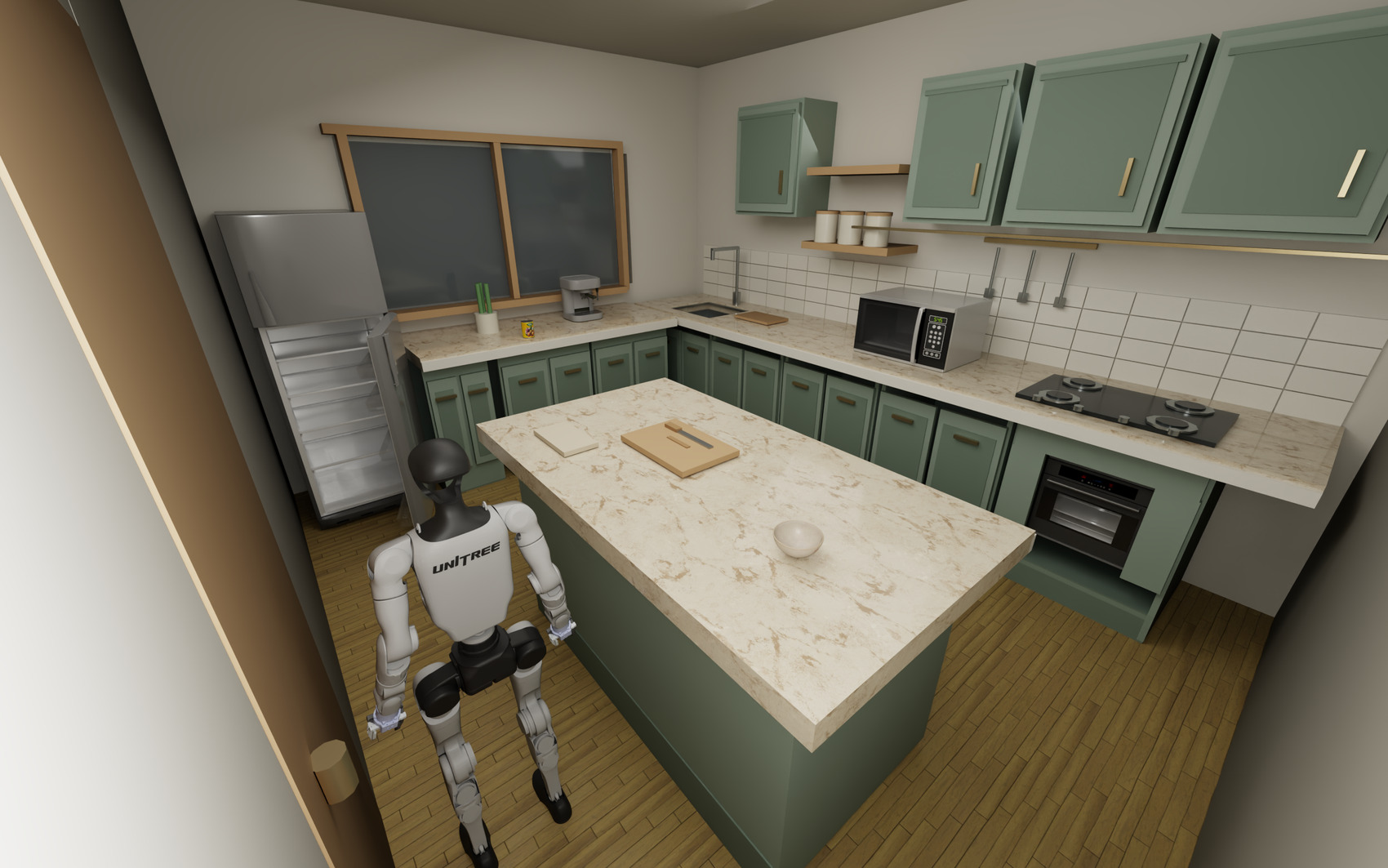} &
\includegraphics[width=.465\textwidth]{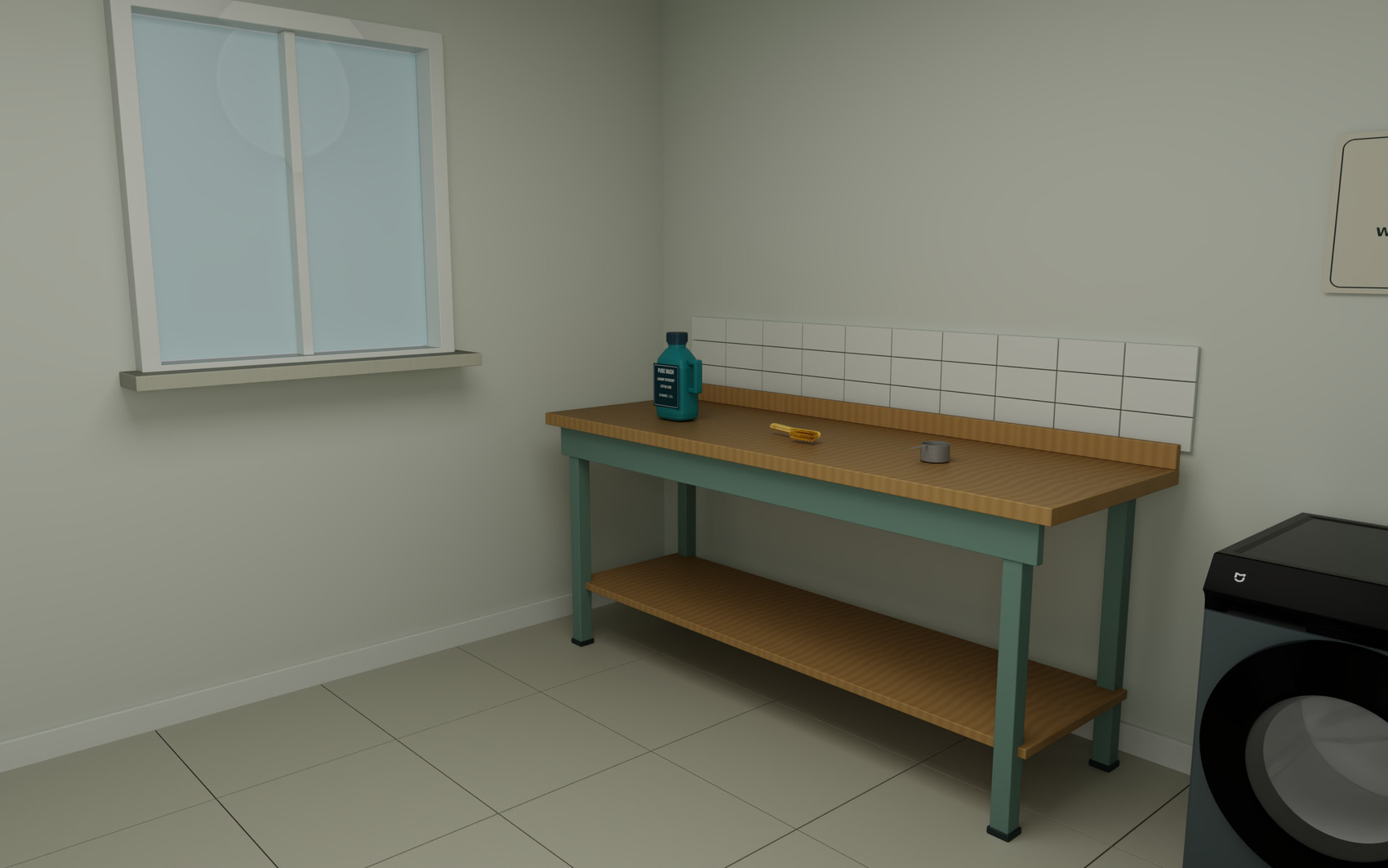} \\
\scriptsize (c) Furnished kitchen with Unitree G1 & \scriptsize (d) Laundry: supplies and work surface \\[4pt]
\includegraphics[width=.465\textwidth]{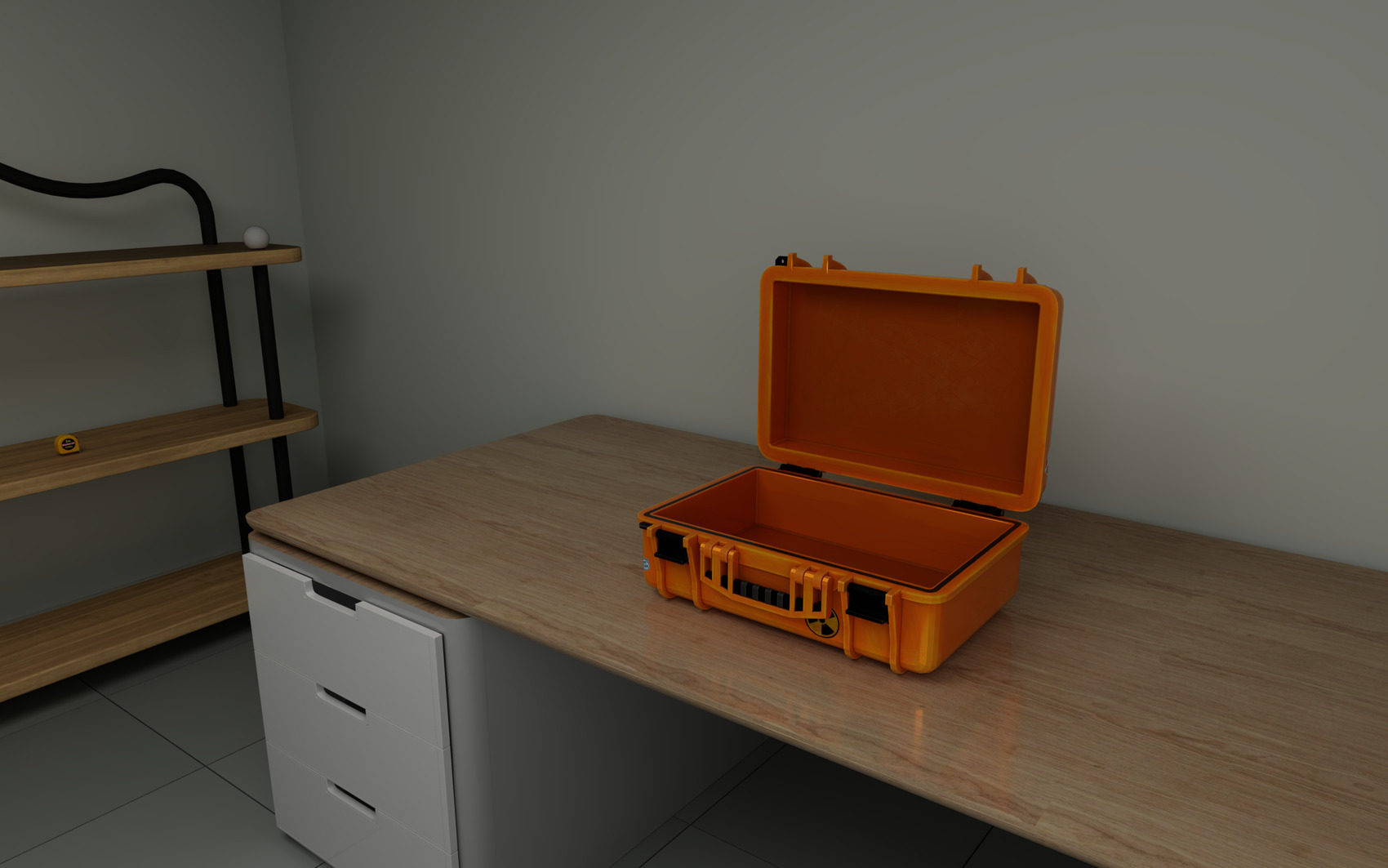} &
\includegraphics[width=.465\textwidth]{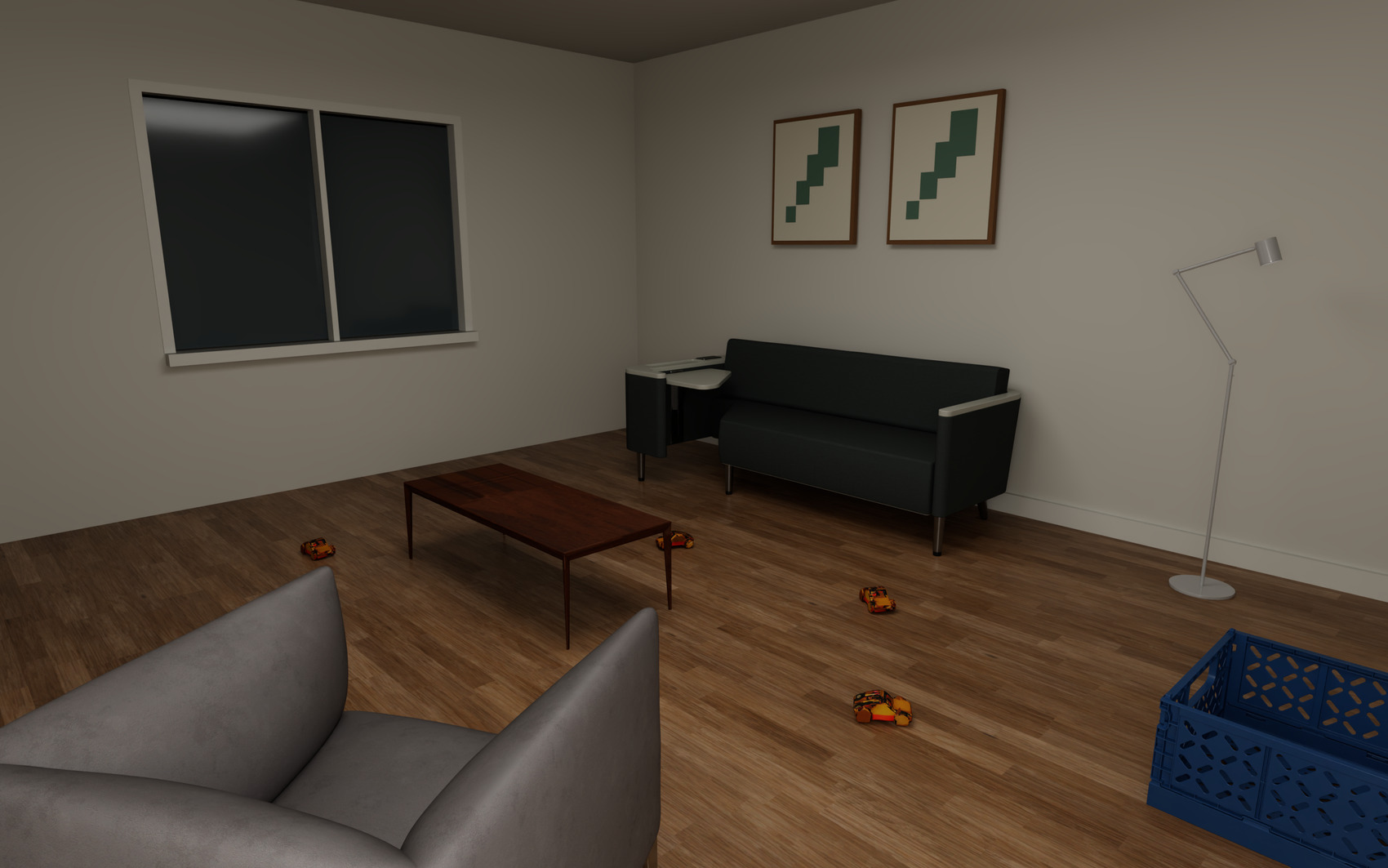} \\
\scriptsize (e) Utility room: tools and packing area & \scriptsize (f) Living room: toys, seating and storage \\[4pt]
\end{tabular}
\caption{\textbf{High-quality interior close views of functional scene areas.} Six generated environments rendered from interior viewpoints with surrounding surfaces, storage, and furnishings intact (no architectural cutaways). Panel~(c) shows a furnished kitchen with Unitree~G1 in its initial pose; panel~(f) shows the living-room toy-cleanup environment. All images use 1024-sample path tracing at $2560\times1600$ (12 diffuse bounces, up to 16 specular/transmission bounces).}
\label{fig:supp-closeup}
\end{figure*}

To complement the main-paper comparisons (\texttt{T11}, \texttt{R02}, \texttt{M02}), Figure~\ref{fig:supp-more-scenes} compares all six scene-generation methods across three additional requests spanning tabletop (\texttt{T01}: place the target apple in a bowl), single-room (\texttt{R14}: organize bedside items), and multi-room (\texttt{M01}: set a dining table for two) scales. For each method and task, two identical interior views compare common reference illumination with source-native illumination. Figure~\ref{fig:supp-closeup} presents high-quality path-traced interior close views of six functional areas: two kitchens (including the furnished Unitree G1 example), a bathroom, a laundry room, a utility room, and a living room. These complement the complete-layout galleries without repeating isolated asset enlargements.

\subsection{Twelve Additional Frozen Environments}
\label{sec:supp-new-gallery}
\begin{figure*}[p]
\centering
\setlength{\tabcolsep}{3pt}
\begin{tabular}{@{}cc@{}}
\includegraphics[width=.465\textwidth]{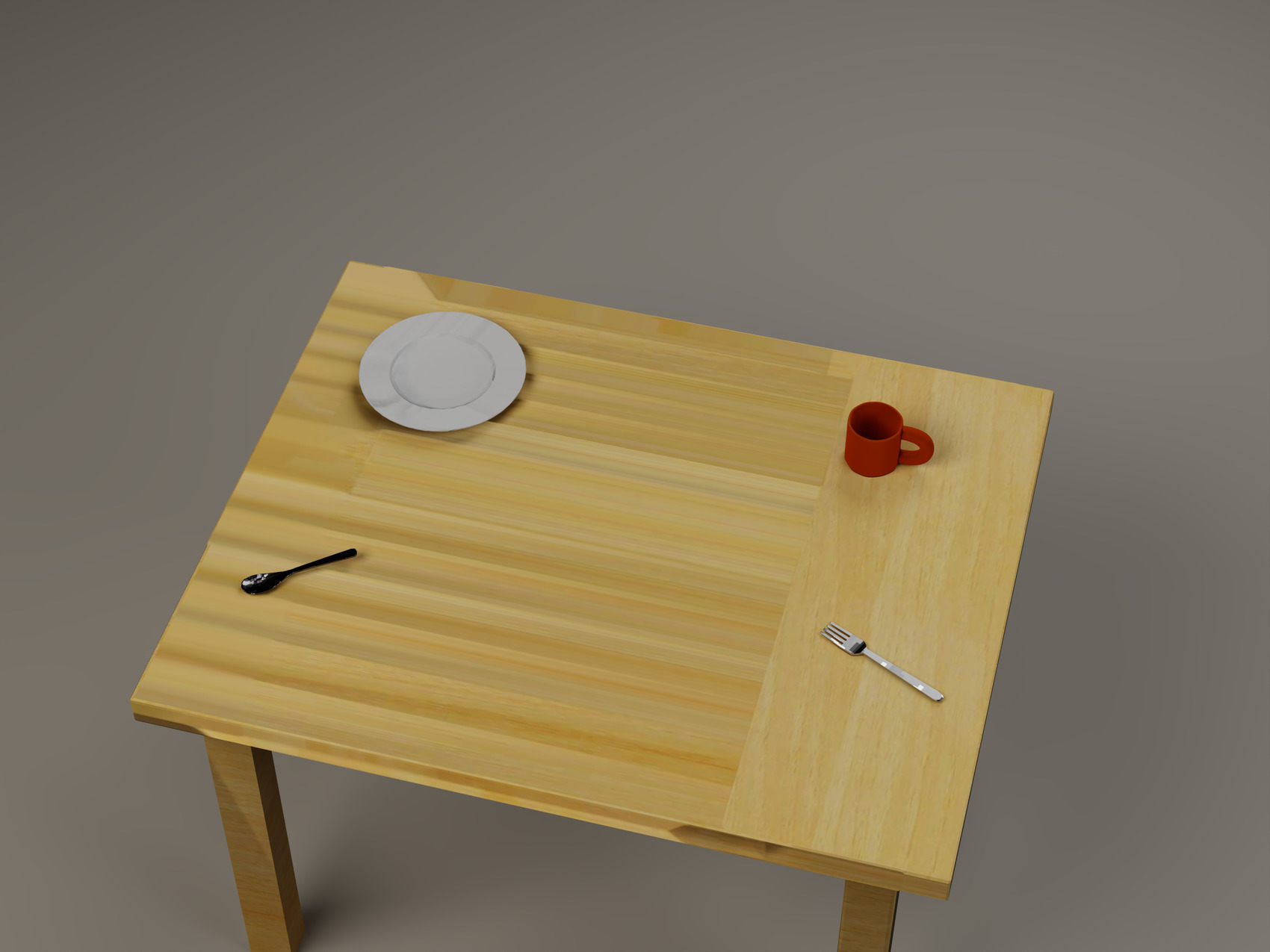} &
\includegraphics[width=.465\textwidth]{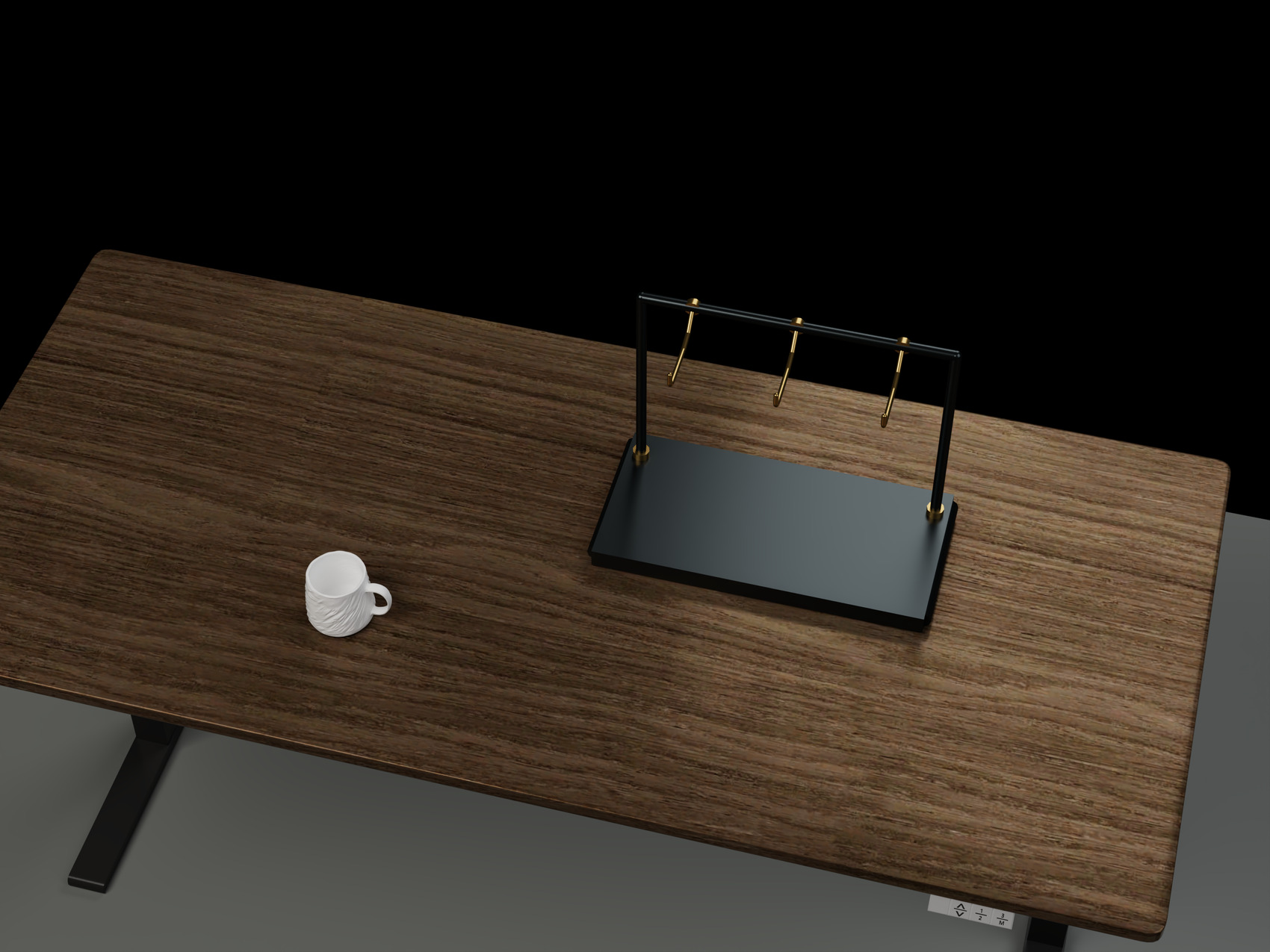} \\
\scriptsize\textbf{T03}: Set a place for one person & \scriptsize\textbf{T06}: Hang a mug on a rack \\[5pt]
\includegraphics[width=.465\textwidth]{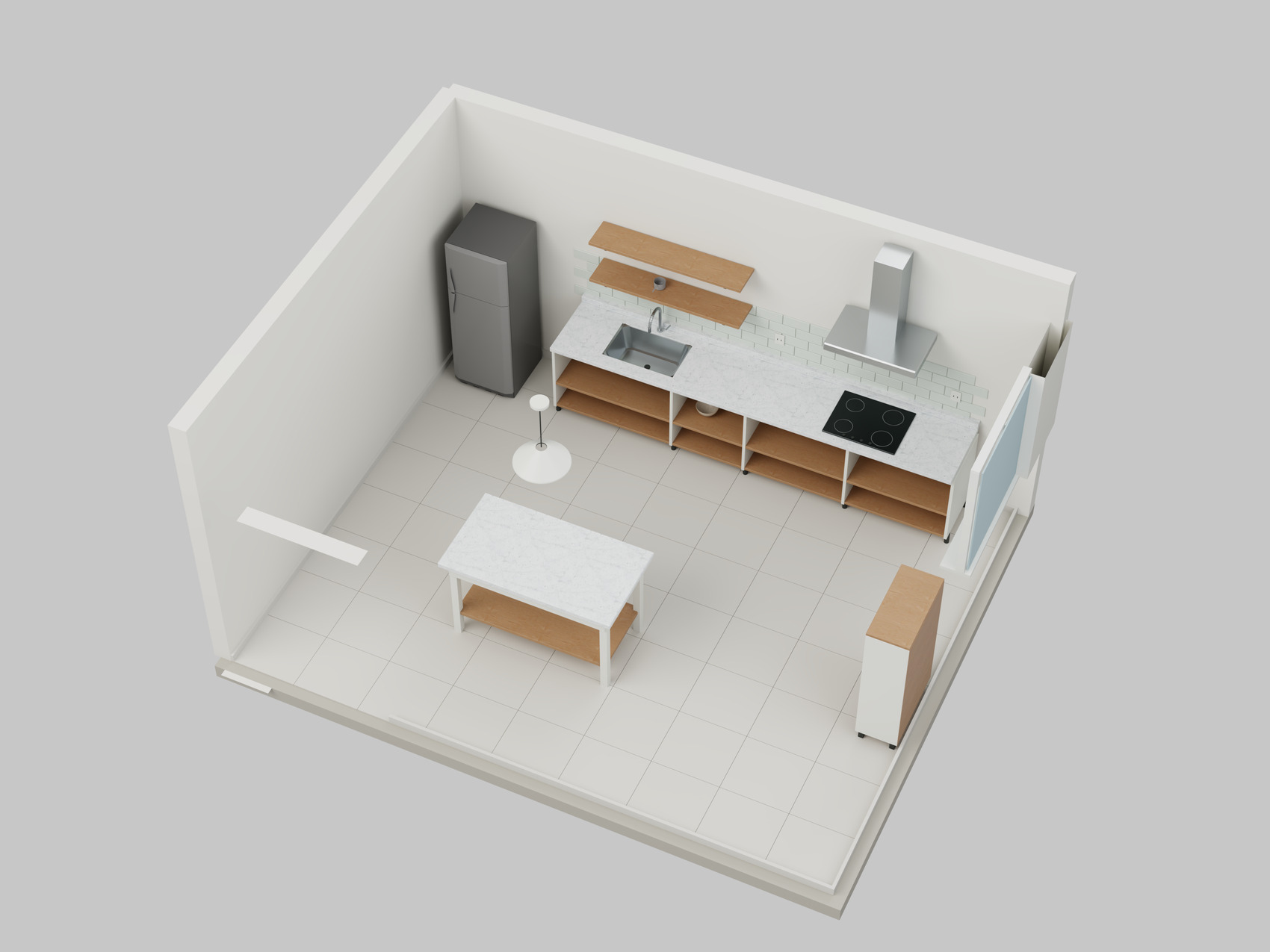} &
\includegraphics[width=.465\textwidth]{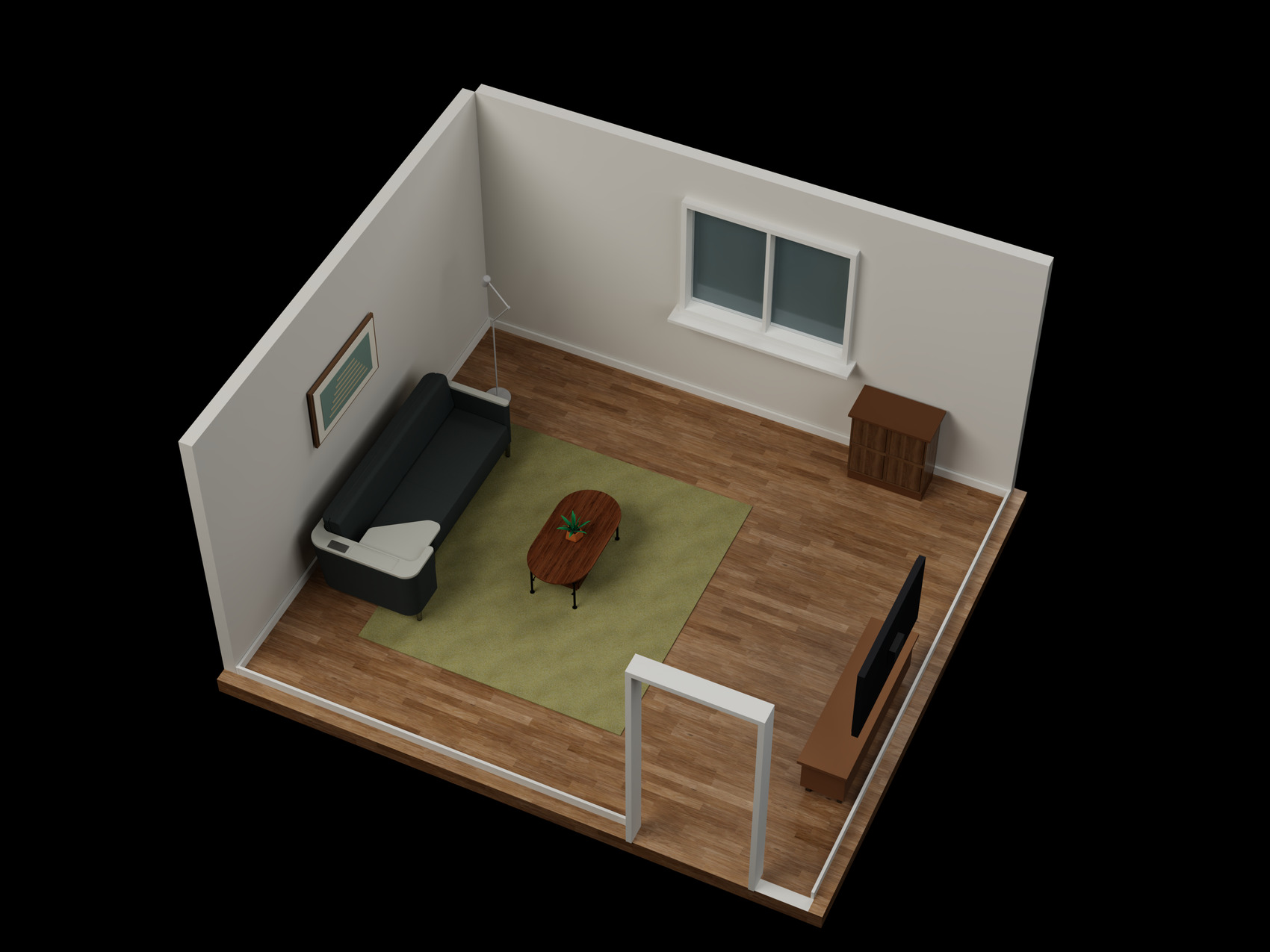} \\
\scriptsize\textbf{R04}: Gather baking utensils & \scriptsize\textbf{R11}: Move a potted plant to a window-side cabinet \\[5pt]
\includegraphics[width=.465\textwidth]{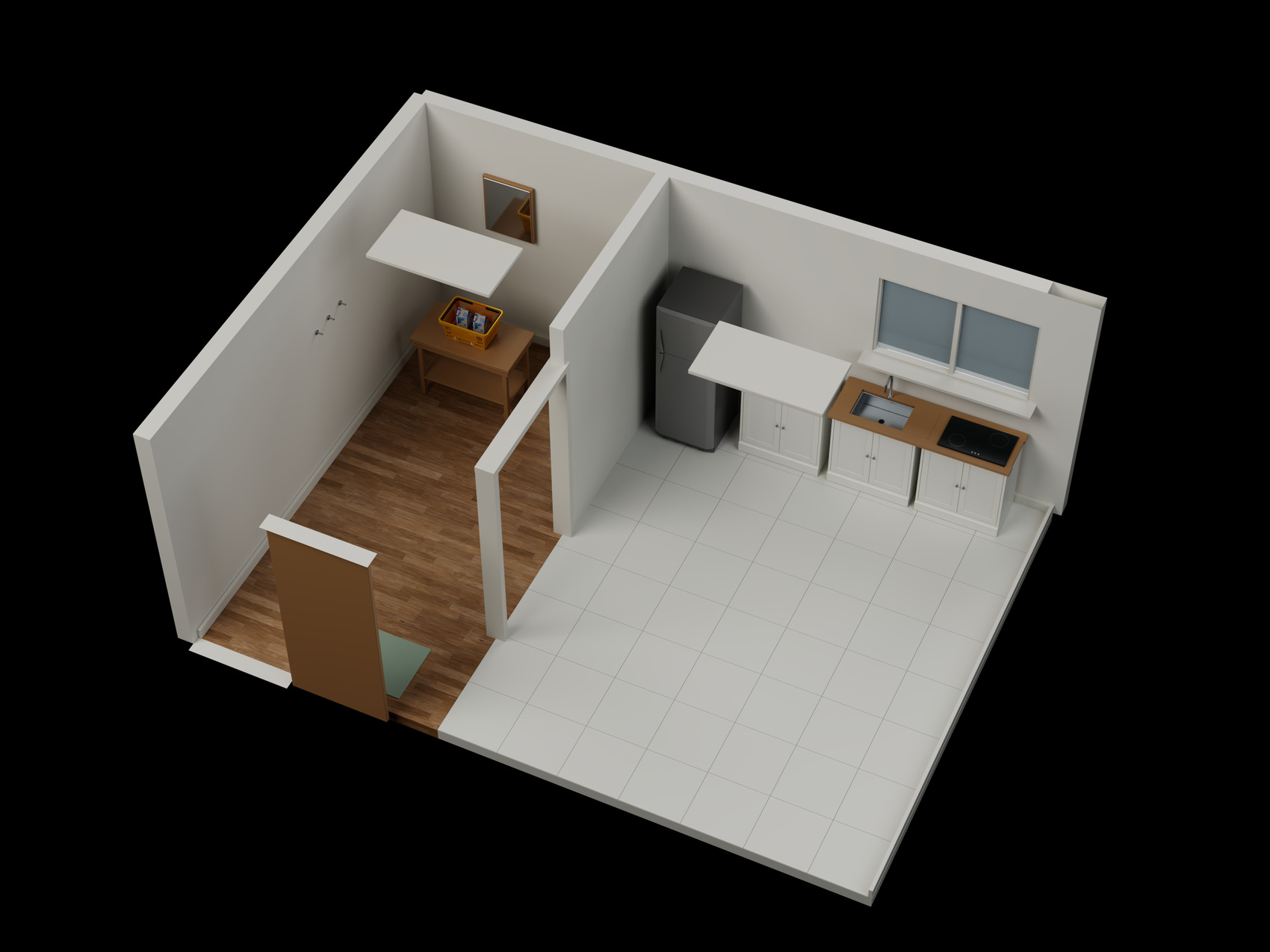} &
\includegraphics[width=.465\textwidth]{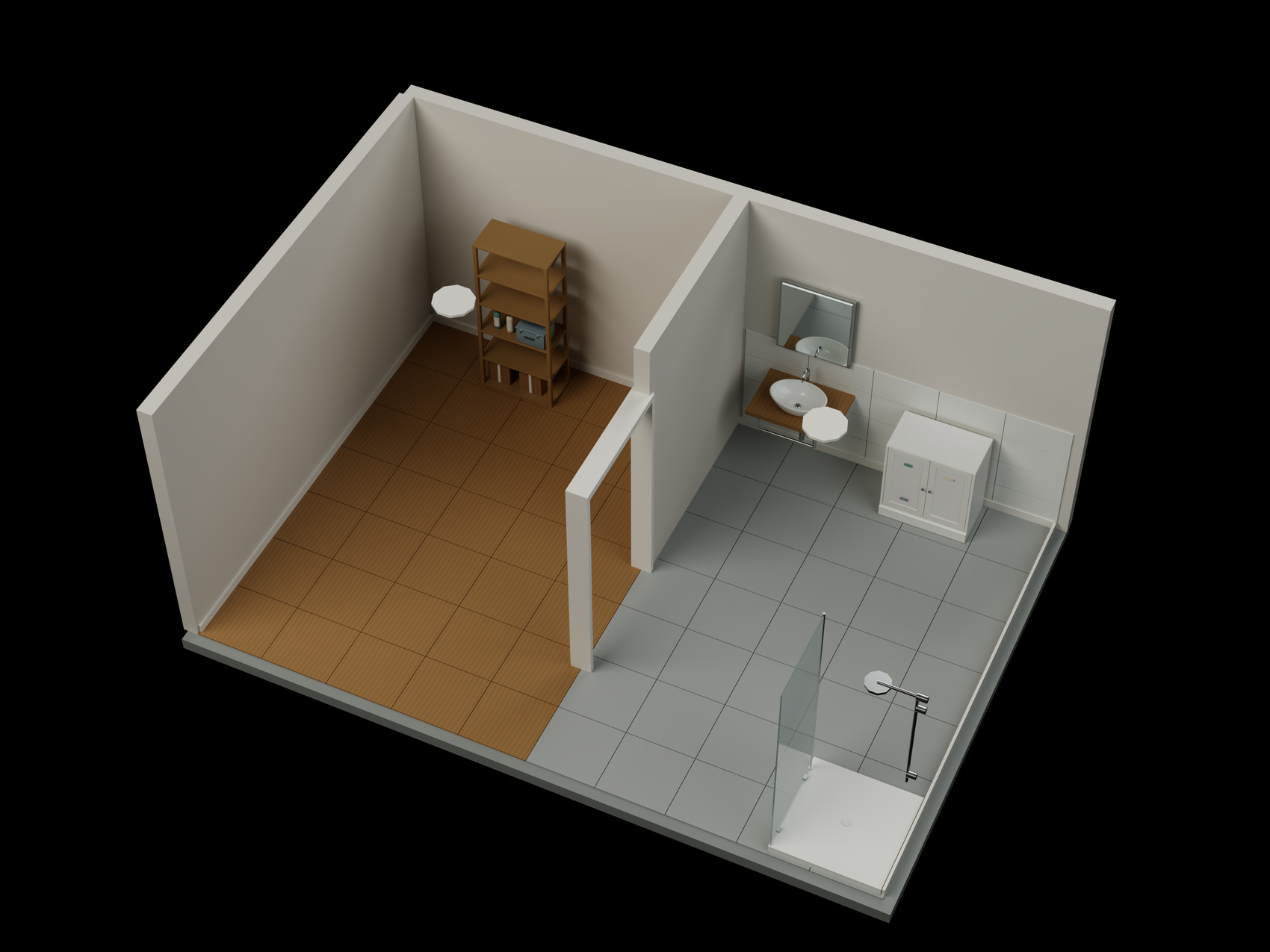} \\
\scriptsize\textbf{M04}: Put away groceries from the entrance & \scriptsize\textbf{M06}: Restock the bathroom from storage \\[5pt]
\end{tabular}
\caption{\textbf{Additional generated environments (a).} Tabletop details (top), single-room interiors (middle), and multi-room cutaways (bottom), rendered from the frozen pre-task scenes at $3000\times2250$ with 1024 path-tracing samples per pixel (camera-facing walls and ceilings hidden for visibility).}
\label{fig:supp-new-gallery-a}
\end{figure*}

\begin{figure*}[p]
\centering
\setlength{\tabcolsep}{3pt}
\begin{tabular}{@{}cc@{}}
\includegraphics[width=.465\textwidth]{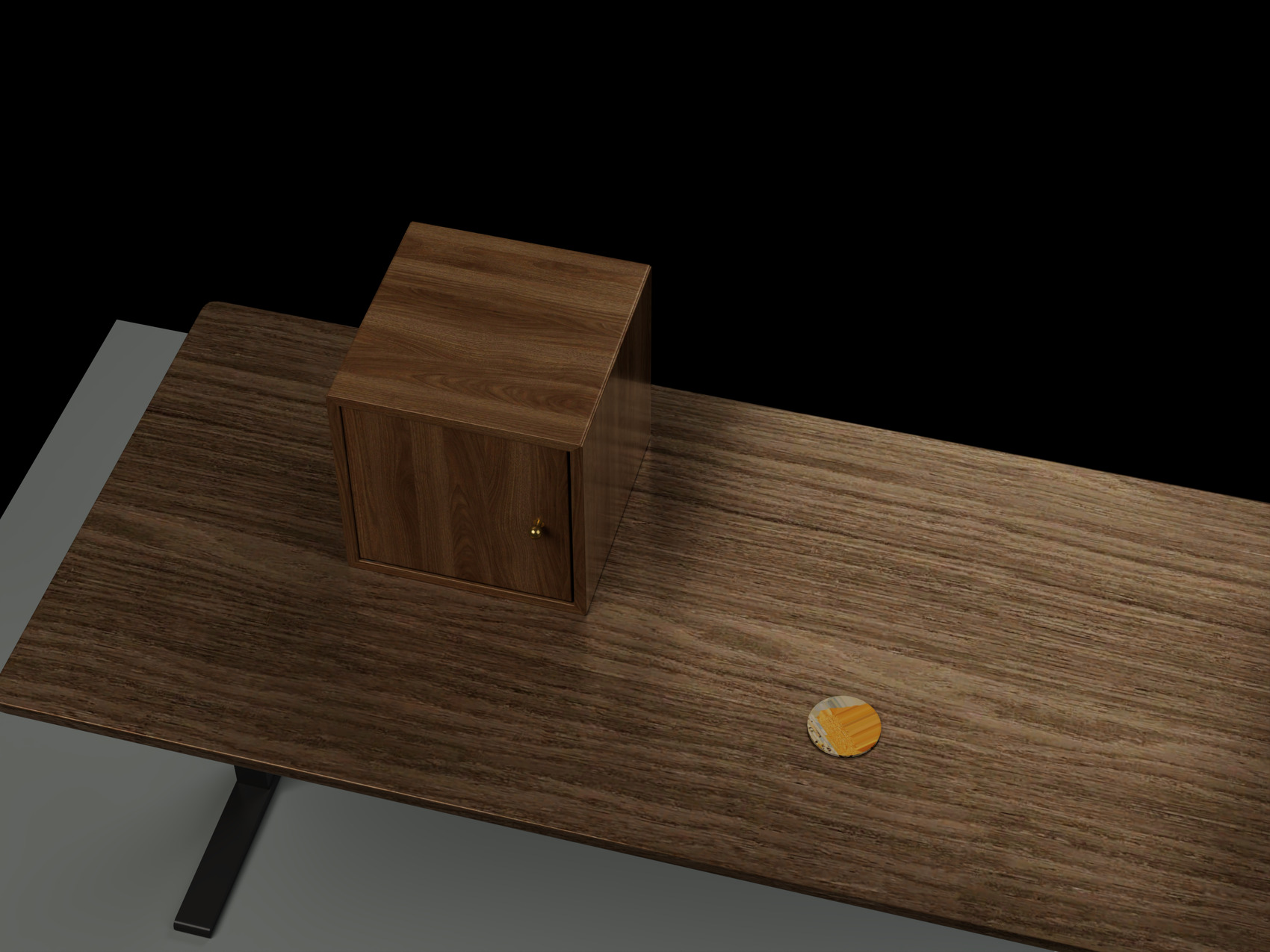} &
\includegraphics[width=.465\textwidth]{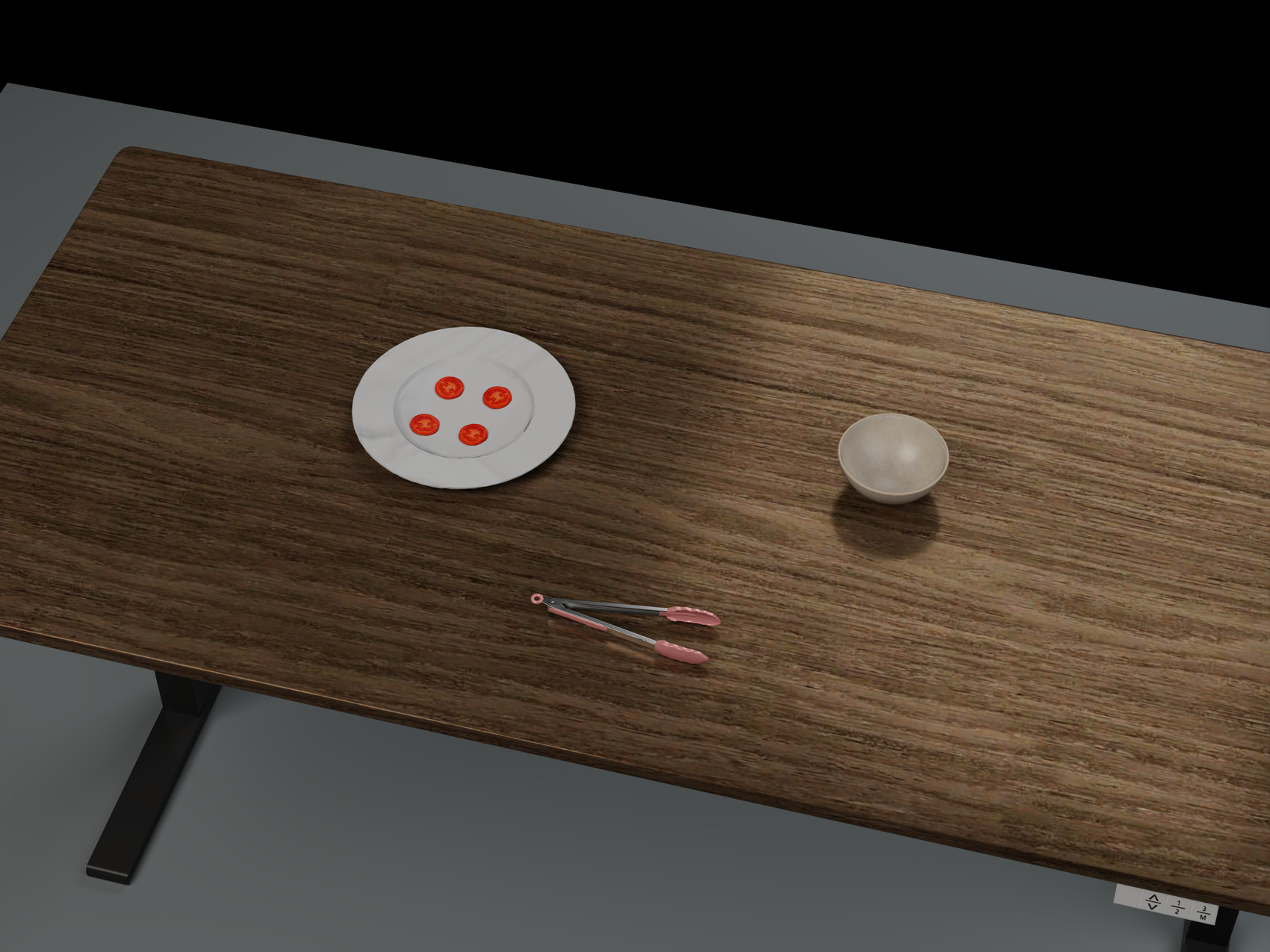} \\
\scriptsize\textbf{T08}: Retrieve a cup from a small cabinet & \scriptsize\textbf{T13}: Transfer objects using kitchen tongs \\[5pt]
\includegraphics[width=.465\textwidth]{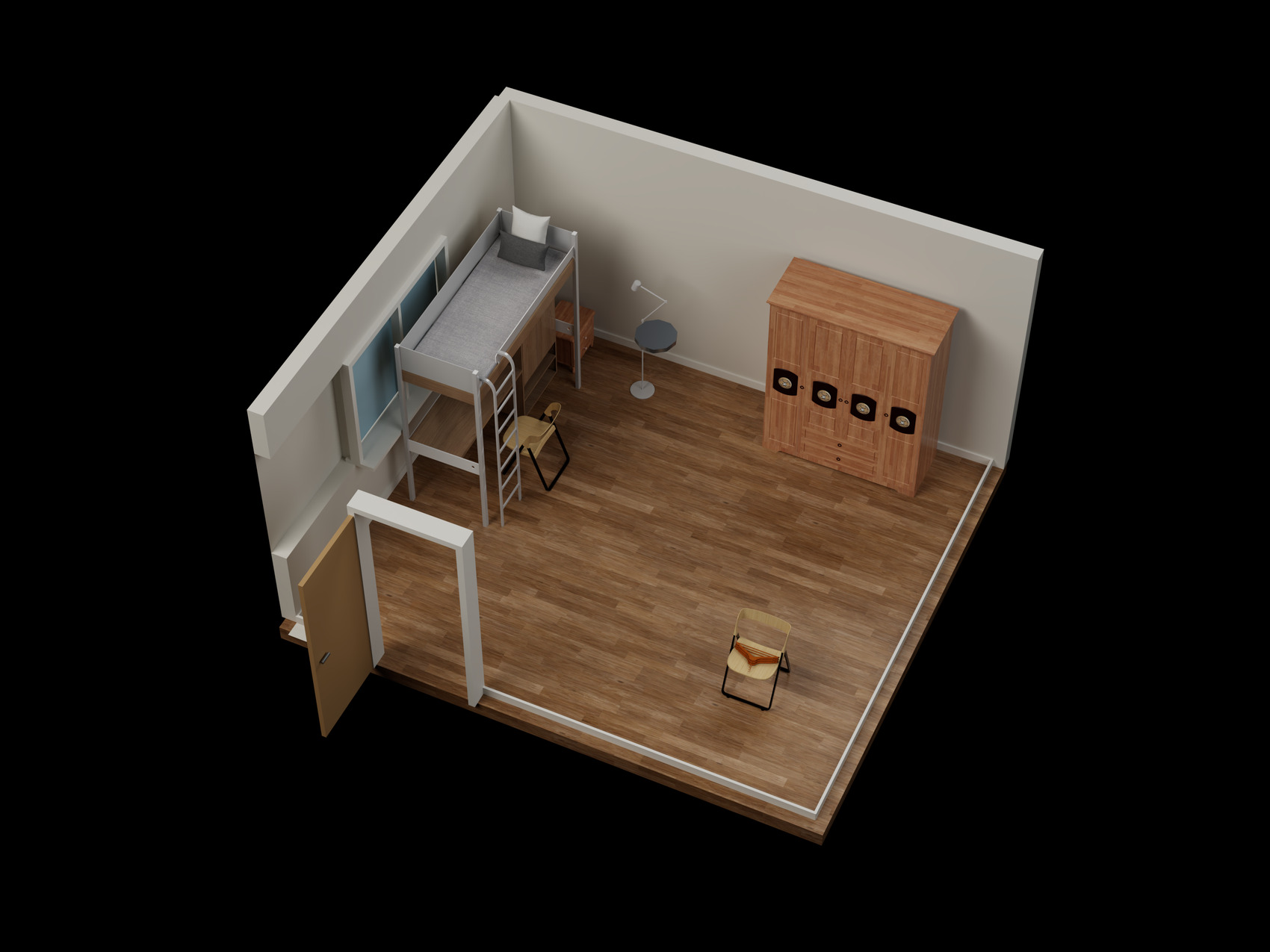} &
\includegraphics[width=.465\textwidth]{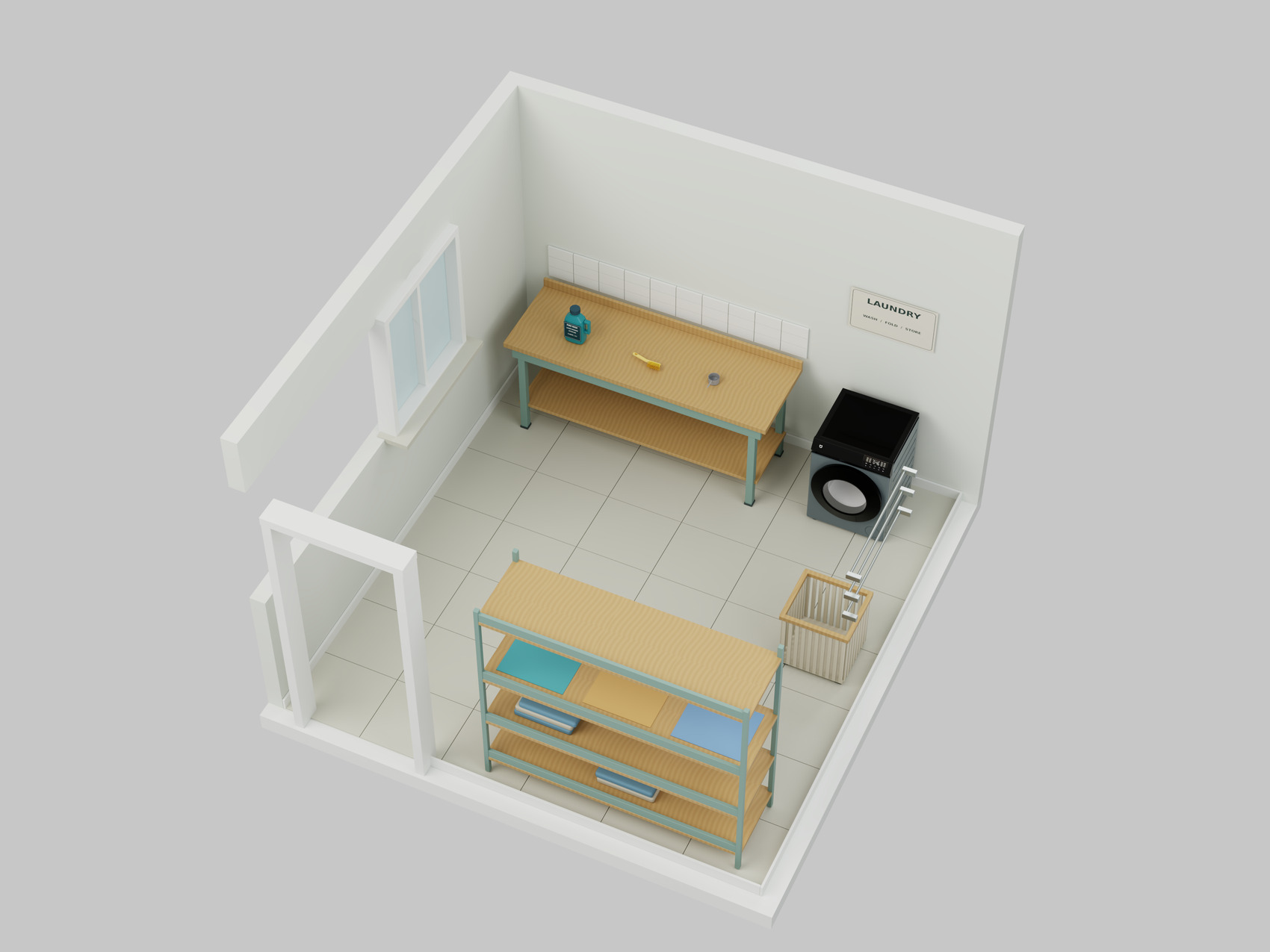} \\
\scriptsize\textbf{R13}: Hang empty clothes hangers & \scriptsize\textbf{R19}: Store laundry-room supplies \\[5pt]
\includegraphics[width=.465\textwidth]{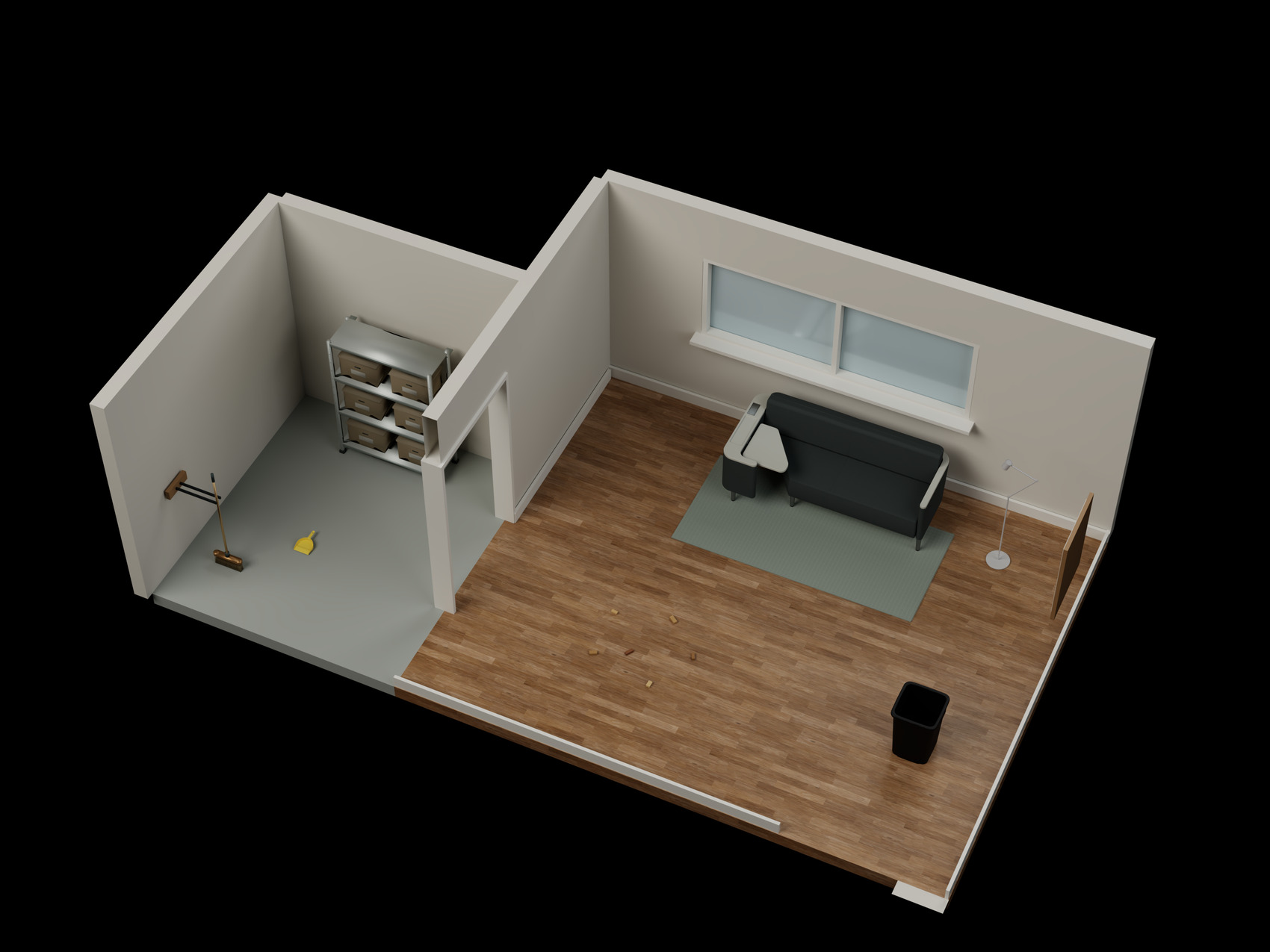} &
\includegraphics[width=.465\textwidth]{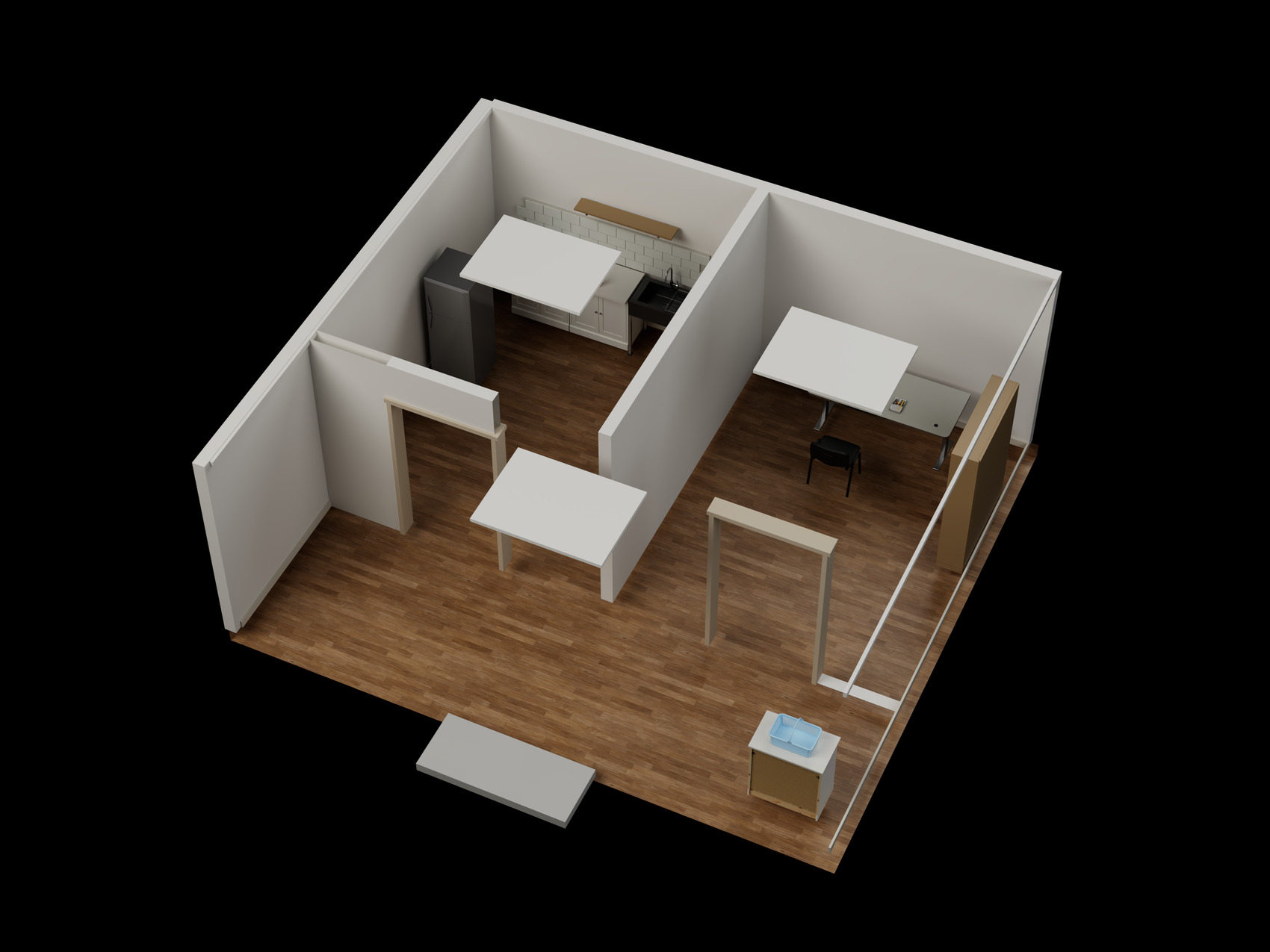} \\
\scriptsize\textbf{M08}: Fetch tools, sweep, and return them & \scriptsize\textbf{M09}: Pack items for leaving home \\[5pt]
\end{tabular}
\caption{\textbf{Additional generated environments (b).} Tabletop details (top), single-room interiors (middle), and multi-room cutaways (bottom), rendered from the frozen pre-task scenes at $3000\times2250$ with 1024 path-tracing samples per pixel (camera-facing walls and ceilings hidden for visibility).}
\label{fig:supp-new-gallery-b}
\end{figure*}

Figures~\ref{fig:supp-new-gallery-a} and~\ref{fig:supp-new-gallery-b} present twelve additional pre-task environments spanning four tabletop workspaces, four single-room interiors, and four multi-room layouts, using close framing for tabletops and architectural cutaways to expose room interiors and inter-room connections.

\definecolor{matchHint}{HTML}{185E83}
\definecolor{readyHint}{HTML}{95501F}
\newcommand{\readinesspanel}[6]{%
  \begin{minipage}[t]{.485\textwidth}
    \raggedright
    \textbf{#1}\hfill{\small\textcolor{matchHint}{\textbf{M}~#2}\quad
    \textcolor{readyHint}{\textbf{R}~#3}}\par\smallskip
    \includegraphics[width=\linewidth]{figures/evidence/match-ready-20261122/#4.jpg}\par\smallskip
    {\small\textcolor{matchHint}{\textbf{M}}\enspace#5\par
    \textcolor{readyHint}{\textbf{R}}\enspace#6\par}
  \end{minipage}%
}

\subsection{Diagnostic Views of MATCH and READY}
\label{sec:supp-match-ready}

\begin{figure*}[p]
\centering
\begin{minipage}{\textwidth}
\textbf{\large T04 \enspace Stack three bowls}\par\smallskip
Start with three separate, unstacked bowls and a distinct destination. The READY checks include native bowl motion, nesting and space at the destination.
\end{minipage}\par\medskip
\readinesspanel{Holodeck}{0.0}{14.3}{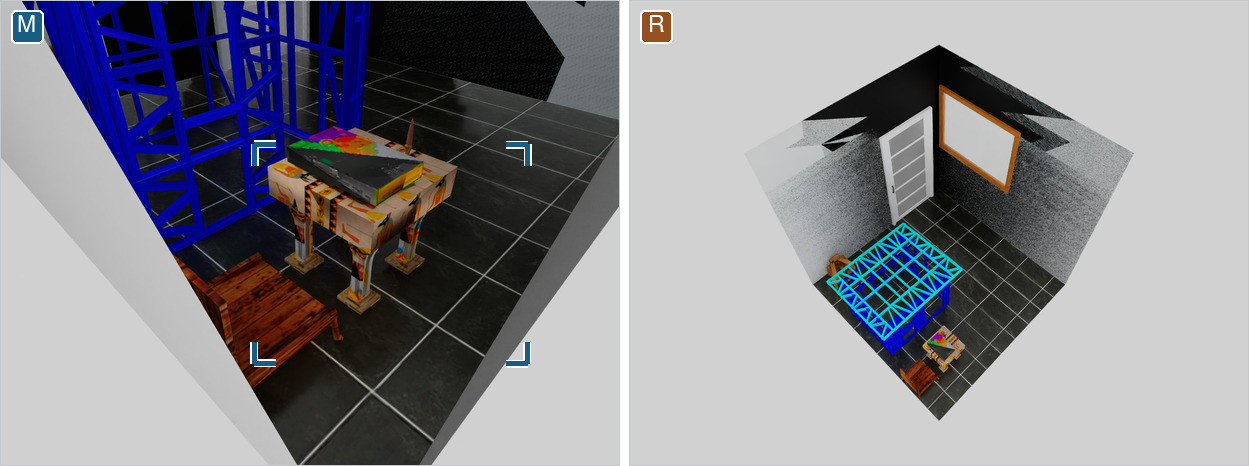}%
  {No set of three bowls or separate stack destination is present.}%
  {The tabletop is visible, but the required bowl bodies and nesting geometry are absent.}%
\hfill%
\readinesspanel{LayoutVLM}{37.5}{0.0}{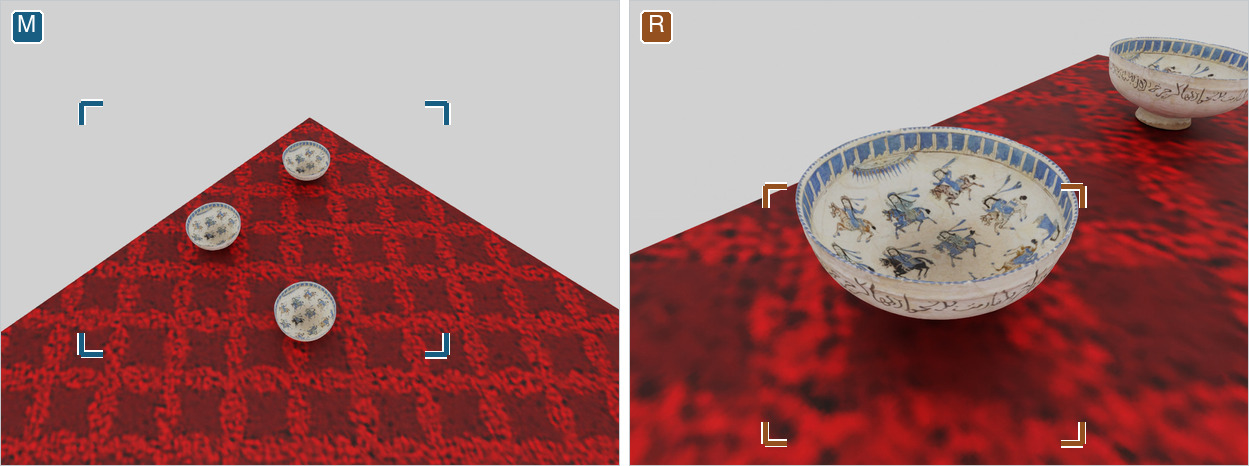}%
  {Three distinct bowls are on a patterned floor surface. The worktable and stack destination are absent.}%
  {The native product supplies neither independent bowl bodies nor bowl collision geometry.}%
\par\medskip
\readinesspanel{SAGE}{29.2}{14.3}{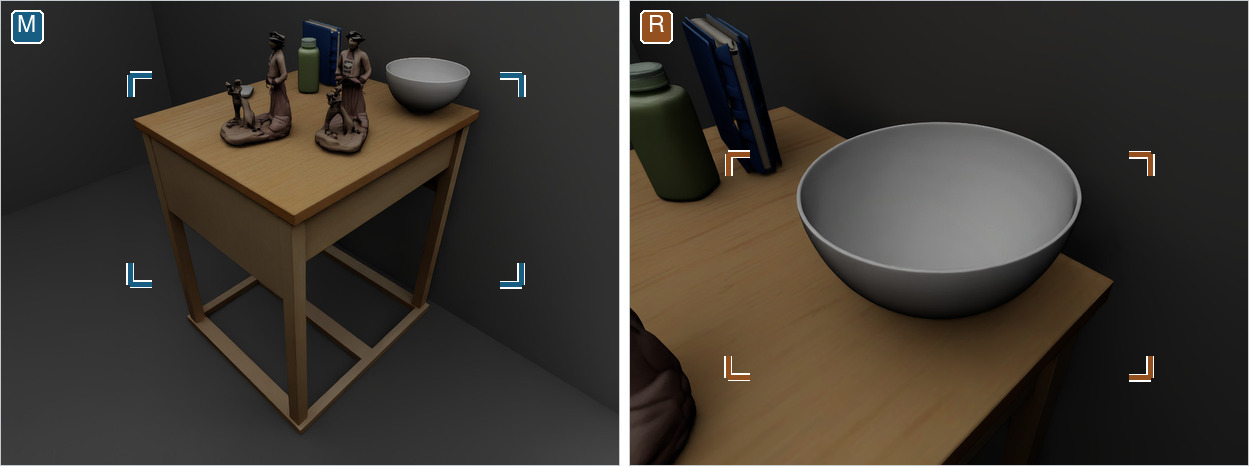}%
  {One bowl appears beside other objects. The full three-bowl set and a distinct destination are missing.}%
  {A visible bowl rim does not supply the three compatible movable bowls required for nesting.}%
\hfill%
\readinesspanel{EmbodiedGen V2}{0.0}{0.0}{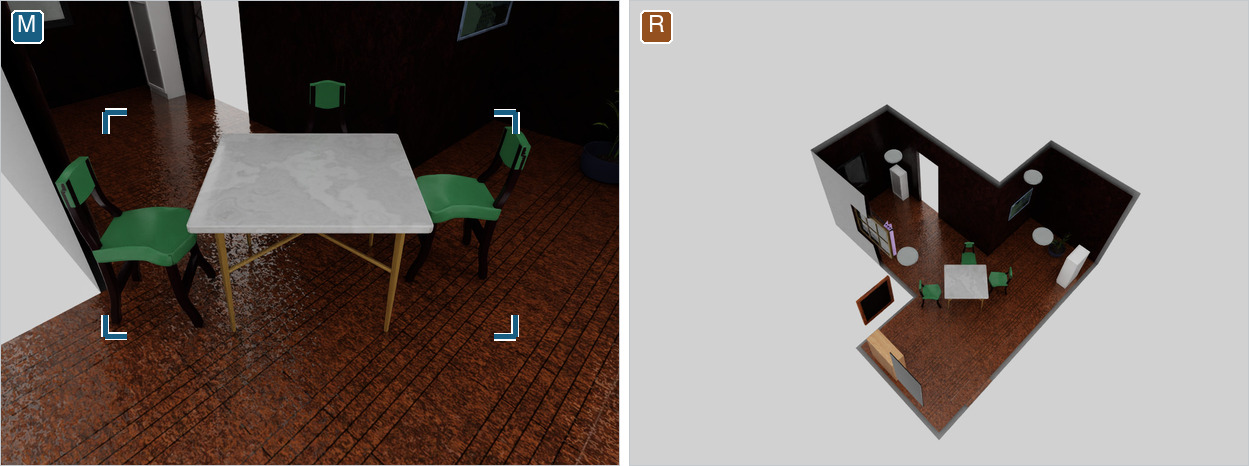}%
  {The furnished dining area has a table and chairs but no set of task bowls.}%
  {The required bowl bodies and a usable destination for stacking are absent.}%
\par\medskip
\readinesspanel{SceneSmith}{66.7}{42.9}{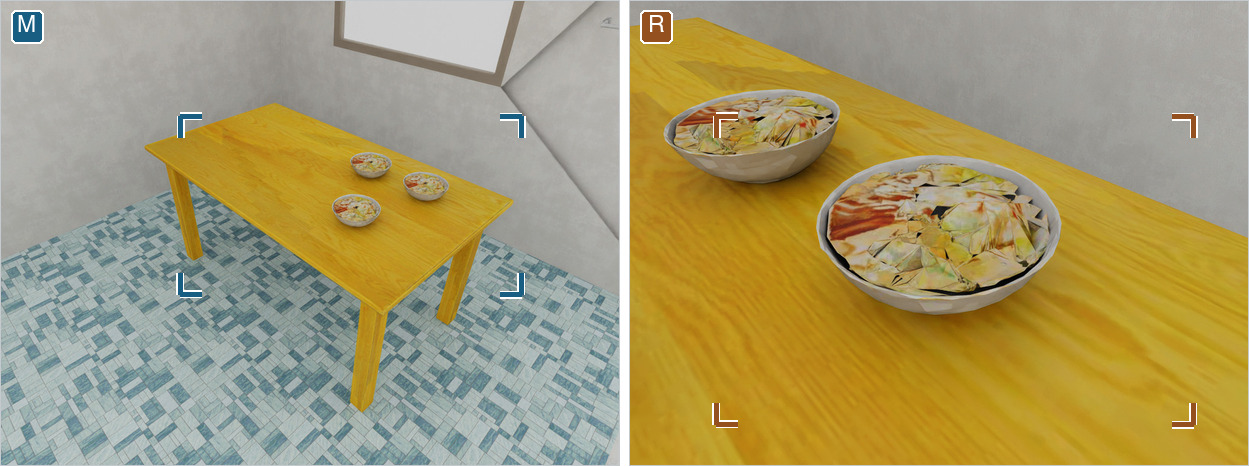}%
  {Three bowls are present on the table. A distinct stack destination is absent.}%
  {The native nesting check passes. The unspecified destination fails the target capacity and access checks.}%
\hfill%
\readinesspanel{EmbodiedStudio}{100.0}{100.0}{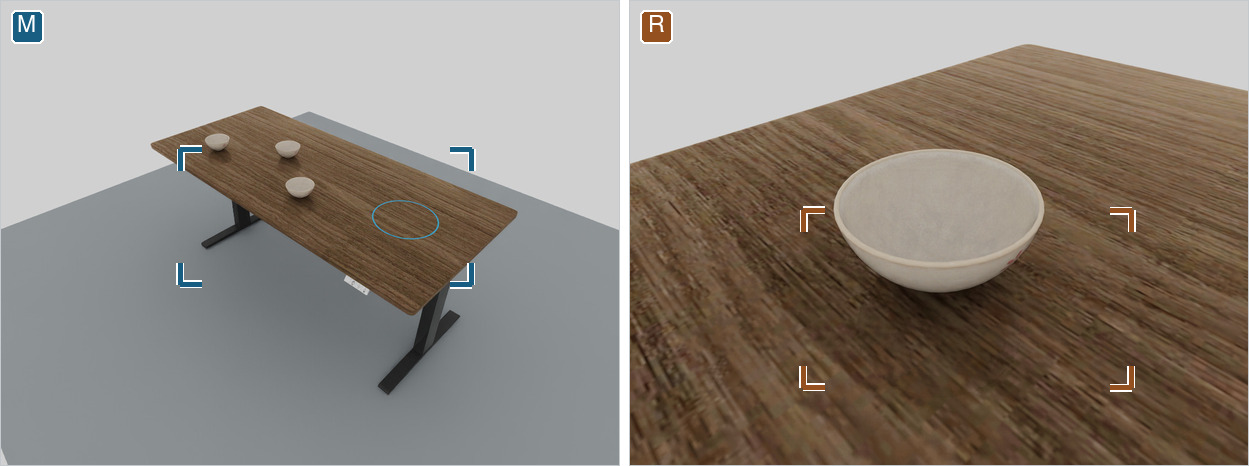}%
  {Three separate, unstacked bowls and a distinct blue target ring occupy one worktable.}%
  {Native shape queries found open cavities about 55 mm deep and a supported nesting configuration.}%
\caption{\textbf{T04: Stack three bowls.} SceneSmith provides compatible bowls but lacks a designated stacking destination. EmbodiedStudio supplies both the bowl set and a separate target.}
\label{fig:ready-T04}
\end{figure*}

\begin{figure*}[p]
\centering
\begin{minipage}{\textwidth}
\textbf{\large T15 \enspace Build a two-level block structure}\par\smallskip
Start with three separate, unstacked blocks on the tabletop. The READY checks include flat contacts, compatible dimensions and a clear building area.
\end{minipage}\par\medskip
\readinesspanel{Holodeck}{41.7}{28.6}{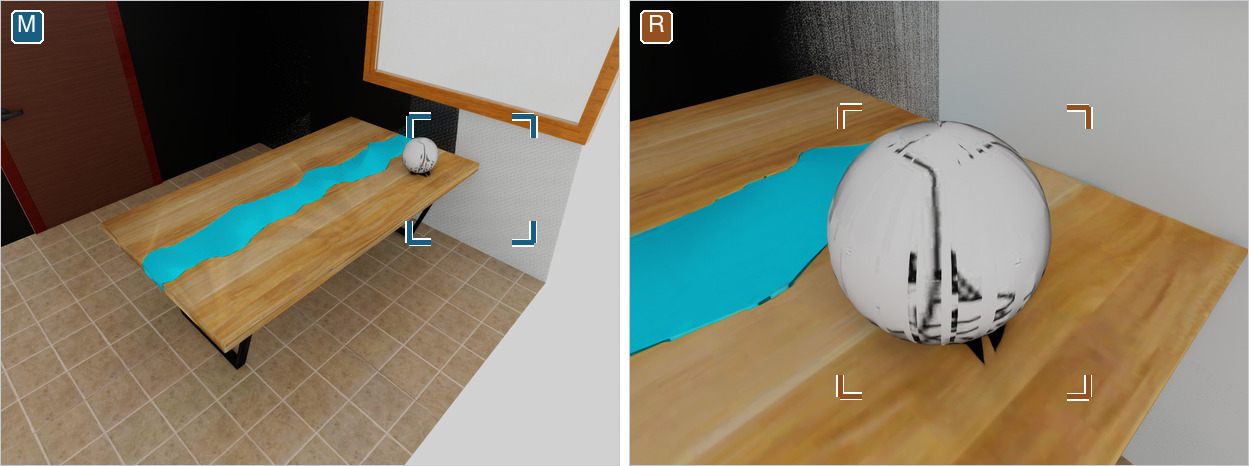}%
  {The block candidates appear as rounded surfaces occupying nearly the same location.}%
  {The source surfaces intersect, and the rounded pieces lack the flat contacts needed for the upper block.}%
\hfill%
\readinesspanel{LayoutVLM}{0.0}{0.0}{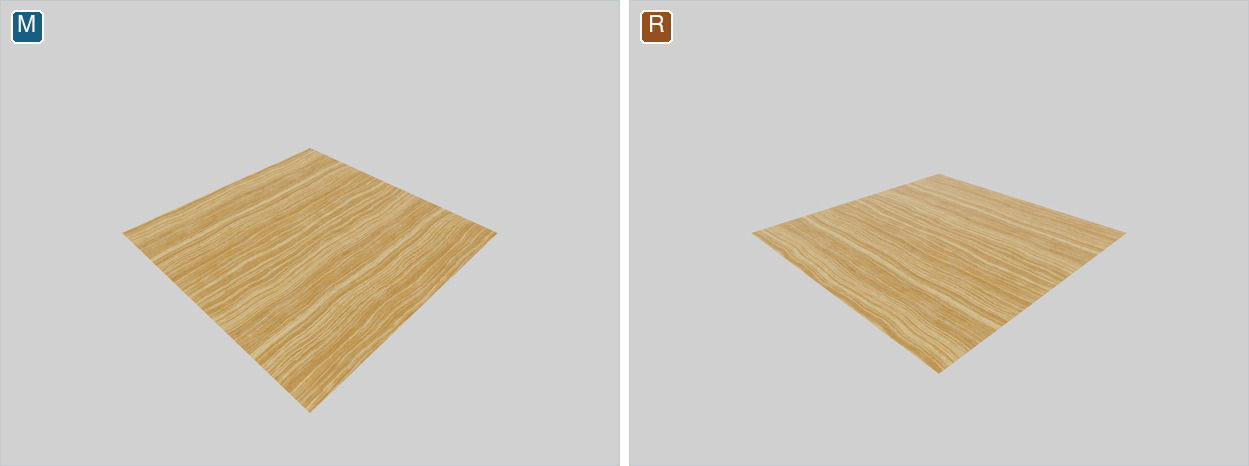}%
  {The delivered scene contains a floor plane with no worktable or blocks.}%
  {The required support and separate movable blocks are absent.}%
\par\medskip
\readinesspanel{SAGE}{100.0}{100.0}{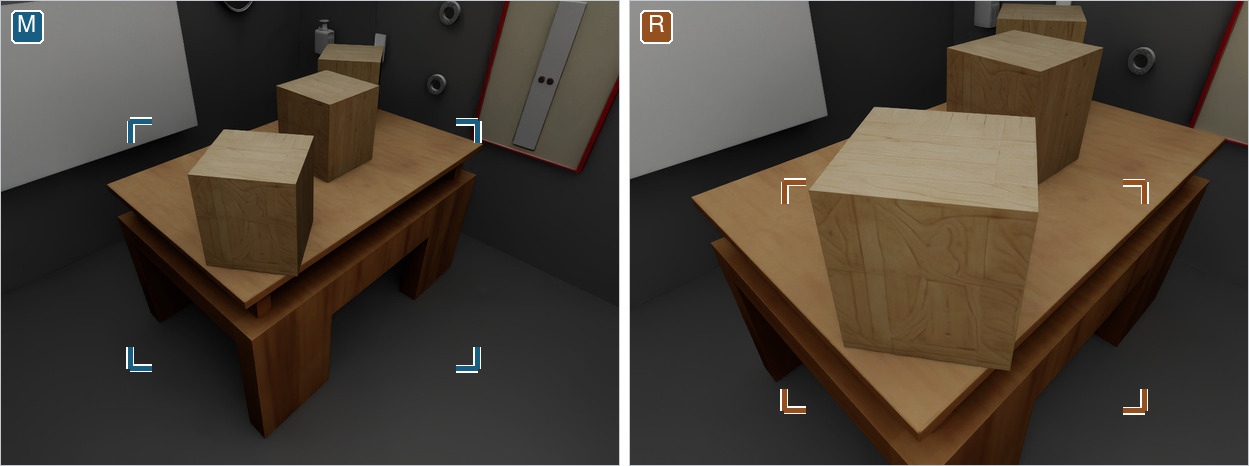}%
  {Three separate blocks start unstacked on the tabletop.}%
  {Flat faces and two lower blocks of about 335 mm in height provide compatible stacking geometry.}%
\hfill%
\readinesspanel{EmbodiedGen V2}{12.5}{0.0}{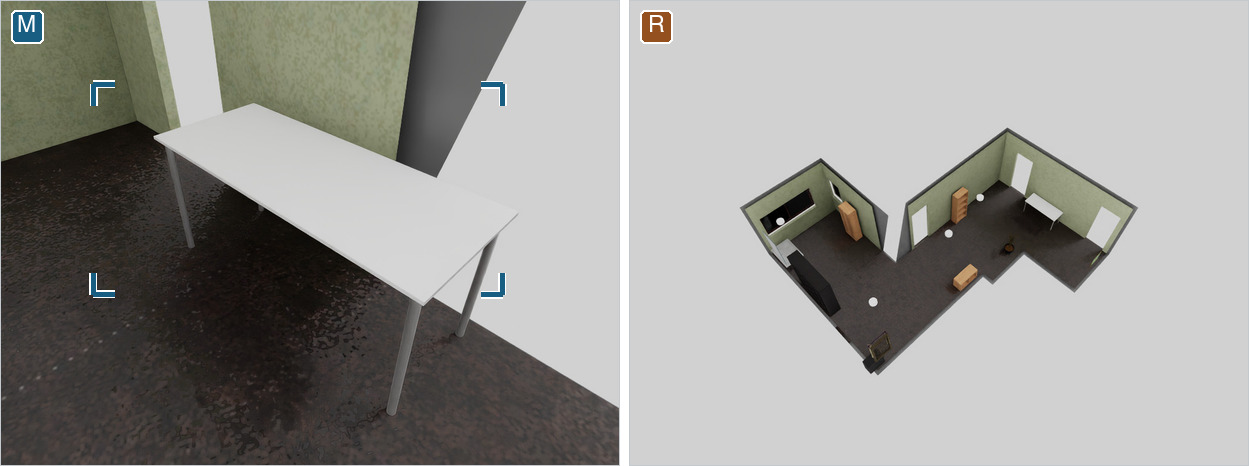}%
  {A table appears in the furnished scene, but the three block objects are missing.}%
  {The task has no block bodies or compatible contact surfaces to support a two-level stack.}%
\par\medskip
\readinesspanel{SceneSmith}{0.0}{0.0}{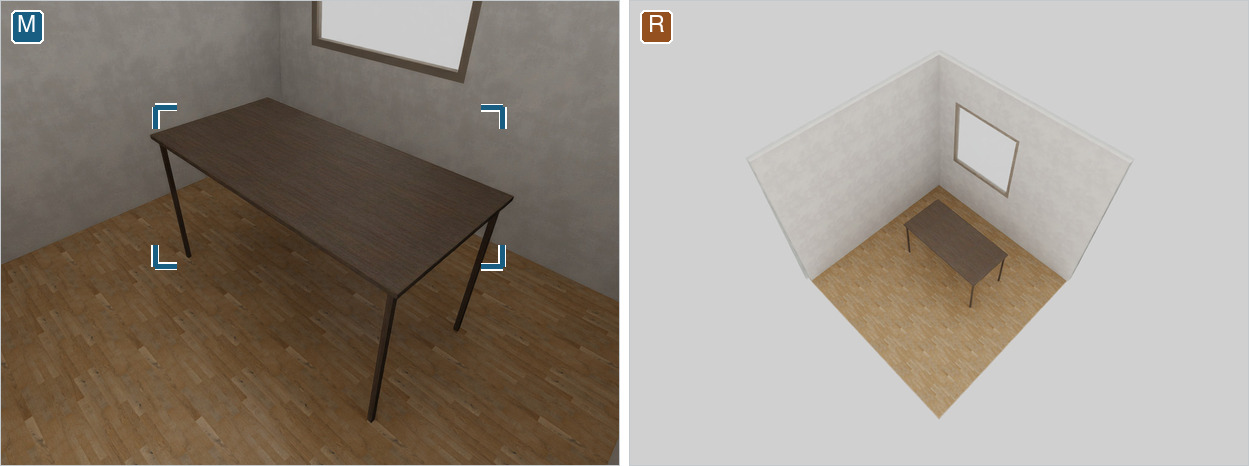}%
  {The room contains a bare table without the three required blocks.}%
  {There are no separate block bodies or candidate block-to-block support surfaces.}%
\hfill%
\readinesspanel{EmbodiedStudio}{100.0}{100.0}{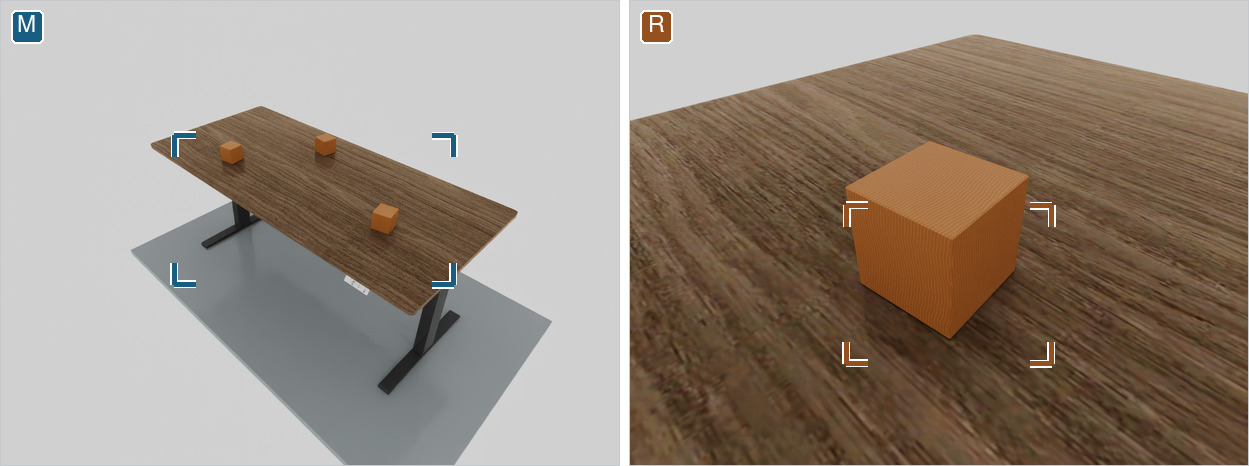}%
  {Three separate blocks start unstacked on the tabletop, with space between them.}%
  {Separate movable bodies, flat contact faces and a clear building area satisfy the tested prerequisites.}%
\caption{\textbf{T15: Build a two-level block structure.} SAGE and EmbodiedStudio satisfy all tested conditions. Other panels show different outcomes with a floor, an empty table or rounded candidate objects.}
\label{fig:ready-T15}
\end{figure*}

\begin{figure*}[p]
\centering
\begin{minipage}{\textwidth}
\textbf{\large R20 \enspace Pack tools in a storage room}\par\smallskip
A screwdriver, wrench, and retracted tape measure start across the workbench and shelving. READY checks whether the three tools can be inserted into the open toolbox and the lid can close.
\end{minipage}\par\medskip
\readinesspanel{Holodeck}{31.2}{0.0}{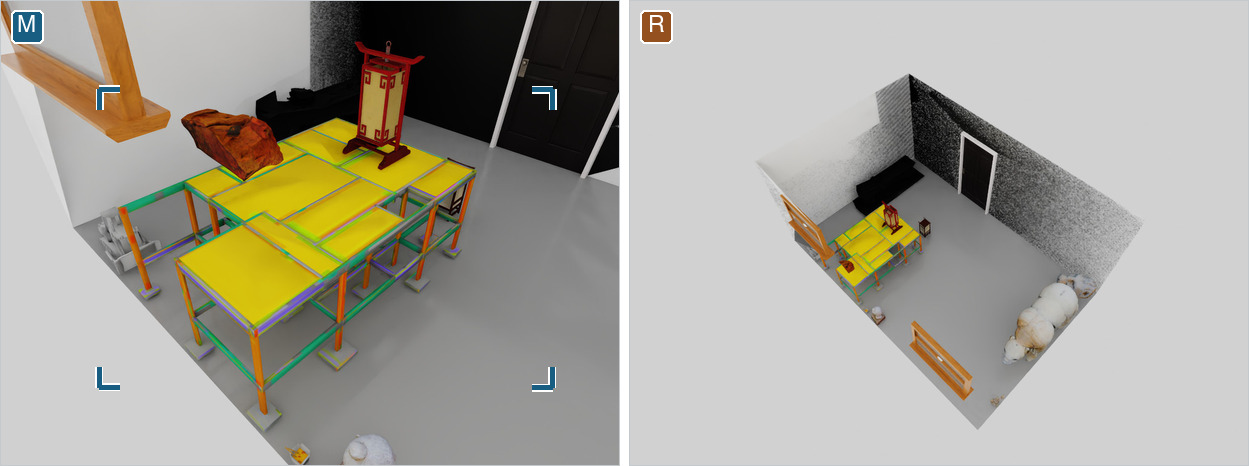}%
  {The yellow work surface is present. It holds decorative objects, including a lantern, and the three tools are absent.}%
  {The scene supplies neither the toolbox opening nor a toolbox lid mechanism.}%
\hfill%
\readinesspanel{LayoutVLM}{56.2}{16.7}{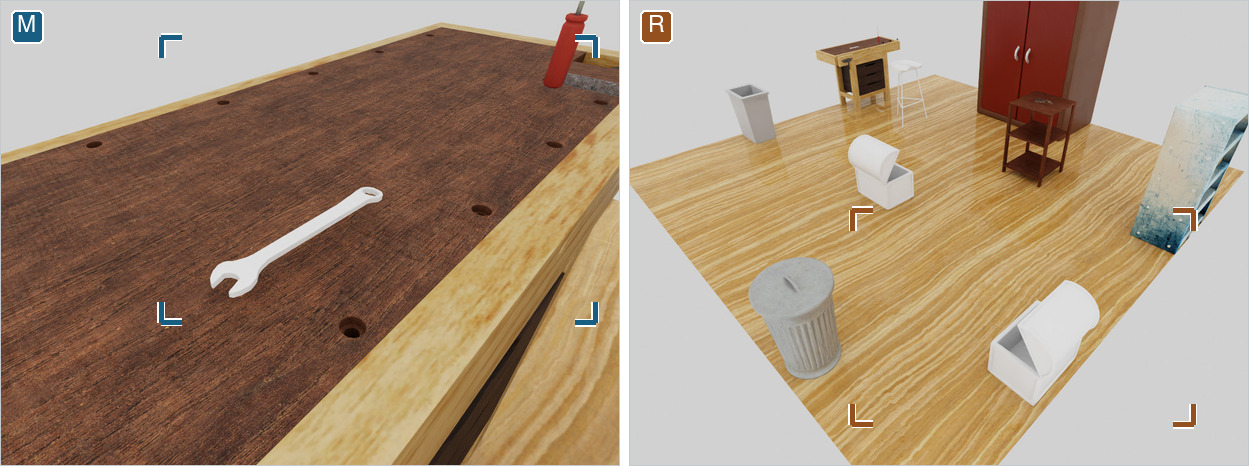}%
  {A screwdriver and wrench appear on the bench. A tape measure is absent.}%
  {The two white containers have open-looking lids and lack a supplied hinge or closure control.}%
\par\medskip
\readinesspanel{SAGE}{42.7}{2.8}{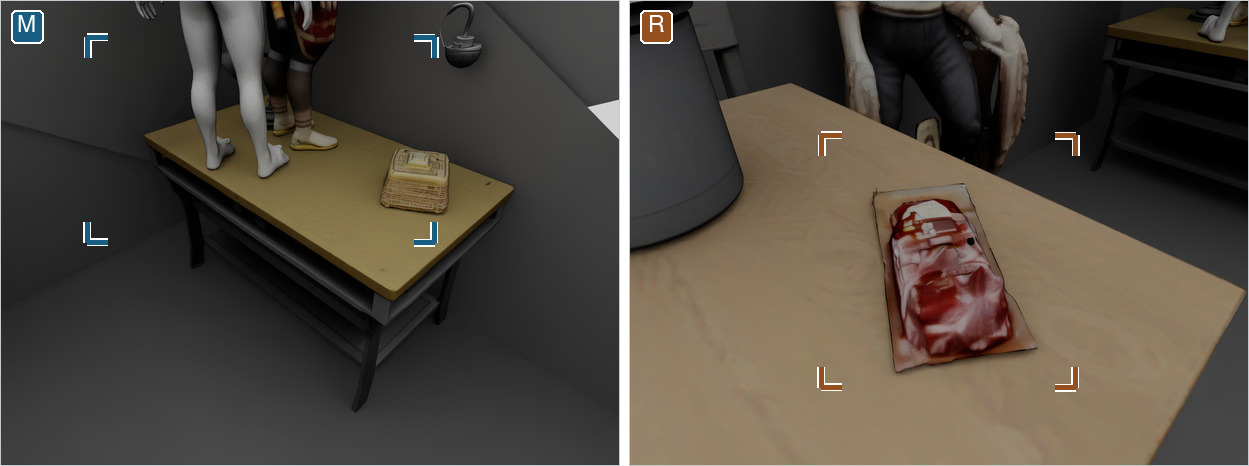}%
  {A workbench and a rack are present. The screwdriver and wrench candidates are figurines.}%
  {The red candidate has no identified receiving opening or native lid mechanism.}%
\hfill%
\readinesspanel{EmbodiedGen V2}{32.3}{0.0}{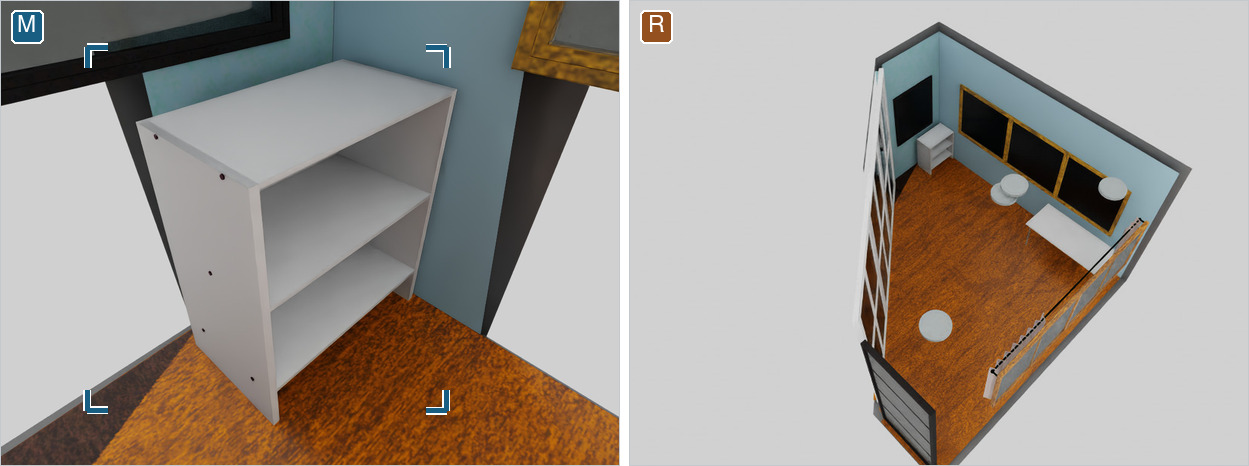}%
  {The room has a table and open shelves. The three tools and toolbox are absent.}%
  {The empty work surfaces provide no tool insertion target or lid to close.}%
\par\medskip
\readinesspanel{SceneSmith}{94.8}{75.0}{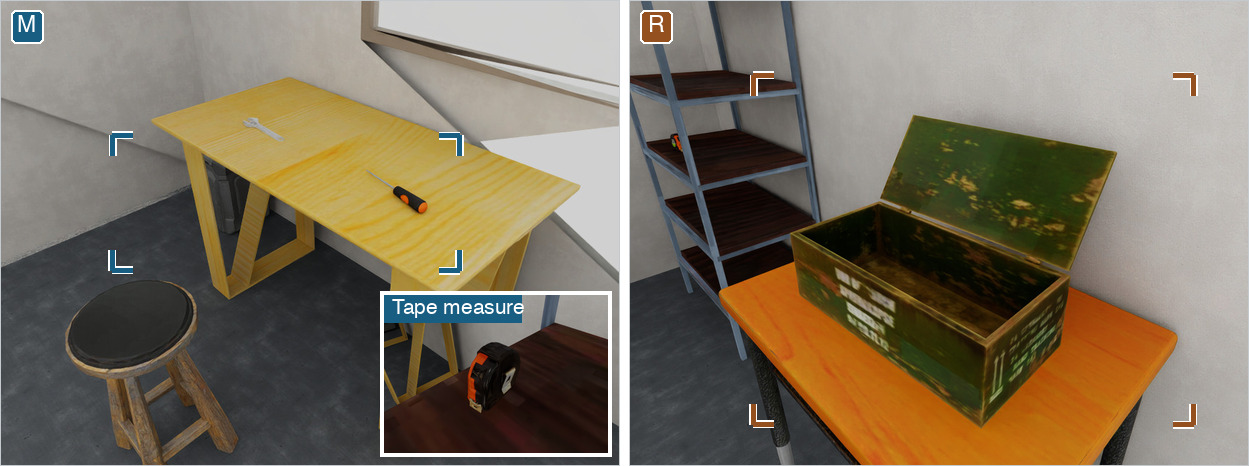}%
  {All three tools and an open toolbox are present. The tape measure starts on the rack.}%
  {The opening fits the tools. The box and lid are one rigid body, so the lid cannot close independently.}%
\hfill%
\readinesspanel{EmbodiedStudio}{100.0}{100.0}{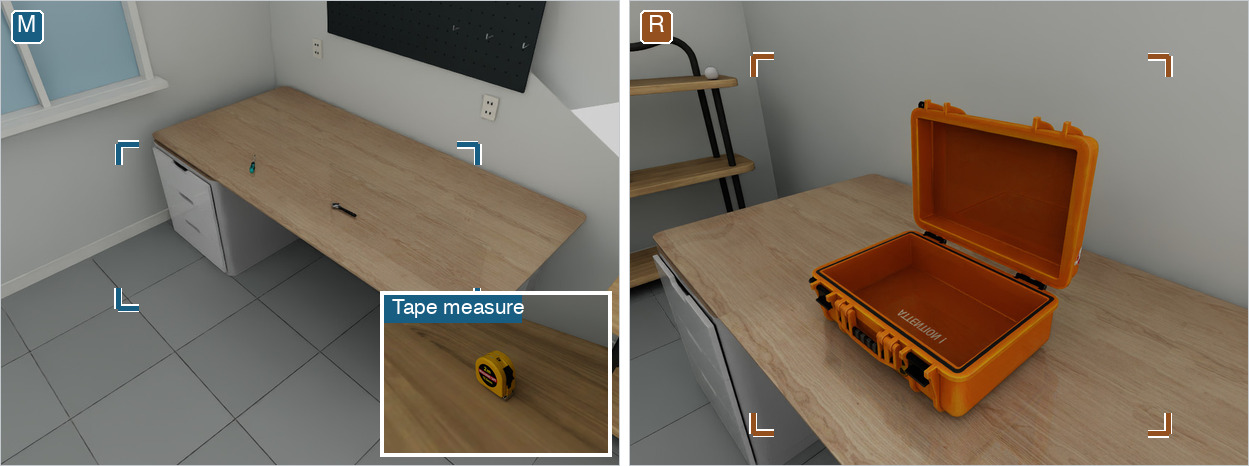}%
  {The bench holds the screwdriver and wrench, and the shelf holds the retracted tape measure.}%
  {A bound hinge provides $90^\circ$ of lid travel. Native clearance checks pass with all three tools inside and the lid closed.}%
\caption{\textbf{R20: Pack tools in a storage room.} Visible tools and an open toolbox can satisfy MATCH. READY also checks whether the native lid can close with all three tools inside.}
\label{fig:ready-R20}
\end{figure*}

\begin{figure*}[p]
\centering
\begin{minipage}{\textwidth}
\textbf{\large R15 \enspace Sort books onto labelled shelves}\par\smallskip
Start with mixed books on the desk and shelf labels that match the visible cover categories. READY checks book motion, shelf support and insertion clearance.
\end{minipage}\par\medskip
\readinesspanel{Holodeck}{15.0}{12.5}{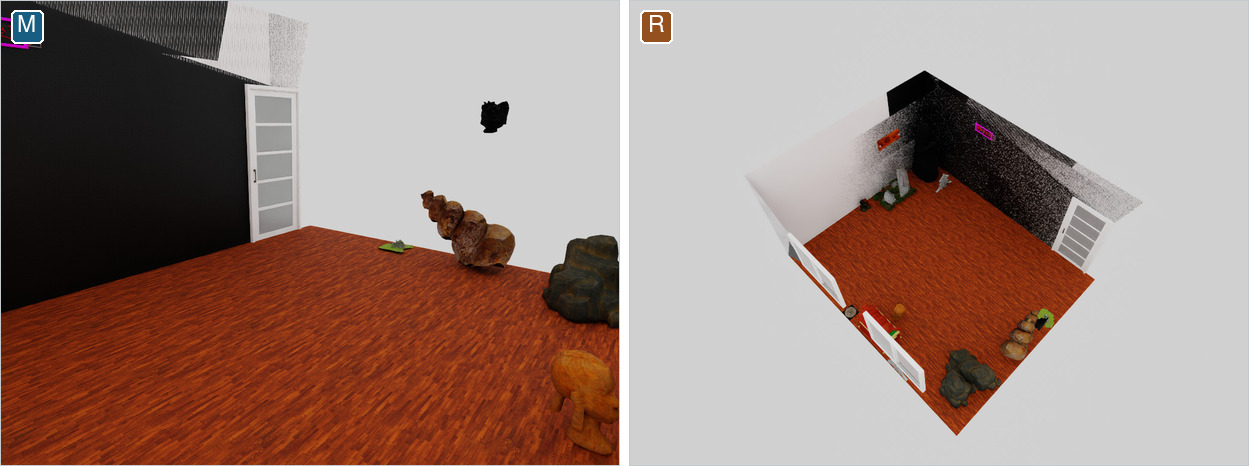}%
  {The room layout lacks the required desk-book set and category-labelled destination shelves.}%
  {Without the book set and its shelves, the book pickup and shelf-placement checks fail.}%
\hfill%
\readinesspanel{LayoutVLM}{58.3}{41.7}{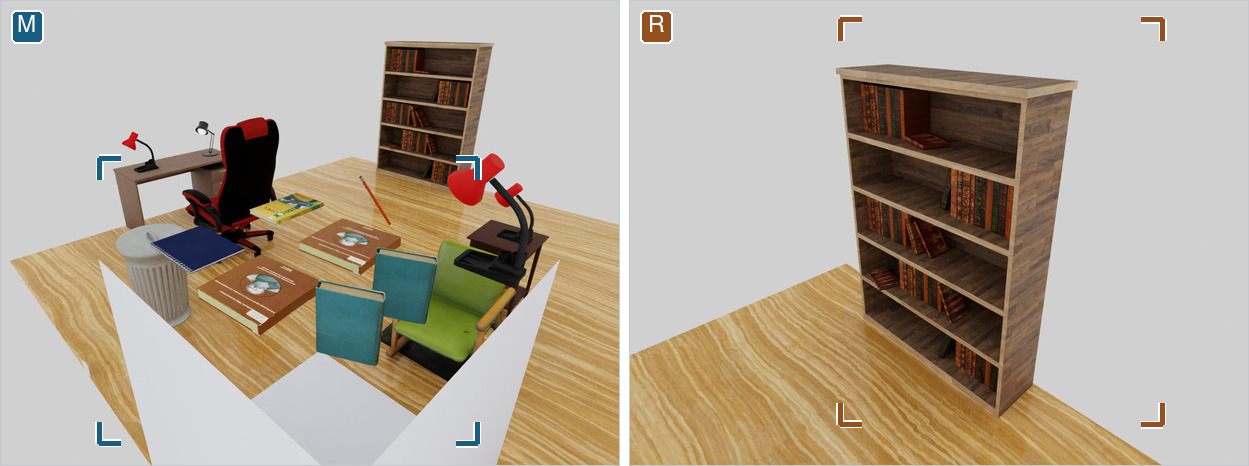}%
  {Visible books and shelving are supplied, but the shelves lack matching category labels.}%
  {The shelf approach passes. The delivered book geometry has no native rigid-body or collision definitions.}%
\par\medskip
\readinesspanel{SAGE}{10.0}{12.5}{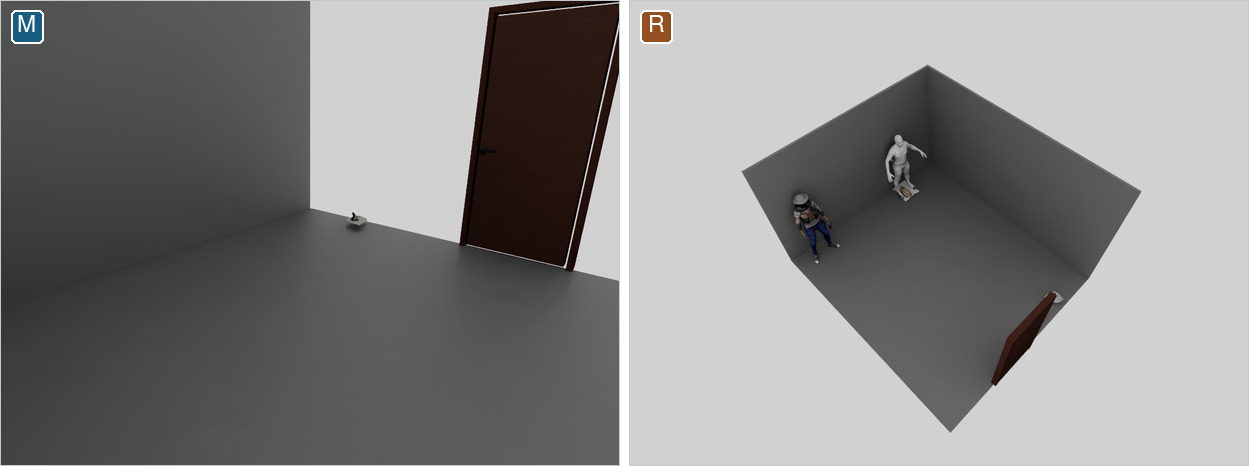}%
  {The output lacks a furnished study with a task book set and a destination bookcase.}%
  {The book handling and shelf placement conditions are unmet.}%
\hfill%
\readinesspanel{EmbodiedGen V2}{31.7}{12.5}{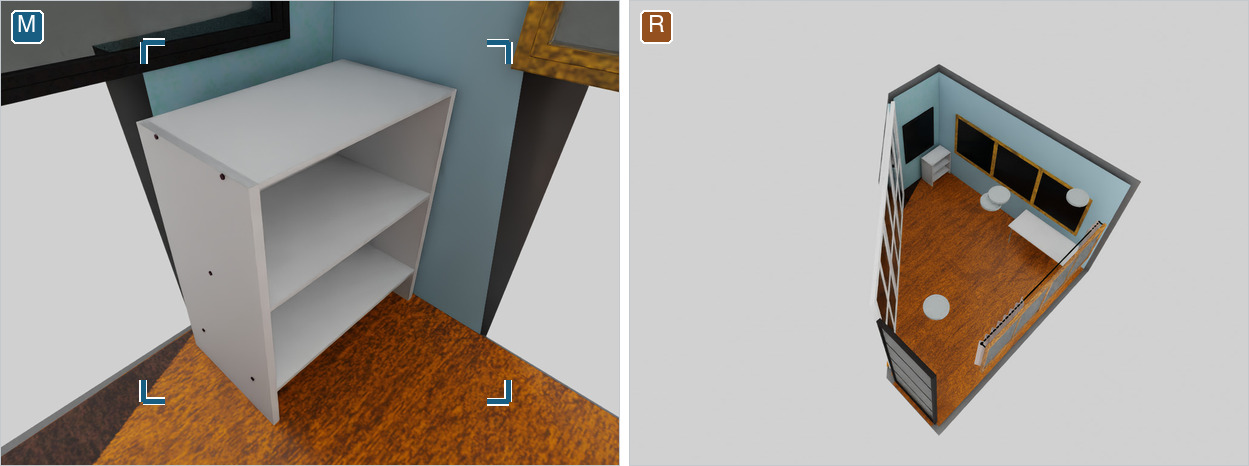}%
  {A small shelf unit is present. The requested mix of labelled books on the desk is missing.}%
  {Some shelves can provide support, but the required book transfer and matching destinations are absent.}%
\par\medskip
\readinesspanel{SceneSmith}{63.3}{75.0}{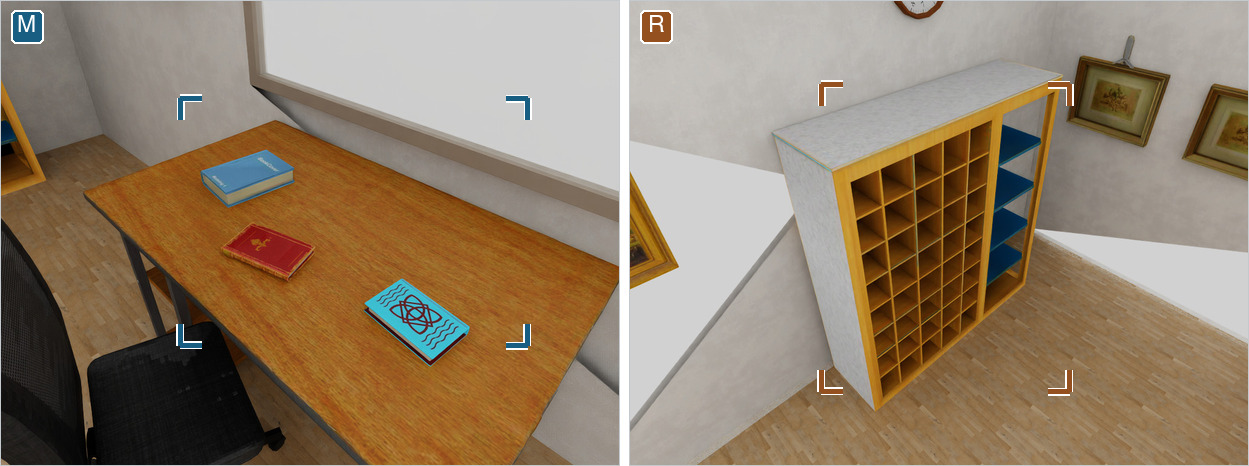}%
  {Three books are on the desk. The separate cubby shelf lacks matching category labels.}%
  {The books are independently movable. Whole-book insertion passes, while shelf-support judgments differ across repeats.}%
\hfill%
\readinesspanel{EmbodiedStudio}{98.3}{87.5}{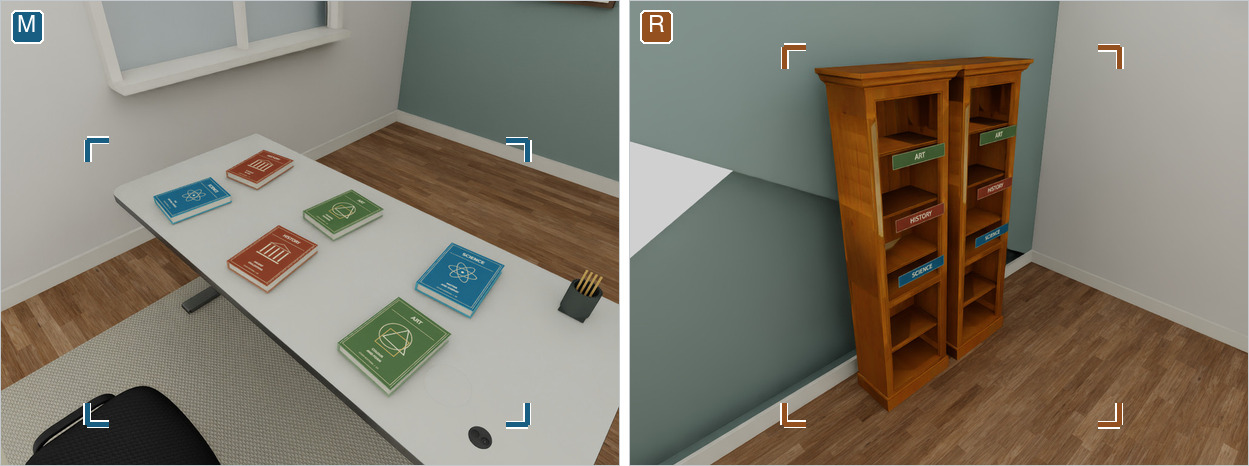}%
  {Six books show cover cues, with matching SCIENCE, HISTORY and ART shelf labels.}%
  {Book insertion passes. The floating shelf labels fail the support check, with a measured gap of about 60 mm.}%
\caption{\textbf{R15: Sort books onto labelled shelves.} SceneSmith provides usable book insertion but lacks matching shelf labels. EmbodiedStudio supplies the labels, while their unsupported placement reduces READY.}
\label{fig:ready-R15}
\end{figure*}

\begin{figure*}[p]
\centering
\begin{minipage}{\textwidth}
\textbf{\large M08 \enspace Fetch tools, sweep, and return them}\par\smallskip
Start with a broom and dustpan in storage and rigid debris on the living-room floor. READY checks the passage, sweep space and tool and bin receiving geometry.
\end{minipage}\par\medskip
\readinesspanel{Holodeck}{60.4}{30.3}{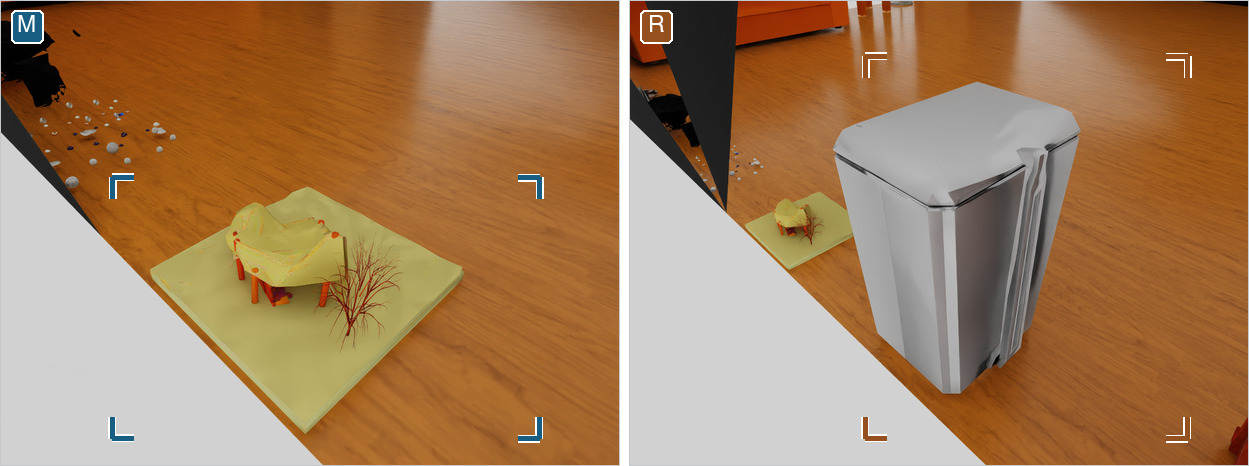}%
  {The two rooms and rigid floor debris are present. The yellow candidate is a decorative snow globe.}%
  {The doorway passes. The required tools are absent and the bin is a closed collision body.}%
\hfill%
\readinesspanel{LayoutVLM}{62.5}{18.2}{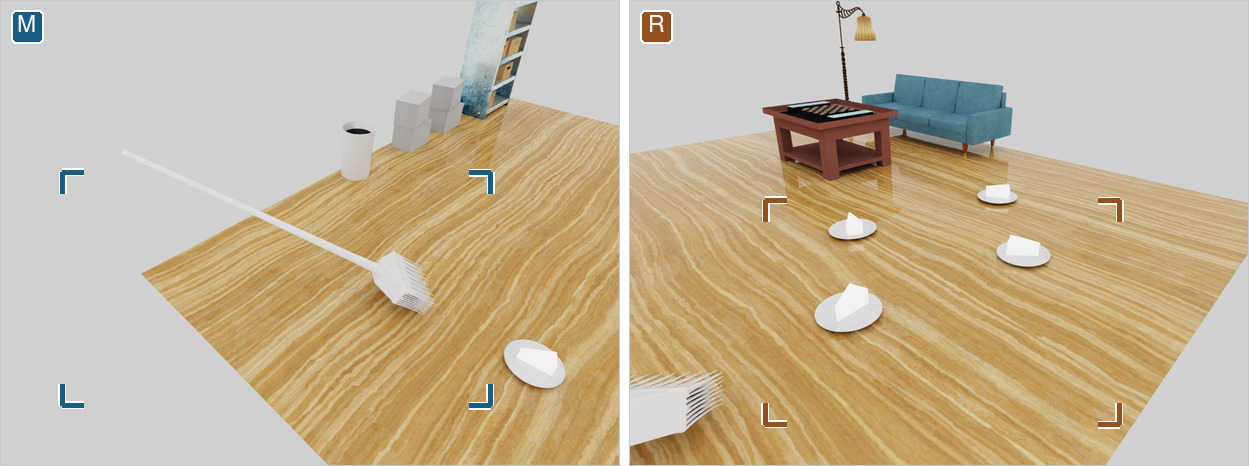}%
  {The broom is in the living area. Separate storage and a dustpan are missing.}%
  {The floor area is clear. The missing storage and pan prevent the retrieval-route and receiving-edge checks.}%
\par\medskip
\readinesspanel{SAGE}{77.1}{51.5}{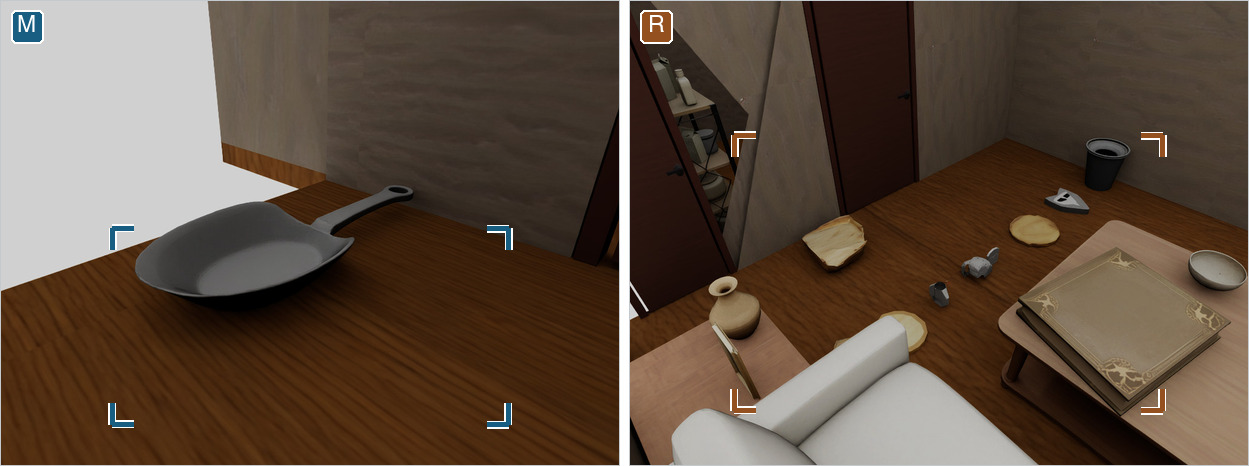}%
  {The storage and living rooms show a long-handled tool, a pan candidate and scattered floor objects.}%
  {The passage passes. The raised rim of the frying-pan-shaped object provides no floor-level collecting edge.}%
\hfill%
\readinesspanel{EmbodiedGen V2}{22.9}{0.0}{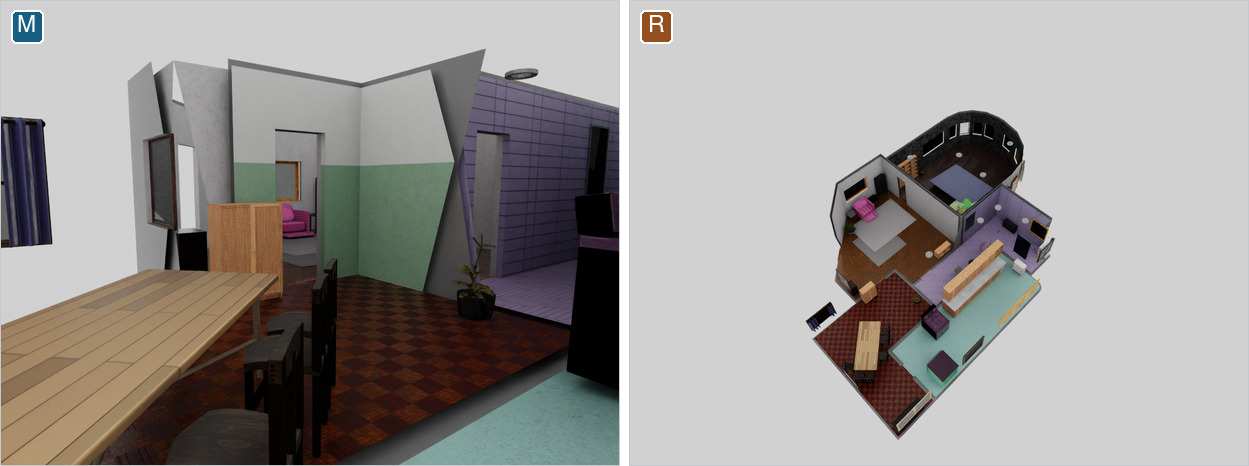}%
  {Furnished rooms are visible. The specified storage room, tools and floor debris are missing.}%
  {The required inter-room retrieval and debris collection geometry are absent.}%
\par\medskip
\readinesspanel{SceneSmith}{52.1}{45.5}{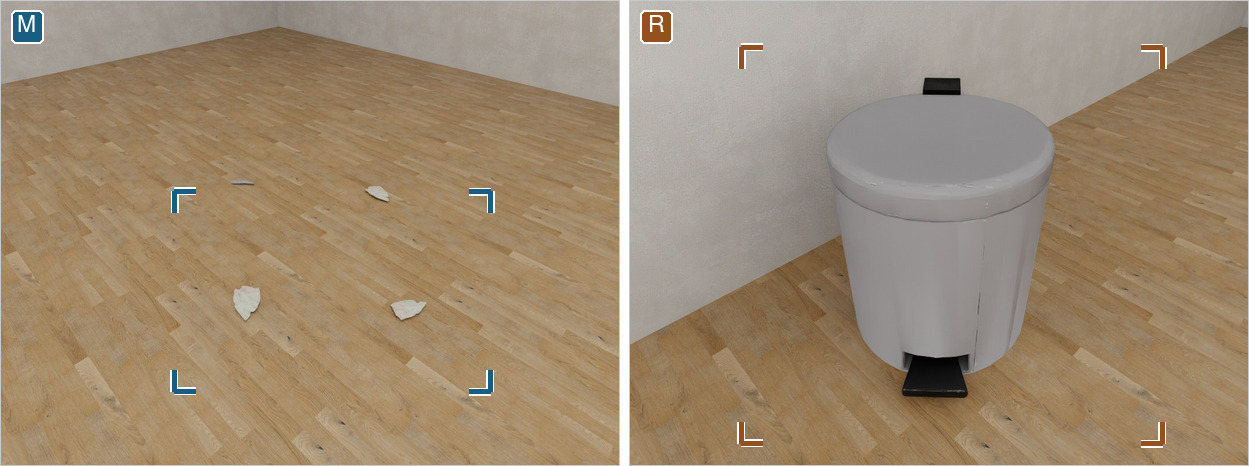}%
  {Floor debris and a lidded bin are visible. The broom and dustpan are absent.}%
  {The inter-room opening passes. The bin is a single body with no available lid motion.}%
\hfill%
\readinesspanel{EmbodiedStudio}{100.0}{100.0}{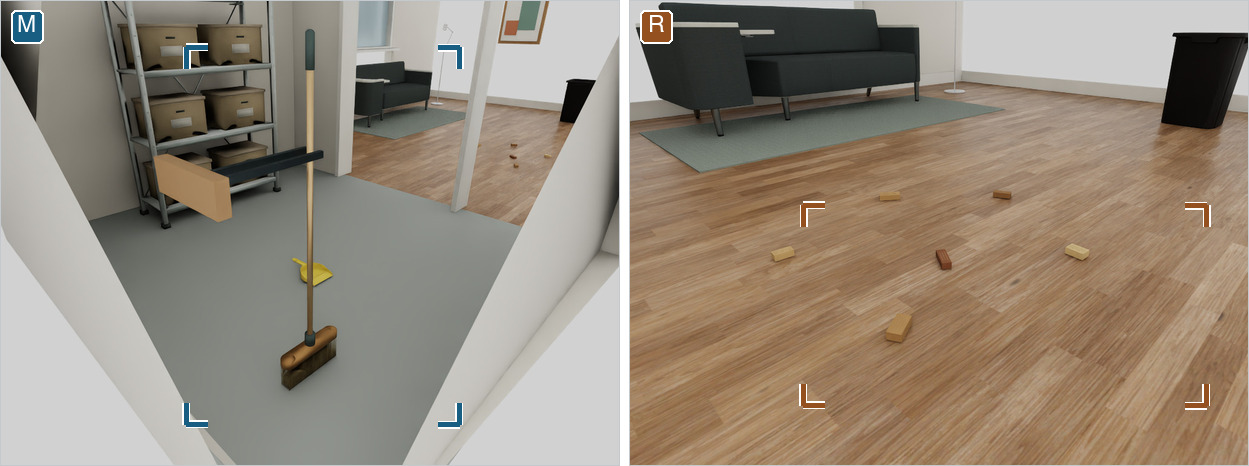}%
  {The broom and dustpan start in storage. Separate debris and the receiving bin are in the living room.}%
  {The doorway is about 1.30 m wide. All six debris probes clear the pan lip in allowed poses.}%
\caption{\textbf{M08: Fetch tools, sweep, and return them.} A connected route and a clear floor can pass even when collection fails. A raised pan rim or a closed bin changes the receiving condition.}
\label{fig:ready-M08}
\end{figure*}

Figures~\ref{fig:ready-T04}--\ref{fig:ready-M08} illustrate five representative tasks across tabletop (\texttt{T04}, \texttt{T15}), single-room (\texttt{R20}, \texttt{R15}), and multi-room (\texttt{M08}) scales where all six methods delivered complete scenes, highlighting distinct visible failure modes in MATCH and READY (including a subtle shelf-label support deduction on EmbodiedStudio in \texttt{R15}).

In each panel, the \textcolor{matchHint}{\textbf{M}} and \textcolor{readyHint}{\textbf{R}} insets localize task objects and operational regions (with a wider view when a requested object is absent), accompanied by summaries of the underlying clauses and measurements. Displayed MATCH and READY scores reflect the full three-repeat benchmark evaluations over complete scene geometry.

\section{Generation Latency Records}
\label{sec:supp-cost}

All 50 EmbodiedStudio requests and all five baselines used \texttt{gpt-6.1-sol} (\texttt{high} reasoning effort) under each method's native stopping rules (e.g., SAGE's 15-call client default) without a fixed token or monetary cap. Table~\ref{tab:gen-cost-all} compares overall wall-clock generation latency across all six methods from recorded runtime logs. Because multi-turn agentic pipelines (SAGE, SceneSmith, and EmbodiedStudio) executed concurrently on the shared GPU workstation, their wall-clock spans reflect end-to-end elapsed time from task launch to completion under concurrent job scheduling, encompassing active reasoning, asset preparation, simulation building, rendering, and review. Across the 50 requests, EmbodiedStudio averages 3.5 revisions / 8.9 scene nodes on Tabletop, 5.0 revisions / 73.2 nodes on Single-room, and 5.0 revisions / 88.8 nodes on Multi-room.

\begin{table}[!tb]
\centering\scriptsize
\caption{\textbf{Cross-method generation latency across the 50-request benchmark} (minutes; mean / median). \emph{Note:} Holodeck reports the 42 completed scenes (all 50 runs including 8 early-exit parser failures: 8.3 / 7.9\,min); LayoutVLM reports logged layout-solver duration; SAGE reports the 49 tasks with completed run logs on their final attempt; EmbodiedGen~V2, SceneSmith, and EmbodiedStudio report end-to-end wall-clock duration across all 50 tasks.}
\label{tab:gen-cost-all}
\setlength{\tabcolsep}{4pt}
\begin{tabular*}{\columnwidth}{@{\extracolsep{\fill}}lc@{}}
\toprule
Method & All Tasks (mean / median) \\
\midrule
Holodeck ($n=42$ completed) & 9.6 / 8.5\,min \\
LayoutVLM (native solver) & 3.2 / 3.2\,min \\
EmbodiedGen V2 & 25.5 / 17.3\,min \\
SAGE ($n=49$ logged runs) & 187.3 / 153.4\,min \\
SceneSmith & 310.0 / 160.7\,min \\
\textbf{EmbodiedStudio (Ours)} & 197.5 / 149.5\,min \\
\bottomrule
\end{tabular*}
\end{table}

\section{Prompt and Skill Excerpts}
\label{sec:supp-prompts}

\paragraph{Task-Request Template.} Each of the 50 requests follows a standardized three-part specification specifying the initial pre-task state, scale scope, and required interactions (excerpted from \texttt{T11}):

\smallskip
\noindent\fbox{\begin{minipage}{\dimexpr\columnwidth-2\fboxsep-2\fboxrule\relax}
\ttfamily\scriptsize\raggedright
Generate a realistic, simulation-ready scene in the initial state BEFORE the task below begins. Create the environment, not a depiction of the completed task. Include the objects and functional interactions needed for the task, with plausible scale, materials, collision geometry, and support. Objects that must be picked up or moved must be separate movable objects. Required openings and moving parts must be usable rather than fixed decorative geometry.\\[3pt]
Scope: One tabletop workspace. All task objects and destinations belong to this workspace; travel between room locations is not required.\\[3pt]
Task: Move the three designated objects from the tray into the storage box, leaving the non-target objects on the tray.
\end{minipage}}
\smallskip

\paragraph{Authoring Skill Summary.} The authoring skill instructs the agent to: (1) declare overall scope, functional roles, and the unified ownership tree before expanding subtrees; (2) follow the four-tier asset sourcing order and verify physical scale and articulation; (3) write coupled \texttt{scene.py}, \texttt{tasks.py}, \texttt{rubrics.py}, and (when requested) \texttt{expert.py} programs using real simulator feedback without deleting or weakening original requirements; and (4) invoke the independent reviewer on the frozen revision prior to delivery.

\onecolumn
\normalsize
\section{Interactive Scene Explorer and Recorded Media}
\label{sec:supp-viewer}

The \href{https://embodiedstudio.github.io/?view=paper\#gallery}{project website} provides an interactive scene gallery for inspecting frozen environments. Figure~\ref{fig:supp-workbench-viewer} shows the EmbodiedStudio workbench interface displaying the method-figure kitchen.

\paragraph{Inspecting task matching and readiness.}
Request-matching cards focus the requested objects, receiving entities, and initial arrangements in the 3D scene, while task-readiness cards highlight supporting work surfaces, container interiors, and connected access areas. For \texttt{T11}, reviewers can inspect the three designated objects and distractors in the tray, the open receiving box, and the surrounding workspace. In the method-figure kitchen, cards frame the food on the island, the refrigerator receiving shelf, and the aisle beside the island; a dedicated interior view temporarily hides the lower door panel to expose the shelf and cavity. Highlighted bounds derive directly from the frozen scene entities and workspace definitions, providing interactive spatial inspection complementary to Section~\ref{sec:supp-scene-eval}.

\paragraph{Walkthrough video.}
The \href{https://embodiedstudio.github.io/?view=paper\#walkthrough}{project walkthrough} presents the EmbodiedStudio workflow, generated scenes across tabletop, single-room, and multi-room scales, and recorded robot executions.

\noindent\begin{minipage}{\textwidth}
  \centering
  \includegraphics[width=\textwidth]{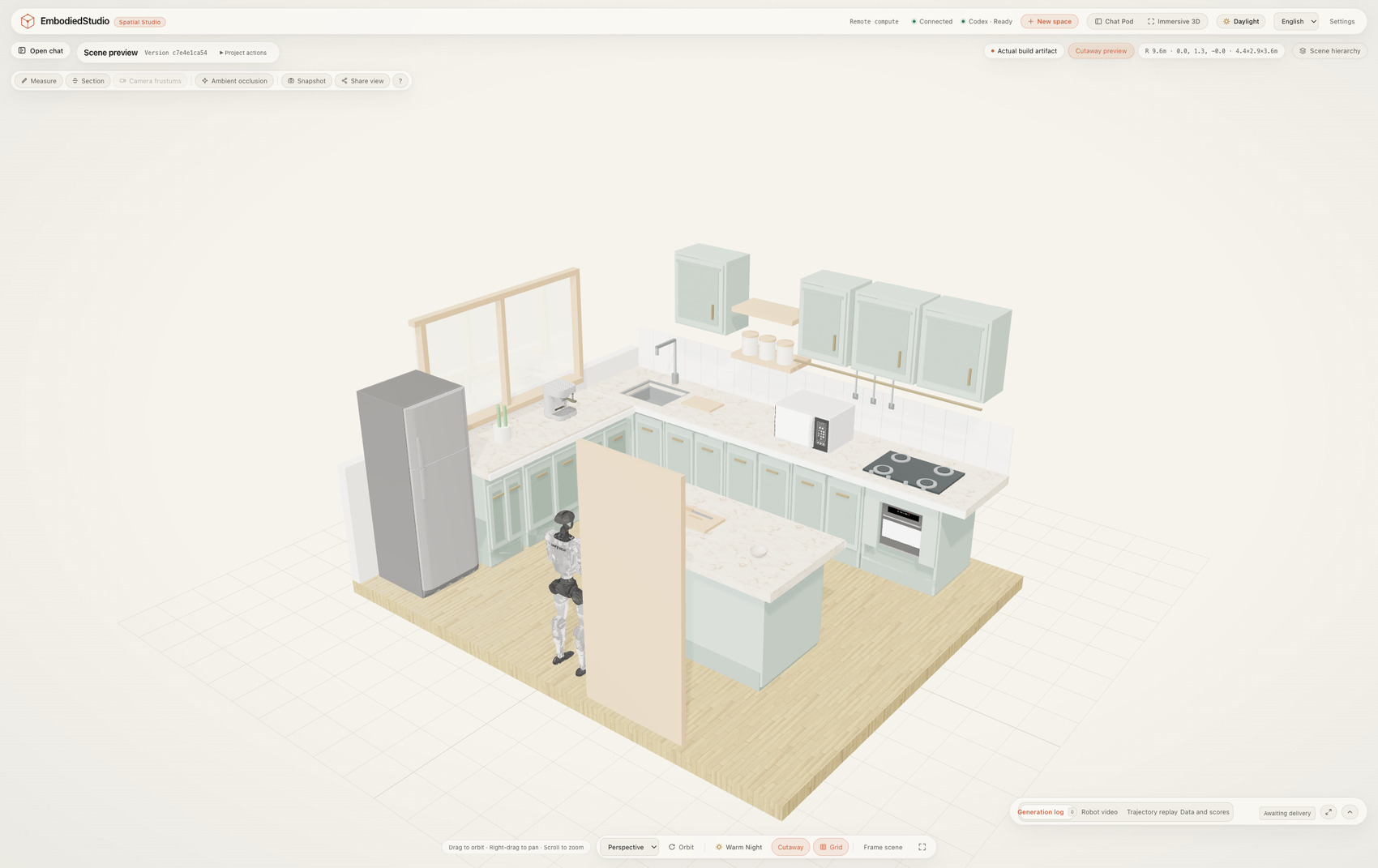}
  \captionof{figure}{\textbf{The EmbodiedStudio workbench interface} displaying the frozen kitchen used in the method figure.}
  \label{fig:supp-workbench-viewer}
\end{minipage}

\end{document}